%% file: main.tex
\documentclass{article}

\usepackage{microtype}
\usepackage{graphicx}
\usepackage{booktabs} 

\usepackage{hyperref}

\usepackage[accepted]{icml2023}

\usepackage{amsmath}
\usepackage{amssymb}
\usepackage{mathtools}
\usepackage{amsthm}

\usepackage[capitalize,noabbrev]{cleveref}

\theoremstyle{plain}

\theoremstyle{definition}

\theoremstyle{remark}

\usepackage[colorinlistoftodos,prependcaption,textsize=tiny,textwidth=1.8cm]{todonotes}
\usepackage{pgfplots}
\DeclareUnicodeCharacter{2212}{−}
\usepgfplotslibrary{groupplots,dateplot}
\usetikzlibrary{patterns,shapes.arrows}
\pgfplotsset{compat=newest}
\pgfplotsset{
    BarStyle/.style={
        ybar interval,
        fill=white!80!black,draw=black
    },
}
\usetikzlibrary{shapes,decorations,arrows,calc,arrows.meta,fit,positioning}
\usetikzlibrary{positioning,calc}
\usetikzlibrary{backgrounds,fit}
\usetikzlibrary{arrows.meta}
\usetikzlibrary{arrows}
\usepackage{caption}
\usepackage{subcaption}
\usepackage[shortlabels]{enumitem}
\usepackage{color,soul}
\usepackage{bm}
\usepackage{xcolor}

\usepackage{array}
\newcolumntype{H}{>{\setbox0=\hbox\bgroup}c<{\egroup}@{}}
\usepackage{tabu}
\newcommand{\slimparagraph}[1]{\textbf{#1~~~}}

\definecolor{limegreen}{HTML}{8DC73E}
\definecolor{thistle}{HTML}{D883B7}
\newcommand{\pfnbotext}[1]{\colorbox{limegreen}{#1}}
\newcommand{\standardbotext}[1]{\colorbox{thistle}{#1}}

\newcommand{\priorparameters}{\phi}
\newcommand{\priorhyperparameters}{\psi}

\def\vx{{\bm{x}}}
\def\vy{{\bm{y}}}

\DeclareMathOperator*{\argmax}{arg\,max}

\icmltitlerunning{PFNs4BO: In-Context Learning for Bayesian Optimization}

\begin{document}

\twocolumn[
\icmltitle{PFNs4BO: In-Context Learning for Bayesian Optimization}



\icmlsetsymbol{equal}{*}

\begin{icmlauthorlist}
\icmlauthor{Samuel M\"uller}{freiburg,priorlabs}
\icmlauthor{Matthias Feurer}{freiburg}
\icmlauthor{Noah Hollmann}{freiburg,priorlabs,charite}
\icmlauthor{Frank Hutter}{freiburg,priorlabs}
\end{icmlauthorlist}

\icmlaffiliation{freiburg}{University of Freiburg, Germany}
\icmlaffiliation{priorlabs}{\href{http://priorlabs.ai/}{Prior Labs}}
\icmlaffiliation{charite}{Charit\'e -- Berlin University of Medicine, Germany}

\icmlcorrespondingauthor{Samuel M\"uller}{muellesa@cs.uni-freiburg.de}

\icmlkeywords{Bayesian optimization, prior-data fitted networks, posterior predictive distribution, Gaussian process, Bayesian neural network, user priors, non-myopic BO, hyperparameter optimization, optimal experimental design}

\vskip 0.3in
]



\printAffiliationsAndNotice{}  

\begin{abstract}
In this paper, we use Prior-data Fitted Networks (PFNs) as a flexible surrogate for Bayesian Optimization (BO).
PFNs are neural processes that are trained to approximate the posterior predictive distribution (PPD) through in-context learning on any prior distribution that can be efficiently sampled from.
We describe how this flexibility can be exploited for surrogate modeling in BO.
We use PFNs to mimic a naive Gaussian process (GP), an advanced GP, and a Bayesian Neural Network (BNN).
In addition, we show how to incorporate further information into the prior, such as allowing hints about the position of optima (user priors), ignoring irrelevant dimensions, and performing non-myopic BO by learning the acquisition function.
The flexibility underlying these extensions opens up vast possibilities for using PFNs for BO. 
We demonstrate the usefulness of PFNs for BO in a large-scale evaluation on artificial GP samples and three different hyperparameter optimization testbeds: HPO-B, Bayesmark, and PD1.
We publish code alongside trained models at \href{https://github.com/automl/PFNs4BO}{github.com/automl/PFNs4BO}.
\end{abstract}

\section{Introduction}
Gaussian processes (GPs) are today\textquotesingle{s} de facto standard surrogate model in Bayesian Optimization (BO;~\citealp{frazier-arxiv18a,garnett-book22a}).
This dominance can be attributed to both their strong performance and their mathematical convenience.
However, a GP can only model priors that can be encoded as a valid kernel function and is jointly normal-distributed.
Moreover, while kernel hyperparameters could be treated in a Bayesian manner with Markov chain Monte Carlo, this is typically not done due to the high computational cost,
even though the fully Bayesian treatment was already shown to yield stronger results a decade ago~\citep{benassi-lion11a,snoek-nips12a,eriksson2021high}.
Instead, GPs are usually fitted using maximum likelihood, often called ``Empirical Bayes''. This makes Bayesian optimization less principled from a Bayesian perspective.

\begin{figure}[t]
    \centering
    \input{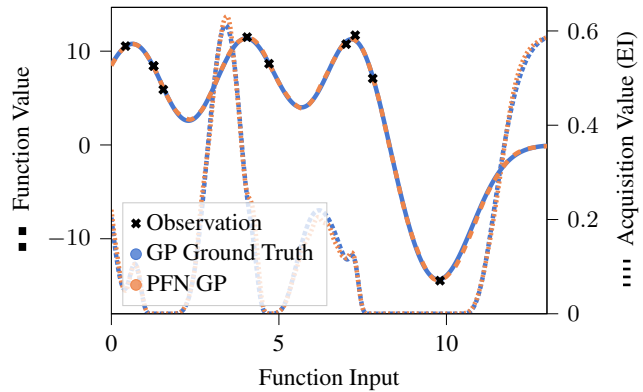}
    \vspace{-.1cm}
    \caption{Our proposed Prior-data Fitted Network almost exactly approximates a Gaussian Process posterior with fixed hyperparameters. We plot the exact and approximated GP prediction for (i) the mean and (ii) expected improvement. For the simple GP model approximated here, a ground truth can be exactly calculated, which is generally not the case, see Section \ref{sec:simple gp}. PFNs, however, can be extended to approximate any prior one can sample from.}
    \label{fig:front_page_illustration}
\end{figure}

The recently proposed Prior-data Fitted Networks (PFNs, \citealp{muller-iclr22a}) show that fast approximate Bayesian inference is possible by training a neural network, more specifically, a specific type of neural process~\citep{garnelo-icml18a}, to mimic the posterior predictive distribution (PPD) in a single forward pass using \emph{in-context learning}.
This is a powerful approach, as it makes approximate Bayesian inference readily usable in novel applications and allows using any prior that we can sample from, e.g.\ a GP kernel and its hyperparameters, or also a Bayesian neural network.
PFNs can be a robust and generalizable PPD approximation method, e.g.\ for tabular data using a prior over different neural architectures and their weights~\citep{hollmann-iclr23a}. 

In this work, we demonstrate the flexibility and effectiveness of PFNs as a Bayesian drop-in replacement for Gaussian Processes.
We perform Bayesian optimization with three different kinds of priors, which we describe in Section \ref{sec:priors}: a GP-based and a BNN-based prior, as well as a prior that mimics the state-of-the-art BO implementation, HEBO~\citep{cowenrivers-jair22a}.
In Section \ref{sec:bopfn}, we show that standard acquisition functions, such as probability of improvement (PI), expected improvement (EI), or upper confidence bound (UCB), are easy to implement analytically in the PFN framework and that PFNs can produce sensible results for BO, as illustrated in Figure~\ref{fig:front_page_illustration} for a PFN trained on a simple GP prior for which results closely follow the ground truth GP posterior. 
We also show that PFNs can readily be combined with gradient-based optimization techniques to optimize input warping \citep{snoek-icml14a} and acquisition functions.
In Section \ref{sec:prior-extensions},
we show how our approach can be extended to allow the user to specify prior beliefs about where optima lie, which in our experiments significantly improves performance when accurate prior beliefs exist, as well as an extension towards non-myopic optimization.

In our experiments, we focus on BO for hyperparameter optimization (HPO, \citealp{feurer-automlbook19a}).
HPO is a crucial task for achieving top performance with machine learning algorithms and a key application of BO~\citep{brochu-arxiv10a,snoek-nips12a,garnett-book22a}. 
We show that PFNs perform strongly across three benchmarks, including tuning hyperparameters of large neural networks.

\section{Bayesian Optimization}
BO is a popular technique to find maxima of (noisy) black-box functions in as few evaluations as possible~\citep{garnett-book22a}. 
BO is an iterative procedure that switches between modeling outcomes based on all observations $D_k$ up to a time-step $k$ with a probabilistic model $\hat{f}(y|x,D_k)$, and using an acquisition function $\alpha(\hat{f}(y|x,D_k))$ to decide which point to query next.
The acquisition function accepts a distribution over possible outcomes, typically the predictive posterior distribution, to trade off exploitation (trying to improve solutions in known good areas) and exploration (reducing the posterior uncertainty in unknown areas).
In this work, we restrict optimization to real- and integer-valued inputs $x$, but BO has been extended to more complex design spaces~\cite{hutter-lion11a,swersky-bayesopt13a,korovina-aistats20a,ru-iclr21a,daulton-neurips22a}.
We show pseudocode for the BO loop in Algorithm~\ref{alg:boloop} (purple depicts the standard setting, and green PFN-based models, introduced in Sections~\ref{sec:bopfn}-\ref{sec:prior-extensions}). 
We refer to \citet{brochu-arxiv10a}, \citet{frazier-arxiv18a} and \citet{garnett-book22a} for thorough introductions to Bayesian optimization. We will now focus on GPs for BO and briefly review other models in Appendix~\ref{sec:more_models}.

\begin{algorithm}[t]
\begin{algorithmic}
    \STATE {\bfseries Input} \standardbotext{hyperparameter prior settings}, \pfnbotext{PFN $q_\theta$ trained} \pfnbotext{on a prior distribution over datasets $p(\mathcal{D})$}, initial observations $D=\{ (\vx_1, y_1),...,(\vx_k, y_k)\}$, search space $\mathcal{X}$, number of BO iterations $K$, black-box function $f$ to optimize, acquisition function $\alpha$
    \STATE {\bfseries Output} {Best observed input $\vx_*$ and response $y*$}
    \FOR{$i\gets k + 1$ {\bfseries to} $K$}
  \STATE \standardbotext{Fit GP model $\hat{f}$ to data $D$}
  \STATE Suggest $\vx \in \argmax_{\hat{\vx} \in \mathcal{X}} \alpha (\hat{\vx}, D, \standardbotext{$\hat{f}$} or$\pfnbotext{$q_\theta(\cdot \mid D)$})
  \STATE Update history with response $D \leftarrow D \cup \{ (\vx, f(\vx))\}$
  \ENDFOR

 \STATE Return best configuration: $\argmax_{(\vx_i,y_i) \in D} y_i$
    
\end{algorithmic}
 \caption{Bayesian optimization with \standardbotext{GPs} or \pfnbotext{PFNs}}
 \label{alg:boloop}
\end{algorithm}

\paragraph{Gaussian Processes}
GPs have been widely adopted as probabilistic surrogates in BO due to their flexibility and analytic tractability~\citep{rasmussen-book06a}. 
However, traditional GPs exhibit a cubic scaling, which limits their application in large-data settings.
They also assume a joint Gaussian distribution of the data introducing a model mismatch for long-tailed data. Furthermore, the predominant use of stationary kernels renders the optimization of nonstationary and heteroscedastic functions problematic.
Two further problems come from the kernel that defines the GP's covariance matrix. First, the model prior must be encoded as a valid kernel function, which complicates representing categorical or hierarchical concepts. Second, the kernel hyperparameters need to be tuned to the data, which suffers from the curse of dimensionality when using Empirical Bayes.

The current state-of-the-art BO method using GPs for hyperparameter optimization is HEBO~\citep{cowenrivers-jair22a}. We describe HEBO and extend on it in Section~\ref{sec:heboprior}.

\section{Bayesian Optimization with PFNs}\label{sec:bopfn}
We will now show how to train and use PFNs, combine common acquisition functions with them, and incorporate gradient-based optimization at suggestion time for acquisition function optimization and input warping~\citep{snoek-icml14a}.
\subsection{Background on Prior-Data Fitted Networks}
\label{sec:prior_fitted_intro}

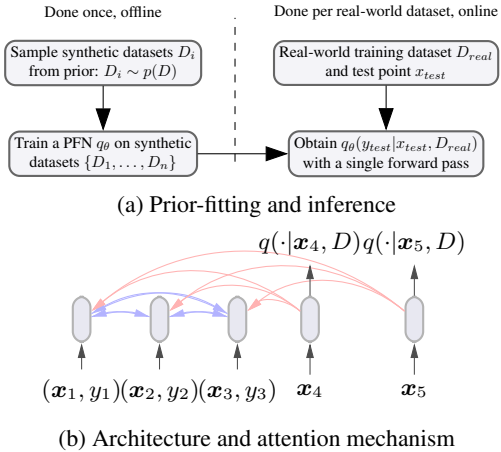
\begin{figure}[t]
     \centering
     \begin{subfigure}[t]{.49\textwidth}
         \centering
         \input{icml2023/figures/pfn_usage}
         \caption{Prior-fitting and inference}
         \label{fig:pfn_usage}
     \end{subfigure}
     \begin{subfigure}[t]{.49\textwidth}
         \centering
         \input{icml2023/figures/transformer_visualization}
         \caption{Architecture and attention mechanism}
         \label{fig:transformer_visualization}
     \end{subfigure}
     \caption{(a) The PFN learns to approximate the PPD of a given prior offline to yield predictions on new observations in a single forward pass.
     (b) The positions representing the training samples $(x_i,y_i)$ can attend only to each other; test positions (for $x_4$ and $x_5$) attend only to the training positions. Plots based on \citet{muller-iclr22a}, with permission.}
     \label{fig:pfn_overview}
\end{figure}

Prior-data Fitted Networks (PFNs, \citealp{muller-iclr22a}) are neural networks trained to approximate the Posterior Predictive Distribution $p(y|\vx,D)$ for supervised learning tasks.
We visualize their use in Figure~\ref{fig:pfn_usage}.

\slimparagraph{Prior-fitting} 
Prior-fitting is the training of a PFN to approximate the PPD and thus perform Bayesian prediction for a particular, chosen, prior.
We assume that there is a sampling scheme for the prior s.t.\ we can sample datasets of inputs and outputs from it: $D \sim p(D)$.
This requirement is easy to satisfy for most priors.
For a GP, for example, this can be achieved by sampling outcomes from the GP prior.
We describe our priors in more detail in Section~\ref{sec:priors}.
We repeatedly sample synthetic datasets $D=\{(\vx_i, y_i)\}_{i \in \{1,\dots,n\}}$ and optimize the PFN\textquotesingle{}s parameters $\theta$ to make predictions for $(\vx_{test},y_{test}) \in D$, conditioned on the rest of the dataset $D_{train} = D \setminus \{(\vx_{test},y_{test})\}$.
Our PFN $q_\theta$ is an approximation to the PPD and thus accepts a training set and a test input and returns a distribution over outcomes for the test input.
The loss of the PFN training is the cross-entropy on the held-out examples
\begin{equation}\small
    \mathbb{E}_{(\vx_{test},y_{test}) \cup D_{train} \sim p(\mathcal{D})}[ - \text{log} \  q_{ \theta}(y_{test} |\vx_{test}, D_{train}) ],
    \label{equation:pfn_loss}
\end{equation}
and minimizing this loss approximates Bayesian prediction~\citep{muller-iclr22a}. 
Crucially, this \textit{synthetic prior-fitting} phase is performed only once for a given prior $p(D)$ as part of algorithm development.
For technical details of our training procedure, we refer to Appendix \ref{appendix:training details}.

\slimparagraph{Real-World Inference} During inference, the trained model is applied to unseen real-world observations (see Figure \ref{fig:pfn_usage}).
For a novel dataset with observations $D$ and test features $x$, feeding $\langle{}D,x\rangle$ as an input to the model trained above yields an approximation to the PPD 
$q_{\theta}(y|\vx,D)$ through in-context learning.

\slimparagraph{Architecture} 
PFNs rely on a Transformer \citep{vaswani-neurips17a} that encodes each feature vector and label as a vector representation, allowing these representations to attend to each other, as depicted in Figure \ref{fig:transformer_visualization}.
They accept a variable-length training set $D$ of feature and label vectors (treated as a set-valued input to exploit permutation invariance) as well as a variable-length query set of feature vectors $\{\vx_{1}, \dots, \vx_{m}\}$ and return estimates of the PPD for each query.
This architecture also fulfills all requirements for a neural process architecture \citep{nguyen2022transformer}.

\slimparagraph{Regression Head} PFNs do not use a standard regression head, but instead make predictions with a discretized distribution, the Riemann distribution. Consequently, the approximated PPD $q$ becomes a discrete distribution.
This allows PFNs to treat regression as a classification problem, shown to be clearly advantageous for PFNs over a classical regression head~\citep{muller-iclr22a}.
Please see Appendix \ref{app:regression head} for details.

\newcommand{\best}{f^*}

\subsection{PFNs as Models in Bayesian Optimization}
We contrast Bayesian optimization with surrogates based on a PFN and a GP with Empirical Bayes in Algorithm~\ref{alg:boloop}; the standard components colored in purple are replaced by the new ones in green.
The trained PFN is passed to the BO algorithm and used in the acquisition function as a replacement of the standard GP model.

While PFNs have an up-front cost of fitting the prior, Empirical Bayes incurs the online cost of fitting the hyperparameters in each iteration.
Crucially, PFNs incur the training cost exactly once per prior, and a single trained PFN can be used for BO on different tasks with varying dimensions, since the data $D$ is passed to the PFN for in-context learning (in a single forward pass). The offline training of the PFN can be compared to developing the code base for GP regression and tuning its hyperparameter spaces: a one-time investment that is then generally applicable.
We list important details for BO with PFNs in Appendix~\ref{sec:bo tricks}.

\subsection{Prior Work using Transformers as Surrogates}
Three recent works propose Transformers as surrogates.

\citet{maraval-arxiv22a} investigate the feasibility of using PFNs in Bayesian Optimization (BO) on toy functions in their early work.
They demonstrate that using GP priors with a PFN can match or outperform GPs while being an order of magnitude faster.
However, their work only evaluates a small set of toy functions and does not leverage the flexibility of PFNs to model extensions to a simple GP model.
Finally, the authors train one PFN per search space dimensionality, while we share one network across all search spaces.
Their paper reports that (i) models trained on uniformly distributed inputs impair predictive accuracy on non-uniform data, and (ii) their PFNs require many novel observations to adapt posteriors. However, they used extremely few epochs for training, and we observe that both of these issues vanish
when increasing the number of epochs for prior-fitting to the magnitude used by \citet{muller-iclr22a}.

Similarly, \citet{nguyen-icml22a} used a Transformer to do Bayesian Optimization on toy functions. Interestingly, they did not use a Riemann distribution to predict the output, but instead parameterized a normal distribution with the outputs of the neural network.

The only application to real-world data was done with the recent
Optformer~\citep{chen-arxiv22a}.
Optformer is a Transformer that utilizes transfer learning on previously recorded BO optimizations performed by other optimizers. As such, it is not a Bayesian method; instead, the Optformer models full BO trajectories, including observations, rather than acting as a surrogate model. Additionally, the Optformer incorporates additional user-provided information, such as the names of the optimization dimensions or of the metric to optimize.

\subsection{Acquisition Functions}
When we condition a PFN on new observations, we obtain an approximation of the PPD in a forward-pass $p(y |\vx, D) \approx q_{ \theta}(y |\vx, D)$ in the form of a Riemann distribution, that is, a piece-wise constant distribution over bins spanning a reasonable output range~\citep{muller-iclr22a}.
The Riemann distribution allows us to calculate the utility for different acquisition functions exactly, e.g., EI, PI, or UCB, as described by \citet{chen-arxiv22a}.
To exemplify the general approach, we outline how to compute $\text{PI}(y)=\int_{-\infty}^{\infty} [y>\best{}]p(y)dy$ for the unbounded Riemann distribution in Appendix \ref{appendix:riemann_acqf}.
We experimented with EI, PI and UCB and found simple EI to work robustly. We show a comparison in Appendix Table \ref{table:acq_ablation_on_hpob}.

\subsection{Gradient-based Optimization at Inference Time}
\label{sec:acq f opt and input warping}
The PFN, being a standard Transformer under the hood, propagates gradients from its outputs to its inputs.
We use gradient ascent to find the acquisition function\textquotesingle{}s maximum and to tune our input warping, which we outline below.

\slimparagraph{Acquisition Function Optimization}
Our acquisition functions are optimized with an extensive initial random search followed by a gradient-based optimization of the strongest candidates, similar to previous work \citep{snoek-nips12a}.
We refer to Appendix \ref{app:acq func opt} for more details.

\slimparagraph{Input Warping}
To improve performance on search spaces with misspecified scales, e.g., a log scale that is not declared as such, we warp features before passing them to the PFN.
This can be necessary as missing log scaling can result in almost all values lying in a range that is difficult for the PFN to handle numerically (e.g., in the lowest percentile of the search space).
We follow \citet{cowenrivers-jair22a} and use the CDF of a Kumaraswamy distribution $w(\cdot; a, b)$, where $a$ and $b$ are tuned per feature, to warp features to
\begin{align}
    w(x;a,b) = 1-(1-x^{a})^{b}.
\end{align}
Traditionally, the parameters $a$ and $b$ are tuned to maximize the likelihood of the observed data, i.e., maximize $p(D)$.
While the data likelihood can be computed with a PFN, this is expensive, as one needs to compute the factorization $p(y_1|\vx_1)p(y_2|\vx_2,\{(\vx_1, y_1)\})\cdots$.
Instead, we use the likelihood $\prod_{(\vx,y) \in D} p(y|\vx,D)$ of observing the same $y$s again, which is cheaper to compute.
We view this as a measure for the amount of noise a PFN uses to explain the data $D$. E.g., if a prediction $p(y_1|x_1,{\{(\vx_i,y_i)\}}_{i \in \{1,\dots,n\}})$ has a high probability, the network does not assign a lot of probability to noise changing the value of $y$ when re-evaluating.
While this approximation works well for our prior, it might fail in other setups.
We ablate the impact of input warping in Appendix \ref{sec:ablation studies}.

\section{Priors for Bayesian Optimization}\label{sec:priors}
In this work we use a set of three different priors to show the flexibility of PFNs as a surrogate.

\subsection{A Simple Prior Based on a Simple GP (GP)}
\label{sec:simple gp}
We use the prior of a simple GP with fixed hyperparameters to show-case our method (see Figure \ref{fig:front_page_illustration} and Figures \ref{fig:full_bo_ei_dumb_gp} and \ref{fig:full bo ei dumb nonsmooth traj}), as it allows us to compare against the ground truth posterior.
Here we use an RBF kernel with zero mean.
During prior-fitting, we sample $N$ inputs $x_i$ uniformly at random from the unit-hypercube.
Then we sample all outputs $\vy$ from the GP prior.
We can simply sample $\vy \sim \mathcal{N}(\mathbf{0}, K)$, where $K_{i,j} = k_{\text{RBF}}(x_i,x_j)$.

\subsection{A HEBO-inspired Prior (HEBO and HEBO$^+$)}\label{sec:heboprior}
HEBO \citep{cowenrivers-jair22a} is a state-of-the-art BO method that won the NeurIPS black-box competition~\citep{turner-neuripscomp21a} and demonstrated excellent performance in a recent empirical evaluation~\citep{eggensperger-neuripsdbt21a}. It performs non-linear input and output warping for robust surrogate modeling and combines it with a well-engineered GP prior.
We use it as a starting point for our own HEBO-inspired prior for PFN training. Our method considers a set of parameters $\priorparameters$: the lengthscale per dimension, the output scale and the noise.
These are modeled as independent random variables subject to some distribution $p(\priorparameters; \priorhyperparameters{})$ that depends on hyperparameters $\priorhyperparameters$.

To sample functions, we perform the following four steps per dataset: \textit{i)} Sample all $\vx$ from a uniform distribution, \textit{ii)} sample the parameters $\priorparameters{} \sim p(\priorparameters{}; \priorhyperparameters{})$,
\textit{iii)} draw the outputs for our dataset
$y \sim \mathcal{N}(0,K(\vx; \priorparameters))$, where $K$ is the covariance matrix defined by the 3/2 Matérn kernel, that uses the sampled $\mathrm{outputscale}$, $\mathrm{lengthscale}$ and $\mathrm{noise}$. For additional details on $\priorhyperparameters{}$, we refer to Appendix \ref{appendix:detail_on_prior}.

We use this prior in a form that is as close to the original HEBO as possible and a variant we dub HEBO$^+$ that extends on HEBO (see Section \ref{sec:ignore feats}) and is tuned to work well with PFNs (see Section \ref{sec:experiments}). 

\subsection{Bayesian Neural Network Prior (BNN Prior)}
We follow previous work~\citep{muller-iclr22a,hollmann-iclr23a} in building our Bayesian Neural Network (BNN) prior.
To sample datasets from this prior, we (i) first sample one network architecture, i.e., the number of layers, the number of hidden nodes, the sparseness, the amount of Gaussian noise added per unit, and the standard deviation with which to sample weights (ii) sample the network's weights from a normal distribution. 
For each dataset, we first sample a BNN and then sample inputs to the BNN $\vx$ uniformly at random from the unit hypercube, feed these through the network, and use the output values as the target $y$.
Additionally, we employed input warping as described in Section \ref{sec: input warping prior extension}.
For more details, see Appendix \ref{appendix:detail_on_prior}.

\section{Prior Extensions}\label{sec:prior-extensions}
PFNs offer versatility, making it simple to explore new priors for BO, such as the above BNN prior, which incorporates a distribution over architectures. In this section, we discuss modifications that can easily be combined with priors, but would be harder to incorporate in traditional GPs.

\subsection{Input Warping}
\label{sec: input warping prior extension}
In addition to using input warping after prior-fitting (see Section~\ref{sec:acq f opt and input warping}), we can include a Bayesian formulation of input warping in the prior directly and include it in prior-fitting.
That is, (i) we sample warping hyperparameters $h_{warp}$ randomly from a predefined distribution and then (ii) warp inputs of a synthetic prior dataset with an exponential transform and hyperparameters $h_{warp}$.
While we found this to be beneficial, it was not powerful enough to remove the need for feature warping after prior-fitting completely.

\subsection{Spurious Dimensions}
\label{sec:ignore feats}
Many real-world tuning tasks contain irrelevant features, which do not influence the output~\citep{bergstra-jmlr12a}.
While it would be hard with traditional GPs and Empirical Bayes to encode a specific chance of a feature being spurious in the prior, for PFNs, we can do this easily by adding irrelevant features at random during prior-fitting.
These features are simply not fed to the GP/NN which generates the outcomes for the dataset.
In our final HEBO$^+$ prior we use 30\% irrelevant features.
We found that this improves performance, especially for large search spaces, and show a steep improvement on the three largest search spaces in HPO-B in Table~\ref{tab:ignore_feats_xgboost} of the Appendix.

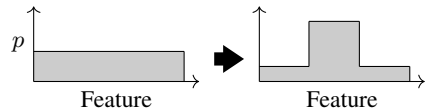
\begin{figure}
    \centering
\begin{tikzpicture}[line cap=corner,every text node part/.style={align=center}]
  \draw[->] (0, 0) -- (2.2, 0) node[below,midway] {\footnotesize Feature};
  \draw[->] (0, 0) -- (0, 1.) node[midway,left] {\footnotesize $p$} ;
  \draw (0,.4) -- (2,.4);
  \draw (2,.4) -- (2,0);
\path[fill=black,fill opacity=0.2] (0,0) -- (0,.4) -- (2,.4) -- (2,0.)  -- cycle;

\draw[-{Triangle[width=10pt,length=8pt]}, line width=5pt](2.4,.3) -- (2.8, .3);

  \draw[->] (3, 0) -- (5.2, 0) node[below,midway] {\footnotesize Feature};
  \draw[->] (3, 0) -- (3, 1.) node[left,midway] {};

  \draw[black] (3,.2) -- (3.66,.2);
  \draw[black] (3.66,.2) -- (3.66,.8);
  \draw[black] (3.66,.8) -- (4.33,.8);
  \draw[black] (4.33,.8) -- (4.33,.2);
  \draw[black] (4.33,.2) -- (5.,.2);
  \draw[black] (5.,.2) -- (5.,0.);
  \path[fill=black,fill opacity=0.2] (3.,0.) -- (3,.2) -- (3.66,.2) -- (3.66,.2) -- (3.66,.8) -- (3.66,.8) -- (4.33,.8) -- (4.33,.8) -- (4.33,.2) -- (4.33,.2) -- (5.,.2) -- (5.,0.)  -- cycle;
\end{tikzpicture}
    \caption{An example of the impact of the user prior on the prior belief about the position of the optimum.}
    \label{fig:user prior mini example}
\vspace{-.4cm}
\end{figure}

\def\intsample{{\mathrm{I}}}
\def\intvar{{\mathrm{I}}}
\def\intset{{\mathcal{I}}}

\subsection{User Priors}
\label{sec:user priors}
The ability to incorporate practitioner\textquotesingle{s} knowledge is pivotal in order to improve the usefulness of automated HPO approaches for Machine Learning as well as in other BO applications.
Previous work has shown that user knowledge can improve BO performance tremendously~\citep{ramachandran-kbs20a, li-arxiv20a, souza-ecml21a, hvarfner-iclr22a}.
One way to incorporate explicit user knowledge  is through a user\textquotesingle{s} belief about the location of the optimum in the search space.
While previous work relies on warping the space \citep{ramachandran-kbs20a}, reweighting the posterior \citep{souza-ecml21a} or reweighting the acquisition function \citep{hvarfner-iclr22a}, PFNs, can integrate these priors in a more direct and sound way.

In this work we let the user define an interval $\intvar \in \mathbf{I}$, in which they belief the optimum lies, and a weighting  $\rho \in [0,1]$, indicating their confidence. This yields a prior:
\begin{align}
    p(D|\rho,\mathrm{I}) = \rho \cdot p(D|m \in \intvar) + (1-\rho) \cdot p(D),\label{eq:conditioned_f}
\end{align}
where $m$ is the maximum of the function underlying the dataset $D$; $y \sim f(x)$ for all $x,y \in D$.
Figure \ref{fig:user prior mini example} shows an exemplary change to the density of the optimum in the prior.

We train a single PFN that can accept any interval $\intsample \in \intset$ and confidence $\rho \in [0,1]$ and adapt its prior on the fly to be $p(D|\rho,\mathrm{I})$.
The extra inputs, $\rho$ and $I$, are fed to the PFN using an extra position with its own linear encoder similar to style embeddings for language models ~\citep{dai-acl19a}.

In Appendix \ref{appendix:user_prior_distribution}, we detail how we build a PFN prior $p(D,\rho,\mathrm{I})$ that is cheap to sample and allows the neural network to adapt to any interval $\intsample$ out of a set of $|\intset|=15$ different intervals per search space dimension.

\subsection{Non-Myopic Acquisition Function Approximation}
\label{sec:non_myopic_acq}
Non-myopic acquisition functions provide an optimal exploration and exploitation strategy, optimizing sampling policies over a rolling horizon. Though promising, computing them is computationally expensive and thus rarely used in practice.
We demonstrate how to use PFNs to enable an effective approximation.
The Knowledge Gradient~\citep{frazier-arxiv18a} measures the predicted improvement in the maximum value of the black-box function by obtaining an additional observation at the candidate point $x$:
\begin{align}
    \alpha_{\text{KG}}(\vx; D) =
\mathbb{E}_{p(y | \vx, D)} [ \tau(D \cup \{(\vx,y)\}) - \tau(D) ],
\end{align}
where $\tau(D) = \max_{\vx \in \mathcal{X}} \mathbb{E}[y | \vx, {D}]$.
Knowledge gradient is the optimal acquisition function choice when optimizing for the mean in the following step.
It is thus an approximation for the non-myopic setting with a one step look-ahead.
Traditionally one can approximate $\alpha_{\text{KG}}$ with a Monte Carlo estimate.
For this, one needs to sample a set of $N$ outcomes $y$ at the current position and an additional set of $M$ positions $x$ at which to evaluate the mean per $y$.
This incurs costs of $M \cdot N$.
Alternatively, this can be solved by a two-level optimization of random batches \citep{wu-neurips16a,wu-neurips17a}.
With a PFN, we can get away without any prediction time optimization, though.
The PFN can directly learn to approximate the $\alpha_{\text{KG}}$.
In Appendix \ref{sec:nonmyopic acq function approx} we detail this method and in Appendix \ref{sec:nonmyopic results} show results, with this non-myopic method outperforming standard EI on four search spaces with few dimensions.

\section{Bayesian optimization experiments}
\label{sec:artificial experiments}
\begin{table}
\centering
\begin{tabular}{lrrr}
\toprule
           \# Features &      1  &      2  &      10 \\
\midrule
\# Wins Empirical Bayes & 206 & 144 & 169 \\
\# Wins PFN & 239 & 154 & 171 \\
\# Ties & \textbf{555} & \textbf{702} & \textbf{660} \\
\bottomrule
\end{tabular}
    \caption{
    BO performance with minimal confounding factors after 50 evaluations. The majority of runs yielded ties, showing PFNs are a strong alternative to Empirical Bayes.}
    \label{fig:toy_results_short}
\end{table}
\vspace{-.2cm}
In this section we analyze the performance of PFNs on prior samples with as few as possible confounding factors.
First, we verify that the PFN predictions match the true GP posterior without Empirical Bayes; as shown in Figures \ref{fig:full_bo_ei_dumb_gp} and \ref{fig:full bo ei dumb nonsmooth traj} in the appendix.
This is not the case when training the PFN on very few ($\leq$100k) prior samples but the approximation becomes tighter with enough samples ($\geq$ 20M).

Next, we consider prediction performance with the original HEBO prior, following their hyper-prior setup as close as possible.
We evaluate the HEBO with Empirical Bayes with our reimplementation of the HEBO model in GPyTorch \citep{gardner-neurips18a}, which has some slight adaptions which we detail in Appendix \ref{sec:approx quality on prior details}.
Table \ref{fig:toy_results} shows that both, PFNs and Empirical Bayes, can approximate the posterior similarly well in terms of likelihood on prior samples.

Finally, we compare BO performance on samples from the HEBO prior.
We create a discrete benchmark with $1\,000$ different datasets per number of dimensions, each containing $1\,000$ evaluations sampled uniformly at random in $[0.0, 1.0]$.
Table \ref{fig:toy_results_short} shows that across dimensionalities, the GP with empirical Bayes and our PFN approximation perform on par, most of the time they find the same maxima.
We refer to Appendix \ref{sec:approx quality on prior details} for a more detailed comparison.

\definecolor{colorRandom}{HTML}{4878d0}
\definecolor{colorRandom_plus_prior}{HTML}{a1c9f4}
\definecolor{colorGP}{HTML}{d65f5f}
\definecolor{colorHEBO}{HTML}{6acc64}
\definecolor{colorDNGO}{HTML}{956cb4}
\definecolor{colorDGP}{HTML}{8c613c}
\definecolor{colorPySOT}{HTML}{dc7ec0}
\definecolor{colorOPT_BEI}{HTML}{956cb4}
\definecolor{colorOPT_REI}{HTML}{8c613c}
\definecolor{colorBOHAMIANN}{HTML}{dc7ec0}
\definecolor{colorHyperopt}{HTML}{797979}
\definecolor{colorPFN_GP}{HTML}{ee854a}
\definecolor{colorPFN_GP_init_at_min}{HTML}{ffb482}
\definecolor{colorPFN_GP_init_at_mid}{HTML}{d0bbff}
\definecolor{colorPFN_GP_plus_prior}{HTML}{d0bbff}
\definecolor{colorPFN_BNN}{HTML}{debb9b}

\newlength\benchmarkplotheight
\setlength{\benchmarkplotheight}{5cm}

\newlength\benchmarkplotwidth
\setlength{\benchmarkplotwidth}{5.5cm}

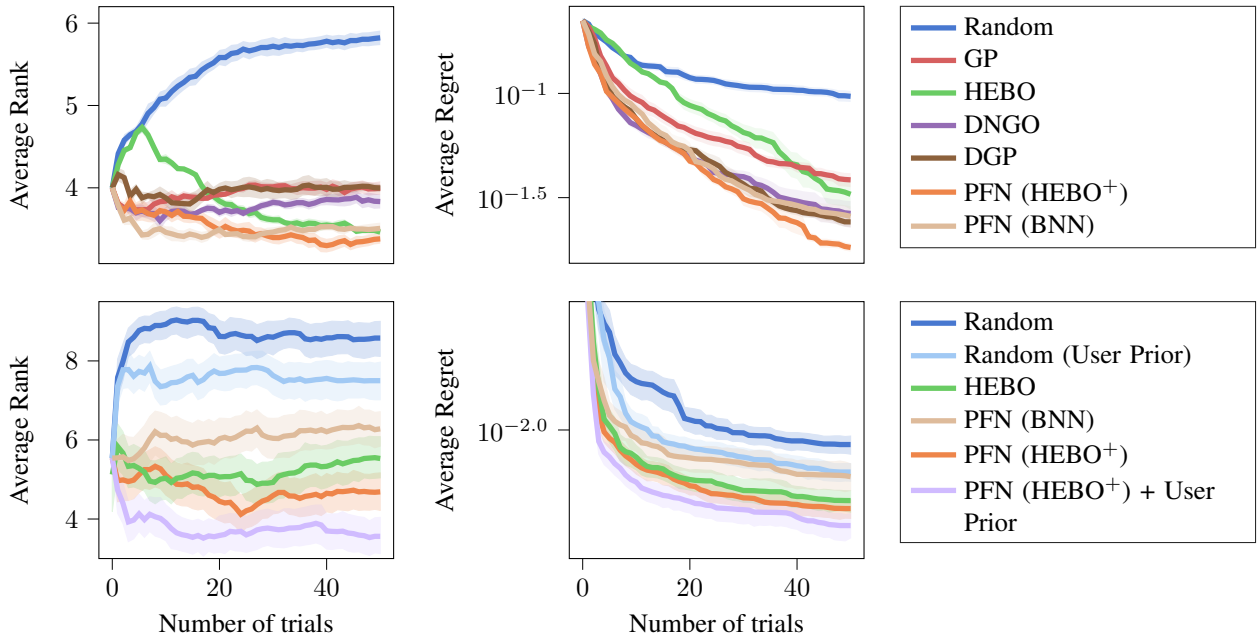
\begin{figure*}[t]
    \begin{subfigure}{1.\textwidth}
        \centering
        \begin{subfigure}[t]{.42\textwidth}
        \vskip 0pt
            \input{icml2023/figures/hpob_main_rank_test.tex}
        \end{subfigure}
        \hspace{-.1\textwidth}
        \begin{subfigure}[t]{.42\textwidth}
        \vskip 0pt
            \input{icml2023/figures/hpob_main_regret_test.tex}
        \end{subfigure}
        \hspace{-.09\textwidth}
        \begin{subfigure}[t]{.33\textwidth}
        \vskip 0pt
            \centering
            \begin{tikzpicture}
                \begin{axis}[%
                hide axis,
                xmin=0,
                xmax=0,
                ymin=0,
                ymax=0,
                legend style={draw=white!15!black,
                legend cell align=center,
                /tikz/every even column/.append style    = {column sep=.5cm, text width=3.7cm}
                },
                legend columns=1
                ]
                \addlegendimage{colorRandom, line width=2pt}
                \addlegendentry{Random};
                \addlegendimage{colorGP, line width=2pt}
                \addlegendentry{GP};
                \addlegendimage{colorHEBO, line width=2pt}
                \addlegendentry{HEBO};
                \addlegendimage{colorDNGO, line width=2pt}
                \addlegendentry{DNGO};
                \addlegendimage{colorDGP, line width=2pt}
                \addlegendentry{DGP};
                
                \addlegendimage{colorPFN_GP, line width=2pt}
                \addlegendentry{PFN (HEBO$^+$)};
                \addlegendimage{colorPFN_BNN, line width=2pt}
                \addlegendentry{PFN (BNN)};
                \end{axis}
            \end{tikzpicture}
        \end{subfigure}
        \vspace{1pt}
    \end{subfigure}
    \begin{subfigure}{1.\textwidth}
        \centering
        \begin{subfigure}[t]{.42\textwidth}
        \vskip 0pt
            \input{icml2023/figures/pd1_ranks_random_hard.tex}
        \end{subfigure}
        \hspace{-.1\textwidth}
        \begin{subfigure}[t]{.42\textwidth}
        \vskip 0pt
            \input{icml2023/figures/pd1_regrets_random_hard.tex}
        \end{subfigure}
        \hspace{-.09\textwidth}
        \begin{subfigure}[t]{.33\textwidth}
        \vskip 0pt
            \centering
            \begin{tikzpicture}
                \begin{axis}[%
                hide axis,
                xmin=0,
                xmax=0,
                ymin=0,
                ymax=0,
                legend style={draw=white!15!black,
                /tikz/every even column/.append style  = {column sep=.5cm, text width=3.7cm},
                },
                legend columns=1
                ]
                \addlegendimage{colorRandom, line width=2pt}
                \addlegendentry{Random};
                \addlegendimage{colorRandom_plus_prior, line width=2pt}
                \addlegendentry{Random (User Prior)};
                \addlegendimage{colorHEBO, line width=2pt}
                \addlegendentry{HEBO};
                \addlegendimage{colorPFN_BNN, line width=2pt}
                \addlegendentry{PFN (BNN)};
                \addlegendimage{colorPFN_GP, line width=2pt}
                \addlegendentry{PFN (HEBO$^+$)};
                \addlegendimage{colorPFN_GP_plus_prior, line width=2pt}
                \addlegendentry[align=left]{PFN (HEBO$^+$) + User Prior};
                \addlegendimage{empty legend}
                \addlegendentry{\ };
                \addlegendimage{empty legend}
                \addlegendentry{\ };
                \end{axis}
            \end{tikzpicture}
        \end{subfigure}
    \end{subfigure}
    \caption{Aggregated average rank and average regret over time on the HPO-B (top) and PD1 (bottom) benchmarks. Shading indicates 95\% confidence intervals. (Top) Aggregate over all HPO-B test search spaces. We can see that both priors seem to yield advantages for the PFN, yielding top performance. We provide per-search-space results in Figure~\ref{fig:hebo_all_search_spaces} in the Appendix. (Bottom) Aggreate over all PD1 tasks, including user priors. We can see that user priors clearly improve performance.}
    \label{fig:hebo_pd1_combined}
\end{figure*}

\begin{figure*}[t]
    \begin{subfigure}{1.\textwidth}
        \centering
        \begin{subfigure}[t]{.42\textwidth}
        \vskip 0pt
            \input{icml2023/figures/bayesmark_rank_plot.tex}
        \end{subfigure}
        \hspace{-.1\textwidth}
        \begin{subfigure}[t]{.42\textwidth}
        \vskip 0pt
            \input{icml2023/figures/bayesmark_mean_plot.tex}
        \end{subfigure}
        \hspace{-.09\textwidth}
        \begin{subfigure}[t]{.33\textwidth}
        \vskip 0pt
            \centering
            \begin{tikzpicture}
                \begin{axis}[%
                hide axis,
                xmin=0,
                xmax=0,
                ymin=0,
                ymax=0,
                legend style={draw=white!15!black,
                legend cell align=center,
                /tikz/every even column/.append style    = {column sep=.5cm, text width=3.7cm}
                },
                legend columns=1
                ]
                \addlegendimage{colorRandom, line width=2pt}
                \addlegendentry{Random};
                \addlegendimage{colorHyperopt, line width=2pt}
                \addlegendentry{HyperOpt};
                \addlegendimage{colorPySOT, line width=2pt}
                \addlegendentry{PySOT};
                \addlegendimage{colorHEBO, line width=2pt}
                \addlegendentry{HEBO};
                \addlegendimage{colorPFN_GP, line width=2pt}
                \addlegendentry{PFN (HEBO$^+$, sobol)};
                \addlegendimage{colorPFN_BNN, line width=2pt}
                \addlegendentry{PFN (BNN, sobol)};
                \addlegendimage{colorPFN_GP_init_at_min, line width=2pt}
                \addlegendentry{PFN (HEBO$^+$, init min)};
                \addlegendimage{colorPFN_GP_init_at_mid, line width=2pt}
                \addlegendentry{PFN (HEBO$^+$, init mid)};
                \end{axis}
            \end{tikzpicture}
        \end{subfigure}
    \end{subfigure}
    \caption{Average rank and average regret over time on the BayesMark Benchmark, where tasks were optimized for accuracy. Shading indicates 95\% confidence intervals. Using a single initial data point works well for the PFN.}
    \label{fig:bayesmark_results}
\end{figure*}
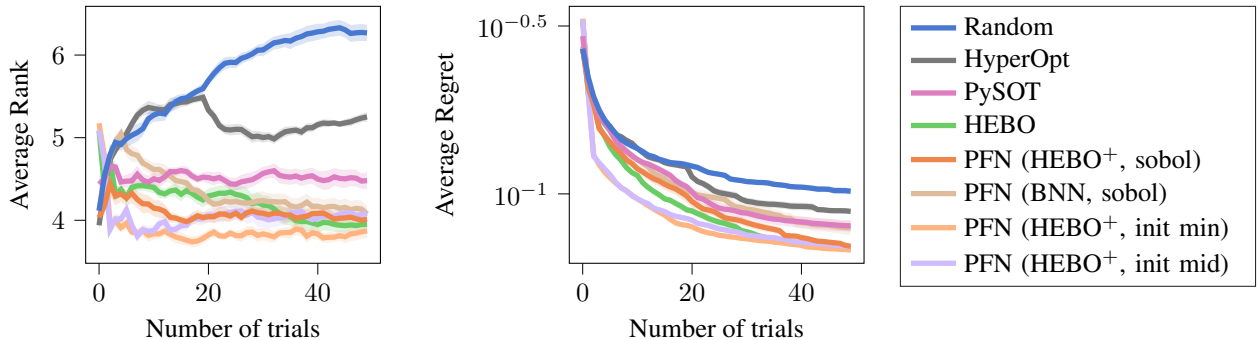

\section{HPO Experiments}
We conduct a large-scale evaluation of PFNs as surrogates for Hyperparameter Optimization (HPO).
We present results on three diverse benchmarks: HPO-B~\citep{pineda-neuripsdbt21a}, a discrete benchmark with a focus on tree-based ML models including XGBoost~\citep{chen-kdd16a}, Bayesmark, a continuous HPO benchmark which was used in the evaluation of the baseline HEBO~\citep{cowenrivers-jair22a}, and PD1~\citep{wang-arxiv21a}, a discrete large neural network tuning benchmark.
We run every optimizer for 50 function evaluations per problem.
HPO-B provides baseline results for 5 repetitions, and we therefore also conduct 5 repetitions, except for Bayesmark, where we conduct 10 repetitions because it is considerably more noisy.
In total we evaluated 105 tasks from a total of 19 search spaces.
We describe a set of ablation studies in Appendix~\ref{sec:ablation studies}, in which we analyse our HEBO-inspired model against a set of HEBO variations to find the changes that improve performance.
Moreover, we study the handling of spurious features and different acquisition functions.

\textbf{Model selection}
Our HEBO and BNN priors contain meta hyperparameters, which we chose once upfront.
This is similar to the GP meta hyperparameters which are chosen once and fixed across tasks, e.g.\ the length scale prior for HEBO.
To determine our global prior hyperparameters, we split off a set of 7 validation search spaces from our largest benchmark, HPO-B, as validation search spaces.
On these we performed one random search for each prior and picked the strongest hyperparameters based on average rank.
We list our search space split in Appendix \ref{appendix:search_space_split}.

\subsection{HPO-B Benchmark}
\label{sec:experiments}

We perform experiments on the 9 test search spaces of the HPO-B benchmark~\citep{pineda-neuripsdbt21a}, which contain a total of 51 different tasks spread across the search spaces.
HPO-B is a discrete benchmark in which the BO tool can only query recorded function evaluations and contains search spaces for Decision trees, SVMs, linear models, random forests and XGBoost.

We compare our method to the following baselines (using the results provided by HPO-B for 1-4): 1) \textit{Random Search}~(RS;~\citealp{bergstra-jmlr12a}), 2) \textit{Gaussian Processes}~(GP;~\citep{jones-jgo98a,snoek-nips12a} 3) \textit{DNGO}~\citep{snoek-icml15a},
4) \textit{Deep Kernel Gaussian Processes}~(DGP;~\citealp{wistuba-iclr21a}),
5) \textit{HEBO}~\citep{cowenrivers-jair22a}. 

We evaluate both PFN priors in Figure~\ref{fig:hebo_pd1_combined} (top).
We can see that the BNN prior performs similar to the best baseline, while the HEBO$^+$ prior outperforms the baselines.
Additionally, in Figure \ref{fig:hpob diff baselines} of the Appendix, we study the impact of input warping on the PFN model and also compare against a BNN baseline and two Optformers~\citep{chen-arxiv22a}.

Early on we found that HPO-B has many instances of wrongly scaled hyperparameters, where a log transformation is missing.
We found that approximately 10\% of the hyperparameters are missing log scaling in HPO-B using a heuristic.
We show one example of such a parameter in the appendix in Figure \ref{fig:show log behaviour}. 
While this likely is a bug in the original development of the benchmark, we still include this benchmark as it is a very important real-world setting.
Users, like the creators of HPO-B, could very well miss to specify a log scaling.
However it is very distinct from the standard BO setting, where user knowledge is used to apply logarithms where needed.
Thus, we used this benchmark to show-case our inference time input warping, as described in Section \ref{sec:acq f opt and input warping}, that can find the right transformation for extremely mis-specified parameters.

\subsection{Continuous HPO on BayesMark}
We use Bayesmark\footnote{\url{https://github.com/uber/bayesmark}} to test the PFN\textquotesingle{s} ability to optimize hyperparameters in continuous search spaces, i.e.\ give the PFN the ability to propose any hyperparameter, not only previously queried hyperparameters and show results in Figure \ref{fig:hebo_pd1_combined}.
We evaluated 10 seeds on 9 methods from scikit-learn~\citep{scikit-learn}, i.e.\ 9 search spaces, training on 4 different datasets with accuracy as evaluation metric. 
This yields a total of 36 different tasks.
We compare our PFNs to the following baselines: 1) HEBO\textquotesingle{s} competition version for BayesMark \citep{turner-neuripscomp21a},
2) HyperOpt~\citep{bergstra-icml13a},
3) PySOT~\citep{PySOT},
and 4) Random Search.
For this benchmark, the question of initial design arises.
First we used $d$ standard Sobol samples, where $d$ is the dimensionality of the search space.
In this setting, we can see that the PFN with a BNN prior is outperformed by HEBO, while outperforming all other baselines, while the HEBO$^+$ prior helps the PFN to perform comparable or slightly worse than HEBO. 

Since our PFN is fully Bayesian, there should be no need for an initial design, as the model will never overfit the data.
We thus experimented with two more initial designs of size one as the PPD cannot be used without a single training point:
We initialize i) at the middle of the search space and ii) in the lower corner of the search space. The first turned out to be slightly better than standard Sobol sampling, while the second was clearly superior to all other approaches. 
These results are surprising, especially considering the bad performance these initializations seem to have at step 0.

\subsection{Neural Network HPO with PD1 and User Priors}\label{appendix:pd1}
While the focus of the previous benchmarks was HPO for traditional machine learning methods, the PD1 benchmark~\citet{wang-arxiv21a} focuses on real-world large neural network setups.
PD1 is a discrete benchmark, considering different relevant tuning tasks e.g.\ of ResNets \citep{he-cvpr16a} or Transformer language models \citep{vaswani-neurips17a}.
All tasks in PD1 share the same five dimensional search space, which is given in Table \ref{tab:userprior} of the Appendix.
For each task we ran 5 seeds with a different (shared) single initial given evaluation.
We do not perform any further initial design for our PFNs.
Figure~\ref{fig:hebo_pd1_combined} (bottom) shows that this time the PFN with BNN prior is clearly outperformed by the HEBO baseline, but that the PFN with the HEBO$^+$ prior performs better.

\slimparagraph{User Prior}
Since PD1 only has a single search space, we used it to evaluated the performance of a simple user prior.
As we lacked experience with the optimizer setup used in PD1, we defined a single generic user prior for the benchmark based on the location of optima across all 18 tasks.
We did not choose to make it a very strong prior, though, as in the most extreme case it places half of the prior weight on optima in one-fourth of the transformed search space.
This generic user prior does not reveal the exact location of optima, but rather provides hints towards them.
The definition can be found in Table \ref{tab:userprior} of the Appendix.

In Figure~\ref{fig:hebo_pd1_combined} (bottom), we can see that the user prior yields strong improvements over our HEBO$^+$ prior.
We contrast our method to a quasi-random search that samples randomly from the user prior.
Although this approach improves upon random search, it is not competitive with any BO method, thus the prior clearly does not make the BO problem trivial.

\section{Conclusion \& Limitations}\label{sec:conclusion}
We showed that PFNs can be trained in a way that they can be powerful BO surrogates and incorporate function priors that would be complicated to model otherwise.

Still, we believe that there is a lot of room to improve upon our work.
Unfortunately, our work has several limitations. First, PFNs tend to work worse for data that has a very low likelihood in the prior distribution, which is not such a big problem for GPs.
Second, we could not find a setting that recovers mislabeled log dimensions, but still performs very strong on correctly specified search spaces. That is why we use input-warping only for HPO-B, but not for the other benchmarks.
Third, compared to GPs, we cannot draw joint samples from multiple data points, which prohibits the straight-forward adaptation of some acquisition functions such as the noisy EI \citep{letham-ba18a}.
Fourth, we have a strong focus on accuracy as evaluation metric in HPO tasks, with only PD-1 having non-accuracy tasks.
Our method seems to work worse for non-accuracy based tasks, as they typically have very different dynamics, even when using a power transformation.
Fourth, PFNs were only shown to perform well up to around 1000 data points \citep{hollmann-iclr23a} so far, but we are hopeful for future improvements.

Furthermore, we plan to extend our work to contain more elaborate priors that can handle non-stationary functions~\citep{assael-arxiv14a,martinez-cantin-bayesopt15a,wabersich-bayesopt16a}, heteroscedastic noise~\citep{griffiths-iopscience22a}, discrete and categorical variables~\citep{daulton-neurips22a}, discontinuous functions~\citep{jenatton-icml17a,levesque-ieee17a} and 100s or 1000s of irrelevant features~\citep{wang-jair16a}. In addition, we expect that it will be easier to use other random processes such as student-t processes~\citep{shah-aistats14a} or other non-Gaussian output distributions,
like log-normals~\citep{eggensperger-ijcai18a}. 

We have demonstrated how user priors can be added into the model, and we plan to extend this to allow more flexible user priors that allow specifying details about the behavior of the black-box function.
Following up on this, a great addition will be learning user priors via transfer learning~\citep{wistuba-ecml16a,rijn-kdd18a,vanschoren-automlbook19a,feurer-arxiv22a} or allowing simpler user priors based on just a few good starting points.

We would like to end this paper by once again highlighting the potential of PFNs as surrogate in BO as they can be easily adapted to model any efficient to sample prior.

\section*{Acknowledgements}
Robert Bosch GmbH is acknowledged for financial support. 
This research was supported by the Deutsche Forschungsgemeinschaft (DFG, German Research Foundation) under grant number 417962828 and by the state of Baden-W\"{u}rttemberg through bwHPC and the German Research Foundation (DFG) through grant no INST 39/963-1 FUGG. The authors acknowledge funding through the research network ``Responsive and Scalable Learning for Robots Assisting Humans'' (ReScaLe) of the University of Freiburg. The ReScaLe project is funded by the Carl Zeiss Foundation.
We acknowledge funding through the European Research Council (ERC) Consolidator Grant ``Deep Learning 2.0'' (grant no.\ 101045765). Funded by the European Union. Views and opinions expressed are however those of the author(s) only and do not necessarily reflect those of the European Union or the ERC. Neither the European Union nor the ERC can be held responsible for them. 
\begin{center}\includegraphics[width=0.3\textwidth]{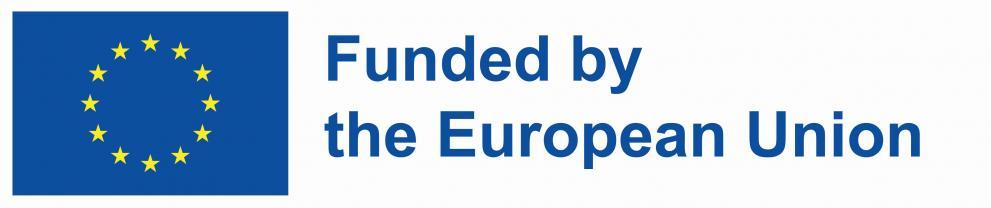}\end{center}

\bibliography{icml2023/strings,icml2023/lib,local,icml2023/proc}
\bibliographystyle{icml2023}

\newpage

\appendix
\onecolumn

\newpage

\section{Search spaces split}\label{appendix:search_space_split}

For the HPO-B Benchmark, we split search spaces into test, validation, and training search spaces. We make sure to group similar algorithms into the same split.
As validation search spaces, we use HPO-B IDs: 5527, 5891, 5906, 5971, 6767, 6766, and 5860.
As test search spaces, we use HPO-B IDs:
5965, 7609, 5889, 6794, 5859, 4796, 7607, 5636, and 5970.

\section{Details on Our Priors}\label{appendix:detail_on_prior}
Below we explain implementation details and the hyperparameters used in our priors. We found all hyperparameters using random search on the distinct validation search spaces of HPO-B, which ensures that we have an unbiased performance on the test set.

\subsection{Our HEBO-inspired Prior}
For our final model, not the one used for the direct comparison with HEBO on artificial data shown in Figure \ref{fig:example_query_points}, we made some adaptions to the HEBO prior. We do not use a linear kernel and use the following priors, found by using BO on the validation search spaces of HPO-B, as described in Section \ref{sec:experiments}.
We use $Gamma(concentration,rate)$ distributions for length- and outputscale hyperparameters:
For the outputscale we use $Gamma(0.8452, 0.3993)$ and for the lengthscale we use $Gamma(1.2107, 1.5212)$.
For the noise we follow the original kernel exactly with $\log(\epsilon) \sim \mathcal{N}(-4.63,0.5)$.

\subsection{Our BNN prior}

We generate the inputs for the BNN via a uniform random sampling scheme from the unit hypercube, akin to the inputs to our GP priors. We employ a multi-layer perceptron (MLP) for the BNN with a $tanh$ activation function. To account for model complexity, the number of layers in the MLP is sampled uniformly from the range of 8 to 15.
Additionally, the number of hidden units per layer is uniformly sampled between 36 and 150.
The weights of the network are sampled from a Normal distribution with 0 mean, and a standard deviation sampled uniformly at random from $[0.089, 0.193]$.
Every weight is zeroed with a probability of $14.5\%$, and the remaining weights are rescaled by a factor of $1/(1. - .145) ** (1 / 2)$ to counteract the changed distribution.

Before any activation, we add a 0-mean Gaussian noise with a standard deviation uniformly sampled between 0.0003 and 0.0014.
We add a noise uniformly at random sampled between 0.0004 and 0.0013.

We used a zero-mean Gaussian for sampling the input warping parameter C1 (C2) with a standard deviation of 0.976 (0.8003).

To be more efficient, we sampled the hyperparameters (the standard deviation of the distributions and the architecture) 16 times per batch, but used a batch size of 128. This is more efficient than sampling these per example, but adds correlation to the gradients inside a batch.

\section{Acquisition Function Optimization}
\label{app:acq func opt}
Acquisition functions are used in Bayesian optimization to determine the next point to sample in the optimization process. These functions take into account the current model of the objective function and the uncertainty of the model to determine the most promising points to sample.

As the PFNs are standard neural networks, they are fully differentiable.
Thus, we can compute derivatives of the acquisition function with respect to the input $\vx$ via backpropagation. We can then apply gradient-based optimization techniques to find the candidate $x$ that maximize the acquisition function given a PFN. 
Here we use standard techniques proposed by ~\citet{snoek-nips12a}.

We first create a set of candidates for $x$ by sampling random candidate positions (N=100,000) and combine them with all observations $\{\vx_1, \dots, \vx_k\}$.
Then, we use the 100 candidates with the highest acquisition function values as candidates for a gradient-based search with scipy\textquotesingle{s}~\citep{scipy-fixed} L-BFGS-B~\citep{zhu-acm97a}.
Finally, we used the candidate with maximum EI after optimization as our proposal.

In the case of integers or boolean hyperparameters we treat the value between the two next legal solutions as the probability of a coin flip and round in a probabilistic fashion. Furthermore, we skip candidates that we had already evaluated to ensure that we do not evaluate the same candidate twice.

\section{BO Tricks}
\label{sec:bo tricks}
We found that the following tricks improve the performance of our PFNs considerably and use them for all our experiments:
\begin{itemize}
    \item We use a power transform to transform the observed outputs to a distribution more similar to a standard-normal, as proposed by \citet{cowenrivers-jair22a}.
    They introduce this to handle heteroscedasticity, which might be the way it is helping the model. We additionally saw that it led to flattening outliers on the non-optimal end of the spectrum, while increasing the differences between points in magnitude close to the optimum, leading to a different exploration/exploitation trade-off.
    \item We found that training on datasets with inputs from the unit hypercube, i.e.\ $\mathcal{X} = [0,1]^d$, and transforming all observation values into a unit hypercube based on their min/max and scaling leads to improved consistency.
\end{itemize}

\section{Training Details}\label{appendix:training details}
We train a single PFN which is shared across search spaces for one particular prior. This PFN is flexible across datasets with a varying number of features.
To do this, we sample datasets with a number of dimensions sampled uniformly at random in $\{1, \dots, 18\}$ during PFN training.
Following \citet{hollmann-iclr23a}, we zero-pad the features when the number of features $k$ is smaller than the maximum number of features $K$.  We also linearly scale the features by $\frac{K}{k}$ to make sure the magnitude of the inputs is similar across different numbers of features.

Our PFNs were trained with the standard PFN settings used by \citet{muller-iclr22a}:
We use an embedding size of 512 and six layers in the transformer.
Our models were trained with Adam \citep{kingma-iclr15a} and cosine-annealing \citep{loshchilov-iclr17a} without any special tricks.
The lr was chosen based on simple grid searches for minimal training loss in $\{1e-3,3e-4,1e-4,5e-5\}$.
We found that other hyperparameters did not have a large impact on final performance.

Our final models, besides studies on smaller budgets like in Figure \ref{fig:full_bo_ei_dumb_gp}, trained for less than 24 hours on a cluster node with eight RTX 2080 Ti GPUs.

\subsection{Regression Heads}
\label{app:regression head}
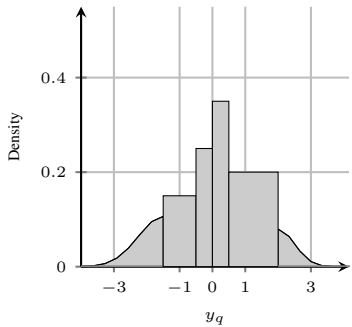
\begin{figure}[h]

     \centering
     \begin{subfigure}[b]{.3\textwidth}
     \input{icml2023/figures/riemann}
     \end{subfigure}
    \caption{A visualisation of the Riemann distribution, with unbounded support. Plot based on \cite{muller-iclr22a}}
    \label{fig:riemann}
    \vspace{-0em}
\end{figure}
\citet{muller-iclr22a} employ a Riemann distribution to model the output of the PFN architecture for regression tasks. Modeling continuous distributions with neural networks is challenging. In order to achieve robust performance in modeling PPDs, they utilized a distribution that is particularly compatible with neural networks. Drawing from the understanding that neural networks excel in classification tasks and taking inspiration from discretizations in distributional reinforcement learning \citep{bellemare2017distributional}, they employed a discretized continuous distribution named \emph{Riemann Distribution}.
\newcommand{\buckets}{\mathbf{B}}
It discretizes the space into buckets $\buckets{}$, which are selected such that each bucket has equal probability in prior-data: $p(y \in b) = 1/|\buckets{}|, \forall b \in \buckets{}$.
A Riemann distribution with unbounded support is utilized, as suggested by \citet{muller-iclr22a}, which replaces the final bar on each side with a suitably scaled half-normal distribution, as shown in Figure \ref{fig:riemann}. For a more precise definition, we direct the reader to \citet{muller-iclr22a}.

\section{Riemann Distribution and Acquisition Function}
\label{appendix:riemann_acqf}
 We outline how to compute the $PI(\best)=\int_{-\infty}^{\infty} [y>\best{}]p(y)dy$ for the unbounded Riemann distribution.

\begin{align}
    &\int_{-\infty}^{\infty} [y>\best{}]p(y)dy\\
    &= \int_{-\infty}^{y_1} [y>\best{}]p(y)dy + \sum_{i=1}^{M} \int_{y_i}^{y_{i+1}} [y>\best{}]p(y)dy + \int_{y_{M+1}}^{\infty} [y>\best{}]p(y)dy\\
    &= (1-F_l(y_1 - \best{})) + \sum_{i=1}^{M} \left. \begin{cases}
    (y_{i+1}-\best{})\frac{p(b_i)}{y_{i+1}-y_i}  &\text{, if } y_i < \best{} < y_{i+1} \\
    [y_i \leq \best{}] p(b_i) &\text{, else}
    \end{cases} \right\} + F_r(\best{} - y_{M+1})\\
    &= (1-F_l(y_1 - \best{})) + \sum_{i=1}^{M} (y_{i+1}-min(y_{i+1},max(\best{},y_{i})))\frac{p(b_i)}{y_{i+1}-y_i} + F_r(\best{} - y_{M+1}),
    \label{eq:riemann_acqf}
\end{align}

where $F_l$ ($F_r$) is the CDF of the half-normal distribution used for the left (right) side.
The acquisition function is divided into three terms: (1) a term that governs the probability mass for values lower than interval $b_1$ and that goes up to values of $y_1$, (2) a term that summarizes over all intervals $b_1, \dots, b_M$, and (3) a term that governs the probability mass for values larger than the interval $b_M$ and that start from values of $y_M+1$.

\section{User Prior Distribution}
\label{appendix:user_prior_distribution}
We will describe our method for a one dimensional search space with a fixed discretization $\intset$ into intervals $\intsample \in \intset$. Nevertheless, the method can be extended to more dimensions and intervals of varying size.
As an approximation to the prior, we will use the estimated maximum $m,m_y = \argmax_{(x,y) \in D} y$.
We will thus from now on assume to be able to sample from a joint distribution of the dataset $D$ and its maximum input $m$: $p(D,m)$.

An example of the effect of the form of the prior over the optimum is given in Figure \ref{fig:user prior mini example}.
We train a single PFN that can be conditioned on arbitrary $\rho$ and $\mathrm{I}$ on a prior of the form $p(D,\rho,\mathrm{I})$.
This yields the training objective:
\begin{align}
    \mathbb{E}_{\{(\vx, y)\} \cup D, \rho, \intsample \sim p(D,\mathbf{\rho},\mathbf{I})}[ - \text{log} \  q_{ \theta}(y |\vx, D, \rho, \intsample) ].
\end{align}
The extra inputs, $\rho$ and $I$, are fed to the PFN using an extra position with its own linear encoder similar to style embeddings for language models ~\citep{dai-acl19a}.
During training we need $p(D,\rho,\intvar)$ to cover many $\rho$ and $\intsample$, preserve Equation \ref{eq:conditioned_f}, and be fast to sample from.
To achieve this we actually sample the dataset $D$ first and then a possible interval $\intsample$.
We actually condition the interval $\intsample$ on our maximum $m$ and not the other way around.
We first sample $D,m \sim p(D,m)$ and $\rho \sim U(0,1)$ independently at random.
Next we sample $\intsample$ using $p(\intvar|\rho,f) = \rho [m \in \intvar] + (1-\rho) p(\intvar)$, where $p(\intvar) = \mathbb{E}_{D,m \sim p(D,m)}[m \in \intvar]$. 
It is easy to see that this distribution has a good coverage of both $I$ and $\rho$.

We can now show, rather easily, that this sampling scheme actually models our definition of $p(D,m|I)$:
\begin{align}
    p(D,m|I) &= \frac{p(I|D,m)p(D,m)}{p(I)} \\
    &= \frac{p(I|m)p(D,m)}{p(I)} \\
    &= \frac{(\rho [m \in I] + (1-\rho) p(I))p(D,m)}{p(I)} \\
    &= \rho\frac{[m \in I]p(D,m)}{p(I)} + (1-\rho)\frac{ p(\mathrm{I})p(D,m)}{p(I)} \\
    &= \rho p(D,m|m \in I) + (1-\rho)p(D,m),
\end{align}
where we assume a dependence on the independently distributed confidence $\rho$ everywhere.
We used $p(D,m|m \in I) = \frac{[m \in I]p(D,m)}{p(I)}$, which is trivial to derive, as well as $p(I) = \mathbb{E}_{f}[[m \in I]]$, which was introduced in Section \ref{sec:user priors}.

In our experiments we approximate $p(\intvar)$ with a simple uniform distribution, as we saw $\mathbb{E}_{m}[m \in \mathrm{I}]$ to be close to a uniform distribution, and use the maximum example in the training set with maximal value as an approximation to the true maximum of the underlying function generating the dataset.
We choose the set of intervals to be $\intset = \{[i/k,(i+1)/k]| k \in \{1,\dots,5\}, i \in \{0,\dots,k-1\}\}$.
In Figure \ref{fig:user prior traj} of the Appendix one can see the impact of our user prior on the optimization behaviour.

\section{Non-Myopic Acquisition Function Approximation}
\label{sec:nonmyopic acq function approx}
In the following we explain how PFNs can learn to directly in one forward pass approximate $\alpha_{\text{KG}}$ as defined in Section \ref{sec:non_myopic_acq}. To do this, we first learn an approximation to the PPD $q_\theta(y|\vx,D)$.
Based on which we can learn an approximation of the distribution of means with the following loss
\begin{align}
    \mathbb{E}_{\{(\vx, y)\} \cup D \sim p(\mathcal{D})}
    [q^{(\mu)}_\theta(\mathbb{E}_{q(y|\vx, D)}[y]|D)],
\end{align}
which yields a $q^{(\mu)}_{\theta^*}$ that is an approximation to the distribution of the random variable $\mathbb{E}_{p(y|\vx, D)}[y]$ with $\vx \sim p(\vx| D)$.
We approximate the maximal mean $\tau(D)$ with the upper 1 per mille interval of $q^{(\mu)}_{\theta^*}(\cdot|D)$.

Based on this in turn, we can finally approximate $\alpha(\vx; D)$ by training a second PFN to approximate the new target $y' = \text{icdf}(q^{(\mu)}_{\theta^*}(\cdot|\{(\vx, y)\} \cup D),.999)$ in the usual manner
\begin{align}
    \mathbb{E}_{\{(\vx, y)\} \cup D \sim p(\mathcal{D})}
    - \text{log} \  q_{level 1, \theta}(y'|\vx, D).
\end{align}

In practice we use the same PFN for both steps and add a style embedding, like for the user priors in Section \ref{sec:user priors}. This style embedding indicates in which mode the PFN should operate, the standard, myopic, setting or the non-myopic setting.

It would be very interesting for future work to generalize this to larger search spaces and multi-step look ahead.

\subsection{A Study on Our Knowledge Gradient Approximation}
\label{sec:nonmyopic results}
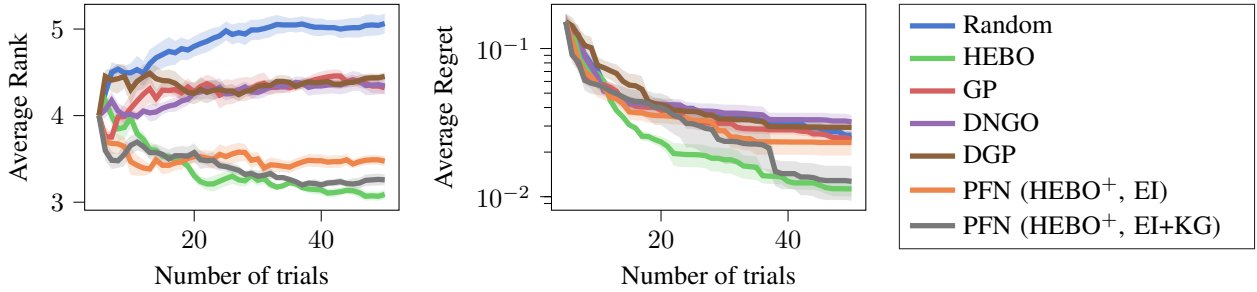
\begin{figure*}[t]
    \begin{subfigure}{1.\textwidth}
        \centering
        \begin{subfigure}[t]{.42\textwidth}
        \vskip 0pt
            \input{icml2023/figures/hpob_nonmyopic_rank_test.tex}
        \end{subfigure}
        \hspace{-.1\textwidth}
        \begin{subfigure}[t]{.42\textwidth}
        \vskip 0pt
            \input{icml2023/figures/hpob_nonmyopic_regret_test.tex}
        \end{subfigure}
        \hspace{-.09\textwidth}
        \begin{subfigure}[t]{.33\textwidth}
        \vskip 0pt
            \centering
            \begin{tikzpicture}
                \begin{axis}[%
                hide axis,
                xmin=0,
                xmax=0,
                ymin=0,
                ymax=0,
                legend style={draw=white!15!black,
                legend cell align=center,
                /tikz/every even column/.append style    = {column sep=.5cm, text width=3.7cm}
                },
                legend columns=1
                ]
                \addlegendimage{colorRandom, line width=2pt}
                \addlegendentry{Random};
                \addlegendimage{colorHEBO, line width=2pt}
                \addlegendentry{HEBO};
                \addlegendimage{colorGP, line width=2pt}
                \addlegendentry{GP};
                \addlegendimage{colorDNGO, line width=2pt}
                \addlegendentry{DNGO};
                \addlegendimage{colorDGP, line width=2pt}
                \addlegendentry{DGP};
                \addlegendimage{colorPFN_GP, line width=2pt}
                \addlegendentry{PFN (HEBO$^+$, EI)};
                \addlegendimage{colorHyperopt, line width=2pt}
                \addlegendentry{PFN (HEBO$^+$, EI+KG)};
                \end{axis}
            \end{tikzpicture}
        \end{subfigure}
    \end{subfigure}
    \caption{Performance over number of trials for the four search spaces of HPO-B with three or less dimensions. Figure \ref{fig:hebo_all_search_spaces} shows per-search-space results.}
    \label{fig:nonmyopic_hpob_results}
\end{figure*}

We found that the PFN with the strong GP prior does not perform very well on very small (three or less dimensions) search spaces.
Additionally, we saw that we could train our Knowledge Gradient (KG) models more successfully to approximate KG for few dimensions.
Thus, we decided to show the impact of Knowledge Gradient for these kinds of search spaces.
In Figure \ref{fig:nonmyopic_hpob_results} we show the impact of KG for the four search spaces with 3 or less dimensions in HPO-B.
We ablated how to exactly use KG for these search spaces on all other search spaces and found that a mixture of KG and EI worked best.
That means, at every step it is a random coin flip whether to use KG or EI.

We can see that this improves performance for small search spaces considerably over plain EI.

\begin{figure*}

\begin{center}
\hspace{-.3cm}
\begin{minipage}[t]{.45\linewidth}
\vspace{0pt}
\centering
\input{MetaLearnWorkshop/figures/ll_ranks_hebo_warp_obsnoise}
\end{minipage}%
\hspace{.3cm}
\begin{minipage}[t]{.45\linewidth}
\vspace{0pt}
\centering
\small
\begin{tabular}{llrrr}
\toprule
          & \# Features &      1  &      2  &      10 \\
{} & Approx. &         &         &         \\
\midrule
Mean Regret & Emp. Bay. &   0.057 &   0.059 &   0.110 \\
          & PFN &   0.054 &   0.052 &   0.124 \\
\# Wins & Emp. Bay. & 206 & 144 & 169 \\
          & PFN & 239 & 154 & 171 \\
Mean Rank & Emp. Bay. &   1.517 &   1.506 &   1.501 \\
          & PFN &   1.483 &   1.494 &   1.499 \\
\bottomrule
\end{tabular}
\end{minipage}
\end{center}
    \caption{Left: For the HEBO prior, fraction of cases in which the PFN gives a higher likelihood to unseen examples than the original Empirical Bayes approximation by \citet{cowenrivers-jair22a}. We aggregate across $1\,000$ datasets drawn from the prior. The optimization of the original Empirical Bayes approximation failed in $4.5\%$ of cases. We ignored these cases to be as fair as possible.
    Right: BO performance after 50 evaluations on the prior with EI over $1\,000$ sampled datasets. The majority of runs yielded ties.}
    \label{fig:toy_results}
\end{figure*}
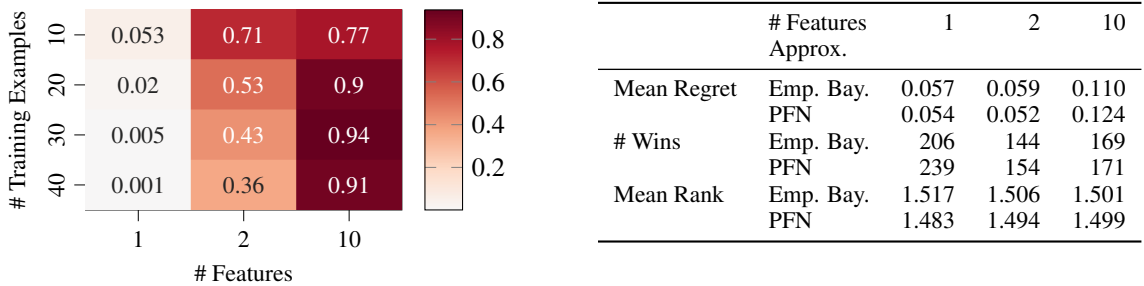

\section{Approximation Quality on Prior Data}
\label{sec:approx quality on prior details}
\slimparagraph{Our Adapted HEBO Prior}
We stayed as close as possible to HEBO, but had to make some adaptions as one cannot sample from the HEBO model as it is.
Specifically, we had to introduce simple priors for the hyperparameters of both the kernels used in HEBO (Matérn and linear kernel), as they did not have a prior attached.
We introduce uniform priors ($\mathcal{U}(0,1)$) for the $\mathrm{lengthscale}$ of both the Matérn and the Linear kernel and for the $\mathrm{variance}$ of the linear kernel.
Next, we trained a PFN using the HEBO prior with probabilistic distributions over the hyperparameters (see Section~\ref{sec:heboprior}) enabling PFNs to approximate hyperpriors.

\slimparagraph{Comparison of the Likelihood of Prior Data}
For the HEBO prior, we compare the resulting PFN's fit vs. the Empirical Bayes approximation, as used in the original HEBO work.
Figure \ref{fig:toy_results} shows that the likelihood assigned to held-out outputs is higher for the PFN in many cases and across a varying number of optimized features;
We hypothesize that this is because HEBO\textquotesingle{s} empirical Bayes approximation becomes too greedy in high dimensions.

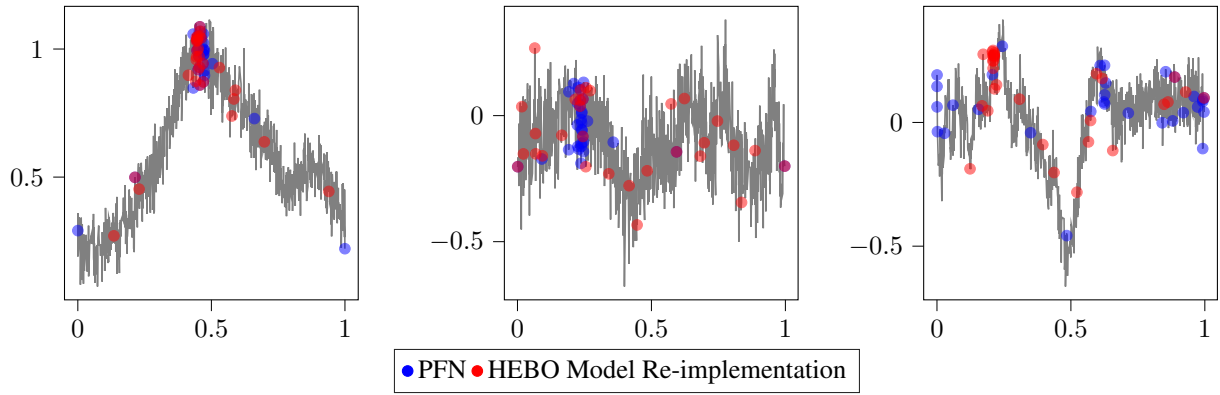
\begin{figure}
    \centering
    \begin{subfigure}{.32\textwidth}
    \input{MetaLearnWorkshop/figures/hebo_on_hebo_example_1}
    \end{subfigure}
    \begin{subfigure}{.32\textwidth}
    \input{MetaLearnWorkshop/figures/hebo_on_hebo_example_2}
    \end{subfigure}
    \begin{subfigure}{.32\textwidth}
    \input{MetaLearnWorkshop/figures/hebo_on_hebo_example_3}
    \end{subfigure}
    \begin{subfigure}{\textwidth}
    \centering
      \begin{tikzpicture} 
\definecolor{color0}{rgb}{0.12156862745098,0.466666666666667,0.705882352941177}
\definecolor{color1}{rgb}{1,0.498039215686275,0.0549019607843137}
\definecolor{color2}{rgb}{0.172549019607843,0.627450980392157,0.172549019607843}
    \begin{axis}[%
    hide axis,
    xmin=0,
    xmax=0,
    ymin=0,
    ymax=0,
    legend style={draw=white!15!black,legend cell align=left,},
    legend columns=3,
    ]
    \addlegendimage{blue, mark=*, only marks}
    \addlegendentry{PFN};
    \addlegendimage{red, mark=*, only marks}
    \addlegendentry{HEBO Model Re-implementation};
    \end{axis}
    \end{tikzpicture} 
    \end{subfigure}
    \caption{Examples of optimization trajectories on functions sampled from the our HEBO prior. We compare to our re-implementation of HEBO }
    \label{fig:example_query_points}
\end{figure}

\begin{figure*}
    \includegraphics[width=\textwidth]{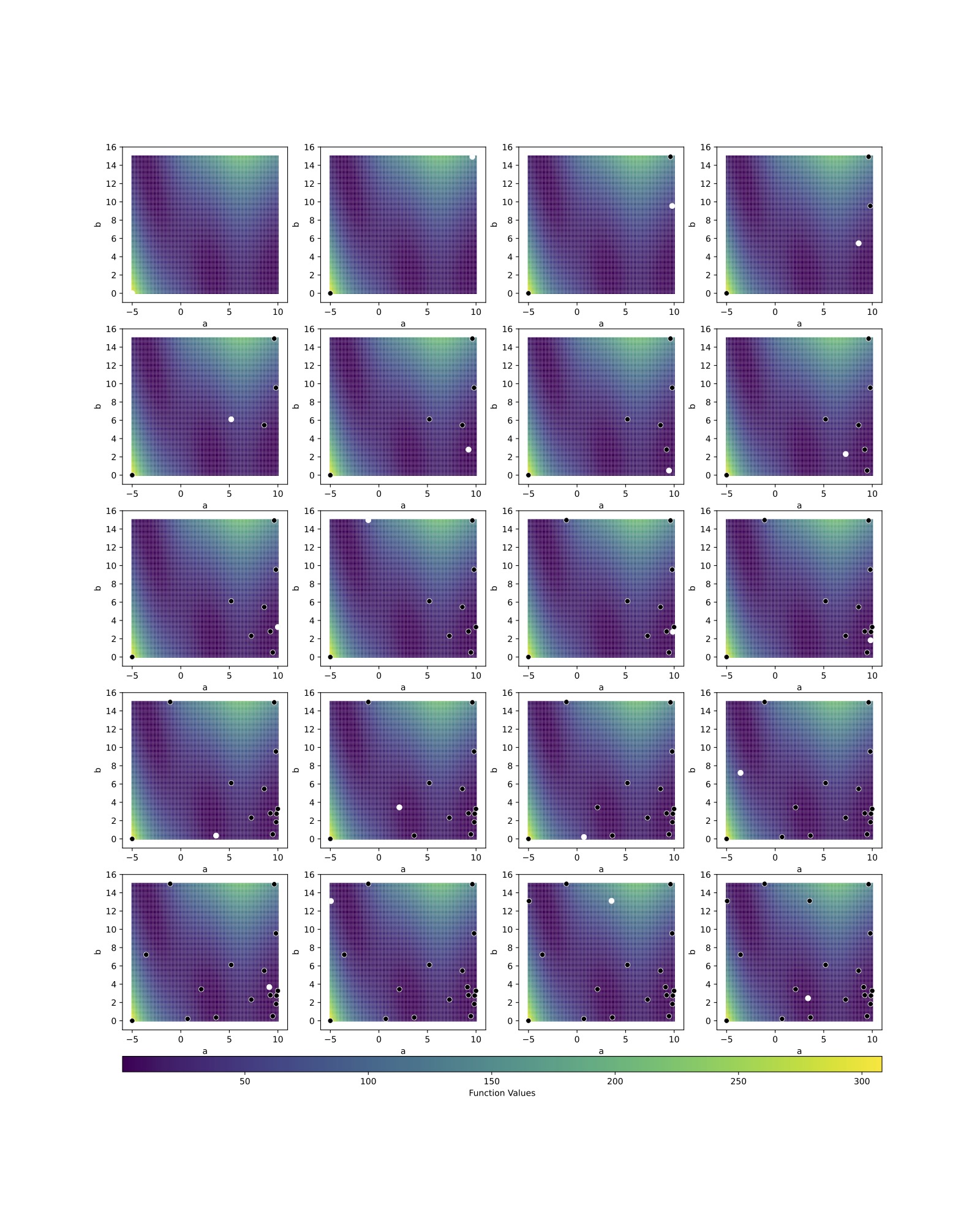}
    \vspace{-2.5cm}
    \caption{This figure showcases the optimization trajectory of a PFN with HEBO$^+$ prior to minimize a Branin function starting from the lower corner of the search space.
    The newly queries point is white in all plots. Figure \ref{fig:2d ei traj HEBO prior start at mid} shows the same experiment, but starting from the center. We can find that for a deterministic, smooth test function both trajectories converge to similar queries fast.}
    \label{fig:2d ei traj HEBO prior start at lower}
\end{figure*}

\begin{figure*}
    \includegraphics[width=\textwidth]{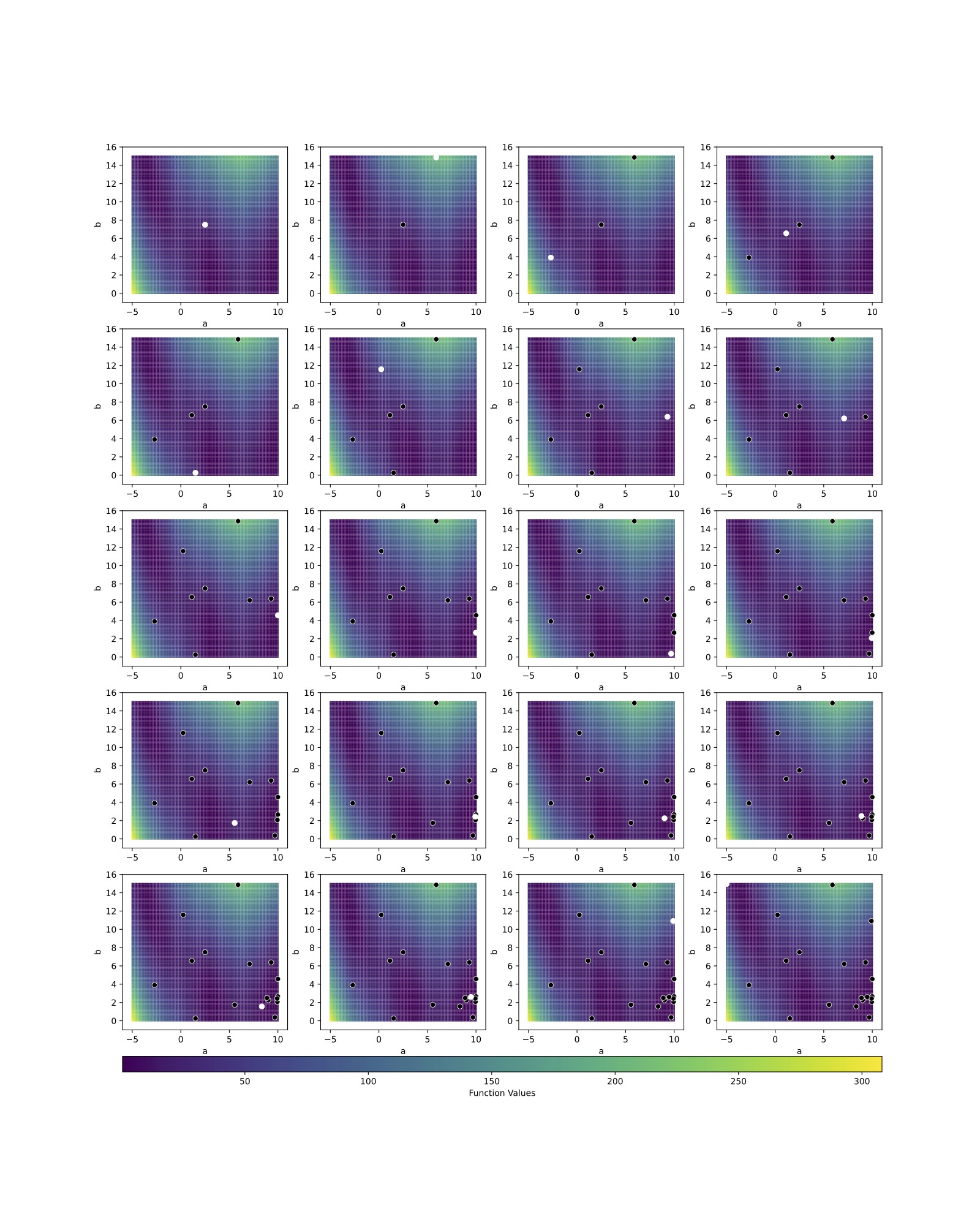}
    \vspace{-2.5cm}
    \caption{This figure showcases the optimization trajectory of a PFN with HEBO$^+$ prior to minimize a Branin function starting from the center of the search space.
    The newly queries point is white in all plots. Figure \ref{fig:2d ei traj HEBO prior start at lower} shows the same experiment, but starting from the lower corner. We can find that for a deterministic, smooth test function both trajectories converge to similar queries fast.}
    \label{fig:2d ei traj HEBO prior start at mid}
\end{figure*}

\slimparagraph{BO comparison}
To assess our method qualitatively, we provide optimization trajectories of the HEBO$^+$ on 1d samples from the prior in Figure \ref{fig:example_query_points} and on a 2d Branin function without initial design in Figures \ref{fig:2d ei traj HEBO prior start at lower} (initialization at lower corner) and \ref{fig:2d ei traj HEBO prior start at mid} (initialization at middle point).

\begin{figure}
    \centering
    \includegraphics[scale=.5]{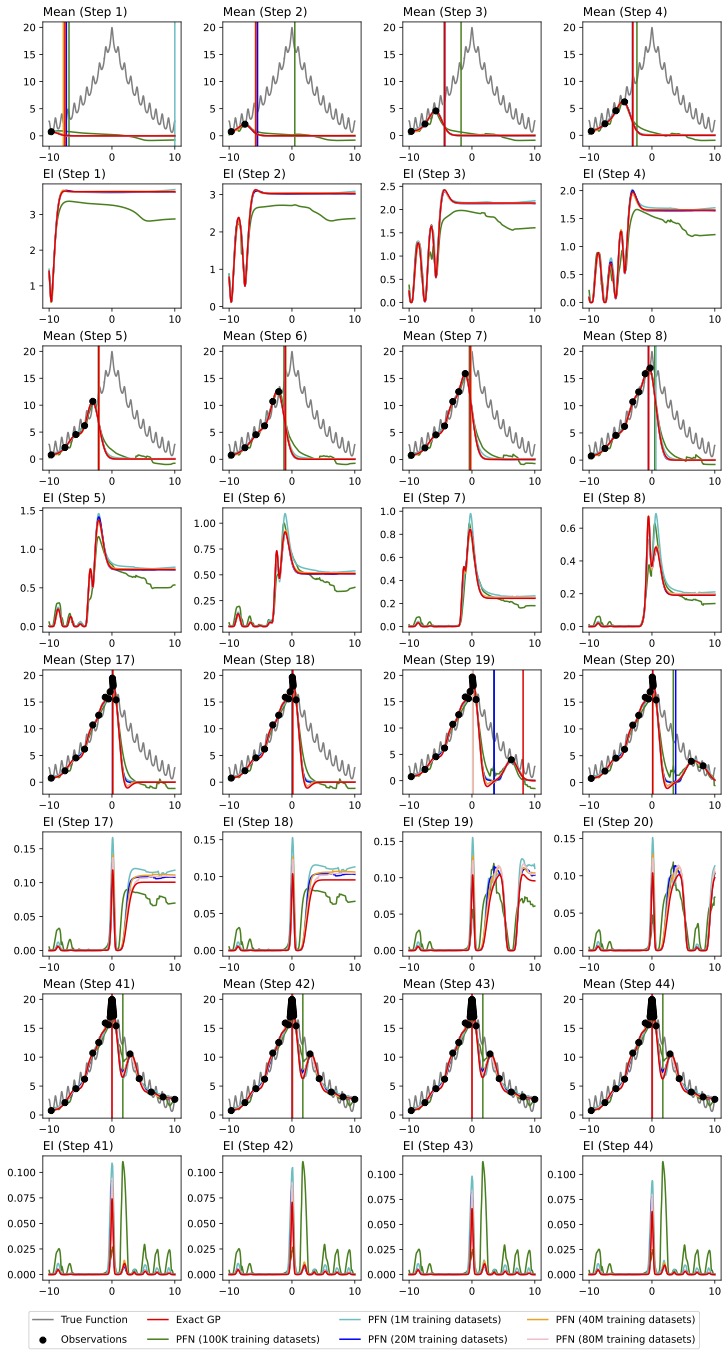}
    \caption{In this figure, we compare models trained on a simple GP prior (with fixed hyper-parameters), thus we can compare to the exact posterior of the GP. We show how PFNs behave differently depending on how much they were trained. Vertical lines mark the maximum of the acquisition function.}
    \label{fig:full_bo_ei_dumb_gp}
\end{figure}

\begin{figure*}
    \hspace{-.5cm}
    \centering
    \includegraphics[scale=.55]{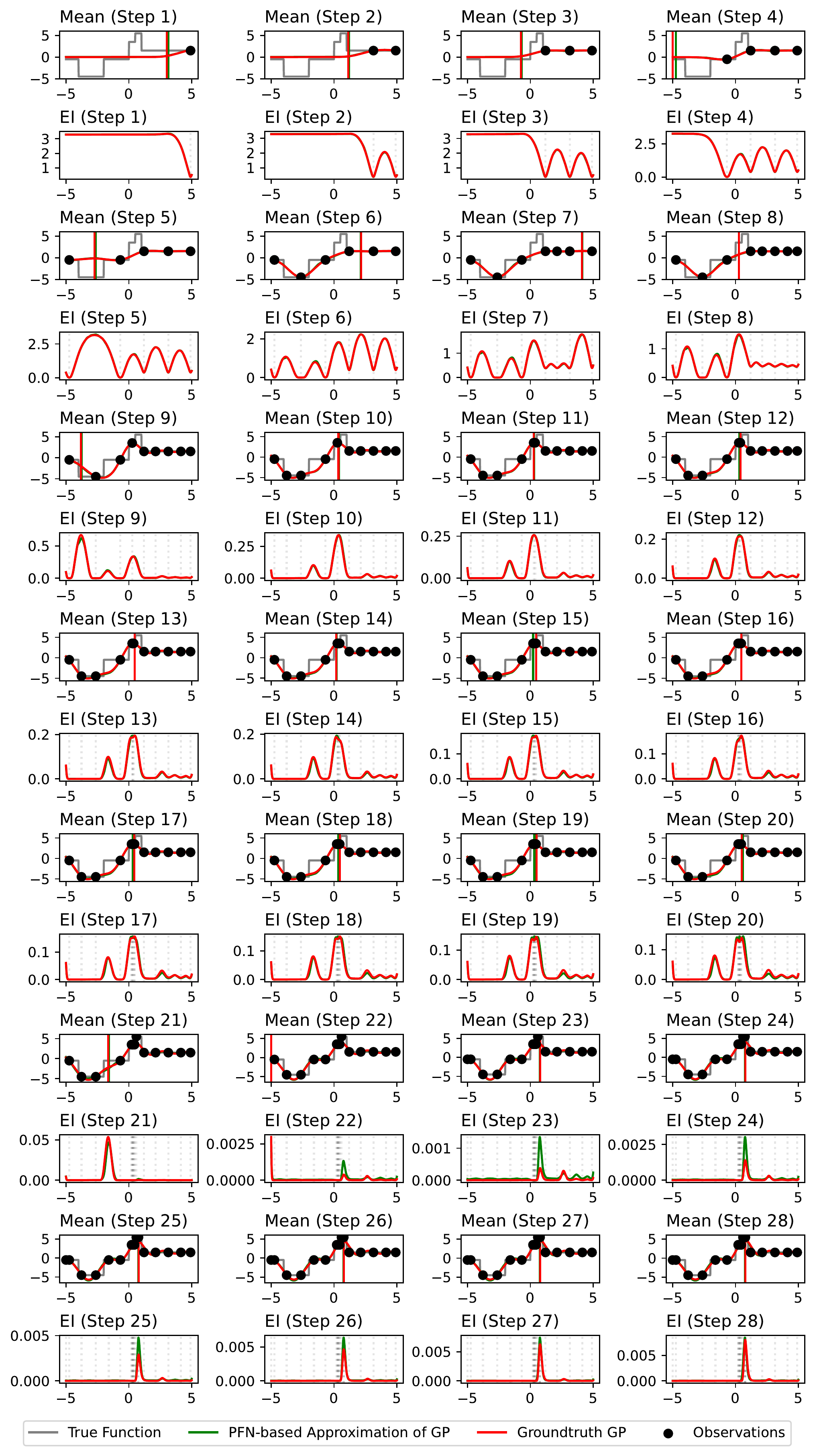}
    \caption{This figure show cases the approximation quality of a PFN on out-of-distribution functions: the to-be-maximized function is discontinuous, while the RBF kernel consists of continuous functions only.
    We show the suggested next evaluation point (horizontal line), the mean and the EI for a RBF-kernel GP and its PFN approximation.
    This is a special case, where we can actually compute the exact posterior with the GP.
    We can see that the mean and EI of the PFN are good approximations.
    The EI diverges in some of the later steps, though, but not the suggested next evaluation point.}
    \label{fig:full bo ei dumb nonsmooth traj}
\end{figure*}

Additionally, we show comparisons of a PFN approximating a simple RBF-kernel GP, for which we thus have the ground truth posterior.
Figure \ref{fig:full_bo_ei_dumb_gp} shows an optimization of a 1d Ackley function and Figure \ref{fig:full bo ei dumb nonsmooth traj} shows an optimization of a non-continuous function that is chosen to show out-of-distribution performance, as RBF-kernel functions are smooth.

\begin{figure*}[t]
    \begin{subfigure}{1.\textwidth}
        \centering
        \begin{subfigure}[t]{.42\textwidth}
        \vskip 0pt
            \input{icml2023/figures/hpob_diffbaselines_rank_test.tex}
        \end{subfigure}
        \hspace{-.1\textwidth}
        \begin{subfigure}[t]{.42\textwidth}
        \vskip 0pt
            \input{icml2023/figures/hpob_diffbaselines_regret_test.tex}
        \end{subfigure}
        \hspace{-.09\textwidth}
        \begin{subfigure}[t]{.33\textwidth}
        \vskip 0pt
            \centering
            \begin{tikzpicture}
                \begin{axis}[%
                hide axis,
                xmin=0,
                xmax=0,
                ymin=0,
                ymax=0,
                legend style={draw=white!15!black,
                legend cell align=center,
                /tikz/every even column/.append style    = {column sep=.5cm, text width=3.7cm}
                },
                legend columns=1
                ]
                \addlegendimage{colorRandom, line width=2pt}
                \addlegendentry{Random};
                \addlegendimage{colorHEBO, line width=2pt}
                \addlegendentry{HEBO};
                \addlegendimage{colorOPT_BEI, line width=2pt}
                \addlegendentry{Optformer-B EI};
                \addlegendimage{colorOPT_REI, line width=2pt}
                \addlegendentry{Optformer-R EI};
                \addlegendimage{colorBOHAMIANN, line width=2pt}
                \addlegendentry{BOHAMIANN};
                \addlegendimage{colorPFN_GP, line width=2pt}
                \addlegendentry{PFN (HEBO$^+$)};
                \addlegendimage{colorPFN_BNN, line width=2pt}
                \addlegendentry{PFN (BNN)};
                \end{axis}
            \end{tikzpicture}
        \end{subfigure}
    \end{subfigure}
    \caption{Average Regret and Rank on the HPO-B Benchmark. We plot a different set of baselines that includes the meta learned Optformer baseline which was trained on real world data. While for few samples the Optformer baselines performs strongly (likely due to prior knowledge of the names of optimized parameters, which is only revealed to the Optformer), this advantage drops quickly.}
    \label{fig:hpob diff baselines}
\end{figure*}
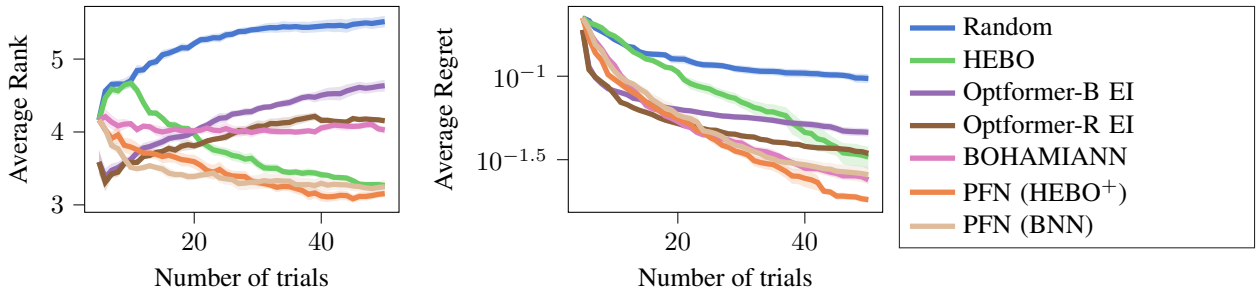


\begin{figure*}
    \centering
    \begin{subfigure}{.45\textwidth}
    \centering
    \includegraphics[scale=0.4]{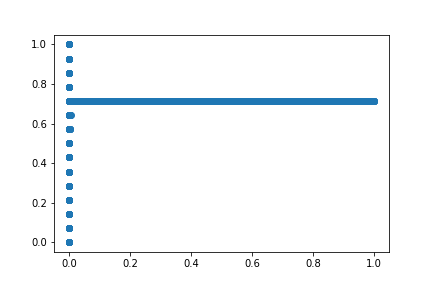}
    \end{subfigure}
    \begin{subfigure}{.45\textwidth}
    \centering
    \includegraphics[scale=0.4]{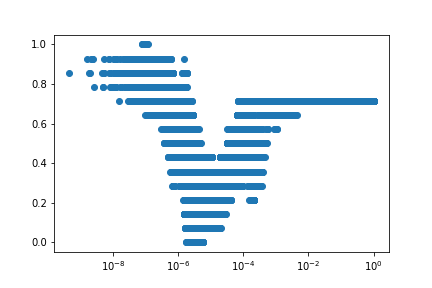}
    \end{subfigure}
    \caption{In this figure we show a typical search space with a missing log encoding in HPO-B. This is the search space 6766, which has two input dimension. The first dimension has almost almost no impact on the outcome (y), thus we plot only the second dimension (x). One can see that this hyperparameter clearly lives on a log scale, with all interesting curvature happening in the lowest 1\% of the interval, a minimum (worst outcome) close to 10-6 and a maximum (best outcome) around 10-7.}
    \label{fig:show log behaviour}
\end{figure*}

\section{User Prior Details}
We list the user prior used for PD1 in Table \ref{tab:userprior}.
We use this one user prior for all different tasks in PD1.
Our PFN, though, supports specifying any user prior at prediction time, i.e.\ you can use the weights we share with your specific user prior.

\begin{table}[]
    \centering
    \begin{tabular}{c|cHHcc}
hyper parameter & $\rho$ & min & max & encoded min & encoded max\\
\hline
lr decay factor& .5& .75& 1.&0.74&0.99\\
lr initial& .25& .5& .75 & .01 & .31\\
lr power& .1& .75& 1. & 1.5 & 2.\\
epoch& .5& .75& 1. & 235 & 299\\
activation fn& .5& .8& 1. & 1 & 1\\
    \end{tabular}
    \caption{$\rho$ defines what percentage of datasets in the prior should be sampled to explicitly have their maximum between the specified minimum and maximum. The rest is sampled as usually.}
    \label{tab:userprior}
\end{table}

\begin{figure*}
    \centering
    \includegraphics[width=\textwidth]{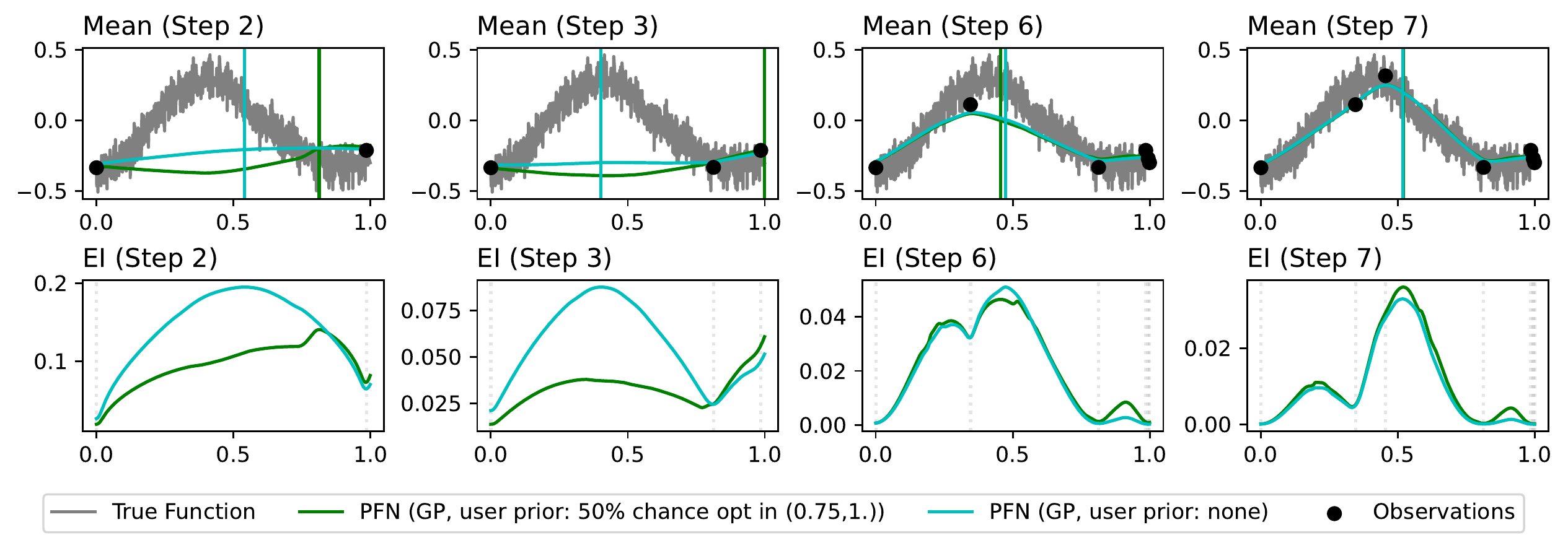}
    \caption{Impact of a user prior on the mean prediction and acquisition function. We can see that adding the user prior for a maximum in the right-most quarter, increases EI and mean in that region.
    The important point we can see here is that the user prior can be overwritten by the data. The algorithm samples much more outside of the specified interval than inside, because the actual curve is higher there.}
    \label{fig:user prior traj}
\end{figure*}

\section{Other BO Surrogates}
\label{sec:more_models}

Another popular model class is Bayesian neural networks (BNNs) or approximations of them~\citep{snoek-icml15a,schilling-ictai15a,springenberg-nips16a,perrone-neurips18a,eggensperger-arxiv20a} because they promise better scaling to large numbers of observations. 
Two popular approaches, that we also compare against in our experiments, are DNGO~\citep{snoek-icml15a} and BOHAMIANN~\citep{springenberg-nips16a}. DNGO trains a standard neural network and then replaces the output layer by a Bayesian linear regression, while BOHAMIANN adapts a mini-batch Monte Carlo method~\citep{chen-icml14a} for a fully Bayesian treatment of the network weights.

Other models that have been investigated aim to strike a middle ground between GPs and BNNs, for example, Deep GPs~\citep{hebbal-cec18a}, Deep Kernel GPs~\citep{wistuba-iclr21a}, or Manifold GPs~\citep{calandra-ijcnn16a}. However, these models suffer from the need to adapt hyperparameters online and, similar to the GP, make a parametric assumption about the distribution of the targets.

Last but not least, trees and ensembles of trees have been used, which despite relying on frequentist uncertainty predictions, yielded strong results and overcame many of the drawbacks of the GPs~\citep{hutter-lion11a,bergstra-nips11a,skopt}.

\section{Ablation Studies}
\label{sec:ablation studies}
In this Section we describe additional ablation studies we performed to further the understanding of our method.
\paragraph{HEBO Variants}
In Section \ref{sec:artificial experiments} we already showed that two variants of HEBO, one with a PFN as posterior approximation and one using MLE-II and a traditional GP, perform very similar on data in the prior.
The original HEBO prior is slightly different, though, compared to the prior used there, e.g.\ there is no prior weighting on the lengthscale and it is unbounded.

To further the understanding in comparison to the full HEBO method, we provide an ablation on HPO-B in Figure \ref{fig:hpob heboabl}.
Here, we additionally show the impact of the acquisition function of HEBO, which is mixture of EI, UCB and PI with added noise, and the initial design.
The default initial design on HPO-B is five steps, but for search spaces with more than five dimensions, HEBO draws more random samples.
While we can see that most of the ablations on HEBO have little impact on performance, we can improve its performance by removing all initial design on top of the standard HPO-B design and replacing the acquisition function with simple EI only.

In Figure \ref{fig:hpob heboabl}, we also show the impact of disabling input warping for our PFN (HEBO$^+$).
While this does make performance worse, the only non-HEBO baseline that can beat the PFN after 50 trials is DGP, even though the search spaces of HPO-B contain many ill-conditioned dimensions with missing log transforms.

\paragraph{Ignoring features}
In Table \ref{tab:ignore_feats_xgboost} we show the impact of adding meaningless features to the prior, as described in Section \ref{sec:ignore feats}, for search spaces with many dimensions.
We can see that performance is improved on average.
At the same time, we found no impact on smaller search spaces.
\paragraph{Ablation of the Acquisition Function}
Table \ref{table:acq_ablation_on_hpob} shows a simple ablation of acquisition functions on the test search spaces of HPO-B. We can see that all EI variants and UCB perform similarly well, while PI seems to be worse choice.

\definecolor{color0}{rgb}{0.12156862745098,0.466666666666667,0.705882352941177}
\definecolor{color1}{rgb}{1,0.498039215686275,0.0549019607843137}
\definecolor{color2}{rgb}{0.172549019607843,0.627450980392157,0.172549019607843}
\definecolor{color3}{rgb}{0.83921568627451,0.152941176470588,0.156862745098039}
\definecolor{color4}{rgb}{0.580392156862745,0.403921568627451,0.741176470588235}
\definecolor{color5}{rgb}{0.549019607843137,0.337254901960784,0.294117647058824}
\definecolor{color6}{rgb}{0.890196078431372,0.466666666666667,0.76078431372549}
\definecolor{color7}{rgb}{0.737254901960784,0.741176470588235,0.133333333333333}
\definecolor{color8}{rgb}{0.0901960784313725,0.745098039215686,0.811764705882353}

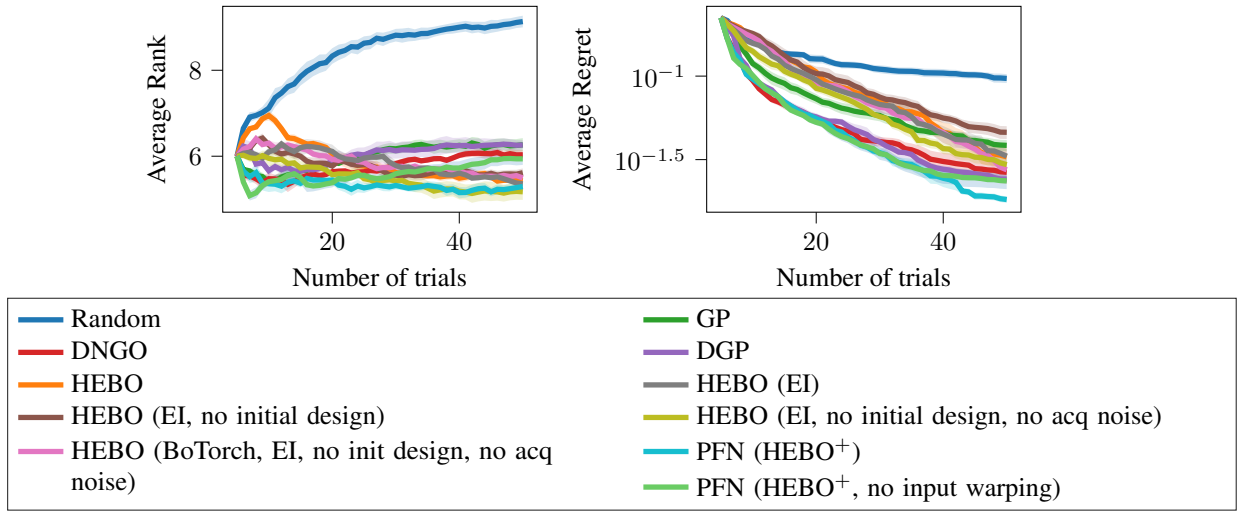
\begin{figure*}[t]
    \begin{subfigure}{1.\textwidth}
        \centering
        \begin{subfigure}[t]{.42\textwidth}
        \vskip 0pt
            \input{icml2023/figures/hpob_heboabl_rank_test.tex}
        \end{subfigure}
        \hspace{-.1\textwidth}
        \begin{subfigure}[t]{.42\textwidth}
        \vskip 0pt
            \input{icml2023/figures/hpob_heboabl_regret_test.tex}
        \end{subfigure}
        \hspace{-.09\textwidth}
        \begin{subfigure}[t]{1.0\textwidth}
        \vskip 0pt
            \centering
            \begin{tikzpicture}
                \begin{axis}[%
                hide axis,
                xmin=0,
                xmax=0,
                ymin=0,
                ymax=0,
                legend style={draw=white!15!black,
                legend cell align=center,
                /tikz/every even column/.append style    = {column sep=.5cm, text width=7cm}
                },
                legend columns=2
                ]
                \addlegendimage{color0, line width=2pt}
\addlegendentry{Random};
\addlegendimage{color2, line width=2pt}
\addlegendentry{GP};
\addlegendimage{color3, line width=2pt}
\addlegendentry{DNGO};
\addlegendimage{color4, line width=2pt}
\addlegendentry{DGP};
\addlegendimage{color1, line width=2pt}
\addlegendentry{HEBO};
\addlegendimage{white!49.8039215686275!black, line width=2pt}
\addlegendentry{HEBO (EI)};
\addlegendimage{color5, line width=2pt}
\addlegendentry{HEBO (EI, no initial design)};
\addlegendimage{color7, line width=2pt}
\addlegendentry{HEBO (EI, no initial design, no acq noise)};
\addlegendimage{color6, line width=2pt}
\addlegendentry{HEBO (BoTorch, EI, no init design, no acq noise)};
\addlegendimage{color8, line width=2pt}
\addlegendentry{PFN (HEBO$^+$)};
\addlegendimage{empty legend}
\addlegendentry{};
\addlegendimage{colorHEBO, line width=2pt}
\addlegendentry{PFN (HEBO$^+$, no input warping)};
                \end{axis}
            \end{tikzpicture}
        \end{subfigure}
    \end{subfigure}
    \caption{Ablation on HPO-B test spaces of different features of the HEBO baseline. Additionally, we show the performance of the PFN (HEBO$^+$) without input warping on HPO-B. We can see that PFN (HEBO$^+$) performs worse without input warping. We can also see that most ablations on HEBO have little impact on the performance, but removing the initial design, but for the default 5 seeds of HPO-B, in addition to using EI without any noise improves performance of the standard HEBO implementation to be more competitive with the PFN (HEBO$^+$).}
     \label{fig:hpob heboabl}
\end{figure*}

\begin{table}[]
    \centering
    \begin{tabular}{lrrHH}
\toprule
Method &  Rank Mean &  Regret Mean &  Final Rank &  Num Wins \\
\midrule
Random & 2.765 & 0.076 & 3.021 & 347 \\
HEBO & 2.406 & \textbf{0.070} & \textbf{2.150} & 531 \\
PFN (HEBO$^+$, no ignored feat\textquotesingle{s}) & 2.639 & 0.103 & 2.636 & 281 \\
PFN (HEBO$^+$, 30\% ignored feat\textquotesingle{s}) & \textbf{2.189} & 0.091 & 2.193 & 868 \\
\bottomrule
\end{tabular}
    \caption{Mean rank and mean regret over time on the three search spaces of HPO-B with the most hyperparameters, which all have XGBoost as method. We can see that for large search spaces it considerably helps to allow the PFN to ignore certain features.}
    \label{tab:ignore_feats_xgboost}
\end{table}

\begin{table}
\centering
\begin{tabular}{HlrrHH}
\toprule
{} & Method &  Rank Mean &  Regret Mean &  Final Rank &  Num Wins \\
\midrule
0 &  Random &  7.658 &  0.125 &  8.235 &  124 \\
1 &  HEBO &  5.467 &  0.092 &  4.962 &  394 \\
2 &  GP &  5.539 &  0.075 &  5.670 &  460 \\
3 &  DNGO &  5.332 &  0.057 &  5.472 &  350 \\
4 &  DGP &  5.604 &  0.059 &  5.690 &  278 \\
\midrule
5 &  EI &  4.969 &  0.053 &  4.831 &  342 \\
6 & EI(predict mean)   &  \textbf{4.883} &  0.059 &  4.872549 &  375 \\
7 &  PI  &  5.227 &  0.071 &  5.084 &  329 \\
8 &  PI (predicted mean)  &  5.385 &  0.071 &  5.341176 &  361 \\
9 &  UCB (0.95 percentile) &  4.932 &  \textbf{0.053} &  4.839216 &  417 \\
\bottomrule
\end{tabular}
\caption{Ablation of different acq functions on HPO-B. We see that we actually did not make the best choice for the test set (but it is the best on the validation search spaces, so we stuck with it).}
\label{table:acq_ablation_on_hpob}
\end{table}

\begin{figure}
\centering
\includegraphics[width=\textwidth]{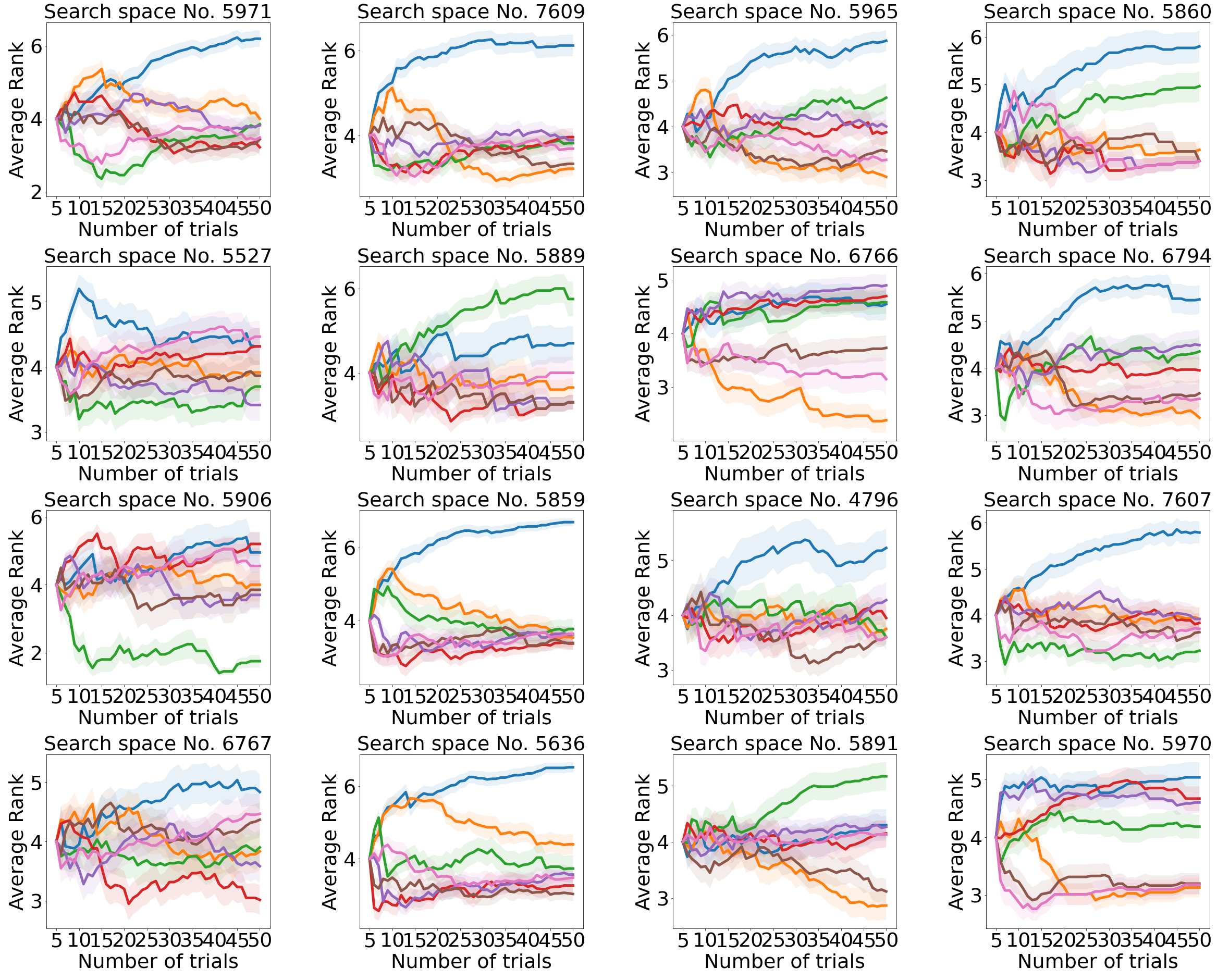}
\begin{tikzpicture} 
\definecolor{color0}{rgb}{0.12156862745098,0.466666666666667,0.705882352941177}
\definecolor{color1}{rgb}{1,0.498039215686275,0.0549019607843137}
\definecolor{color2}{rgb}{0.172549019607843,0.627450980392157,0.172549019607843}
\definecolor{color3}{rgb}{0.83921568627451,0.152941176470588,0.156862745098039}
\definecolor{color4}{rgb}{0.580392156862745,0.403921568627451,0.741176470588235}
\definecolor{color5}{rgb}{0.549019607843137,0.337254901960784,0.294117647058824}
\definecolor{color6}{rgb}{0.890196078431372,0.466666666666667,0.76078431372549}
\definecolor{color7}{rgb}{0.737254901960784,0.741176470588235,0.133333333333333}
    \begin{axis}[%
    hide axis,
    xmin=0,
    xmax=0,
    ymin=0,
    ymax=0,
    legend style={draw=white!15!black,legend cell align=center},
    legend columns=3,
    ]

\addlegendimage {line width=2pt, color0}
\addlegendentry{Random};
\addlegendimage {line width=2pt, color1}
\addlegendentry{HEBO};
\addlegendimage {line width=2pt, color2}
\addlegendentry{GP};
\addlegendimage {line width=2pt, color3}
\addlegendentry{DNGO};
\addlegendimage {line width=2pt, color4}
\addlegendentry{DGP};
\addlegendimage {line width=2pt, color5}
\addlegendentry{HEBO (GP)};
\addlegendimage {line width=2pt, color6}
\addlegendentry{HEBO (BNN)};
    \end{axis}
\end{tikzpicture}
    \caption{Average Ranks across search spaces. We can see that the PFN generally works well, but also has clear failure modes for some search spaces, which need further investigation.}
    \label{fig:hebo_all_search_spaces}
\end{figure}

\newpage

\section{Software Versions}
In this section we list the versions of the optimizers that we benchmarked:

\begin{itemize}
    \item HEBO: we use the archived submission to the NeurIPS 2020 Black-Box optimization challenge that is available at \url{https://github.com/huawei-noah/HEBO/tree/master/HEBO/archived_submissions/hebo} for BayesMark (HEBO\_0.0.8\_a376313), as it is optmized for the challenge, and adapted the public release to support discrete benchmarks as well as our ablations. We publish our adapted HEBO implementation with the supplementary material. It is branched off from  \url{https://github.com/huawei-noah/HEBO/tree/405dc4ceb93a79f0d1f0eaa24f5458dd26de1d05}. 
    \item GP: We used the results from \citet{pineda-neuripsdbt21a} that are available at \url{https://github.com/releaunifreiburg/HPO-B/tree/main/results}.
    \item DNGO: We used the results from \citet{pineda-neuripsdbt21a} that are available at \url{https://github.com/releaunifreiburg/HPO-B/tree/main/results}.
    \item DGP: We used the results from \citet{pineda-neuripsdbt21a} that are available at \url{https://github.com/releaunifreiburg/HPO-B/tree/main/results}.
    \item HyperOpt: We used the implementation coming with BayesMark (\url{https://github.com/uber/bayesmark/tree/8c420e935718f0d6867153b781e58943ecaf2338}), which is HyperOpt\_0.2.7\_a376313.
    \item PySOT: We used the implementation coming with BayesMark (\url{https://github.com/uber/bayesmark/tree/8c420e935718f0d6867153b781e58943ecaf2338}), which is PySOT\_0.3.3\_a376313.
\end{itemize}

Moreover, we used the following versions of the benchmark platforms:

\sloppy
\begin{itemize}
    \item HPO-B: \href{https://github.com/releaunifreiburg/HPO-B/tree/f5d415e45012544c61b0e334a42aa69f6aae5d7f}{https://github.com/releaunifreiburg/HPO-B/tree/f5d415e45012544c61b0e334a42aa69f6aae5d7f} with mode="v3"
    \item BayesMark: \href{https://github.com/uber/bayesmark/tree/8c420e935718f0d6867153b781e58943ecaf2338}{https://github.com/uber/bayesmark/tree/8c420e935718f0d6867153b781e58943ecaf2338}
\end{itemize}

\end{document}

%% file: icml2023/figures/pfn_usage.tex
\newcommand{\lastidx}{n+1}
\newcommand{\evalidx}{n+1}
\newcommand{\dataset}{\mathbf{D}}
\newcommand{\setofpairs}[2]{\{(x_1^{#1}, y_1^{#1}),\dots,(x_{#2}^{#1}, y_{#2}^{#1})\}}
\newcommand{\seq}[1]{$\setofpairs{#1}{\lastidx{}}$}
\definecolor{MidnightBlue}{rgb}{0.1, 0.1, 0.44}
\begin{tikzpicture}[remember picture,
  inner/.style={circle,draw=blue!50,fill=blue!20,thick,inner sep=3pt},
  scale=0.6, every node/.style={transform shape},
]
\tikzstyle{bigbox} = [thick, fill=white, rounded corners, rectangle,draw=black!30]
\tikzstyle{sample} = [thick, minimum size=0.6cm, rounded corners,rectangle, fill=Salmon!10,draw=black!30]
\tikzstyle{loss} = [thick, minimum size=0.6cm, rounded corners,rectangle, fill=PineGreen!10,draw=black!30]
\tikzstyle{pfmbox} = [thick,minimum size=0.6cm, rounded corners,rectangle, fill=MidnightBlue!10,draw=black!30]
\tikzstyle{box} = [thick, minimum size=0.6cm, rounded corners,rectangle,fill=MidnightBlue!5,draw=black!60]

\node[] (desc_offline) {Done once, offline};
\node [below=1em of desc_offline, align=center, box] (sample_datasets) {Sample synthetic datasets $D_i$ \\ from prior: $D_i \sim p(D)$};
\node [below=of sample_datasets, align=center, box] (train) {Train a PFN $q_\theta$ on synthetic \\ datasets $\{D_1,\dots,D_n\}$};
\draw [-{Triangle[length=3mm, width=2mm]}] (sample_datasets) -- (train);

\node[right=2em of train] (bottom_of_line) {};
\node [] (top_of_line) at (bottom_of_line|-desc_offline) {};
\draw [dashed] (bottom_of_line)++(0,-.5em) to (top_of_line)++(0,0em);

\node[right=3em of bottom_of_line, anchor=west, align=center, box] (predict) {Obtain $q_\theta(y_{test}|x_{test},D_{real})$\\with a single forward pass};
\node[] (desc_online) at (desc_offline-|predict) {Done per real-world dataset, online};
\draw [-{Triangle[length=3mm, width=2mm]}] (train) -- (predict);
\node[anchor=center, align=center, box] (evalinput) at (sample_datasets-|predict) {Real-world training dataset $D_{real}$\\and test point $x_{test}$};
\draw [-{Triangle[length=3mm, width=2mm]}] (evalinput) -- (predict);



\end{tikzpicture}

%% file: icml2023/figures/transformer_visualization.tex
\newcommand{\lastidx}{n+1}
\newcommand{\evalidx}{n+1}
\newcommand{\setofpairs}[2]{\{(x_1^{#1}, y_1^{#1}),\dots,(x_{#2}^{#1}, y_{#2}^{#1})\}}
\newcommand{\seq}[1]{$\setofpairs{#1}{\lastidx{}}$}
\newcommand{\xypair}[1]{$(\vx_{#1},y_{#1})$}
\newcommand{\qattbend}{40}
\def\innerdistance{2.2em}
\def\outdistane{1.2em}
\def\indistane{1.em}
\definecolor{MidnightBlue}{rgb}{0.1, 0.1, 0.44}
\begin{tikzpicture}[remember picture,
  inner/.style={circle,draw=blue!50,fill=blue!20,thick,inner sep=3pt},
]
\tikzstyle{bigbox} = [thick, fill=white, rounded corners, rectangle,draw=black!30]
\tikzstyle{sample} = [thick, minimum size=0.6cm, rounded corners,rectangle, fill=Salmon!10,draw=black!30]
\tikzstyle{loss} = [thick, minimum size=0.6cm, rounded corners,rectangle, fill=PineGreen!10,draw=black!30]
\tikzstyle{timestep} = [thick,minimum width=0.2cm,minimum height=.6cm, rounded corners,rectangle, fill=MidnightBlue!10,draw=black!30]
\tikzstyle{querytimestep} = [thick,minimum width=0.2cm,minimum height=.6cm, rounded corners,rectangle, fill=MidnightBlue!10,draw=black!30]
\tikzstyle{innerattention} = [draw=blue!30,-{Triangle[length=2mm, width=1mm]},]
\tikzstyle{queryattention} = [draw=red!30,-{Triangle[length=2mm, width=1mm]},]
\tikzstyle{output} = [draw=black!70,-{Triangle[length=2mm, width=1mm]},]
\tikzstyle{input} = [draw=black!70,-{Triangle[length=2mm, width=1mm]},]
\tikzstyle{every node} = [font=\footnotesize,]

\node[timestep] (t1) {};
\node[timestep, right=\innerdistance of t1] (t2) {};
\node[timestep, right=\innerdistance of t2] (t3) {};

\draw [ innerattention] (t1) to [bend left=15] (t2);
\draw [ innerattention] (t1) to [bend left=30] (t3);
\draw [ innerattention] (t2) to [bend right=15] (t1);
\draw [ innerattention] (t2) to [bend left=15] (t3);
\draw [ innerattention] (t3) to [bend right=30] (t1);
\draw [ innerattention] (t3) to [bend right=15] (t2);

\node[below=\indistane of t1] (i1) {\xypair{1}};
\draw [ input] (i1) -- (t1);
\node[below=\indistane of t2] (i2) {\xypair{2}};
\draw [ input] (i2) -- (t2);
\node[below=\indistane of t3] (i3) {\xypair{3}};
\draw [ input] (i3) -- (t3);


\node[querytimestep, right=2em of t3] (q1) {};
\node[querytimestep, right=3.2em of q1] (q2) {};

\draw [ queryattention] (q1) to [bend right=40] (t1);
\draw [ queryattention] (q1) to [bend right=40] (t2);
\draw [ queryattention] (q1) to [bend right=40] (t3);

\draw [ queryattention] (q2) to [bend right=40] (t1);
\draw [ queryattention] (q2) to [bend right=40] (t2);
\draw [ queryattention] (q2) to [bend right=40] (t3);
\node[above=\outdistane of q2] (d2) {$q(\cdot|\vx_5,D)$};

\node[above=\outdistane of q1] (d1) {$q(\cdot|\vx_4,D)$};
\draw [ output] (q1) -- (d1);
\node[] at (q1|-i1) (i4) {$\vx_4$};
\draw [ input] (i4) -- (q1);
\draw [ output] (q2) -- (d2);
\node[] at (q2|-i1) (i5) {$\vx_5$};
\draw [ input] (i5) -- (q2);
\end{tikzpicture}

%% file: icml2023/figures/hpob_main_rank_test.tex
\begin{tikzpicture}

\begin{axis}[
legend cell align={left},
legend columns=4,
legend style={
  fill opacity=0.8,
  draw opacity=1,
  text opacity=1,
  at={(0.55,-0.15)},
  anchor=south,
  draw=white!80!black
},
tick align=outside,
tick pos=left,
x grid style={white!69.0196078431373!black},
xmin=2.75, xmax=52.25,
xtick style={color=black},
y grid style={white!69.0196078431373!black},
xlabel={},
xticklabels={},
xtick={},
xmajorticks=false,
ylabel={Average Rank},
ymin=3.08007932297102, ymax=6.2,
ytick style={color=black},
ytick={4,5,6},
yticklabels={$\ \ 4$,$\ 5$,$\ 6$},
height=\benchmarkplotheight,
width=\benchmarkplotwidth,
]
\path [fill=colorRandom, fill opacity=0.2]
(axis cs:5,4)
--(axis cs:5,4)
--(axis cs:6,4.36129535887869)
--(axis cs:7,4.50668709173121)
--(axis cs:8,4.59112966309675)
--(axis cs:9,4.62418672732167)
--(axis cs:10,4.67326665823629)
--(axis cs:11,4.78502009416003)
--(axis cs:12,4.89125464150845)
--(axis cs:13,4.98530638243992)
--(axis cs:14,4.98553142619795)
--(axis cs:15,5.05604598096114)
--(axis cs:16,5.12306810626748)
--(axis cs:17,5.18344906721857)
--(axis cs:18,5.23651181835481)
--(axis cs:19,5.24622200744965)
--(axis cs:20,5.32713809094532)
--(axis cs:21,5.38964741564542)
--(axis cs:22,5.43286253831808)
--(axis cs:23,5.49001362075382)
--(axis cs:24,5.47607091264568)
--(axis cs:25,5.52659506597349)
--(axis cs:26,5.54223751783769)
--(axis cs:27,5.58627982132035)
--(axis cs:28,5.55931542831994)
--(axis cs:29,5.58999261482255)
--(axis cs:30,5.60890976378005)
--(axis cs:31,5.594800831219)
--(axis cs:32,5.60800456652998)
--(axis cs:33,5.60110777031135)
--(axis cs:34,5.62349429524135)
--(axis cs:35,5.62843553709701)
--(axis cs:36,5.65362140863675)
--(axis cs:37,5.64595290758783)
--(axis cs:38,5.66751805233952)
--(axis cs:39,5.67223827050387)
--(axis cs:40,5.68363805936465)
--(axis cs:41,5.70081044377453)
--(axis cs:42,5.67835344413008)
--(axis cs:43,5.69101549112646)
--(axis cs:44,5.67018077372303)
--(axis cs:45,5.69509965574158)
--(axis cs:46,5.70529116771574)
--(axis cs:47,5.7210932673686)
--(axis cs:48,5.71859007354921)
--(axis cs:49,5.72766152249793)
--(axis cs:50,5.73332698236364)
--(axis cs:50,5.90981027253832)
--(axis cs:50,5.90981027253832)
--(axis cs:49,5.89978945789422)
--(axis cs:48,5.88925306370569)
--(axis cs:47,5.8867498698863)
--(axis cs:46,5.87117942051955)
--(axis cs:45,5.88137093249371)
--(axis cs:44,5.85138785372795)
--(axis cs:43,5.85800411671668)
--(axis cs:42,5.84713675194835)
--(axis cs:41,5.85997386995096)
--(axis cs:40,5.84185213671378)
--(axis cs:39,5.82972251380985)
--(axis cs:38,5.82267802609185)
--(axis cs:37,5.80894905319648)
--(axis cs:36,5.80912368940246)
--(axis cs:35,5.8107801491775)
--(axis cs:34,5.83140766554296)
--(axis cs:33,5.80673536694355)
--(axis cs:32,5.83121111974453)
--(axis cs:31,5.8130423060359)
--(axis cs:30,5.79893337347486)
--(axis cs:29,5.77079169890294)
--(axis cs:28,5.7661747677585)
--(axis cs:27,5.78234762966004)
--(axis cs:26,5.74403699196623)
--(axis cs:25,5.73615003206573)
--(axis cs:24,5.68863496970726)
--(axis cs:23,5.67469226159912)
--(axis cs:22,5.60243157932898)
--(axis cs:21,5.56721532945262)
--(axis cs:20,5.53168543846645)
--(axis cs:19,5.46358191411898)
--(axis cs:18,5.44584112282166)
--(axis cs:17,5.37733524650692)
--(axis cs:16,5.34359856039919)
--(axis cs:15,5.2851304896271)
--(axis cs:14,5.1948607306648)
--(axis cs:13,5.16763479403066)
--(axis cs:12,5.05776496633469)
--(axis cs:11,5.02282304309487)
--(axis cs:10,4.87575294960685)
--(axis cs:9,4.74836229228618)
--(axis cs:8,4.69906641533462)
--(axis cs:7,4.65017565336683)
--(axis cs:6,4.47007719014092)
--(axis cs:5,4)
--cycle;

\path [fill=colorHEBO, fill opacity=0.2]
(axis cs:5,4)
--(axis cs:5,4)
--(axis cs:6,4.25241098014562)
--(axis cs:7,4.4156530101865)
--(axis cs:8,4.44995795574579)
--(axis cs:9,4.61871651828826)
--(axis cs:10,4.66741484651031)
--(axis cs:11,4.57318582230796)
--(axis cs:12,4.43591389037651)
--(axis cs:13,4.30429919565004)
--(axis cs:14,4.28815712553119)
--(axis cs:15,4.25509656577917)
--(axis cs:16,4.21649354960455)
--(axis cs:17,4.20951633643383)
--(axis cs:18,4.18670358934927)
--(axis cs:19,4.12888156838092)
--(axis cs:20,3.98531148725361)
--(axis cs:21,3.88047098388465)
--(axis cs:22,3.83929615971606)
--(axis cs:23,3.76910607582136)
--(axis cs:24,3.75573118164385)
--(axis cs:25,3.74231570204288)
--(axis cs:26,3.69205098776749)
--(axis cs:27,3.67861753129503)
--(axis cs:28,3.63316129622799)
--(axis cs:29,3.64380379622182)
--(axis cs:30,3.58504824369061)
--(axis cs:31,3.57178553469359)
--(axis cs:32,3.56567909609524)
--(axis cs:33,3.55819462580559)
--(axis cs:34,3.54555781019428)
--(axis cs:35,3.51252271082531)
--(axis cs:36,3.51944420465132)
--(axis cs:37,3.49572735777381)
--(axis cs:38,3.51355939336308)
--(axis cs:39,3.51601592552841)
--(axis cs:40,3.47936286031949)
--(axis cs:41,3.49781782610751)
--(axis cs:42,3.4927788821407)
--(axis cs:43,3.49648498004742)
--(axis cs:44,3.52475972993867)
--(axis cs:45,3.48748808677513)
--(axis cs:46,3.43369720182909)
--(axis cs:47,3.45341434731753)
--(axis cs:48,3.45232090622666)
--(axis cs:49,3.46176433987253)
--(axis cs:50,3.43763202656055)
--(axis cs:50,3.48785816951788)
--(axis cs:50,3.48785816951788)
--(axis cs:49,3.50294154248041)
--(axis cs:48,3.51630654475373)
--(axis cs:47,3.51129153503541)
--(axis cs:46,3.5114008373866)
--(axis cs:45,3.57133544263664)
--(axis cs:44,3.61641674064956)
--(axis cs:43,3.60939737289376)
--(axis cs:42,3.58565249040832)
--(axis cs:41,3.60022138957876)
--(axis cs:40,3.60299008085698)
--(axis cs:39,3.61731740780492)
--(axis cs:38,3.62761707722516)
--(axis cs:37,3.60623342653992)
--(axis cs:36,3.62957540319181)
--(axis cs:35,3.60904591662567)
--(axis cs:34,3.60738336627631)
--(axis cs:33,3.66533478595912)
--(axis cs:32,3.65785031566947)
--(axis cs:31,3.67135172020837)
--(axis cs:30,3.69730469748586)
--(axis cs:29,3.76403934103309)
--(axis cs:28,3.73938772337986)
--(axis cs:27,3.78804913537164)
--(axis cs:26,3.79422352203643)
--(axis cs:25,3.84984116070222)
--(axis cs:24,3.8481903869836)
--(axis cs:23,3.90932529672766)
--(axis cs:22,3.95286070302904)
--(axis cs:21,3.98619568278202)
--(axis cs:20,4.06566890490325)
--(axis cs:19,4.18876549044261)
--(axis cs:18,4.20937484202328)
--(axis cs:17,4.24930719297794)
--(axis cs:16,4.25017311706212)
--(axis cs:15,4.34882500284828)
--(axis cs:14,4.40988209015508)
--(axis cs:13,4.38589688278133)
--(axis cs:12,4.54447826648623)
--(axis cs:11,4.71308868749596)
--(axis cs:10,4.79925182015635)
--(axis cs:9,4.71069524641763)
--(axis cs:8,4.5265126324895)
--(axis cs:7,4.49415091138213)
--(axis cs:6,4.31229490220732)
--(axis cs:5,4)
--cycle;

\path [fill=colorGP, fill opacity=0.2]
(axis cs:5,4)
--(axis cs:5,4)
--(axis cs:6,3.77690647911914)
--(axis cs:7,3.74024367580845)
--(axis cs:8,3.65541832877145)
--(axis cs:9,3.66355101898638)
--(axis cs:10,3.646824322262)
--(axis cs:11,3.64298438232319)
--(axis cs:12,3.69997634663501)
--(axis cs:13,3.72198144266994)
--(axis cs:14,3.74796241733221)
--(axis cs:15,3.80779684960962)
--(axis cs:16,3.80865665370263)
--(axis cs:17,3.84285888090492)
--(axis cs:18,3.84080017683828)
--(axis cs:19,3.82846248131108)
--(axis cs:20,3.84180392156863)
--(axis cs:21,3.84582753569584)
--(axis cs:22,3.88099320546241)
--(axis cs:23,3.90167067382099)
--(axis cs:24,3.90138546967048)
--(axis cs:25,3.92421603821247)
--(axis cs:26,3.94964705882353)
--(axis cs:27,3.94272870789823)
--(axis cs:28,3.98028628188513)
--(axis cs:29,3.97482164750963)
--(axis cs:30,3.9614108063703)
--(axis cs:31,3.96192231221918)
--(axis cs:32,3.96642220053434)
--(axis cs:33,3.97114419442459)
--(axis cs:34,3.98264309346659)
--(axis cs:35,3.93985629703693)
--(axis cs:36,3.91149755262848)
--(axis cs:37,3.9212823576113)
--(axis cs:38,3.9120287472323)
--(axis cs:39,3.9066645579884)
--(axis cs:40,3.93080898550873)
--(axis cs:41,3.8942424253489)
--(axis cs:42,3.93562847876179)
--(axis cs:43,3.8996258010111)
--(axis cs:44,3.85531505151398)
--(axis cs:45,3.90450035995563)
--(axis cs:46,3.92420148523588)
--(axis cs:47,3.93025005210355)
--(axis cs:48,3.90467682267264)
--(axis cs:49,3.91228363829781)
--(axis cs:50,3.91389457881674)
--(axis cs:50,4.07826228392836)
--(axis cs:50,4.07826228392836)
--(axis cs:49,4.06810851856494)
--(axis cs:48,4.06787219693521)
--(axis cs:47,4.0893577910337)
--(axis cs:46,4.07579851476412)
--(axis cs:45,4.06412709102476)
--(axis cs:44,4.03095945828994)
--(axis cs:43,4.0846879244791)
--(axis cs:42,4.10358720751272)
--(axis cs:41,4.06654188837659)
--(axis cs:40,4.08879885762852)
--(axis cs:39,4.0854923047567)
--(axis cs:38,4.0801281155128)
--(axis cs:37,4.08263921101615)
--(axis cs:36,4.06497303560681)
--(axis cs:35,4.04445742845327)
--(axis cs:34,4.06833729869028)
--(axis cs:33,4.07591462910482)
--(axis cs:32,4.05710721123036)
--(axis cs:31,4.06160709954552)
--(axis cs:30,4.06604017402185)
--(axis cs:29,4.09184501915704)
--(axis cs:28,4.08245881615408)
--(axis cs:27,4.04550658621941)
--(axis cs:26,4.01113725490196)
--(axis cs:25,3.97774474610126)
--(axis cs:24,3.94567335385893)
--(axis cs:23,3.92970187519862)
--(axis cs:22,3.93861463767484)
--(axis cs:21,3.9031920721473)
--(axis cs:20,3.90329411764706)
--(axis cs:19,3.94016496966931)
--(axis cs:18,3.93959198002447)
--(axis cs:17,3.96106268772253)
--(axis cs:16,3.92075511100326)
--(axis cs:15,3.97259530725312)
--(axis cs:14,3.93046895521681)
--(axis cs:13,3.94076365536928)
--(axis cs:12,3.90394522199244)
--(axis cs:11,3.83152542159837)
--(axis cs:10,3.81984234440467)
--(axis cs:9,3.80311564768029)
--(axis cs:8,3.84654245554227)
--(axis cs:7,3.77348181438763)
--(axis cs:6,3.83877979539066)
--(axis cs:5,4)
--cycle;

\path [fill=colorDNGO, fill opacity=0.2]
(axis cs:5,4)
--(axis cs:5,4)
--(axis cs:6,3.68813986985734)
--(axis cs:7,3.70402172095302)
--(axis cs:8,3.69405695853208)
--(axis cs:9,3.60201965893718)
--(axis cs:10,3.64306131911275)
--(axis cs:11,3.6225974799772)
--(axis cs:12,3.58804064465161)
--(axis cs:13,3.49121816873777)
--(axis cs:14,3.59153906757119)
--(axis cs:15,3.64216672858761)
--(axis cs:16,3.65109131591207)
--(axis cs:17,3.64241426491906)
--(axis cs:18,3.62184934511576)
--(axis cs:19,3.6613597414076)
--(axis cs:20,3.69271361126994)
--(axis cs:21,3.6992471710976)
--(axis cs:22,3.69648103467032)
--(axis cs:23,3.64616245463647)
--(axis cs:24,3.67597971650004)
--(axis cs:25,3.64557637050188)
--(axis cs:26,3.67405881498061)
--(axis cs:27,3.64956046716822)
--(axis cs:28,3.67709268246684)
--(axis cs:29,3.70977065724532)
--(axis cs:30,3.73040657543742)
--(axis cs:31,3.75596294594378)
--(axis cs:32,3.77401405195219)
--(axis cs:33,3.77434797812287)
--(axis cs:34,3.73985629703693)
--(axis cs:35,3.74153669293524)
--(axis cs:36,3.76434144485646)
--(axis cs:37,3.76269566448952)
--(axis cs:38,3.73597522618534)
--(axis cs:39,3.75807765987783)
--(axis cs:40,3.78547365038716)
--(axis cs:41,3.78860866501787)
--(axis cs:42,3.78464503896563)
--(axis cs:43,3.77617456992812)
--(axis cs:44,3.77406974528521)
--(axis cs:45,3.78466567574835)
--(axis cs:46,3.81147635568195)
--(axis cs:47,3.75215739405605)
--(axis cs:48,3.76750121199965)
--(axis cs:49,3.74515581953193)
--(axis cs:50,3.74891543143624)
--(axis cs:50,3.91775123523043)
--(axis cs:50,3.91775123523043)
--(axis cs:49,3.92935398438964)
--(axis cs:48,3.94230270956898)
--(axis cs:47,3.92235240986551)
--(axis cs:46,3.96107266392589)
--(axis cs:45,3.95258922621243)
--(axis cs:44,3.95534201942067)
--(axis cs:43,3.92970778301306)
--(axis cs:42,3.925158882603)
--(axis cs:41,3.92903839380566)
--(axis cs:40,3.90472242804422)
--(axis cs:39,3.87329488914178)
--(axis cs:38,3.86010320518721)
--(axis cs:37,3.87652002178499)
--(axis cs:36,3.87487424141805)
--(axis cs:35,3.86238487569221)
--(axis cs:34,3.84445742845327)
--(axis cs:33,3.85702457089674)
--(axis cs:32,3.87304477157722)
--(axis cs:31,3.84403705405622)
--(axis cs:30,3.83429930691552)
--(axis cs:29,3.78826855844095)
--(axis cs:28,3.77780927831747)
--(axis cs:27,3.73083168969452)
--(axis cs:26,3.7730000085488)
--(axis cs:25,3.77403147263538)
--(axis cs:24,3.78284381291172)
--(axis cs:23,3.75775911399098)
--(axis cs:22,3.7976366123885)
--(axis cs:21,3.80663518184358)
--(axis cs:20,3.76218834951438)
--(axis cs:19,3.71903241545515)
--(axis cs:18,3.73109183135483)
--(axis cs:17,3.78895828410054)
--(axis cs:16,3.79204593898989)
--(axis cs:15,3.78528425180455)
--(axis cs:14,3.76532367752685)
--(axis cs:13,3.70878183126223)
--(axis cs:12,3.78842994358369)
--(axis cs:11,3.7538731082581)
--(axis cs:10,3.80007593578921)
--(axis cs:9,3.79013720380792)
--(axis cs:8,3.80006068852674)
--(axis cs:7,3.7979390633607)
--(axis cs:6,3.83342875759364)
--(axis cs:5,4)
--cycle;

\path [fill=colorDGP, fill opacity=0.2]
(axis cs:5,4)
--(axis cs:5,4)
--(axis cs:6,4.07682253849345)
--(axis cs:7,4.03748260708751)
--(axis cs:8,3.81978055089083)
--(axis cs:9,3.91554422515291)
--(axis cs:10,3.80314118474308)
--(axis cs:11,3.86312272506181)
--(axis cs:12,3.81190980095112)
--(axis cs:13,3.82172471405536)
--(axis cs:14,3.7802539033925)
--(axis cs:15,3.71450338348471)
--(axis cs:16,3.70781236467506)
--(axis cs:17,3.69192300672175)
--(axis cs:18,3.67579457712466)
--(axis cs:19,3.74369869023152)
--(axis cs:20,3.81308123342345)
--(axis cs:21,3.86734308343948)
--(axis cs:22,3.8749066848604)
--(axis cs:23,3.90427953453696)
--(axis cs:24,3.91013296385013)
--(axis cs:25,3.87429805396635)
--(axis cs:26,3.89354289558306)
--(axis cs:27,3.88663357528412)
--(axis cs:28,3.88835271811954)
--(axis cs:29,3.87762568497836)
--(axis cs:30,3.86559757706521)
--(axis cs:31,3.8757662597218)
--(axis cs:32,3.85442790696719)
--(axis cs:33,3.87033492216841)
--(axis cs:34,3.87422580126354)
--(axis cs:35,3.8555887707719)
--(axis cs:36,3.86304844123507)
--(axis cs:37,3.90552569617277)
--(axis cs:38,3.91126909878912)
--(axis cs:39,3.90090995809219)
--(axis cs:40,3.89935574322737)
--(axis cs:41,3.89299612447773)
--(axis cs:42,3.88537938544429)
--(axis cs:43,3.89019216348012)
--(axis cs:44,3.91851540835379)
--(axis cs:45,3.93742486929977)
--(axis cs:46,3.94947504999716)
--(axis cs:47,3.96038112036363)
--(axis cs:48,3.96641643913477)
--(axis cs:49,3.95094684663239)
--(axis cs:50,3.95039361254544)
--(axis cs:50,4.04568481882711)
--(axis cs:50,4.04568481882711)
--(axis cs:49,4.05297472199506)
--(axis cs:48,4.06887767851229)
--(axis cs:47,4.07883456591088)
--(axis cs:46,4.07405436176755)
--(axis cs:45,4.07433983658258)
--(axis cs:44,4.09324929752857)
--(axis cs:43,4.07843528750028)
--(axis cs:42,4.07932649690865)
--(axis cs:41,4.11876858140462)
--(axis cs:40,4.10848739402753)
--(axis cs:39,4.11477631641761)
--(axis cs:38,4.15539756787754)
--(axis cs:37,4.14153312735664)
--(axis cs:36,4.10165744111787)
--(axis cs:35,4.08950926844379)
--(axis cs:34,4.07871537520705)
--(axis cs:33,4.06299841116492)
--(axis cs:32,4.09459170087595)
--(axis cs:31,4.10854746576839)
--(axis cs:30,4.08734359940537)
--(axis cs:29,4.11845274639419)
--(axis cs:28,4.13517669364516)
--(axis cs:27,4.12513113059824)
--(axis cs:26,4.08684926127969)
--(axis cs:25,4.07864312250424)
--(axis cs:24,4.08594546752242)
--(axis cs:23,4.06042634781598)
--(axis cs:22,4.01136782494353)
--(axis cs:21,4.08559809303111)
--(axis cs:20,4.02221288422361)
--(axis cs:19,3.96218366270965)
--(axis cs:18,3.93204856013024)
--(axis cs:17,3.92376326778805)
--(axis cs:16,3.9196386157171)
--(axis cs:15,3.92079073416235)
--(axis cs:14,3.97268727307809)
--(axis cs:13,4.01749097221914)
--(axis cs:12,3.97240392453908)
--(axis cs:11,3.94472041219309)
--(axis cs:10,3.95372156035496)
--(axis cs:9,4.04916165720003)
--(axis cs:8,3.98021944910917)
--(axis cs:7,4.2095762164419)
--(axis cs:6,4.24474608895753)
--(axis cs:5,4)
--cycle;

\path [fill=colorPFN_GP, fill opacity=0.2]
(axis cs:5,4)
--(axis cs:5,4)
--(axis cs:6,3.72573615382945)
--(axis cs:7,3.63181939203328)
--(axis cs:8,3.6953648621649)
--(axis cs:9,3.59135739981007)
--(axis cs:10,3.60148246920777)
--(axis cs:11,3.58286743152784)
--(axis cs:12,3.57890334508442)
--(axis cs:13,3.62388908357415)
--(axis cs:14,3.63197645396627)
--(axis cs:15,3.5802704863303)
--(axis cs:16,3.61509079042738)
--(axis cs:17,3.57874381778504)
--(axis cs:18,3.57033858752927)
--(axis cs:19,3.57604246255376)
--(axis cs:20,3.54559083549498)
--(axis cs:21,3.47083489853334)
--(axis cs:22,3.42292753795352)
--(axis cs:23,3.38125063196017)
--(axis cs:24,3.41580972489071)
--(axis cs:25,3.40231060764075)
--(axis cs:26,3.38868202675753)
--(axis cs:27,3.39137652693215)
--(axis cs:28,3.3625431806749)
--(axis cs:29,3.33389607941754)
--(axis cs:30,3.36683379953434)
--(axis cs:31,3.34203227078622)
--(axis cs:32,3.3191388619224)
--(axis cs:33,3.31172178748521)
--(axis cs:34,3.33471999518263)
--(axis cs:35,3.35403335192465)
--(axis cs:36,3.32015085193328)
--(axis cs:37,3.31145956539211)
--(axis cs:38,3.26749814340405)
--(axis cs:39,3.28130921260742)
--(axis cs:40,3.21867212395267)
--(axis cs:41,3.21482841580756)
--(axis cs:42,3.23062989796997)
--(axis cs:43,3.25726748490887)
--(axis cs:44,3.28358240018131)
--(axis cs:45,3.24077581244713)
--(axis cs:46,3.26501832758896)
--(axis cs:47,3.27996332616176)
--(axis cs:48,3.29031212949545)
--(axis cs:49,3.31239823757233)
--(axis cs:50,3.32076776803422)
--(axis cs:50,3.44001654569127)
--(axis cs:50,3.44001654569127)
--(axis cs:49,3.43662137027081)
--(axis cs:48,3.42733492932808)
--(axis cs:47,3.41415432089706)
--(axis cs:46,3.41733461358751)
--(axis cs:45,3.38275359931757)
--(axis cs:44,3.43406465864222)
--(axis cs:43,3.4250854562676)
--(axis cs:42,3.39289951379473)
--(axis cs:41,3.38125001556499)
--(axis cs:40,3.40093571918459)
--(axis cs:39,3.45202412072591)
--(axis cs:38,3.45407048404693)
--(axis cs:37,3.48069729735299)
--(axis cs:36,3.49945699120397)
--(axis cs:35,3.50479017748712)
--(axis cs:34,3.48488784795462)
--(axis cs:33,3.48043507525989)
--(axis cs:32,3.4730180008227)
--(axis cs:31,3.48934027823339)
--(axis cs:30,3.51944071026958)
--(axis cs:29,3.52492744999422)
--(axis cs:28,3.58647642716824)
--(axis cs:27,3.57725092404824)
--(axis cs:26,3.56425914971306)
--(axis cs:25,3.58984625510434)
--(axis cs:24,3.6469353731485)
--(axis cs:23,3.59129838764767)
--(axis cs:22,3.6319744228308)
--(axis cs:21,3.66642000342745)
--(axis cs:20,3.71323269391679)
--(axis cs:19,3.69454577274036)
--(axis cs:18,3.73162219678446)
--(axis cs:17,3.73106010378359)
--(axis cs:16,3.81236018996478)
--(axis cs:15,3.74914127837558)
--(axis cs:14,3.78370982054354)
--(axis cs:13,3.82316973995527)
--(axis cs:12,3.75442998824891)
--(axis cs:11,3.79360315670745)
--(axis cs:10,3.86126262883145)
--(axis cs:9,3.82432887469973)
--(axis cs:8,3.97522337312922)
--(axis cs:7,3.82308256875103)
--(axis cs:6,3.85857757166074)
--(axis cs:5,4)
--cycle;

\path [fill=colorPFN_BNN, fill opacity=0.2]
(axis cs:5,4)
--(axis cs:5,4)
--(axis cs:6,3.69700404845168)
--(axis cs:7,3.5476271732405)
--(axis cs:8,3.5497241180262)
--(axis cs:9,3.47982188963705)
--(axis cs:10,3.37319600827687)
--(axis cs:11,3.36597900603211)
--(axis cs:12,3.39735756805604)
--(axis cs:13,3.44385295221728)
--(axis cs:14,3.38553136523129)
--(axis cs:15,3.35100595417634)
--(axis cs:16,3.3503566851988)
--(axis cs:17,3.33778225794748)
--(axis cs:18,3.3667340768927)
--(axis cs:19,3.36135842555461)
--(axis cs:20,3.33012324888381)
--(axis cs:21,3.35440058517671)
--(axis cs:22,3.41616643706039)
--(axis cs:23,3.44107059561063)
--(axis cs:24,3.40136360524514)
--(axis cs:25,3.40065003849344)
--(axis cs:26,3.39457058194185)
--(axis cs:27,3.35358816513961)
--(axis cs:28,3.34263967880122)
--(axis cs:29,3.35478163220134)
--(axis cs:30,3.38915153980206)
--(axis cs:31,3.3937368889453)
--(axis cs:32,3.40456248768496)
--(axis cs:33,3.43199464325116)
--(axis cs:34,3.43551783291482)
--(axis cs:35,3.45496413797051)
--(axis cs:36,3.42875041248651)
--(axis cs:37,3.42734850875684)
--(axis cs:38,3.44607332282922)
--(axis cs:39,3.44696740759305)
--(axis cs:40,3.47694624918908)
--(axis cs:41,3.48359878026954)
--(axis cs:42,3.48095164780918)
--(axis cs:43,3.44441630040922)
--(axis cs:44,3.43750367375537)
--(axis cs:45,3.4686769838751)
--(axis cs:46,3.46205206018634)
--(axis cs:47,3.44894265235108)
--(axis cs:48,3.42807604214652)
--(axis cs:49,3.43023885426095)
--(axis cs:50,3.43795507604573)
--(axis cs:50,3.57773119846407)
--(axis cs:50,3.57773119846407)
--(axis cs:49,3.56976114573905)
--(axis cs:48,3.56015925197113)
--(axis cs:47,3.55105734764892)
--(axis cs:46,3.53794793981366)
--(axis cs:45,3.5548524278896)
--(axis cs:44,3.55465318898973)
--(axis cs:43,3.55950526821823)
--(axis cs:42,3.57787188160259)
--(axis cs:41,3.57130318051477)
--(axis cs:40,3.57795571159523)
--(axis cs:39,3.54518945515205)
--(axis cs:38,3.54608353991588)
--(axis cs:37,3.53343580496865)
--(axis cs:36,3.55948488163114)
--(axis cs:35,3.5920946855589)
--(axis cs:34,3.54879589257538)
--(axis cs:33,3.53271123910178)
--(axis cs:32,3.52092770839347)
--(axis cs:31,3.51606703262333)
--(axis cs:30,3.48927983274696)
--(axis cs:29,3.45698307368101)
--(axis cs:28,3.46912502708113)
--(axis cs:27,3.46209810937019)
--(axis cs:26,3.49170392786207)
--(axis cs:25,3.47778133405558)
--(axis cs:24,3.46530306142153)
--(axis cs:23,3.54324312987957)
--(axis cs:22,3.50148062176314)
--(axis cs:21,3.47697196384289)
--(axis cs:20,3.46595518248873)
--(axis cs:19,3.4857003979748)
--(axis cs:18,3.51169729565632)
--(axis cs:17,3.48182558518977)
--(axis cs:16,3.48885900107571)
--(axis cs:15,3.53134698700013)
--(axis cs:14,3.53211569359224)
--(axis cs:13,3.56399018503762)
--(axis cs:12,3.5751914515518)
--(axis cs:11,3.50460922926201)
--(axis cs:10,3.48170595250745)
--(axis cs:9,3.57900163977472)
--(axis cs:8,3.71694254864047)
--(axis cs:7,3.6680591012693)
--(axis cs:6,3.86378026527381)
--(axis cs:5,4)
--cycle;

\addplot [line width=2pt, colorRandom]
table {%
5 4
6 4.4156862745098
7 4.57843137254902
8 4.64509803921569
9 4.68627450980392
10 4.77450980392157
11 4.90392156862745
12 4.97450980392157
13 5.07647058823529
14 5.09019607843137
15 5.17058823529412
16 5.23333333333333
17 5.28039215686274
18 5.34117647058823
19 5.35490196078431
20 5.42941176470588
21 5.47843137254902
22 5.51764705882353
23 5.58235294117647
24 5.58235294117647
25 5.63137254901961
26 5.64313725490196
27 5.6843137254902
28 5.66274509803922
29 5.68039215686275
30 5.70392156862745
31 5.70392156862745
32 5.71960784313726
33 5.70392156862745
34 5.72745098039216
35 5.71960784313726
36 5.73137254901961
37 5.72745098039216
38 5.74509803921569
39 5.75098039215686
40 5.76274509803922
41 5.78039215686274
42 5.76274509803922
43 5.77450980392157
44 5.76078431372549
45 5.78823529411765
46 5.78823529411765
47 5.80392156862745
48 5.80392156862745
49 5.81372549019608
50 5.82156862745098
};
\addplot [line width=2pt, colorHEBO]
table {%
5 4
6 4.28235294117647
7 4.45490196078431
8 4.48823529411765
9 4.66470588235294
10 4.73333333333333
11 4.64313725490196
12 4.49019607843137
13 4.34509803921569
14 4.34901960784314
15 4.30196078431373
16 4.23333333333333
17 4.22941176470588
18 4.19803921568627
19 4.15882352941177
20 4.02549019607843
21 3.93333333333333
22 3.89607843137255
23 3.83921568627451
24 3.80196078431373
25 3.79607843137255
26 3.74313725490196
27 3.73333333333333
28 3.68627450980392
29 3.70392156862745
30 3.64117647058824
31 3.62156862745098
32 3.61176470588235
33 3.61176470588235
34 3.57647058823529
35 3.56078431372549
36 3.57450980392157
37 3.55098039215686
38 3.57058823529412
39 3.56666666666667
40 3.54117647058824
41 3.54901960784314
42 3.53921568627451
43 3.55294117647059
44 3.57058823529412
45 3.52941176470588
46 3.47254901960784
47 3.48235294117647
48 3.4843137254902
49 3.48235294117647
50 3.46274509803922
};
\addplot [line width=2pt, colorGP]
table {%
5 4
6 3.8078431372549
7 3.75686274509804
8 3.75098039215686
9 3.73333333333333
10 3.73333333333333
11 3.73725490196078
12 3.80196078431373
13 3.83137254901961
14 3.83921568627451
15 3.89019607843137
16 3.86470588235294
17 3.90196078431373
18 3.89019607843137
19 3.8843137254902
20 3.87254901960784
21 3.87450980392157
22 3.90980392156863
23 3.9156862745098
24 3.92352941176471
25 3.95098039215686
26 3.98039215686275
27 3.99411764705882
28 4.03137254901961
29 4.03333333333333
30 4.01372549019608
31 4.01176470588235
32 4.01176470588235
33 4.02352941176471
34 4.02549019607843
35 3.9921568627451
36 3.98823529411765
37 4.00196078431373
38 3.99607843137255
39 3.99607843137255
40 4.00980392156863
41 3.98039215686275
42 4.01960784313725
43 3.9921568627451
44 3.94313725490196
45 3.9843137254902
46 4
47 4.00980392156863
48 3.98627450980392
49 3.99019607843137
50 3.99607843137255
};
\addplot [line width=2pt, colorDNGO]
table {%
5 4
6 3.76078431372549
7 3.75098039215686
8 3.74705882352941
9 3.69607843137255
10 3.72156862745098
11 3.68823529411765
12 3.68823529411765
13 3.6
14 3.67843137254902
15 3.71372549019608
16 3.72156862745098
17 3.7156862745098
18 3.67647058823529
19 3.69019607843137
20 3.72745098039216
21 3.75294117647059
22 3.74705882352941
23 3.70196078431373
24 3.72941176470588
25 3.70980392156863
26 3.72352941176471
27 3.69019607843137
28 3.72745098039216
29 3.74901960784314
30 3.78235294117647
31 3.8
32 3.82352941176471
33 3.8156862745098
34 3.7921568627451
35 3.80196078431373
36 3.81960784313726
37 3.81960784313726
38 3.79803921568627
39 3.8156862745098
40 3.84509803921569
41 3.85882352941176
42 3.85490196078431
43 3.85294117647059
44 3.86470588235294
45 3.86862745098039
46 3.88627450980392
47 3.83725490196078
48 3.85490196078431
49 3.83725490196078
50 3.83333333333333
};
\addplot [line width=2pt, colorDGP]
table {%
5 4
6 4.16078431372549
7 4.12352941176471
8 3.9
9 3.98235294117647
10 3.87843137254902
11 3.90392156862745
12 3.8921568627451
13 3.91960784313725
14 3.87647058823529
15 3.81764705882353
16 3.81372549019608
17 3.8078431372549
18 3.80392156862745
19 3.85294117647059
20 3.91764705882353
21 3.97647058823529
22 3.94313725490196
23 3.98235294117647
24 3.99803921568627
25 3.97647058823529
26 3.99019607843137
27 4.00588235294118
28 4.01176470588235
29 3.99803921568627
30 3.97647058823529
31 3.9921568627451
32 3.97450980392157
33 3.96666666666667
34 3.97647058823529
35 3.97254901960784
36 3.98235294117647
37 4.02352941176471
38 4.03333333333333
39 4.0078431372549
40 4.00392156862745
41 4.00588235294118
42 3.98235294117647
43 3.9843137254902
44 4.00588235294118
45 4.00588235294118
46 4.01176470588235
47 4.01960784313725
48 4.01764705882353
49 4.00196078431373
50 3.99803921568627
};
\addplot [line width=2pt, colorPFN_GP]
table {%
5 4
6 3.7921568627451
7 3.72745098039216
8 3.83529411764706
9 3.7078431372549
10 3.73137254901961
11 3.68823529411765
12 3.66666666666667
13 3.72352941176471
14 3.7078431372549
15 3.66470588235294
16 3.71372549019608
17 3.65490196078431
18 3.65098039215686
19 3.63529411764706
20 3.62941176470588
21 3.56862745098039
22 3.52745098039216
23 3.48627450980392
24 3.53137254901961
25 3.49607843137255
26 3.47647058823529
27 3.4843137254902
28 3.47450980392157
29 3.42941176470588
30 3.44313725490196
31 3.4156862745098
32 3.39607843137255
33 3.39607843137255
34 3.40980392156863
35 3.42941176470588
36 3.40980392156863
37 3.39607843137255
38 3.36078431372549
39 3.36666666666667
40 3.30980392156863
41 3.29803921568627
42 3.31176470588235
43 3.34117647058824
44 3.35882352941176
45 3.31176470588235
46 3.34117647058824
47 3.34705882352941
48 3.35882352941176
49 3.37450980392157
50 3.38039215686275
};
\addplot [line width=2pt, colorPFN_BNN]
table {%
5 4
6 3.78039215686274
7 3.6078431372549
8 3.63333333333333
9 3.52941176470588
10 3.42745098039216
11 3.43529411764706
12 3.48627450980392
13 3.50392156862745
14 3.45882352941176
15 3.44117647058824
16 3.41960784313725
17 3.40980392156863
18 3.43921568627451
19 3.42352941176471
20 3.39803921568627
21 3.4156862745098
22 3.45882352941176
23 3.4921568627451
24 3.43333333333333
25 3.43921568627451
26 3.44313725490196
27 3.4078431372549
28 3.40588235294118
29 3.40588235294118
30 3.43921568627451
31 3.45490196078431
32 3.46274509803922
33 3.48235294117647
34 3.4921568627451
35 3.52352941176471
36 3.49411764705882
37 3.48039215686275
38 3.49607843137255
39 3.49607843137255
40 3.52745098039216
41 3.52745098039216
42 3.52941176470588
43 3.50196078431373
44 3.49607843137255
45 3.51176470588235
46 3.5
47 3.5
48 3.49411764705882
49 3.5
50 3.5078431372549
};
\end{axis}

\end{tikzpicture}

%% file: icml2023/figures/hpob_main_regret_test.tex
\begin{tikzpicture}

\begin{axis}[
log basis y={10},
tick align=outside,
tick pos=left,
x grid style={white!69.0196078431373!black},
xmin=2.75, xmax=52.25,
xtick style={color=black},
y grid style={white!69.0196078431373!black},
xlabel={Number of trials},
ylabel={Average Regret},
ymin=0.0153171731609242, ymax=0.261636933606157,
ymode=log,
ytick style={color=black},
xlabel={},
xticklabels={},
xtick={},
xmajorticks=false,
height=\benchmarkplotheight,
width=\benchmarkplotwidth,
]
\path [fill=colorRandom, fill opacity=0.1]
(axis cs:5,0.229972229042153)
--(axis cs:5,0.21710238749041)
--(axis cs:6,0.210551931890934)
--(axis cs:7,0.191336856479292)
--(axis cs:8,0.18398611450507)
--(axis cs:9,0.173174426009038)
--(axis cs:10,0.161566672085494)
--(axis cs:11,0.153221170289856)
--(axis cs:12,0.147103225495095)
--(axis cs:13,0.144964010471113)
--(axis cs:14,0.133513953193397)
--(axis cs:15,0.130869269073569)
--(axis cs:16,0.129588704326338)
--(axis cs:17,0.129056064670601)
--(axis cs:18,0.128240983502576)
--(axis cs:19,0.119655546102882)
--(axis cs:20,0.119261341744938)
--(axis cs:21,0.118952916746962)
--(axis cs:22,0.115740984667637)
--(axis cs:23,0.113822815372513)
--(axis cs:24,0.111239996489407)
--(axis cs:25,0.111109586652343)
--(axis cs:26,0.110677655794738)
--(axis cs:27,0.110054406629553)
--(axis cs:28,0.10869636607477)
--(axis cs:29,0.105425670492834)
--(axis cs:30,0.104148807039482)
--(axis cs:31,0.103036857102853)
--(axis cs:32,0.102326505972465)
--(axis cs:33,0.101512284262623)
--(axis cs:34,0.101273140087157)
--(axis cs:35,0.101095535190606)
--(axis cs:36,0.100374825444411)
--(axis cs:37,0.0987592666410219)
--(axis cs:38,0.0985667843139294)
--(axis cs:39,0.0983883405881294)
--(axis cs:40,0.0982740935933329)
--(axis cs:41,0.097971176092753)
--(axis cs:42,0.0978147028919637)
--(axis cs:43,0.0970845745081627)
--(axis cs:44,0.0967919127817516)
--(axis cs:45,0.0967251933978113)
--(axis cs:46,0.0964900529287663)
--(axis cs:47,0.0948039725845666)
--(axis cs:48,0.0914377093499991)
--(axis cs:49,0.0913116376456255)
--(axis cs:50,0.0909194074778135)
--(axis cs:50,0.102760161966289)
--(axis cs:50,0.102760161966289)
--(axis cs:49,0.10355391050267)
--(axis cs:48,0.103618894177447)
--(axis cs:47,0.105118030426644)
--(axis cs:46,0.105890409431554)
--(axis cs:45,0.106714359948635)
--(axis cs:44,0.106830096253732)
--(axis cs:43,0.107401874324678)
--(axis cs:42,0.109019289222573)
--(axis cs:41,0.109232186712493)
--(axis cs:40,0.110249516016526)
--(axis cs:39,0.110511645921156)
--(axis cs:38,0.110593493901512)
--(axis cs:37,0.110726131211684)
--(axis cs:36,0.112007148271516)
--(axis cs:35,0.112514679946367)
--(axis cs:34,0.112607841546255)
--(axis cs:33,0.112700097536489)
--(axis cs:32,0.113557523551764)
--(axis cs:31,0.115252610748441)
--(axis cs:30,0.115924337046734)
--(axis cs:29,0.116760957112317)
--(axis cs:28,0.118213697427976)
--(axis cs:27,0.120825416295935)
--(axis cs:26,0.121896892724957)
--(axis cs:25,0.122429135447138)
--(axis cs:24,0.122801304845827)
--(axis cs:23,0.124100798337417)
--(axis cs:22,0.129089778757046)
--(axis cs:21,0.133831452125081)
--(axis cs:20,0.133983446483229)
--(axis cs:19,0.134736088401152)
--(axis cs:18,0.139933467453242)
--(axis cs:17,0.14085611822347)
--(axis cs:16,0.141347747166577)
--(axis cs:15,0.142075058190751)
--(axis cs:14,0.145751665683148)
--(axis cs:13,0.152858007901119)
--(axis cs:12,0.15497155554015)
--(axis cs:11,0.164415684662798)
--(axis cs:10,0.170072803621721)
--(axis cs:9,0.180017514871002)
--(axis cs:8,0.186814074102229)
--(axis cs:7,0.196881831709089)
--(axis cs:6,0.219660583406081)
--(axis cs:5,0.229972229042153)
--cycle;

\path [fill=colorHEBO, fill opacity=0.1]
(axis cs:5,0.229972229042153)
--(axis cs:5,0.21710238749041)
--(axis cs:6,0.206767140505162)
--(axis cs:7,0.200269236542652)
--(axis cs:8,0.191893881430571)
--(axis cs:9,0.178401295834877)
--(axis cs:10,0.172228709564284)
--(axis cs:11,0.160242156579528)
--(axis cs:12,0.148288228185233)
--(axis cs:13,0.13706877128598)
--(axis cs:14,0.131755403377912)
--(axis cs:15,0.124634753528544)
--(axis cs:16,0.122314300804427)
--(axis cs:17,0.117386177238596)
--(axis cs:18,0.109443737179489)
--(axis cs:19,0.107964035869148)
--(axis cs:20,0.101478655615687)
--(axis cs:21,0.0894026065920785)
--(axis cs:22,0.0834904358748244)
--(axis cs:23,0.0810612376894862)
--(axis cs:24,0.0792019900308486)
--(axis cs:25,0.077272045739271)
--(axis cs:26,0.0743406880548449)
--(axis cs:27,0.0723679242034906)
--(axis cs:28,0.0692501331132998)
--(axis cs:29,0.0679576455343032)
--(axis cs:30,0.0642696604071299)
--(axis cs:31,0.0619732061149486)
--(axis cs:32,0.0588216188244668)
--(axis cs:33,0.0575674376163326)
--(axis cs:34,0.0543787145588783)
--(axis cs:35,0.0542253016853686)
--(axis cs:36,0.052423616783675)
--(axis cs:37,0.0507972753430667)
--(axis cs:38,0.0447286066464856)
--(axis cs:39,0.0427350322893294)
--(axis cs:40,0.0396577344033624)
--(axis cs:41,0.0388375831557457)
--(axis cs:42,0.0353897302858142)
--(axis cs:43,0.0352365203911538)
--(axis cs:44,0.0342058190248711)
--(axis cs:45,0.0333112179082447)
--(axis cs:46,0.0297218006534051)
--(axis cs:47,0.02925041183829)
--(axis cs:48,0.0289366390239416)
--(axis cs:49,0.0287281894276406)
--(axis cs:50,0.027704668884422)
--(axis cs:50,0.0379314778610262)
--(axis cs:50,0.0379314778610262)
--(axis cs:49,0.0385168128715617)
--(axis cs:48,0.0393725565609319)
--(axis cs:47,0.0395320769678211)
--(axis cs:46,0.039810107941024)
--(axis cs:45,0.0419644805299724)
--(axis cs:44,0.0441549254466072)
--(axis cs:43,0.047486456652137)
--(axis cs:42,0.0489356418855603)
--(axis cs:41,0.0519426310609125)
--(axis cs:40,0.0524295108410772)
--(axis cs:39,0.0560439699596165)
--(axis cs:38,0.0601576956040919)
--(axis cs:37,0.0653047631088551)
--(axis cs:36,0.0665071595620215)
--(axis cs:35,0.0679698668069191)
--(axis cs:34,0.0681561666136415)
--(axis cs:33,0.0698212639436179)
--(axis cs:32,0.0711194692498258)
--(axis cs:31,0.0735111490708296)
--(axis cs:30,0.0756630863393933)
--(axis cs:29,0.0796081418458316)
--(axis cs:28,0.0812864229024938)
--(axis cs:27,0.0834186826816302)
--(axis cs:26,0.0858884492261974)
--(axis cs:25,0.0903980956630372)
--(axis cs:24,0.0918888972060752)
--(axis cs:23,0.0939932055801463)
--(axis cs:22,0.0972281217787287)
--(axis cs:21,0.101010725685203)
--(axis cs:20,0.110119388570072)
--(axis cs:19,0.115691538281368)
--(axis cs:18,0.117482158326457)
--(axis cs:17,0.120996473127716)
--(axis cs:16,0.126426727255947)
--(axis cs:15,0.128761250870552)
--(axis cs:14,0.135914421443853)
--(axis cs:13,0.142248332662099)
--(axis cs:12,0.152950954979235)
--(axis cs:11,0.166301876686783)
--(axis cs:10,0.175840787690108)
--(axis cs:9,0.183015318333832)
--(axis cs:8,0.198153077650713)
--(axis cs:7,0.205652428193418)
--(axis cs:6,0.211559225297995)
--(axis cs:5,0.229972229042153)
--cycle;

\path [fill=colorGP, fill opacity=0.1]
(axis cs:5,0.229972229042153)
--(axis cs:5,0.21710238749041)
--(axis cs:6,0.200309700446158)
--(axis cs:7,0.172642371797726)
--(axis cs:8,0.143085183754668)
--(axis cs:9,0.126310835209599)
--(axis cs:10,0.111781544013711)
--(axis cs:11,0.104655985146712)
--(axis cs:12,0.0954930207679072)
--(axis cs:13,0.0887366931882569)
--(axis cs:14,0.0855630412434489)
--(axis cs:15,0.0824404849151502)
--(axis cs:16,0.0784176091392623)
--(axis cs:17,0.0768319952522663)
--(axis cs:18,0.0733269428278585)
--(axis cs:19,0.0704938371835569)
--(axis cs:20,0.0681138170508396)
--(axis cs:21,0.0653267719096829)
--(axis cs:22,0.0641226789496801)
--(axis cs:23,0.0614158250634304)
--(axis cs:24,0.0593493787929961)
--(axis cs:25,0.0582443865917597)
--(axis cs:26,0.0568004506012404)
--(axis cs:27,0.0558423988981941)
--(axis cs:28,0.0553442456490849)
--(axis cs:29,0.0544104545936577)
--(axis cs:30,0.0536323661592069)
--(axis cs:31,0.0517465276050149)
--(axis cs:32,0.0507336042348001)
--(axis cs:33,0.0496351817565472)
--(axis cs:34,0.0484386970533811)
--(axis cs:35,0.0463463660032869)
--(axis cs:36,0.0444048631919866)
--(axis cs:37,0.0441747349819475)
--(axis cs:38,0.0436653520525761)
--(axis cs:39,0.0435353016910894)
--(axis cs:40,0.0432458253624085)
--(axis cs:41,0.0417667708372308)
--(axis cs:42,0.0413897535859784)
--(axis cs:43,0.0410878132291635)
--(axis cs:44,0.0384449215455703)
--(axis cs:45,0.0383730844351158)
--(axis cs:46,0.0381951880110114)
--(axis cs:47,0.0377054313396284)
--(axis cs:48,0.0358720138674735)
--(axis cs:49,0.0357519998850474)
--(axis cs:50,0.035500023917921)
--(axis cs:50,0.0414785838658191)
--(axis cs:50,0.0414785838658191)
--(axis cs:49,0.0415709542449192)
--(axis cs:48,0.0420595451521691)
--(axis cs:47,0.0427076898693663)
--(axis cs:46,0.0430664602143396)
--(axis cs:45,0.0440027892782668)
--(axis cs:44,0.0440341908261541)
--(axis cs:43,0.0464411404990339)
--(axis cs:42,0.0466965208392798)
--(axis cs:41,0.0470079338929005)
--(axis cs:40,0.0481436676301634)
--(axis cs:39,0.0484414747204803)
--(axis cs:38,0.0485096306910097)
--(axis cs:37,0.0496542885881978)
--(axis cs:36,0.0502130159902322)
--(axis cs:35,0.0512639533496915)
--(axis cs:34,0.0527011403688273)
--(axis cs:33,0.0581637371210983)
--(axis cs:32,0.0592435461024564)
--(axis cs:31,0.0599278840365871)
--(axis cs:30,0.0625667481418208)
--(axis cs:29,0.0634059744093781)
--(axis cs:28,0.0649432925605156)
--(axis cs:27,0.0656355828718669)
--(axis cs:26,0.066546855501863)
--(axis cs:25,0.0680904189583013)
--(axis cs:24,0.0692755030002074)
--(axis cs:23,0.0712119580874435)
--(axis cs:22,0.0732718818914506)
--(axis cs:21,0.0742441783889251)
--(axis cs:20,0.0781083754973152)
--(axis cs:19,0.0816074647508167)
--(axis cs:18,0.0853813580601277)
--(axis cs:17,0.0894613136000751)
--(axis cs:16,0.0919571557343814)
--(axis cs:15,0.0984375207764523)
--(axis cs:14,0.101314513339932)
--(axis cs:13,0.105894208072124)
--(axis cs:12,0.112036054961722)
--(axis cs:11,0.119415719534675)
--(axis cs:10,0.125867140901724)
--(axis cs:9,0.144155130987797)
--(axis cs:8,0.159529649430728)
--(axis cs:7,0.195512879499695)
--(axis cs:6,0.216515565909631)
--(axis cs:5,0.229972229042153)
--cycle;

\path [fill=colorDNGO, fill opacity=0.1]
(axis cs:5,0.229972229042153)
--(axis cs:5,0.21710238749041)
--(axis cs:6,0.162822433883316)
--(axis cs:7,0.139539209343873)
--(axis cs:8,0.119571311398833)
--(axis cs:9,0.100476845568446)
--(axis cs:10,0.0893018934461284)
--(axis cs:11,0.0803298380565958)
--(axis cs:12,0.0740084907070383)
--(axis cs:13,0.0691975747032489)
--(axis cs:14,0.0667881520659029)
--(axis cs:15,0.0632844731668084)
--(axis cs:16,0.0608492413866825)
--(axis cs:17,0.0595477303379055)
--(axis cs:18,0.0576543564766146)
--(axis cs:19,0.0560464667854456)
--(axis cs:20,0.0531854374906936)
--(axis cs:21,0.0512052405958865)
--(axis cs:22,0.0492029111925153)
--(axis cs:23,0.0471314549058904)
--(axis cs:24,0.0456495261502267)
--(axis cs:25,0.0445806292061675)
--(axis cs:26,0.0429737634933274)
--(axis cs:27,0.0391345630836159)
--(axis cs:28,0.0385586734985247)
--(axis cs:29,0.0382896549072748)
--(axis cs:30,0.0378521453828476)
--(axis cs:31,0.037282788838564)
--(axis cs:32,0.0370896090319186)
--(axis cs:33,0.0359482533891422)
--(axis cs:34,0.0334080363755105)
--(axis cs:35,0.0328944874116617)
--(axis cs:36,0.031585738063792)
--(axis cs:37,0.0297598304211245)
--(axis cs:38,0.0278133937588502)
--(axis cs:39,0.0275146103385441)
--(axis cs:40,0.0268849253014019)
--(axis cs:41,0.0263855668282113)
--(axis cs:42,0.0258795309129917)
--(axis cs:43,0.0255397533143883)
--(axis cs:44,0.0251083782107108)
--(axis cs:45,0.0240934740366982)
--(axis cs:46,0.0240210187995755)
--(axis cs:47,0.0237854981920791)
--(axis cs:48,0.0235908514780228)
--(axis cs:49,0.022900708361541)
--(axis cs:50,0.0226726475809663)
--(axis cs:50,0.0304509290957804)
--(axis cs:50,0.0304509290957804)
--(axis cs:49,0.030586024081862)
--(axis cs:48,0.0309909742704933)
--(axis cs:47,0.0310815325475244)
--(axis cs:46,0.0315222446493229)
--(axis cs:45,0.0317484951879258)
--(axis cs:44,0.0334800878216948)
--(axis cs:43,0.0336703436352408)
--(axis cs:42,0.033977415475616)
--(axis cs:41,0.0344604031528995)
--(axis cs:40,0.0347316896577118)
--(axis cs:39,0.0351477971959214)
--(axis cs:38,0.0358081237048363)
--(axis cs:37,0.0373876359444048)
--(axis cs:36,0.0386093155613666)
--(axis cs:35,0.0395452191217254)
--(axis cs:34,0.0401061779547068)
--(axis cs:33,0.0419654791329775)
--(axis cs:32,0.0426141380932124)
--(axis cs:31,0.0428505553013442)
--(axis cs:30,0.0432421031163497)
--(axis cs:29,0.0436759421273459)
--(axis cs:28,0.0440008281216884)
--(axis cs:27,0.0444384004489777)
--(axis cs:26,0.047499593792225)
--(axis cs:25,0.0489702167312815)
--(axis cs:24,0.0500176989533606)
--(axis cs:23,0.0516316461742082)
--(axis cs:22,0.0535708308479751)
--(axis cs:21,0.0566039244409512)
--(axis cs:20,0.0580441963217054)
--(axis cs:19,0.0599638590031843)
--(axis cs:18,0.0635423323752902)
--(axis cs:17,0.0661238700227293)
--(axis cs:16,0.068964215405761)
--(axis cs:15,0.0712236826674407)
--(axis cs:14,0.073434782842495)
--(axis cs:13,0.0756301511979598)
--(axis cs:12,0.0822357917195428)
--(axis cs:11,0.0877323577035962)
--(axis cs:10,0.0988879735817272)
--(axis cs:9,0.106758356254475)
--(axis cs:8,0.128464580868728)
--(axis cs:7,0.147419012832314)
--(axis cs:6,0.17427990721908)
--(axis cs:5,0.229972229042153)
--cycle;

\path [fill=colorDGP, fill opacity=0.1]
(axis cs:5,0.229972229042153)
--(axis cs:5,0.21710238749041)
--(axis cs:6,0.192438086603204)
--(axis cs:7,0.163135106791521)
--(axis cs:8,0.119625231628889)
--(axis cs:9,0.10954466016472)
--(axis cs:10,0.0937855362746626)
--(axis cs:11,0.0890990037746094)
--(axis cs:12,0.0842994862998966)
--(axis cs:13,0.0799689788924252)
--(axis cs:14,0.0726775602503533)
--(axis cs:15,0.0655934380150659)
--(axis cs:16,0.0621944001395246)
--(axis cs:17,0.0596705456249492)
--(axis cs:18,0.0569088993841469)
--(axis cs:19,0.0549917600015294)
--(axis cs:20,0.0541024777120235)
--(axis cs:21,0.0523562637838312)
--(axis cs:22,0.0512910081960112)
--(axis cs:23,0.051222519473203)
--(axis cs:24,0.0507033876968981)
--(axis cs:25,0.0467973465556716)
--(axis cs:26,0.0438422893914974)
--(axis cs:27,0.0418949328471505)
--(axis cs:28,0.0402381110817726)
--(axis cs:29,0.0384319661379432)
--(axis cs:30,0.0370051257173546)
--(axis cs:31,0.0346885178358028)
--(axis cs:32,0.033225398384863)
--(axis cs:33,0.0318517743847542)
--(axis cs:34,0.0316051412611633)
--(axis cs:35,0.0307112364747519)
--(axis cs:36,0.0295322073598076)
--(axis cs:37,0.0288332784607667)
--(axis cs:38,0.0274153342706747)
--(axis cs:39,0.0269797810369456)
--(axis cs:40,0.0263390858289064)
--(axis cs:41,0.0260036035843466)
--(axis cs:42,0.0254192389036474)
--(axis cs:43,0.0253202532230446)
--(axis cs:44,0.0252477695192634)
--(axis cs:45,0.0249521589147681)
--(axis cs:46,0.0245294272177282)
--(axis cs:47,0.0243543125057462)
--(axis cs:48,0.0232888186061312)
--(axis cs:49,0.0229145814456802)
--(axis cs:50,0.0228942178006145)
--(axis cs:50,0.0254795572847339)
--(axis cs:50,0.0254795572847339)
--(axis cs:49,0.0255265864974801)
--(axis cs:48,0.0261582366865377)
--(axis cs:47,0.0266473738663484)
--(axis cs:46,0.0267822617066741)
--(axis cs:45,0.0270721209743868)
--(axis cs:44,0.0275465811126066)
--(axis cs:43,0.0278234829607847)
--(axis cs:42,0.0281733588256886)
--(axis cs:41,0.0288011766382653)
--(axis cs:40,0.0290497796374978)
--(axis cs:39,0.0293921884948253)
--(axis cs:38,0.03032349596395)
--(axis cs:37,0.031582222422669)
--(axis cs:36,0.0333953496127332)
--(axis cs:35,0.0373155792881315)
--(axis cs:34,0.0380465709692343)
--(axis cs:33,0.0382738921160702)
--(axis cs:32,0.0402838636228005)
--(axis cs:31,0.0417920818360335)
--(axis cs:30,0.0436489653083839)
--(axis cs:29,0.0451033453794043)
--(axis cs:28,0.0491933628936531)
--(axis cs:27,0.0506854029036714)
--(axis cs:26,0.0516807720148754)
--(axis cs:25,0.0534535113869909)
--(axis cs:24,0.05548073340081)
--(axis cs:23,0.0561134128982804)
--(axis cs:22,0.0561513687802884)
--(axis cs:21,0.0577997163810045)
--(axis cs:20,0.0598857271602295)
--(axis cs:19,0.0613799246777615)
--(axis cs:18,0.0648945680269161)
--(axis cs:17,0.0670137810815454)
--(axis cs:16,0.0701719251138507)
--(axis cs:15,0.074197467777458)
--(axis cs:14,0.0782485709745389)
--(axis cs:13,0.0854567139715733)
--(axis cs:12,0.0884150537893443)
--(axis cs:11,0.0929517090876777)
--(axis cs:10,0.100657285931842)
--(axis cs:9,0.121865114036954)
--(axis cs:8,0.135250051873942)
--(axis cs:7,0.183693637788156)
--(axis cs:6,0.205858249564612)
--(axis cs:5,0.229972229042153)
--cycle;

\path [fill=colorPFN_GP, fill opacity=0.1]
(axis cs:5,0.229972229042153)
--(axis cs:5,0.21710238749041)
--(axis cs:6,0.160278831014592)
--(axis cs:7,0.130782163847823)
--(axis cs:8,0.116476611076302)
--(axis cs:9,0.0964332893365131)
--(axis cs:10,0.0909970940271918)
--(axis cs:11,0.0848937200910018)
--(axis cs:12,0.0799838492545874)
--(axis cs:13,0.0753719536792764)
--(axis cs:14,0.0725192375026625)
--(axis cs:15,0.0662305828369612)
--(axis cs:16,0.0631898957810735)
--(axis cs:17,0.0597904014311965)
--(axis cs:18,0.0566694875673723)
--(axis cs:19,0.0549615438692448)
--(axis cs:20,0.0524587204684884)
--(axis cs:21,0.0502854282954943)
--(axis cs:22,0.0485177375759072)
--(axis cs:23,0.0454456648919411)
--(axis cs:24,0.0444372309565432)
--(axis cs:25,0.0407273515835956)
--(axis cs:26,0.0384756222783664)
--(axis cs:27,0.0367077372183817)
--(axis cs:28,0.0357680803128359)
--(axis cs:29,0.0338661626496755)
--(axis cs:30,0.0328506051745369)
--(axis cs:31,0.0317215049957481)
--(axis cs:32,0.0288762770262921)
--(axis cs:33,0.0282302502599518)
--(axis cs:34,0.0276126265046207)
--(axis cs:35,0.027171861652372)
--(axis cs:36,0.0259207942838157)
--(axis cs:37,0.0244240034290502)
--(axis cs:38,0.0226805049232498)
--(axis cs:39,0.0225551593726975)
--(axis cs:40,0.0216690437706978)
--(axis cs:41,0.0211328043231009)
--(axis cs:42,0.020691153889118)
--(axis cs:43,0.0195500586840891)
--(axis cs:44,0.019347499671286)
--(axis cs:45,0.0182153763867652)
--(axis cs:46,0.0182153763867652)
--(axis cs:47,0.0181595108243169)
--(axis cs:48,0.0180503633167558)
--(axis cs:49,0.0176131802485801)
--(axis cs:50,0.0174261833006113)
--(axis cs:50,0.0190491751478515)
--(axis cs:50,0.0190491751478515)
--(axis cs:49,0.0192911309590044)
--(axis cs:48,0.0197975457074311)
--(axis cs:47,0.019864637828914)
--(axis cs:46,0.020095003759592)
--(axis cs:45,0.020095003759592)
--(axis cs:44,0.0211877494332459)
--(axis cs:43,0.0214391028494531)
--(axis cs:42,0.0255267584376508)
--(axis cs:41,0.0268282171956197)
--(axis cs:40,0.0272343979720654)
--(axis cs:39,0.0277372755738238)
--(axis cs:38,0.027855769914311)
--(axis cs:37,0.0294444760602151)
--(axis cs:36,0.0305886549822034)
--(axis cs:35,0.0320135843531685)
--(axis cs:34,0.0322725951091343)
--(axis cs:33,0.0326056884009755)
--(axis cs:32,0.0335400814715294)
--(axis cs:31,0.0358461008734134)
--(axis cs:30,0.0368527717228782)
--(axis cs:29,0.0374051929806618)
--(axis cs:28,0.0393245500594882)
--(axis cs:27,0.0404841037423493)
--(axis cs:26,0.0439548761617103)
--(axis cs:25,0.0456213709484763)
--(axis cs:24,0.0480627903424171)
--(axis cs:23,0.0492098431367699)
--(axis cs:22,0.054460064143617)
--(axis cs:21,0.0567772872821245)
--(axis cs:20,0.0588415103598554)
--(axis cs:19,0.0613160108730878)
--(axis cs:18,0.0625221929845047)
--(axis cs:17,0.0652598812884552)
--(axis cs:16,0.0694508371730795)
--(axis cs:15,0.0740021478936112)
--(axis cs:14,0.0787037147583354)
--(axis cs:13,0.0841041113492363)
--(axis cs:12,0.0885640030903338)
--(axis cs:11,0.0949998034039603)
--(axis cs:10,0.100816834998825)
--(axis cs:9,0.108115914008586)
--(axis cs:8,0.133084280281661)
--(axis cs:7,0.145019877547662)
--(axis cs:6,0.175050770556326)
--(axis cs:5,0.229972229042153)
--cycle;

\path [fill=colorPFN_BNN, fill opacity=0.1]
(axis cs:5,0.229972229042153)
--(axis cs:5,0.21710238749041)
--(axis cs:6,0.183594967492045)
--(axis cs:7,0.144762447909839)
--(axis cs:8,0.133849891657266)
--(axis cs:9,0.11462367856701)
--(axis cs:10,0.100715892464588)
--(axis cs:11,0.0926872358859926)
--(axis cs:12,0.0889921381211661)
--(axis cs:13,0.0861495072458914)
--(axis cs:14,0.0793350105495573)
--(axis cs:15,0.0742156476452009)
--(axis cs:16,0.0661353879138803)
--(axis cs:17,0.0616343097238552)
--(axis cs:18,0.0579970783460914)
--(axis cs:19,0.0564389749899866)
--(axis cs:20,0.0532713786660544)
--(axis cs:21,0.051352854979348)
--(axis cs:22,0.0502335523502818)
--(axis cs:23,0.0480985102273837)
--(axis cs:24,0.0447505973773237)
--(axis cs:25,0.0440795334630975)
--(axis cs:26,0.0424967914435655)
--(axis cs:27,0.0402867393694132)
--(axis cs:28,0.03753282371629)
--(axis cs:29,0.0365076825310318)
--(axis cs:30,0.0360734312457983)
--(axis cs:31,0.0336938069260269)
--(axis cs:32,0.0327469639836636)
--(axis cs:33,0.0316070492430778)
--(axis cs:34,0.0298969099962349)
--(axis cs:35,0.0294530965083396)
--(axis cs:36,0.0282636705310643)
--(axis cs:37,0.0279108359802767)
--(axis cs:38,0.0276635479891179)
--(axis cs:39,0.0273891226297093)
--(axis cs:40,0.0271182534142053)
--(axis cs:41,0.0266494209042222)
--(axis cs:42,0.0260714939404244)
--(axis cs:43,0.0254125569830256)
--(axis cs:44,0.0252551900182939)
--(axis cs:45,0.0242681928573477)
--(axis cs:46,0.0240153468174058)
--(axis cs:47,0.0237422661506803)
--(axis cs:48,0.0234220815226007)
--(axis cs:49,0.0231110108017196)
--(axis cs:50,0.0226515868825414)
--(axis cs:50,0.0287594111470262)
--(axis cs:50,0.0287594111470262)
--(axis cs:49,0.0290843987232345)
--(axis cs:48,0.0293239019762014)
--(axis cs:47,0.0294701117731199)
--(axis cs:46,0.0295670690004524)
--(axis cs:45,0.0297311055903109)
--(axis cs:44,0.0302775056948107)
--(axis cs:43,0.0303977180892438)
--(axis cs:42,0.030959132179896)
--(axis cs:41,0.0313764433924255)
--(axis cs:40,0.03188812808349)
--(axis cs:39,0.0322496083433775)
--(axis cs:38,0.0328556033655389)
--(axis cs:37,0.0330652085291479)
--(axis cs:36,0.0334618898824823)
--(axis cs:35,0.0345975069817636)
--(axis cs:34,0.0355643080299272)
--(axis cs:33,0.0370100657106305)
--(axis cs:32,0.0380777137789999)
--(axis cs:31,0.0396342344038521)
--(axis cs:30,0.0400867399381067)
--(axis cs:29,0.0408714224779201)
--(axis cs:28,0.0430712454033133)
--(axis cs:27,0.0466118090440831)
--(axis cs:26,0.0478926530683019)
--(axis cs:25,0.048669258382749)
--(axis cs:24,0.0494693973336476)
--(axis cs:23,0.0582870080837376)
--(axis cs:22,0.0612775534728455)
--(axis cs:21,0.062178627040015)
--(axis cs:20,0.0639110246020631)
--(axis cs:19,0.0690984449767098)
--(axis cs:18,0.0714178860155361)
--(axis cs:17,0.075396850561149)
--(axis cs:16,0.0785996259794174)
--(axis cs:15,0.0849003956145668)
--(axis cs:14,0.0884687144954101)
--(axis cs:13,0.0931491585352917)
--(axis cs:12,0.0968654627409627)
--(axis cs:11,0.10495921660506)
--(axis cs:10,0.114553056517067)
--(axis cs:9,0.127804744396634)
--(axis cs:8,0.15088390448601)
--(axis cs:7,0.164539810221328)
--(axis cs:6,0.201746124399505)
--(axis cs:5,0.229972229042153)
--cycle;

\addplot [line width=2pt, colorRandom]
table {%
5 0.223537308266282
6 0.215106257648508
7 0.19410934409419
8 0.185400094303649
9 0.17659597044002
10 0.165819737853608
11 0.158818427476327
12 0.151037390517623
13 0.148911009186116
14 0.139632809438272
15 0.13647216363216
16 0.135468225746457
17 0.134956091447036
18 0.134087225477909
19 0.127195817252017
20 0.126622394114084
21 0.126392184436021
22 0.122415381712342
23 0.118961806854965
24 0.117020650667617
25 0.11676936104974
26 0.116287274259848
27 0.115439911462744
28 0.113455031751373
29 0.111093313802576
30 0.110036572043108
31 0.109144733925647
32 0.107942014762115
33 0.107106190899556
34 0.106940490816706
35 0.106805107568487
36 0.106190986857964
37 0.104742698926353
38 0.104580139107721
39 0.104449993254643
40 0.104261804804929
41 0.103601681402623
42 0.103416996057268
43 0.10224322441642
44 0.101811004517742
45 0.101719776673223
46 0.10119023118016
47 0.0999610015056053
48 0.0975283017637233
49 0.0974327740741479
50 0.0968397847220513
};
\addplot [line width=2pt, colorHEBO]
table {%
5 0.223537308266282
6 0.209163182901579
7 0.202960832368035
8 0.195023479540642
9 0.180708307084355
10 0.174034748627196
11 0.163272016633155
12 0.150619591582234
13 0.139658551974039
14 0.133834912410882
15 0.126698002199548
16 0.124370514030187
17 0.119191325183156
18 0.113462947752973
19 0.111827787075258
20 0.105799022092879
21 0.0952066661386408
22 0.0903592788267765
23 0.0875272216348163
24 0.0855454436184619
25 0.0838350707011541
26 0.0801145686405212
27 0.0778933034425604
28 0.0752682780078968
29 0.0737828936900674
30 0.0699663733732616
31 0.0677421775928891
32 0.0649705440371463
33 0.0636943507799752
34 0.0612674405862599
35 0.0610975842461438
36 0.0594653881728482
37 0.0580510192259609
38 0.0524431511252887
39 0.049389501124473
40 0.0460436226222198
41 0.0453901071083291
42 0.0421626860856872
43 0.0413614885216454
44 0.0391803722357391
45 0.0376378492191086
46 0.0347659542972145
47 0.0343912444030556
48 0.0341545977924367
49 0.0336225011496011
50 0.0328180733727241
};
\addplot [line width=2pt, colorGP]
table {%
5 0.223537308266282
6 0.208412633177894
7 0.18407762564871
8 0.151307416592698
9 0.135232983098698
10 0.118824342457717
11 0.112035852340694
12 0.103764537864815
13 0.0973154506301903
14 0.0934387772916905
15 0.0904390028458012
16 0.0851873824368218
17 0.0831466544261707
18 0.0793541504439931
19 0.0760506509671868
20 0.0731110962740774
21 0.069785475149304
22 0.0686972804205654
23 0.0663138915754369
24 0.0643124408966017
25 0.0631674027750305
26 0.0616736530515517
27 0.0607389908850305
28 0.0601437691048003
29 0.0589082145015179
30 0.0580995571505138
31 0.055837205820801
32 0.0549885751686283
33 0.0538994594388228
34 0.0505699187111042
35 0.0488051596764892
36 0.0473089395911094
37 0.0469145117850727
38 0.0460874913717929
39 0.0459883882057849
40 0.045694746496286
41 0.0443873523650657
42 0.0440431372126291
43 0.0437644768640987
44 0.0412395561858622
45 0.0411879368566913
46 0.0406308241126755
47 0.0402065606044973
48 0.0389657795098213
49 0.0386614770649833
50 0.0384893038918701
};
\addplot [line width=2pt, colorDNGO]
table {%
5 0.223537308266282
6 0.168551170551198
7 0.143479111088093
8 0.12401794613378
9 0.10361760091146
10 0.0940949335139278
11 0.084031097880096
12 0.0781221412132905
13 0.0724138629506043
14 0.0701114674541989
15 0.0672540779171246
16 0.0649067283962217
17 0.0628358001803174
18 0.0605983444259524
19 0.058005162894315
20 0.0556148169061995
21 0.0539045825184188
22 0.0513868710202452
23 0.0493815505400493
24 0.0478336125517936
25 0.0467754229687245
26 0.0452366786427762
27 0.0417864817662968
28 0.0412797508101065
29 0.0409827985173103
30 0.0405471242495987
31 0.0400666720699541
32 0.0398518735625655
33 0.0389568662610598
34 0.0367571071651087
35 0.0362198532666936
36 0.0350975268125793
37 0.0335737331827646
38 0.0318107587318433
39 0.0313312037672327
40 0.0308083074795569
41 0.0304229849905554
42 0.0299284731943038
43 0.0296050484748145
44 0.0292942330162028
45 0.027920984612312
46 0.0277716317244492
47 0.0274335153698018
48 0.0272909128742581
49 0.0267433662217015
50 0.0265617883383734
};
\addplot [line width=2pt, colorDGP]
table {%
5 0.223537308266282
6 0.199148168083908
7 0.173414372289838
8 0.127437641751415
9 0.115704887100837
10 0.0972214111032524
11 0.0910253564311435
12 0.0863572700446205
13 0.0827128464319992
14 0.0754630656124461
15 0.069895452896262
16 0.0661831626266876
17 0.0633421633532473
18 0.0609017337055315
19 0.0581858423396455
20 0.0569941024361265
21 0.0550779900824178
22 0.0537211884881498
23 0.0536679661857417
24 0.053092060548854
25 0.0501254289713313
26 0.0477615307031864
27 0.0462901678754109
28 0.0447157369877129
29 0.0417676557586737
30 0.0403270455128693
31 0.0382402998359182
32 0.0367546310038317
33 0.0350628332504122
34 0.0348258561151988
35 0.0340134078814417
36 0.0314637784862704
37 0.0302077504417178
38 0.0288694151173123
39 0.0281859847658855
40 0.0276944327332021
41 0.027402390111306
42 0.026796298864668
43 0.0265718680919146
44 0.026397175315935
45 0.0260121399445774
46 0.0256558444622012
47 0.0255008431860473
48 0.0247235276463344
49 0.0242205839715802
50 0.0241868875426742
};
\addplot [line width=2pt, colorPFN_GP]
table {%
5 0.223537308266282
6 0.167664800785459
7 0.137901020697743
8 0.124780445678982
9 0.10227460167255
10 0.0959069645130084
11 0.089946761747481
12 0.0842739261724606
13 0.0797380325142564
14 0.0756114761304989
15 0.0701163653652862
16 0.0663203664770765
17 0.0625251413598259
18 0.0595958402759385
19 0.0581387773711663
20 0.0556501154141719
21 0.0535313577888094
22 0.0514889008597621
23 0.0473277540143555
24 0.0462500106494802
25 0.0431743612660359
26 0.0412152492200384
27 0.0385959204803655
28 0.0375463151861621
29 0.0356356778151686
30 0.0348516884487076
31 0.0337838029345808
32 0.0312081792489108
33 0.0304179693304636
34 0.0299426108068775
35 0.0295927230027702
36 0.0282547246330095
37 0.0269342397446327
38 0.0252681374187804
39 0.0251462174732606
40 0.0244517208713816
41 0.0239805107593603
42 0.0231089561633844
43 0.0204945807667711
44 0.020267624552266
45 0.0191551900731786
46 0.0191551900731786
47 0.0190120743266155
48 0.0189239545120935
49 0.0184521556037923
50 0.0182376792242314
};
\addplot [line width=2pt, colorPFN_BNN]
table {%
5 0.223537308266282
6 0.192670545945775
7 0.154651129065583
8 0.142366898071638
9 0.121214211481822
10 0.107634474490827
11 0.0988232262455261
12 0.0929288004310644
13 0.0896493328905916
14 0.0839018625224837
15 0.0795580216298839
16 0.0723675069466488
17 0.0685155801425021
18 0.0647074821808137
19 0.0627687099833482
20 0.0585912016340588
21 0.0567657410096815
22 0.0557555529115636
23 0.0531927591555606
24 0.0471099973554856
25 0.0463743959229232
26 0.0451947222559337
27 0.0434492742067481
28 0.0403020345598016
29 0.038689552504476
30 0.0380800855919525
31 0.0366640206649395
32 0.0354123388813317
33 0.0343085574768542
34 0.0327306090130811
35 0.0320253017450516
36 0.0308627802067733
37 0.0304880222547123
38 0.0302595756773284
39 0.0298193654865434
40 0.0295031907488476
41 0.0290129321483239
42 0.0285153130601602
43 0.0279051375361347
44 0.0277663478565523
45 0.0269996492238293
46 0.0267912079089291
47 0.0266061889619001
48 0.026372991749401
49 0.0260977047624771
50 0.0257054990147838
};
\end{axis}

\end{tikzpicture}

%% file: icml2023/figures/pd1_ranks_random_hard.tex
\begin{tikzpicture}

\colorlet{color0}{colorRandom}
\colorlet{color1}{colorPFN_GP}
\colorlet{color2}{colorHEBO}
\colorlet{color3}{colorPFN_GP_plus_prior}
\colorlet{color4}{colorRandom_plus_prior}
\colorlet{color5}{colorPFN_BNN}

\begin{axis}[
legend cell align={left},
legend style={
  fill opacity=0.8,
  draw opacity=1,
  text opacity=1,
  at={(0.91,0.5)},
  anchor=east,
  draw=white!80!black
},
tick align=outside,
tick pos=left,
x grid style={white!69.0196078431373!black},
xmin=-2.5, xmax=52.5,
xtick style={color=black},
y grid style={white!69.0196078431373!black},
xlabel={Number of trials},
ylabel={Average Rank},
ymin=3., ymax=9.5,
ytick style={color=black},
ytick={4,6, 8},
yticklabels={$\ \ 4$,$\ 6$,$\ 8$},
height=\benchmarkplotheight,
width=\benchmarkplotwidth,
]
\path [draw=color0, fill=color0, opacity=0.2]
(axis cs:0,5.63541666666667)
--(axis cs:0,5.44791666666667)
--(axis cs:1,7.06744791666667)
--(axis cs:2,7.51041666666667)
--(axis cs:3,8.109375)
--(axis cs:4,8.21354166666667)
--(axis cs:5,8.41145833333333)
--(axis cs:6,8.46354166666667)
--(axis cs:7,8.453125)
--(axis cs:8,8.54166666666667)
--(axis cs:9,8.536328125)
--(axis cs:10,8.56770833333333)
--(axis cs:11,8.66145833333333)
--(axis cs:12,8.68229166666667)
--(axis cs:13,8.60403645833333)
--(axis cs:14,8.57265625)
--(axis cs:15,8.625)
--(axis cs:16,8.66666666666667)
--(axis cs:17,8.64049479166667)
--(axis cs:18,8.49479166666667)
--(axis cs:19,8.43736979166667)
--(axis cs:20,8.202734375)
--(axis cs:21,8.17161458333333)
--(axis cs:22,8.21875)
--(axis cs:23,8.28645833333333)
--(axis cs:24,8.23919270833333)
--(axis cs:25,8.13541666666667)
--(axis cs:26,8.15065104166667)
--(axis cs:27,8.06236979166667)
--(axis cs:28,8.109375)
--(axis cs:29,8.203125)
--(axis cs:30,8.15091145833333)
--(axis cs:31,8.24973958333333)
--(axis cs:32,8.31236979166667)
--(axis cs:33,8.33841145833333)
--(axis cs:34,8.32291666666667)
--(axis cs:35,8.29140625)
--(axis cs:36,8.130078125)
--(axis cs:37,8.07291666666667)
--(axis cs:38,8.109375)
--(axis cs:39,8.09361979166667)
--(axis cs:40,8.12473958333333)
--(axis cs:41,8.140625)
--(axis cs:42,8.176171875)
--(axis cs:43,8.18229166666667)
--(axis cs:44,8.18203125)
--(axis cs:45,8.10403645833333)
--(axis cs:46,8.08333333333333)
--(axis cs:47,8.06236979166667)
--(axis cs:48,8.08841145833333)
--(axis cs:49,8.12486979166667)
--(axis cs:50,8.11458333333333)
--(axis cs:50,9)
--(axis cs:50,9)
--(axis cs:49,8.99479166666667)
--(axis cs:48,8.953125)
--(axis cs:47,8.95846354166667)
--(axis cs:46,8.969140625)
--(axis cs:45,8.948046875)
--(axis cs:44,9.015625)
--(axis cs:43,8.99479166666667)
--(axis cs:42,9)
--(axis cs:41,9.00520833333333)
--(axis cs:40,8.97942708333333)
--(axis cs:39,8.94791666666667)
--(axis cs:38,8.94817708333333)
--(axis cs:37,8.92200520833333)
--(axis cs:36,8.96875)
--(axis cs:35,9.041796875)
--(axis cs:34,9.06783854166667)
--(axis cs:33,9.08333333333333)
--(axis cs:32,9.02096354166667)
--(axis cs:31,9.01041666666667)
--(axis cs:30,9.00013020833333)
--(axis cs:29,9.00546875)
--(axis cs:28,8.95833333333333)
--(axis cs:27,8.91666666666667)
--(axis cs:26,9.00520833333333)
--(axis cs:25,9.03138020833333)
--(axis cs:24,9.06783854166667)
--(axis cs:23,9.078125)
--(axis cs:22,9.09895833333333)
--(axis cs:21,8.97916666666667)
--(axis cs:20,9.02083333333333)
--(axis cs:19,9.15625)
--(axis cs:18,9.19791666666667)
--(axis cs:17,9.3125)
--(axis cs:16,9.359375)
--(axis cs:15,9.35416666666667)
--(axis cs:14,9.31783854166667)
--(axis cs:13,9.328125)
--(axis cs:12,9.36458333333333)
--(axis cs:11,9.338671875)
--(axis cs:10,9.26041666666667)
--(axis cs:9,9.25)
--(axis cs:8,9.20846354166667)
--(axis cs:7,9.13020833333333)
--(axis cs:6,9.13541666666667)
--(axis cs:5,9.09388020833333)
--(axis cs:4,8.95833333333333)
--(axis cs:3,8.828125)
--(axis cs:2,8.4375)
--(axis cs:1,8.04700520833333)
--(axis cs:0,5.63541666666667)
--cycle;

\path [draw=color1, fill=color1, opacity=0.2]
(axis cs:0,5.64583333333333)
--(axis cs:0,5.4375)
--(axis cs:1,4.43736979166667)
--(axis cs:2,4.46328125)
--(axis cs:3,4.46875)
--(axis cs:4,4.51536458333333)
--(axis cs:5,4.66653645833333)
--(axis cs:6,4.77604166666667)
--(axis cs:7,4.75)
--(axis cs:8,4.8125)
--(axis cs:9,4.765625)
--(axis cs:10,4.60403645833333)
--(axis cs:11,4.51041666666667)
--(axis cs:12,4.37486979166667)
--(axis cs:13,4.33333333333333)
--(axis cs:14,4.32786458333333)
--(axis cs:15,4.26041666666667)
--(axis cs:16,4.22916666666667)
--(axis cs:17,4.25494791666667)
--(axis cs:18,4.10924479166667)
--(axis cs:19,4.02083333333333)
--(axis cs:20,3.958203125)
--(axis cs:21,3.90104166666667)
--(axis cs:22,3.91666666666667)
--(axis cs:23,3.80716145833333)
--(axis cs:24,3.65104166666667)
--(axis cs:25,3.77591145833333)
--(axis cs:26,3.81236979166667)
--(axis cs:27,3.890625)
--(axis cs:28,4.01549479166667)
--(axis cs:29,4.02578125)
--(axis cs:30,4.13020833333333)
--(axis cs:31,4.15104166666667)
--(axis cs:32,4.21848958333333)
--(axis cs:33,4.11979166666667)
--(axis cs:34,4.098828125)
--(axis cs:35,4.05716145833333)
--(axis cs:36,4.07291666666667)
--(axis cs:37,4.093359375)
--(axis cs:38,4.12994791666667)
--(axis cs:39,4.08841145833333)
--(axis cs:40,4.18229166666667)
--(axis cs:41,4.15611979166667)
--(axis cs:42,4.234375)
--(axis cs:43,4.22369791666667)
--(axis cs:44,4.20833333333333)
--(axis cs:45,4.25520833333333)
--(axis cs:46,4.21875)
--(axis cs:47,4.22890625)
--(axis cs:48,4.22369791666667)
--(axis cs:49,4.24466145833333)
--(axis cs:50,4.21848958333333)
--(axis cs:50,5.16692708333333)
--(axis cs:50,5.16692708333333)
--(axis cs:49,5.15638020833333)
--(axis cs:48,5.13059895833333)
--(axis cs:47,5.13541666666667)
--(axis cs:46,5.1875)
--(axis cs:45,5.22916666666667)
--(axis cs:44,5.140625)
--(axis cs:43,5.15104166666667)
--(axis cs:42,5.1828125)
--(axis cs:41,5.09908854166667)
--(axis cs:40,5.151171875)
--(axis cs:39,5.057421875)
--(axis cs:38,5.14583333333333)
--(axis cs:37,5.104296875)
--(axis cs:36,5.047265625)
--(axis cs:35,5.00533854166667)
--(axis cs:34,5.041796875)
--(axis cs:33,5.01575520833333)
--(axis cs:32,5.18763020833333)
--(axis cs:31,5.171875)
--(axis cs:30,5.09895833333333)
--(axis cs:29,5.00533854166667)
--(axis cs:28,4.91666666666667)
--(axis cs:27,4.828125)
--(axis cs:26,4.72916666666667)
--(axis cs:25,4.65182291666667)
--(axis cs:24,4.60950520833333)
--(axis cs:23,4.74479166666667)
--(axis cs:22,4.94270833333333)
--(axis cs:21,4.91666666666667)
--(axis cs:20,4.94791666666667)
--(axis cs:19,5.09375)
--(axis cs:18,5.10416666666667)
--(axis cs:17,5.27096354166667)
--(axis cs:16,5.32291666666667)
--(axis cs:15,5.35455729166667)
--(axis cs:14,5.41666666666667)
--(axis cs:13,5.38020833333333)
--(axis cs:12,5.42213541666667)
--(axis cs:11,5.50026041666667)
--(axis cs:10,5.65130208333333)
--(axis cs:9,5.77083333333333)
--(axis cs:8,5.828125)
--(axis cs:7,5.75)
--(axis cs:6,5.67708333333333)
--(axis cs:5,5.578515625)
--(axis cs:4,5.463671875)
--(axis cs:3,5.41158854166667)
--(axis cs:2,5.47408854166667)
--(axis cs:1,5.45833333333333)
--(axis cs:0,5.64583333333333)
--cycle;

\path [draw=color2, fill=color2, opacity=0.2]
(axis cs:0,5.96875)
--(axis cs:0,4.1875)
--(axis cs:1,5.26041666666667)
--(axis cs:2,5.125)
--(axis cs:3,4.79166666666667)
--(axis cs:4,4.81223958333333)
--(axis cs:5,4.76015625)
--(axis cs:6,4.70807291666667)
--(axis cs:7,4.46848958333333)
--(axis cs:8,4.38515625)
--(axis cs:9,4.47916666666667)
--(axis cs:10,4.49973958333333)
--(axis cs:11,4.38541666666667)
--(axis cs:12,4.53125)
--(axis cs:13,4.52083333333333)
--(axis cs:14,4.61458333333333)
--(axis cs:15,4.66666666666667)
--(axis cs:16,4.60416666666667)
--(axis cs:17,4.58307291666667)
--(axis cs:18,4.61458333333333)
--(axis cs:19,4.64583333333333)
--(axis cs:20,4.5625)
--(axis cs:21,4.61432291666667)
--(axis cs:22,4.58333333333333)
--(axis cs:23,4.60416666666667)
--(axis cs:24,4.69739583333333)
--(axis cs:25,4.47916666666667)
--(axis cs:26,4.4375)
--(axis cs:27,4.34375)
--(axis cs:28,4.40598958333333)
--(axis cs:29,4.40625)
--(axis cs:30,4.42708333333333)
--(axis cs:31,4.52083333333333)
--(axis cs:32,4.53098958333333)
--(axis cs:33,4.66666666666667)
--(axis cs:34,4.625)
--(axis cs:35,4.64583333333333)
--(axis cs:36,4.77083333333333)
--(axis cs:37,4.82265625)
--(axis cs:38,4.82265625)
--(axis cs:39,4.76041666666667)
--(axis cs:40,4.79166666666667)
--(axis cs:41,4.85390625)
--(axis cs:42,4.91640625)
--(axis cs:43,4.85416666666667)
--(axis cs:44,4.88541666666667)
--(axis cs:45,4.97916666666667)
--(axis cs:46,4.94765625)
--(axis cs:47,5.04140625)
--(axis cs:48,5.02083333333333)
--(axis cs:49,5.03098958333333)
--(axis cs:50,5.04140625)
--(axis cs:50,6.08333333333333)
--(axis cs:50,6.08333333333333)
--(axis cs:49,6.08333333333333)
--(axis cs:48,6.09401041666667)
--(axis cs:47,6.05208333333333)
--(axis cs:46,6.05234375)
--(axis cs:45,6.02083333333333)
--(axis cs:44,5.96875)
--(axis cs:43,5.96875)
--(axis cs:42,5.98984375)
--(axis cs:41,5.91666666666667)
--(axis cs:40,5.86458333333333)
--(axis cs:39,5.84401041666667)
--(axis cs:38,5.88541666666667)
--(axis cs:37,5.875)
--(axis cs:36,5.92708333333333)
--(axis cs:35,5.76041666666667)
--(axis cs:34,5.70833333333333)
--(axis cs:33,5.6875)
--(axis cs:32,5.67734375)
--(axis cs:31,5.65625)
--(axis cs:30,5.46875)
--(axis cs:29,5.43776041666667)
--(axis cs:28,5.4375)
--(axis cs:27,5.375)
--(axis cs:26,5.48958333333333)
--(axis cs:25,5.51041666666667)
--(axis cs:24,5.64583333333333)
--(axis cs:23,5.59375)
--(axis cs:22,5.58359375)
--(axis cs:21,5.55234375)
--(axis cs:20,5.51067708333333)
--(axis cs:19,5.59401041666667)
--(axis cs:18,5.57291666666667)
--(axis cs:17,5.53151041666667)
--(axis cs:16,5.66666666666667)
--(axis cs:15,5.6875)
--(axis cs:14,5.64583333333333)
--(axis cs:13,5.55208333333333)
--(axis cs:12,5.52083333333333)
--(axis cs:11,5.38567708333333)
--(axis cs:10,5.57291666666667)
--(axis cs:9,5.52083333333333)
--(axis cs:8,5.46875)
--(axis cs:7,5.47916666666667)
--(axis cs:6,5.72942708333333)
--(axis cs:5,5.81276041666667)
--(axis cs:4,5.88541666666667)
--(axis cs:3,5.82291666666667)
--(axis cs:2,6.20885416666667)
--(axis cs:1,6.40625)
--(axis cs:0,5.96875)
--cycle;

\path [draw=color3, fill=color3, opacity=0.2]
(axis cs:0,5.625)
--(axis cs:0,5.44791666666667)
--(axis cs:1,4.140625)
--(axis cs:2,3.88541666666667)
--(axis cs:3,3.426953125)
--(axis cs:4,3.49986979166667)
--(axis cs:5,3.6875)
--(axis cs:6,3.536328125)
--(axis cs:7,3.6875)
--(axis cs:8,3.56770833333333)
--(axis cs:9,3.52552083333333)
--(axis cs:10,3.44791666666667)
--(axis cs:11,3.25)
--(axis cs:12,3.22916666666667)
--(axis cs:13,3.18736979166667)
--(axis cs:14,3.15625)
--(axis cs:15,3.10924479166667)
--(axis cs:16,3.18216145833333)
--(axis cs:17,3.0625)
--(axis cs:18,3.18229166666667)
--(axis cs:19,3.13541666666667)
--(axis cs:20,3.161328125)
--(axis cs:21,3.124609375)
--(axis cs:22,3.23424479166667)
--(axis cs:23,3.192578125)
--(axis cs:24,3.1875)
--(axis cs:25,3.15611979166667)
--(axis cs:26,3.19270833333333)
--(axis cs:27,3.22903645833333)
--(axis cs:28,3.29674479166667)
--(axis cs:29,3.333203125)
--(axis cs:30,3.29166666666667)
--(axis cs:31,3.28111979166667)
--(axis cs:32,3.27057291666667)
--(axis cs:33,3.296875)
--(axis cs:34,3.33294270833333)
--(axis cs:35,3.31770833333333)
--(axis cs:36,3.34348958333333)
--(axis cs:37,3.364453125)
--(axis cs:38,3.395703125)
--(axis cs:39,3.348828125)
--(axis cs:40,3.2390625)
--(axis cs:41,3.22916666666667)
--(axis cs:42,3.24947916666667)
--(axis cs:43,3.25494791666667)
--(axis cs:44,3.1875)
--(axis cs:45,3.16145833333333)
--(axis cs:46,3.171875)
--(axis cs:47,3.11966145833333)
--(axis cs:48,3.07799479166667)
--(axis cs:49,3.14583333333333)
--(axis cs:50,3.12486979166667)
--(axis cs:50,4.04166666666667)
--(axis cs:50,4.04166666666667)
--(axis cs:49,4.03138020833333)
--(axis cs:48,3.95325520833333)
--(axis cs:47,4.041796875)
--(axis cs:46,4.05729166666667)
--(axis cs:45,4.088671875)
--(axis cs:44,4.11979166666667)
--(axis cs:43,4.1875)
--(axis cs:42,4.13033854166667)
--(axis cs:41,4.13020833333333)
--(axis cs:40,4.12526041666667)
--(axis cs:39,4.354296875)
--(axis cs:38,4.35416666666667)
--(axis cs:37,4.33359375)
--(axis cs:36,4.296875)
--(axis cs:35,4.27083333333333)
--(axis cs:34,4.23958333333333)
--(axis cs:33,4.260546875)
--(axis cs:32,4.25533854166667)
--(axis cs:31,4.18229166666667)
--(axis cs:30,4.20846354166667)
--(axis cs:29,4.25013020833333)
--(axis cs:28,4.22408854166667)
--(axis cs:27,4.19283854166667)
--(axis cs:26,4.125)
--(axis cs:25,4.046875)
--(axis cs:24,4.18255208333333)
--(axis cs:23,4.125)
--(axis cs:22,4.140625)
--(axis cs:21,4.104296875)
--(axis cs:20,4.07825520833333)
--(axis cs:19,3.96354166666667)
--(axis cs:18,4.00533854166667)
--(axis cs:17,3.932421875)
--(axis cs:16,3.994921875)
--(axis cs:15,3.95325520833333)
--(axis cs:14,3.98971354166667)
--(axis cs:13,4.03658854166667)
--(axis cs:12,4.11458333333333)
--(axis cs:11,4.24479166666667)
--(axis cs:10,4.359375)
--(axis cs:9,4.541796875)
--(axis cs:8,4.53658854166667)
--(axis cs:7,4.56276041666667)
--(axis cs:6,4.46354166666667)
--(axis cs:5,4.62513020833333)
--(axis cs:4,4.494921875)
--(axis cs:3,4.43763020833333)
--(axis cs:2,4.90638020833333)
--(axis cs:1,5.22916666666667)
--(axis cs:0,5.625)
--cycle;

\path [draw=color4, fill=color4, opacity=0.2]
(axis cs:0,5.64583333333333)
--(axis cs:0,5.43723958333333)
--(axis cs:1,6.90625)
--(axis cs:2,7.30729166666667)
--(axis cs:3,7.32278645833333)
--(axis cs:4,7.17708333333333)
--(axis cs:5,7.31223958333333)
--(axis cs:6,7.23372395833333)
--(axis cs:7,7.46875)
--(axis cs:8,7.06236979166667)
--(axis cs:9,6.88528645833333)
--(axis cs:10,6.90078125)
--(axis cs:11,7.01549479166667)
--(axis cs:12,6.99986979166667)
--(axis cs:13,7.05729166666667)
--(axis cs:14,7.161328125)
--(axis cs:15,7.22395833333333)
--(axis cs:16,7.0203125)
--(axis cs:17,7.07799479166667)
--(axis cs:18,7.23411458333333)
--(axis cs:19,7.30729166666667)
--(axis cs:20,7.26549479166667)
--(axis cs:21,7.24453125)
--(axis cs:22,7.26041666666667)
--(axis cs:23,7.328125)
--(axis cs:24,7.33854166666667)
--(axis cs:25,7.30729166666667)
--(axis cs:26,7.39049479166667)
--(axis cs:27,7.40091145833333)
--(axis cs:28,7.34361979166667)
--(axis cs:29,7.28645833333333)
--(axis cs:30,7.15104166666667)
--(axis cs:31,7.083203125)
--(axis cs:32,7.02591145833333)
--(axis cs:33,7.062109375)
--(axis cs:34,7.030859375)
--(axis cs:35,7.083203125)
--(axis cs:36,6.97916666666667)
--(axis cs:37,7.03619791666667)
--(axis cs:38,7.04674479166667)
--(axis cs:39,7.01549479166667)
--(axis cs:40,7.0625)
--(axis cs:41,7.01028645833333)
--(axis cs:42,6.989453125)
--(axis cs:43,7.01002604166667)
--(axis cs:44,7.06236979166667)
--(axis cs:45,7.0625)
--(axis cs:46,7.0625)
--(axis cs:47,7.072265625)
--(axis cs:48,7.03111979166667)
--(axis cs:49,7.015625)
--(axis cs:50,7.015625)
--(axis cs:50,7.96888020833333)
--(axis cs:50,7.96888020833333)
--(axis cs:49,7.94270833333333)
--(axis cs:48,7.96888020833333)
--(axis cs:47,7.94791666666667)
--(axis cs:46,7.953125)
--(axis cs:45,7.953125)
--(axis cs:44,7.979296875)
--(axis cs:43,7.9375)
--(axis cs:42,7.948046875)
--(axis cs:41,7.994921875)
--(axis cs:40,8.01041666666667)
--(axis cs:39,7.994921875)
--(axis cs:38,7.984375)
--(axis cs:37,7.92721354166667)
--(axis cs:36,7.93229166666667)
--(axis cs:35,7.95846354166667)
--(axis cs:34,7.94791666666667)
--(axis cs:33,7.96393229166667)
--(axis cs:32,7.91171875)
--(axis cs:31,7.99479166666667)
--(axis cs:30,8.06263020833333)
--(axis cs:29,8.14596354166667)
--(axis cs:28,8.24505208333333)
--(axis cs:27,8.24479166666667)
--(axis cs:26,8.22916666666667)
--(axis cs:25,8.21888020833333)
--(axis cs:24,8.171875)
--(axis cs:23,8.16666666666667)
--(axis cs:22,8.13541666666667)
--(axis cs:21,8.11979166666667)
--(axis cs:20,8.09427083333333)
--(axis cs:19,8.18229166666667)
--(axis cs:18,8.0625)
--(axis cs:17,7.95833333333333)
--(axis cs:16,7.89075520833333)
--(axis cs:15,8.041796875)
--(axis cs:14,8.03658854166667)
--(axis cs:13,7.94270833333333)
--(axis cs:12,7.89583333333333)
--(axis cs:11,7.85950520833333)
--(axis cs:10,7.8125)
--(axis cs:9,7.765625)
--(axis cs:8,7.91666666666667)
--(axis cs:7,8.23958333333333)
--(axis cs:6,8.078125)
--(axis cs:5,8.140625)
--(axis cs:4,8.03125)
--(axis cs:3,8.23958333333333)
--(axis cs:2,8.17721354166667)
--(axis cs:1,7.8125)
--(axis cs:0,5.64583333333333)
--cycle;

\path [draw=color5, fill=color5, opacity=0.2]
(axis cs:0,5.64583333333333)
--(axis cs:0,5.44765625)
--(axis cs:1,5.05208333333333)
--(axis cs:2,5.01549479166667)
--(axis cs:3,4.93697916666667)
--(axis cs:4,4.99466145833333)
--(axis cs:5,5.078125)
--(axis cs:6,5.21354166666667)
--(axis cs:7,5.4375)
--(axis cs:8,5.630078125)
--(axis cs:9,5.53125)
--(axis cs:10,5.5625)
--(axis cs:11,5.426953125)
--(axis cs:12,5.47395833333333)
--(axis cs:13,5.41653645833333)
--(axis cs:14,5.38541666666667)
--(axis cs:15,5.359375)
--(axis cs:16,5.38528645833333)
--(axis cs:17,5.37486979166667)
--(axis cs:18,5.45299479166667)
--(axis cs:19,5.38502604166667)
--(axis cs:20,5.489453125)
--(axis cs:21,5.56197916666667)
--(axis cs:22,5.54166666666667)
--(axis cs:23,5.58333333333333)
--(axis cs:24,5.692578125)
--(axis cs:25,5.74453125)
--(axis cs:26,5.71861979166667)
--(axis cs:27,5.83333333333333)
--(axis cs:28,5.64049479166667)
--(axis cs:29,5.54674479166667)
--(axis cs:30,5.53645833333333)
--(axis cs:31,5.55703125)
--(axis cs:32,5.56731770833333)
--(axis cs:33,5.55716145833333)
--(axis cs:34,5.64049479166667)
--(axis cs:35,5.72903645833333)
--(axis cs:36,5.755078125)
--(axis cs:37,5.74479166666667)
--(axis cs:38,5.78111979166667)
--(axis cs:39,5.786328125)
--(axis cs:40,5.770703125)
--(axis cs:41,5.78645833333333)
--(axis cs:42,5.83333333333333)
--(axis cs:43,5.86419270833333)
--(axis cs:44,5.86979166666667)
--(axis cs:45,5.89049479166667)
--(axis cs:46,5.921875)
--(axis cs:47,5.875)
--(axis cs:48,5.86979166666667)
--(axis cs:49,5.80208333333333)
--(axis cs:50,5.84361979166667)
--(axis cs:50,6.713671875)
--(axis cs:50,6.713671875)
--(axis cs:49,6.698046875)
--(axis cs:48,6.75520833333333)
--(axis cs:47,6.75520833333333)
--(axis cs:46,6.84895833333333)
--(axis cs:45,6.760546875)
--(axis cs:44,6.734375)
--(axis cs:43,6.75026041666667)
--(axis cs:42,6.76041666666667)
--(axis cs:41,6.68776041666667)
--(axis cs:40,6.635546875)
--(axis cs:39,6.72408854166667)
--(axis cs:38,6.67200520833333)
--(axis cs:37,6.70846354166667)
--(axis cs:36,6.73984375)
--(axis cs:35,6.75013020833333)
--(axis cs:34,6.64583333333333)
--(axis cs:33,6.63541666666667)
--(axis cs:32,6.61458333333333)
--(axis cs:31,6.55221354166667)
--(axis cs:30,6.54700520833333)
--(axis cs:29,6.55221354166667)
--(axis cs:28,6.635546875)
--(axis cs:27,6.760546875)
--(axis cs:26,6.67721354166667)
--(axis cs:25,6.73450520833333)
--(axis cs:24,6.61471354166667)
--(axis cs:23,6.59908854166667)
--(axis cs:22,6.53645833333333)
--(axis cs:21,6.546875)
--(axis cs:20,6.55221354166667)
--(axis cs:19,6.45325520833333)
--(axis cs:18,6.52096354166667)
--(axis cs:17,6.448046875)
--(axis cs:16,6.50520833333333)
--(axis cs:15,6.416796875)
--(axis cs:14,6.46354166666667)
--(axis cs:13,6.52096354166667)
--(axis cs:12,6.59908854166667)
--(axis cs:11,6.55221354166667)
--(axis cs:10,6.64583333333333)
--(axis cs:9,6.698046875)
--(axis cs:8,6.72916666666667)
--(axis cs:7,6.59375)
--(axis cs:6,6.36497395833333)
--(axis cs:5,6.20859375)
--(axis cs:4,6.06263020833333)
--(axis cs:3,6.07825520833333)
--(axis cs:2,6.14075520833333)
--(axis cs:1,6.01575520833333)
--(axis cs:0,5.64583333333333)
--cycle;

\addplot [line width=2pt, color0, forget plot]
table {%
0 5.54166666666667
1 7.58333333333333
2 7.98958333333333
3 8.47916666666667
4 8.59895833333333
5 8.77083333333333
6 8.82291666666667
7 8.8125
8 8.89583333333333
9 8.89583333333333
10 8.92708333333333
11 9.01041666666667
12 9.03645833333333
13 9
14 8.984375
15 9.02604166666667
16 9.02604166666667
17 9
18 8.83854166666667
19 8.82291666666667
20 8.61979166666667
21 8.61458333333333
22 8.671875
23 8.69270833333333
24 8.65104166666667
25 8.61979166666667
26 8.61979166666667
27 8.515625
28 8.56770833333333
29 8.61458333333333
30 8.59895833333333
31 8.65104166666667
32 8.69791666666667
33 8.71354166666667
34 8.71354166666667
35 8.6875
36 8.55729166666667
37 8.53645833333333
38 8.56770833333333
39 8.546875
40 8.58333333333333
41 8.58333333333333
42 8.61458333333333
43 8.61458333333333
44 8.609375
45 8.53645833333333
46 8.546875
47 8.546875
48 8.55729166666667
49 8.578125
50 8.578125
};
\addplot [line width=2pt, color1, forget plot]
table {%
0 5.54166666666667
1 4.96875
2 4.98958333333333
3 4.94791666666667
4 4.984375
5 5.125
6 5.25
7 5.25
8 5.33333333333333
9 5.27083333333333
10 5.10416666666667
11 4.99479166666667
12 4.88020833333333
13 4.86979166666667
14 4.890625
15 4.80208333333333
16 4.8125
17 4.75520833333333
18 4.60416666666667
19 4.55208333333333
20 4.44791666666667
21 4.40625
22 4.421875
23 4.28645833333333
24 4.11979166666667
25 4.20833333333333
26 4.25520833333333
27 4.36458333333333
28 4.453125
29 4.5
30 4.59375
31 4.671875
32 4.671875
33 4.58333333333333
34 4.578125
35 4.52083333333333
36 4.57291666666667
37 4.59375
38 4.625
39 4.56770833333333
40 4.65104166666667
41 4.63541666666667
42 4.69791666666667
43 4.69270833333333
44 4.70833333333333
45 4.72916666666667
46 4.6875
47 4.69791666666667
48 4.68229166666667
49 4.67708333333333
50 4.6875
};
\addplot [line width=2pt, color2, forget plot]
table {%
0 5.125
1 5.83333333333333
2 5.66666666666667
3 5.33333333333333
4 5.35416666666667
5 5.28125
6 5.1875
7 4.95833333333333
8 4.92708333333333
9 5.01041666666667
10 5.03125
11 4.92708333333333
12 5.03125
13 5.05208333333333
14 5.10416666666667
15 5.16666666666667
16 5.14583333333333
17 5.0625
18 5.10416666666667
19 5.10416666666667
20 5.0625
21 5.07291666666667
22 5.07291666666667
23 5.10416666666667
24 5.14583333333333
25 4.97916666666667
26 4.96875
27 4.875
28 4.90625
29 4.90625
30 4.92708333333333
31 5.08333333333333
32 5.125
33 5.15625
34 5.16666666666667
35 5.21875
36 5.34375
37 5.36458333333333
38 5.36458333333333
39 5.32291666666667
40 5.33333333333333
41 5.375
42 5.44791666666667
43 5.39583333333333
44 5.4375
45 5.47916666666667
46 5.48958333333333
47 5.53125
48 5.54166666666667
49 5.55208333333333
50 5.53125
};
\addplot [line width=2pt, color3, forget plot]
table {%
0 5.54166666666667
1 4.734375
2 4.39583333333333
3 3.92708333333333
4 3.97395833333333
5 4.13020833333333
6 3.984375
7 4.11979166666667
8 4.05729166666667
9 4.015625
10 3.90625
11 3.76041666666667
12 3.65104166666667
13 3.60416666666667
14 3.578125
15 3.53645833333333
16 3.58333333333333
17 3.515625
18 3.57291666666667
19 3.55208333333333
20 3.58333333333333
21 3.609375
22 3.6875
23 3.671875
24 3.6875
25 3.59895833333333
26 3.65104166666667
27 3.72395833333333
28 3.75520833333333
29 3.77083333333333
30 3.75
31 3.72916666666667
32 3.73958333333333
33 3.77083333333333
34 3.77604166666667
35 3.78645833333333
36 3.83333333333333
37 3.85416666666667
38 3.890625
39 3.84895833333333
40 3.69791666666667
41 3.6875
42 3.68229166666667
43 3.69270833333333
44 3.64583333333333
45 3.63020833333333
46 3.59375
47 3.578125
48 3.53125
49 3.5625
50 3.5625
};
\addplot [line width=2pt, color4, forget plot]
table {%
0 5.54166666666667
1 7.36458333333333
2 7.77604166666667
3 7.78125
4 7.61458333333333
5 7.734375
6 7.69270833333333
7 7.86458333333333
8 7.484375
9 7.34895833333333
10 7.375
11 7.44791666666667
12 7.45833333333333
13 7.50520833333333
14 7.59375
15 7.63541666666667
16 7.45833333333333
17 7.52604166666667
18 7.66145833333333
19 7.74479166666667
20 7.68229166666667
21 7.6875
22 7.72395833333333
23 7.765625
24 7.765625
25 7.77083333333333
26 7.828125
27 7.83333333333333
28 7.80729166666667
29 7.75520833333333
30 7.63020833333333
31 7.55729166666667
32 7.47916666666667
33 7.52083333333333
34 7.51041666666667
35 7.546875
36 7.48958333333333
37 7.50520833333333
38 7.52604166666667
39 7.53645833333333
40 7.55729166666667
41 7.52604166666667
42 7.5
43 7.5
44 7.51041666666667
45 7.52083333333333
46 7.51041666666667
47 7.49479166666667
48 7.5
49 7.5
50 7.5
};
\addplot [line width=2pt, color5, forget plot]
table {%
0 5.54166666666667
1 5.546875
2 5.57291666666667
3 5.47916666666667
4 5.52604166666667
5 5.64583333333333
6 5.796875
7 6.02604166666667
8 6.20833333333333
9 6.11458333333333
10 6.11979166666667
11 6.015625
12 6.04166666666667
13 5.95833333333333
14 5.94791666666667
15 5.86979166666667
16 5.921875
17 5.94270833333333
18 5.96354166666667
19 5.921875
20 6.03125
21 6.06770833333333
22 6.02083333333333
23 6.06770833333333
24 6.14583333333333
25 6.22395833333333
26 6.21875
27 6.30208333333333
28 6.125
29 6.06770833333333
30 6.04166666666667
31 6.03645833333333
32 6.09375
33 6.109375
34 6.125
35 6.234375
36 6.234375
37 6.25
38 6.21875
39 6.26041666666667
40 6.20833333333333
41 6.25520833333333
42 6.28645833333333
43 6.31770833333333
44 6.3125
45 6.31770833333333
46 6.36458333333333
47 6.30729166666667
48 6.32291666666667
49 6.26041666666667
50 6.28125
};
\end{axis}

\end{tikzpicture}

%% file: icml2023/figures/pd1_regrets_random_hard.tex
\begin{tikzpicture}

\colorlet{color0}{colorRandom}
\colorlet{color1}{colorPFN_GP}
\colorlet{color2}{colorHEBO}
\colorlet{color3}{colorPFN_GP_plus_prior}
\colorlet{color4}{colorRandom_plus_prior}
\colorlet{color5}{colorPFN_BNN}

\begin{axis}[
log basis y={10},
tick align=outside,
tick pos=left,
x grid style={white!69.0196078431373!black},
xmin=-2.5, xmax=52.5,
xtick style={color=black},
y grid style={white!69.0196078431373!black},
xlabel={Number of trials},
ylabel={Average Regret},
ymin=0.00114631042120513, ymax=0.08690370725498,
ymode=log,
ytick style={color=black},
ytick={0.01},
yticklabels={$10^{-2.0}$},
height=\benchmarkplotheight,
width=\benchmarkplotwidth,
]
\path [draw=color0, fill=color0, opacity=0.2]
(axis cs:0,0.673495035235962)
--(axis cs:0,0.559886212309632)
--(axis cs:1,0.235216831405857)
--(axis cs:2,0.103391822988696)
--(axis cs:3,0.0542187717418446)
--(axis cs:4,0.0386247566862232)
--(axis cs:5,0.0331901736700236)
--(axis cs:6,0.0253133496628412)
--(axis cs:7,0.0230093903264851)
--(axis cs:8,0.0203269420474078)
--(axis cs:9,0.0181869057572673)
--(axis cs:10,0.01739056902955)
--(axis cs:11,0.0165650341416436)
--(axis cs:12,0.0166864142852997)
--(axis cs:13,0.0161804286875806)
--(axis cs:14,0.0150736031119893)
--(axis cs:15,0.0141496532851444)
--(axis cs:16,0.013873491703869)
--(axis cs:17,0.0132908154253643)
--(axis cs:18,0.0113019204218084)
--(axis cs:19,0.00993710417432757)
--(axis cs:20,0.00967614907936259)
--(axis cs:21,0.00931352788737746)
--(axis cs:22,0.00934139413414793)
--(axis cs:23,0.00897427442976437)
--(axis cs:24,0.0089166877674983)
--(axis cs:25,0.00855649467131965)
--(axis cs:26,0.00852137065310012)
--(axis cs:27,0.00850395323773629)
--(axis cs:28,0.00822944711428493)
--(axis cs:29,0.00824708239047568)
--(axis cs:30,0.00815993752254144)
--(axis cs:31,0.00787689897115419)
--(axis cs:32,0.00776302140376182)
--(axis cs:33,0.00780435194116937)
--(axis cs:34,0.00775820045401207)
--(axis cs:35,0.00764858011844886)
--(axis cs:36,0.00739018612958134)
--(axis cs:37,0.00736729966857202)
--(axis cs:38,0.00714860094711255)
--(axis cs:39,0.00705866557289411)
--(axis cs:40,0.00716216911640614)
--(axis cs:41,0.00711646594470535)
--(axis cs:42,0.00708503404721064)
--(axis cs:43,0.00692576708204734)
--(axis cs:44,0.00684713479671625)
--(axis cs:45,0.00681387000265357)
--(axis cs:46,0.00676007451201208)
--(axis cs:47,0.00674257292765585)
--(axis cs:48,0.00676413737070968)
--(axis cs:49,0.00674009364064686)
--(axis cs:50,0.00665252723629905)
--(axis cs:50,0.00903082996497957)
--(axis cs:50,0.00903082996497957)
--(axis cs:49,0.00896383559219947)
--(axis cs:48,0.00906249734991468)
--(axis cs:47,0.00905886038500702)
--(axis cs:46,0.00901423399786536)
--(axis cs:45,0.00900895946150113)
--(axis cs:44,0.00918521452316226)
--(axis cs:43,0.00929070651359606)
--(axis cs:42,0.00942839944225858)
--(axis cs:41,0.00948646342877911)
--(axis cs:40,0.00958004949702709)
--(axis cs:39,0.00957532250076722)
--(axis cs:38,0.00959976637962774)
--(axis cs:37,0.00980122864603998)
--(axis cs:36,0.00994186680138904)
--(axis cs:35,0.0104323842281296)
--(axis cs:34,0.0104043777884155)
--(axis cs:33,0.0104241553318204)
--(axis cs:32,0.0104667767289112)
--(axis cs:31,0.0106272292270836)
--(axis cs:30,0.0110508308371491)
--(axis cs:29,0.0109931214653781)
--(axis cs:28,0.0109182553949519)
--(axis cs:27,0.0119414349869126)
--(axis cs:26,0.0120818230493636)
--(axis cs:25,0.0122028823857092)
--(axis cs:24,0.0128016445881655)
--(axis cs:23,0.0128257046799904)
--(axis cs:22,0.0139612829761104)
--(axis cs:21,0.0142447657568591)
--(axis cs:20,0.0144418743476492)
--(axis cs:19,0.014662191836033)
--(axis cs:18,0.019361887063693)
--(axis cs:17,0.0232882608518352)
--(axis cs:16,0.0236026840863854)
--(axis cs:15,0.0241141636350438)
--(axis cs:14,0.0254695969384603)
--(axis cs:13,0.0268013357399166)
--(axis cs:12,0.0269478278388253)
--(axis cs:11,0.0278318683296657)
--(axis cs:10,0.0292685163050833)
--(axis cs:9,0.0302152937767481)
--(axis cs:8,0.0334176878212915)
--(axis cs:7,0.0402020614203659)
--(axis cs:6,0.0491938925275309)
--(axis cs:5,0.0749473450243189)
--(axis cs:4,0.0865208110415428)
--(axis cs:3,0.105321317733127)
--(axis cs:2,0.204183065167772)
--(axis cs:1,0.361564456588356)
--(axis cs:0,0.673495035235962)
--cycle;

\path [draw=color1, fill=color1, opacity=0.2]
(axis cs:0,0.673133388861248)
--(axis cs:0,0.562200314924381)
--(axis cs:1,0.0572676687067969)
--(axis cs:2,0.0166637283983411)
--(axis cs:3,0.0103109310413269)
--(axis cs:4,0.00812959685559487)
--(axis cs:5,0.00735302966043988)
--(axis cs:6,0.00672733079483086)
--(axis cs:7,0.00599535288058855)
--(axis cs:8,0.00547180437471911)
--(axis cs:9,0.0051629187441521)
--(axis cs:10,0.00475088778632823)
--(axis cs:11,0.00450780803473302)
--(axis cs:12,0.00424719702890076)
--(axis cs:13,0.00426405239279834)
--(axis cs:14,0.0041788233759378)
--(axis cs:15,0.00396703242140815)
--(axis cs:16,0.00391455427298415)
--(axis cs:17,0.00371253493392206)
--(axis cs:18,0.00355742383361441)
--(axis cs:19,0.00349270234541445)
--(axis cs:20,0.00334382554992446)
--(axis cs:21,0.0033483442006022)
--(axis cs:22,0.00317341093927304)
--(axis cs:23,0.00302974034213415)
--(axis cs:24,0.0029307601581092)
--(axis cs:25,0.00292825771103961)
--(axis cs:26,0.00288851218575241)
--(axis cs:27,0.00285974874311079)
--(axis cs:28,0.00281536643257038)
--(axis cs:29,0.00277941588244471)
--(axis cs:30,0.00270606684058593)
--(axis cs:31,0.00265509811829236)
--(axis cs:32,0.00271094100705326)
--(axis cs:33,0.00259276795477467)
--(axis cs:34,0.00250481597018604)
--(axis cs:35,0.00247512364131373)
--(axis cs:36,0.00243565332046054)
--(axis cs:37,0.00247255256867196)
--(axis cs:38,0.00245588780914117)
--(axis cs:39,0.00240256921150867)
--(axis cs:40,0.00239177275049226)
--(axis cs:41,0.00235004662962289)
--(axis cs:42,0.00237503817986966)
--(axis cs:43,0.0023263541144648)
--(axis cs:44,0.00232618847911822)
--(axis cs:45,0.00230168614150801)
--(axis cs:46,0.00228786378774695)
--(axis cs:47,0.00224686020818983)
--(axis cs:48,0.00222078134480806)
--(axis cs:49,0.00226464508809501)
--(axis cs:50,0.0022577304844682)
--(axis cs:50,0.00309537188550804)
--(axis cs:50,0.00309537188550804)
--(axis cs:49,0.00307821893356231)
--(axis cs:48,0.00314027954550345)
--(axis cs:47,0.00313584463773007)
--(axis cs:46,0.00311526918621265)
--(axis cs:45,0.00318827978926256)
--(axis cs:44,0.0032100272735811)
--(axis cs:43,0.00319168415897024)
--(axis cs:42,0.00322352514154408)
--(axis cs:41,0.00325664070372131)
--(axis cs:40,0.00329232893576849)
--(axis cs:39,0.00331014924197519)
--(axis cs:38,0.00335049024027876)
--(axis cs:37,0.00340600472659939)
--(axis cs:36,0.00340028747880888)
--(axis cs:35,0.00340276553552608)
--(axis cs:34,0.00341249674094194)
--(axis cs:33,0.00350265149195175)
--(axis cs:32,0.00362860888622368)
--(axis cs:31,0.00367147824458072)
--(axis cs:30,0.00368057291397737)
--(axis cs:29,0.00371818902702212)
--(axis cs:28,0.00382653774393516)
--(axis cs:27,0.00383619764039062)
--(axis cs:26,0.00383319800243857)
--(axis cs:25,0.00391443555330965)
--(axis cs:24,0.00394711043667477)
--(axis cs:23,0.00408218644680583)
--(axis cs:22,0.0041520755723741)
--(axis cs:21,0.00444778575805944)
--(axis cs:20,0.00450044688306714)
--(axis cs:19,0.00467951868940455)
--(axis cs:18,0.00490032337272653)
--(axis cs:17,0.0050205142586125)
--(axis cs:16,0.00547334682210175)
--(axis cs:15,0.00563605479813705)
--(axis cs:14,0.00577226922654124)
--(axis cs:13,0.0058823219399086)
--(axis cs:12,0.00593449289356083)
--(axis cs:11,0.00625612637427748)
--(axis cs:10,0.00655394579707211)
--(axis cs:9,0.00702614979774143)
--(axis cs:8,0.0076172081205408)
--(axis cs:7,0.00838005480836784)
--(axis cs:6,0.00970178517026382)
--(axis cs:5,0.0102708711705919)
--(axis cs:4,0.0116151026007369)
--(axis cs:3,0.0168957600108235)
--(axis cs:2,0.0554729327315691)
--(axis cs:1,0.139966762900208)
--(axis cs:0,0.673133388861248)
--cycle;

\path [draw=color2, fill=color2, opacity=0.2]
(axis cs:0,0.674201155203568)
--(axis cs:0,0.556541481773906)
--(axis cs:1,0.122550886140755)
--(axis cs:2,0.0228066484444792)
--(axis cs:3,0.011200149174349)
--(axis cs:4,0.00921263742031249)
--(axis cs:5,0.00808389334432292)
--(axis cs:6,0.00706739439471354)
--(axis cs:7,0.00589035131317709)
--(axis cs:8,0.00553661253739584)
--(axis cs:9,0.00536754170416667)
--(axis cs:10,0.00510811668080729)
--(axis cs:11,0.00468711432677083)
--(axis cs:12,0.004530710771875)
--(axis cs:13,0.00451537608414062)
--(axis cs:14,0.00445578313697917)
--(axis cs:15,0.00431294545994791)
--(axis cs:16,0.00413278130791666)
--(axis cs:17,0.00394594246934896)
--(axis cs:18,0.00373068814294271)
--(axis cs:19,0.00369705369317708)
--(axis cs:20,0.00363417434643228)
--(axis cs:21,0.00354839419828125)
--(axis cs:22,0.00351699354453125)
--(axis cs:23,0.00346779640984375)
--(axis cs:24,0.0034913321754948)
--(axis cs:25,0.00340281710122395)
--(axis cs:26,0.00327082084041665)
--(axis cs:27,0.00313676566791666)
--(axis cs:28,0.00306001833408854)
--(axis cs:29,0.0030309462301302)
--(axis cs:30,0.0029604443113802)
--(axis cs:31,0.00295547401692708)
--(axis cs:32,0.00292659507760416)
--(axis cs:33,0.00285004430893229)
--(axis cs:34,0.00286754760195312)
--(axis cs:35,0.00284105805010416)
--(axis cs:36,0.00284345756223957)
--(axis cs:37,0.00289417562434895)
--(axis cs:38,0.00279821389984375)
--(axis cs:39,0.00272994326145833)
--(axis cs:40,0.00270478883596355)
--(axis cs:41,0.00271579197476562)
--(axis cs:42,0.00269330078994791)
--(axis cs:43,0.00263167462184895)
--(axis cs:44,0.00257207957885417)
--(axis cs:45,0.00259866351054688)
--(axis cs:46,0.00250917350276042)
--(axis cs:47,0.00254486649864583)
--(axis cs:48,0.00256827040143229)
--(axis cs:49,0.00254602155489583)
--(axis cs:50,0.00250006652666666)
--(axis cs:50,0.00358743918005209)
--(axis cs:50,0.00358743918005209)
--(axis cs:49,0.00359789350106771)
--(axis cs:48,0.00361979856554688)
--(axis cs:47,0.00361852063958333)
--(axis cs:46,0.00366079392684895)
--(axis cs:45,0.00368636241325521)
--(axis cs:44,0.00364269912817708)
--(axis cs:43,0.00372559576846354)
--(axis cs:42,0.00374047631460936)
--(axis cs:41,0.0038284780109375)
--(axis cs:40,0.003860151381875)
--(axis cs:39,0.00391407089447916)
--(axis cs:38,0.00418867547377604)
--(axis cs:37,0.00429434282007812)
--(axis cs:36,0.00432671121028646)
--(axis cs:35,0.00439131374328125)
--(axis cs:34,0.00438374734710937)
--(axis cs:33,0.00436397800828124)
--(axis cs:32,0.00440221136963541)
--(axis cs:31,0.00437411406080729)
--(axis cs:30,0.00434456120046874)
--(axis cs:29,0.00443727806216145)
--(axis cs:28,0.0046219679595052)
--(axis cs:27,0.00458627490177083)
--(axis cs:26,0.00477958092940105)
--(axis cs:25,0.00489160598466145)
--(axis cs:24,0.00507629255434896)
--(axis cs:23,0.00508006646249999)
--(axis cs:22,0.00510802421526041)
--(axis cs:21,0.00512663714619791)
--(axis cs:20,0.00509842102588541)
--(axis cs:19,0.00520461181843751)
--(axis cs:18,0.00534322068570313)
--(axis cs:17,0.00547681937473959)
--(axis cs:16,0.00576406889382812)
--(axis cs:15,0.00608998892213542)
--(axis cs:14,0.00624886428114583)
--(axis cs:13,0.00631212121674479)
--(axis cs:12,0.00638408346664062)
--(axis cs:11,0.00713836710242188)
--(axis cs:10,0.0077144025626823)
--(axis cs:9,0.00857449262223958)
--(axis cs:8,0.00862271699276041)
--(axis cs:7,0.0092149626896875)
--(axis cs:6,0.012308441010651)
--(axis cs:5,0.0140893216651042)
--(axis cs:4,0.0154335317566406)
--(axis cs:3,0.0331201015413542)
--(axis cs:2,0.0729615854329167)
--(axis cs:1,0.224276291576198)
--(axis cs:0,0.674201155203568)
--cycle;

\path [draw=color3, fill=color3, opacity=0.2]
(axis cs:0,0.671915999549563)
--(axis cs:0,0.56445996196929)
--(axis cs:1,0.0505108214137491)
--(axis cs:2,0.0130241873689804)
--(axis cs:3,0.00666903252655973)
--(axis cs:4,0.00559133642566756)
--(axis cs:5,0.00514707862657717)
--(axis cs:6,0.00467642932090484)
--(axis cs:7,0.00433905001314534)
--(axis cs:8,0.003912831700002)
--(axis cs:9,0.00364701057195239)
--(axis cs:10,0.00342893115717402)
--(axis cs:11,0.00315756364959278)
--(axis cs:12,0.00300449618544457)
--(axis cs:13,0.00292829353477223)
--(axis cs:14,0.00283852633031318)
--(axis cs:15,0.00270681337868422)
--(axis cs:16,0.00273237114414976)
--(axis cs:17,0.00259395434533318)
--(axis cs:18,0.00259798957596442)
--(axis cs:19,0.00251654872715947)
--(axis cs:20,0.00246947540705325)
--(axis cs:21,0.00232842407255076)
--(axis cs:22,0.00230925423558239)
--(axis cs:23,0.00226243750765647)
--(axis cs:24,0.00225606240052085)
--(axis cs:25,0.002159713233323)
--(axis cs:26,0.00214543118057185)
--(axis cs:27,0.00213175208565346)
--(axis cs:28,0.00212184562147666)
--(axis cs:29,0.00207360242373098)
--(axis cs:30,0.00205406704084795)
--(axis cs:31,0.00206919051886935)
--(axis cs:32,0.0019765765982384)
--(axis cs:33,0.00201580528755347)
--(axis cs:34,0.00200187048400523)
--(axis cs:35,0.00201777665536172)
--(axis cs:36,0.0019755810981253)
--(axis cs:37,0.00198867342503098)
--(axis cs:38,0.0019620807649604)
--(axis cs:39,0.00190577227873474)
--(axis cs:40,0.00187698612620854)
--(axis cs:41,0.00178396691123072)
--(axis cs:42,0.00174560420822479)
--(axis cs:43,0.00174350135600978)
--(axis cs:44,0.00168207192988643)
--(axis cs:45,0.0016359818000809)
--(axis cs:46,0.00164134256188424)
--(axis cs:47,0.00157303585159712)
--(axis cs:48,0.00155821247992811)
--(axis cs:49,0.00154474189663367)
--(axis cs:50,0.00162285117479276)
--(axis cs:50,0.00243988243226035)
--(axis cs:50,0.00243988243226035)
--(axis cs:49,0.00240800425385309)
--(axis cs:48,0.00246133449664735)
--(axis cs:47,0.00244955015326956)
--(axis cs:46,0.00249860641203576)
--(axis cs:45,0.00254165620420289)
--(axis cs:44,0.00258774623094696)
--(axis cs:43,0.0026152821709671)
--(axis cs:42,0.00261621019439494)
--(axis cs:41,0.00267315623304991)
--(axis cs:40,0.00278143443582067)
--(axis cs:39,0.00299467577018156)
--(axis cs:38,0.00303578885704517)
--(axis cs:37,0.00299481460214312)
--(axis cs:36,0.00303979290218384)
--(axis cs:35,0.00305620276483151)
--(axis cs:34,0.00303100934242856)
--(axis cs:33,0.00302791555169762)
--(axis cs:32,0.00301796600933002)
--(axis cs:31,0.00317876792700815)
--(axis cs:30,0.00320392395633625)
--(axis cs:29,0.00320773065356525)
--(axis cs:28,0.00322502964236495)
--(axis cs:27,0.00325925931172799)
--(axis cs:26,0.00325123962552358)
--(axis cs:25,0.00332142336753219)
--(axis cs:24,0.00342617328607441)
--(axis cs:23,0.00344538104283307)
--(axis cs:22,0.00355653594208883)
--(axis cs:21,0.00346166172256413)
--(axis cs:20,0.00360485172482498)
--(axis cs:19,0.0036791237096598)
--(axis cs:18,0.00380985070277415)
--(axis cs:17,0.00380501471282941)
--(axis cs:16,0.00391907688928955)
--(axis cs:15,0.00391381361147889)
--(axis cs:14,0.00406536091222669)
--(axis cs:13,0.00414886716302937)
--(axis cs:12,0.00429437574965851)
--(axis cs:11,0.00442154128860095)
--(axis cs:10,0.0048671980121181)
--(axis cs:9,0.00515900786636538)
--(axis cs:8,0.00556466717177971)
--(axis cs:7,0.00620742728764117)
--(axis cs:6,0.00668983592939275)
--(axis cs:5,0.00769342366126262)
--(axis cs:4,0.00820747275508643)
--(axis cs:3,0.0101316014732186)
--(axis cs:2,0.024690100431556)
--(axis cs:1,0.115448919636314)
--(axis cs:0,0.671915999549563)
--cycle;

\path [draw=color4, fill=color4, opacity=0.2]
(axis cs:0,0.675086396894493)
--(axis cs:0,0.56325345790905)
--(axis cs:1,0.223995994467559)
--(axis cs:2,0.0884288538885094)
--(axis cs:3,0.0510166674007487)
--(axis cs:4,0.0225057066785451)
--(axis cs:5,0.0195309785319675)
--(axis cs:6,0.0142074550664206)
--(axis cs:7,0.0123120083148853)
--(axis cs:8,0.0106881084376054)
--(axis cs:9,0.00928027110594276)
--(axis cs:10,0.00890787011646588)
--(axis cs:11,0.0084103380427099)
--(axis cs:12,0.00808307099918899)
--(axis cs:13,0.00763398996692938)
--(axis cs:14,0.00762652485444314)
--(axis cs:15,0.00703321717409254)
--(axis cs:16,0.00688675755044605)
--(axis cs:17,0.00670684651523811)
--(axis cs:18,0.00650365600679213)
--(axis cs:19,0.00649156875472226)
--(axis cs:20,0.00620088473138614)
--(axis cs:21,0.00613034129859757)
--(axis cs:22,0.00618539635450323)
--(axis cs:23,0.00597454642929037)
--(axis cs:24,0.00593717703409156)
--(axis cs:25,0.00584847502918299)
--(axis cs:26,0.00579355049857009)
--(axis cs:27,0.00556485781120161)
--(axis cs:28,0.00552314251963008)
--(axis cs:29,0.00541376108340641)
--(axis cs:30,0.00528736873724494)
--(axis cs:31,0.00526574694549807)
--(axis cs:32,0.00514132357299168)
--(axis cs:33,0.00501762038272446)
--(axis cs:34,0.00508782831952449)
--(axis cs:35,0.00501821157078059)
--(axis cs:36,0.00482524785758074)
--(axis cs:37,0.00478407063226177)
--(axis cs:38,0.00474966803461708)
--(axis cs:39,0.00477176851369136)
--(axis cs:40,0.00469338952375143)
--(axis cs:41,0.00454437425197118)
--(axis cs:42,0.00435387563873082)
--(axis cs:43,0.00440044728985325)
--(axis cs:44,0.00437713083670149)
--(axis cs:45,0.00426852715495984)
--(axis cs:46,0.0042397645245013)
--(axis cs:47,0.00422441377863931)
--(axis cs:48,0.00419993617297246)
--(axis cs:49,0.00423562371824944)
--(axis cs:50,0.00421989823980279)
--(axis cs:50,0.00567948592995794)
--(axis cs:50,0.00567948592995794)
--(axis cs:49,0.0056917100885211)
--(axis cs:48,0.00566611649993093)
--(axis cs:47,0.00566877019020921)
--(axis cs:46,0.00580088975219008)
--(axis cs:45,0.0059530378837293)
--(axis cs:44,0.00613910447128962)
--(axis cs:43,0.00616480652962785)
--(axis cs:42,0.00617211343280159)
--(axis cs:41,0.00634489293711258)
--(axis cs:40,0.006432831710253)
--(axis cs:39,0.0064860627550267)
--(axis cs:38,0.00663829610718535)
--(axis cs:37,0.00665422402711124)
--(axis cs:36,0.00668018648017187)
--(axis cs:35,0.00704550611103949)
--(axis cs:34,0.00711366996080861)
--(axis cs:33,0.00712387244889906)
--(axis cs:32,0.00710634870080739)
--(axis cs:31,0.00723216886430024)
--(axis cs:30,0.00738639358028216)
--(axis cs:29,0.0074114572108813)
--(axis cs:28,0.00770053938703958)
--(axis cs:27,0.00771526217475608)
--(axis cs:26,0.00800727939451134)
--(axis cs:25,0.00811634052664551)
--(axis cs:24,0.00815048792276853)
--(axis cs:23,0.00814496407458282)
--(axis cs:22,0.00842295775031116)
--(axis cs:21,0.00842683868837923)
--(axis cs:20,0.00859620346592244)
--(axis cs:19,0.00887493072277875)
--(axis cs:18,0.00886489825467915)
--(axis cs:17,0.00910624306664547)
--(axis cs:16,0.0092626200138281)
--(axis cs:15,0.00943332482400438)
--(axis cs:14,0.0104631846336058)
--(axis cs:13,0.0105626551271631)
--(axis cs:12,0.0123607732633502)
--(axis cs:11,0.0129711552677634)
--(axis cs:10,0.0132480503004085)
--(axis cs:9,0.0142036904042563)
--(axis cs:8,0.0163619709133955)
--(axis cs:7,0.0244284351352663)
--(axis cs:6,0.027253206601286)
--(axis cs:5,0.0565958741450785)
--(axis cs:4,0.0743428774252628)
--(axis cs:3,0.124491357042274)
--(axis cs:2,0.18193151363887)
--(axis cs:1,0.359946878654283)
--(axis cs:0,0.675086396894493)
--cycle;

\path [draw=color5, fill=color5, opacity=0.2]
(axis cs:0,0.676687904452012)
--(axis cs:0,0.562171750369561)
--(axis cs:1,0.0730271724152433)
--(axis cs:2,0.0221187847445557)
--(axis cs:3,0.0137967572760469)
--(axis cs:4,0.0107564743157913)
--(axis cs:5,0.0100181518849967)
--(axis cs:6,0.00910911201572929)
--(axis cs:7,0.0084502917903979)
--(axis cs:8,0.00840049169136009)
--(axis cs:9,0.00725713964620075)
--(axis cs:10,0.00700549555862087)
--(axis cs:11,0.00637150109817308)
--(axis cs:12,0.00631138568757657)
--(axis cs:13,0.00603122830733083)
--(axis cs:14,0.00596820995025974)
--(axis cs:15,0.00552131937733078)
--(axis cs:16,0.00523084289521606)
--(axis cs:17,0.00516276227108368)
--(axis cs:18,0.00506022910698792)
--(axis cs:19,0.00509658058804007)
--(axis cs:20,0.00495651153364286)
--(axis cs:21,0.00490875109142807)
--(axis cs:22,0.00491837085158842)
--(axis cs:23,0.00487482413163206)
--(axis cs:24,0.00490460071459449)
--(axis cs:25,0.00474403688567388)
--(axis cs:26,0.00472319324552198)
--(axis cs:27,0.00467430230594876)
--(axis cs:28,0.00461845492016681)
--(axis cs:29,0.00466629634316016)
--(axis cs:30,0.00446243998119362)
--(axis cs:31,0.00446152842308222)
--(axis cs:32,0.00435777765224766)
--(axis cs:33,0.00430715524313563)
--(axis cs:34,0.00433522436818063)
--(axis cs:35,0.00425431114161961)
--(axis cs:36,0.00413484741068277)
--(axis cs:37,0.00397238964166047)
--(axis cs:38,0.00382986379428463)
--(axis cs:39,0.00381125528286165)
--(axis cs:40,0.00370562598401541)
--(axis cs:41,0.00374527051755007)
--(axis cs:42,0.00371327933114715)
--(axis cs:43,0.0036561985009783)
--(axis cs:44,0.0036852461652067)
--(axis cs:45,0.00368855902656263)
--(axis cs:46,0.00366434280272349)
--(axis cs:47,0.0036349900630074)
--(axis cs:48,0.00361073434708954)
--(axis cs:49,0.00354154844103162)
--(axis cs:50,0.00361455280524592)
--(axis cs:50,0.00565794979776238)
--(axis cs:50,0.00565794979776238)
--(axis cs:49,0.00583268319214297)
--(axis cs:48,0.00593051363185137)
--(axis cs:47,0.00585789728308)
--(axis cs:46,0.00587037172783488)
--(axis cs:45,0.00591291503733367)
--(axis cs:44,0.00586929900675755)
--(axis cs:43,0.00596178183404496)
--(axis cs:42,0.00600737149984206)
--(axis cs:41,0.00602471497646356)
--(axis cs:40,0.00595388863135978)
--(axis cs:39,0.00613153270809292)
--(axis cs:38,0.00614166818852079)
--(axis cs:37,0.00631080500234551)
--(axis cs:36,0.00646372664274916)
--(axis cs:35,0.00693103268338984)
--(axis cs:34,0.00698875627269678)
--(axis cs:33,0.00709429808788573)
--(axis cs:32,0.00703923054966198)
--(axis cs:31,0.00698026162030483)
--(axis cs:30,0.00717024938748854)
--(axis cs:29,0.00726832125700169)
--(axis cs:28,0.00726721867077646)
--(axis cs:27,0.00739211483714884)
--(axis cs:26,0.00734780199900052)
--(axis cs:25,0.00751700494964091)
--(axis cs:24,0.0075654793099501)
--(axis cs:23,0.00757984084613226)
--(axis cs:22,0.00751796043606978)
--(axis cs:21,0.00754508447846466)
--(axis cs:20,0.00777397178413637)
--(axis cs:19,0.00788149024458994)
--(axis cs:18,0.00793667374443708)
--(axis cs:17,0.00791710328965755)
--(axis cs:16,0.00820446478176709)
--(axis cs:15,0.0089251061409357)
--(axis cs:14,0.0094489607239155)
--(axis cs:13,0.00950111762439935)
--(axis cs:12,0.00989107541979849)
--(axis cs:11,0.00993211866308675)
--(axis cs:10,0.0107308843317162)
--(axis cs:9,0.0113071061096552)
--(axis cs:8,0.0127647199769248)
--(axis cs:7,0.0137477934201672)
--(axis cs:6,0.0145622315902213)
--(axis cs:5,0.0160907001983056)
--(axis cs:4,0.0241454203709851)
--(axis cs:3,0.0318663192011502)
--(axis cs:2,0.0451404465848749)
--(axis cs:1,0.161357623631398)
--(axis cs:0,0.676687904452012)
--cycle;

\addplot [line width=2.0, color0, forget plot]
table {%
0 0.619448229526432
1 0.298291483337594
2 0.150043195275037
3 0.075836841585959
4 0.060166218421335
5 0.0519364073663701
6 0.0358189442451794
7 0.0309498237392306
8 0.026117945891718
9 0.0238322757130861
10 0.0225644150695204
11 0.0218878818622231
12 0.0215881791533033
13 0.0212921115516126
14 0.0200893512873351
15 0.0188989254592359
16 0.0181032978255053
17 0.0177345474817852
18 0.0149904548757771
19 0.0121431136760612
20 0.0118507038437823
21 0.0115641275056699
22 0.0115090677150587
23 0.0107373949397106
24 0.010684269195646
25 0.0102734988191227
26 0.0101697575557728
27 0.0100446213293572
28 0.00956501938487087
29 0.00955043611511583
30 0.00952052513842776
31 0.00913385180220003
32 0.00910468526268993
33 0.00906667302136614
34 0.00906667302136614
35 0.00897634528661762
36 0.0086509290718535
37 0.00854760504101708
38 0.00838718845282986
39 0.00831843822037175
40 0.00831843822037175
41 0.00831843822037175
42 0.00820405830403164
43 0.00808964486007884
44 0.00798053622951105
45 0.00789511966913933
46 0.00789303599009906
47 0.00788053577850257
48 0.00782667428960397
49 0.00782667428960397
50 0.00782667428960397
};
\addplot [line width=2.0, color1, forget plot]
table {%
0 0.619448229526432
1 0.0972843614983822
2 0.0335459104452654
3 0.0132822175767024
4 0.00966842832128201
5 0.00870136531154309
6 0.00811771290461217
7 0.00710862613677973
8 0.00647576078236098
9 0.00606477992693578
10 0.00560851692616934
11 0.00528286819120243
12 0.0050888538306951
13 0.00499761712511375
14 0.00489704422116274
15 0.00473500278552368
16 0.00464539713621134
17 0.00436838579138115
18 0.00420788910945251
19 0.0040855307078361
20 0.00390537908385192
21 0.00385932052592431
22 0.00365103131865456
23 0.003556432228138
24 0.00342279194007312
25 0.00341011105189715
26 0.00337644948398068
27 0.00334016795044139
28 0.00332499841312562
29 0.00325692463129753
30 0.0031877667899926
31 0.0031778058293958
32 0.00315392811556652
33 0.0030541221557557
34 0.00296310865715141
35 0.00293123972997064
36 0.00292165797288214
37 0.0029154761640727
38 0.00290768534029518
39 0.00286333917414143
40 0.00285229679281507
41 0.00279292249515647
42 0.00277200017829731
43 0.00275655869478975
44 0.00275655869478975
45 0.00274257022971903
46 0.0027061114344497
47 0.00267277815774078
48 0.00266771859263375
49 0.0026611713949342
50 0.0026611713949342
};
\addplot [line width=2.0, color2, forget plot]
table {%
0 0.619448229641667
1 0.1748094187625
2 0.0438646572458333
3 0.0201230551572917
4 0.0119627768885417
5 0.0106718377989583
6 0.009409599646875
7 0.0074468211375
8 0.00693795791979166
9 0.00675998137916667
10 0.00632916745625
11 0.00577226424166666
12 0.005420706175
13 0.00530916712604166
14 0.00530429696770833
15 0.005189490065625
16 0.00487821454479166
17 0.004678382703125
18 0.00451380790104167
19 0.00441228322395833
20 0.00433594592291666
21 0.00427896942708333
22 0.004260137
23 0.00424195045729166
24 0.00423573511458333
25 0.00406618846562499
26 0.00399527989687499
27 0.00387146427083333
28 0.00379280680729166
29 0.00369855900208333
30 0.00360638243541666
31 0.00360300484583333
32 0.00359423350624999
33 0.00357110839791666
34 0.00355763624166666
35 0.003533178778125
36 0.00353281783125
37 0.00352968315416666
38 0.00345706410625
39 0.00330498077291666
40 0.00326021090833333
41 0.00326021090833333
42 0.00322062757499999
43 0.00316594728333333
44 0.00312547109270833
45 0.003111482996875
46 0.003089607996875
47 0.003055232996875
48 0.00305314966354167
49 0.00304689966354166
50 0.00304481633020833
};
\addplot [line width=2.0, color3, forget plot]
table {%
0 0.619448229526432
1 0.0797764482233424
2 0.0182465527773896
3 0.00824908216257885
4 0.00687201075340305
5 0.00634890254050488
6 0.0056685043445229
7 0.00525776997913912
8 0.00468495288948213
9 0.00433663855701679
10 0.00415525293062125
11 0.00376175827741618
12 0.00363745201463495
13 0.00351524991999064
14 0.00342552940756077
15 0.00332351295361911
16 0.00328376783142482
17 0.00320090775152042
18 0.00316841483821466
19 0.00307225391974046
20 0.00299668656125659
21 0.00290025866364434
22 0.00288254987532888
23 0.00281900884047145
24 0.00279572763875121
25 0.00269078559016183
26 0.00267679650420938
27 0.00266012955541407
28 0.00264509987170492
29 0.00262426742747421
30 0.00261087500885124
31 0.00259004318550224
32 0.00251013384595508
33 0.00250395141626392
34 0.00249720243200654
35 0.00249681531225636
36 0.00249384004707132
37 0.00248536314899717
38 0.00247447474633649
39 0.00241831226979687
40 0.00230468036070461
41 0.00221926442121461
42 0.00216676204239323
43 0.0021625953051944
44 0.00210887537732715
45 0.00208387495413417
46 0.00203283350999151
47 0.00201825024023647
48 0.00199950023328259
49 0.00199950023328259
50 0.00199801197980836
};
\addplot [line width=2.0, color4, forget plot]
table {%
0 0.619448229526432
1 0.28980446628845
2 0.134330872200342
3 0.0849720227866495
4 0.0454799914989372
5 0.033725390469258
6 0.0199345493325094
7 0.0175150546191632
8 0.0132548076667884
9 0.0115051234087347
10 0.0109323296021421
11 0.0104918773542841
12 0.0101319068989157
13 0.00906606036633248
14 0.00895912032037968
15 0.00819253411849334
16 0.00800825828770791
17 0.007887424771587
18 0.00768034023006752
19 0.00761617334643994
20 0.00738786644697184
21 0.00724471161226426
22 0.00719588019659116
23 0.00707151650225117
24 0.00699234911635513
25 0.00691362876529489
26 0.00683947499925886
27 0.00662756993209317
28 0.00655986278091862
29 0.00639153321047619
30 0.00632278235713635
31 0.00622666490038231
32 0.00607549696624274
33 0.00605987247784927
34 0.00604456277648602
35 0.00600724716444805
36 0.00571849618812397
37 0.00567641282538568
38 0.00567432976722712
39 0.00558809892048433
40 0.00547372257421409
41 0.00541122275799507
42 0.00525101354430114
43 0.00524684680710231
44 0.00522601312110816
45 0.00510518022586897
46 0.00498851344694688
47 0.00495101343303913
48 0.00492601363072787
49 0.00492497241208945
50 0.00492497241208945
};
\addplot [line width=2.0, color5, forget plot]
table {%
0 0.619448229526432
1 0.115009292830626
2 0.0322063484796384
3 0.0225467865301668
4 0.0164518738422294
5 0.0127958338964482
6 0.0115756830457349
7 0.010899411540081
8 0.010505641991496
9 0.00913234243874743
10 0.00873238518297667
11 0.00805541549583271
12 0.00795095556189611
13 0.00768240761096273
14 0.00754748628869647
15 0.00717330742090935
16 0.00658315376152589
17 0.00644560828268523
18 0.00636092063744857
19 0.00631675359567001
20 0.00625136233329767
21 0.00616419674913083
22 0.00611944918294742
23 0.00610551908075804
24 0.00609097011471783
25 0.00601854845345015
26 0.00596518398036554
27 0.00593694193373118
28 0.00585222293396785
29 0.00581334270000452
30 0.00571449429502084
31 0.00559488577852641
32 0.0055747061914205
33 0.00553303944031392
34 0.00551635898734127
35 0.0054858586386839
36 0.00518547761648888
37 0.0050225144819418
38 0.00497251301467413
39 0.00485182882060602
40 0.00473920832782978
41 0.00473432695577537
42 0.00472766054878627
43 0.00471828523486847
44 0.00469682694186762
45 0.00469113236998513
46 0.00468164948830996
47 0.00466393588816121
48 0.00465579535275692
49 0.00462281784127151
50 0.00455615128785366
};

\end{axis}

\end{tikzpicture}

%% file: icml2023/figures/bayesmark_rank_plot.tex
\begin{tikzpicture}

\colorlet{color0}{colorHEBO}
\colorlet{color1}{colorHyperopt}
\colorlet{color2}{colorPFN_BNN}
\colorlet{color3}{colorPFN_GP_init_at_min}
\colorlet{color4}{colorPFN_GP_init_at_mid}
\colorlet{color5}{colorPFN_GP}
\colorlet{color6}{colorPySOT}
\colorlet{color7}{colorRandom}
\begin{axis}[
legend cell align={left},
legend style={
  fill opacity=0.8,
  draw opacity=1,
  text opacity=1,
  at={(0.03,0.97)},
  anchor=north west,
  draw=white!80!black
},
tick align=outside,
tick pos=left,
x grid style={white!69.0196078431373!black},
xmin=-2.45, xmax=51.45,
xtick style={color=black},
y grid style={white!69.0196078431373!black},
ytick style={color=black},
height=\benchmarkplotheight,
width=\benchmarkplotwidth,
xlabel={Number of trials},
ylabel={Average Rank},
]

\path [fill=color0, fill opacity=0.2]
(axis cs:0,5.15074319203299)
--(axis cs:0,5.01314569685589)
--(axis cs:1,4.31760404411776)
--(axis cs:2,4.67559269623225)
--(axis cs:3,4.23489036657659)
--(axis cs:4,4.30409048708115)
--(axis cs:5,4.21145491296395)
--(axis cs:6,4.34703056684977)
--(axis cs:7,4.35079676355605)
--(axis cs:8,4.36277251113799)
--(axis cs:9,4.34415536526323)
--(axis cs:10,4.32398423007313)
--(axis cs:11,4.26952479743627)
--(axis cs:12,4.25906263457621)
--(axis cs:13,4.28965373311316)
--(axis cs:14,4.27961512453412)
--(axis cs:15,4.2315711043525)
--(axis cs:16,4.28545867282874)
--(axis cs:17,4.24760300885969)
--(axis cs:18,4.19428635469086)
--(axis cs:19,4.15581355627104)
--(axis cs:20,4.20217355868938)
--(axis cs:21,4.21471608166369)
--(axis cs:22,4.23239894248282)
--(axis cs:23,4.23133985792572)
--(axis cs:24,4.23487017304734)
--(axis cs:25,4.20497958524704)
--(axis cs:26,4.1830838038868)
--(axis cs:27,4.15238737366004)
--(axis cs:28,4.08413961443011)
--(axis cs:29,4.09848961023579)
--(axis cs:30,4.06720255482889)
--(axis cs:31,4.05990894148531)
--(axis cs:32,4.03598312773329)
--(axis cs:33,3.97584676722628)
--(axis cs:34,3.94308214471952)
--(axis cs:35,3.93677639506911)
--(axis cs:36,3.89428586944162)
--(axis cs:37,3.86713844955065)
--(axis cs:38,3.87199707406185)
--(axis cs:39,3.86701325338956)
--(axis cs:40,3.85119296356363)
--(axis cs:41,3.87822126706684)
--(axis cs:42,3.87213118706536)
--(axis cs:43,3.85987429731785)
--(axis cs:44,3.85432169824861)
--(axis cs:45,3.84464072576992)
--(axis cs:46,3.84864800189447)
--(axis cs:47,3.84781286336204)
--(axis cs:48,3.86153057541033)
--(axis cs:49,3.86747884510669)
--(axis cs:49,4.03529893267109)
--(axis cs:49,4.03529893267109)
--(axis cs:48,4.03291386903411)
--(axis cs:47,4.02440935886018)
--(axis cs:46,4.03746310921665)
--(axis cs:45,4.02480371867453)
--(axis cs:44,4.02623385730695)
--(axis cs:43,4.04012570268215)
--(axis cs:42,4.06675770182353)
--(axis cs:41,4.07177873293316)
--(axis cs:40,4.04325148088081)
--(axis cs:39,4.06632007994377)
--(axis cs:38,4.09744737038259)
--(axis cs:37,4.10230599489379)
--(axis cs:36,4.09738079722505)
--(axis cs:35,4.149334716042)
--(axis cs:34,4.15691785528048)
--(axis cs:33,4.18248656610705)
--(axis cs:32,4.24179465004448)
--(axis cs:31,4.29842439184802)
--(axis cs:30,4.29946411183778)
--(axis cs:29,4.33484372309754)
--(axis cs:28,4.31863816334767)
--(axis cs:27,4.39483484856218)
--(axis cs:26,4.4169161961132)
--(axis cs:25,4.42279819253074)
--(axis cs:24,4.44846316028599)
--(axis cs:23,4.42143791985206)
--(axis cs:22,4.42315661307273)
--(axis cs:21,4.40472836278075)
--(axis cs:20,4.38671533019951)
--(axis cs:19,4.33585311039563)
--(axis cs:18,4.36404697864248)
--(axis cs:17,4.40517476891809)
--(axis cs:16,4.43398577161571)
--(axis cs:15,4.39898445120305)
--(axis cs:14,4.45371820879921)
--(axis cs:13,4.41867960022017)
--(axis cs:12,4.37982625431267)
--(axis cs:11,4.40547520256373)
--(axis cs:10,4.47323799214909)
--(axis cs:9,4.47528907918122)
--(axis cs:8,4.4761163777509)
--(axis cs:7,4.48253656977729)
--(axis cs:6,4.45019165537245)
--(axis cs:5,4.33854508703605)
--(axis cs:4,4.4403539573633)
--(axis cs:3,4.39566518897897)
--(axis cs:2,4.79107397043442)
--(axis cs:1,4.44628484477113)
--(axis cs:0,5.15074319203299)
--cycle;

\path [fill=color1, fill opacity=0.2]
(axis cs:0,4.02144002530418)
--(axis cs:0,3.85633775247359)
--(axis cs:1,4.31822050689924)
--(axis cs:2,4.63491555378361)
--(axis cs:3,4.79021995020901)
--(axis cs:4,4.87540900928861)
--(axis cs:5,4.98381533950378)
--(axis cs:6,5.10310758908419)
--(axis cs:7,5.20407311800973)
--(axis cs:8,5.26539042561132)
--(axis cs:9,5.28574764673408)
--(axis cs:10,5.27972781812729)
--(axis cs:11,5.27864333612258)
--(axis cs:12,5.26384997756812)
--(axis cs:13,5.31805042191738)
--(axis cs:14,5.33569032444239)
--(axis cs:15,5.36994816498554)
--(axis cs:16,5.38016978994429)
--(axis cs:17,5.41017230585363)
--(axis cs:18,5.42807519654049)
--(axis cs:19,5.4368880312274)
--(axis cs:20,5.2786191057786)
--(axis cs:21,5.1883177054503)
--(axis cs:22,5.09442548022594)
--(axis cs:23,5.04105273072011)
--(axis cs:24,5.02701571243317)
--(axis cs:25,5.03473473439175)
--(axis cs:26,5.03950171158426)
--(axis cs:27,4.98701175141237)
--(axis cs:28,4.93997303245721)
--(axis cs:29,4.95374971742323)
--(axis cs:30,4.93838105758551)
--(axis cs:31,4.95653035641451)
--(axis cs:32,4.92635451475747)
--(axis cs:33,4.96850878089078)
--(axis cs:34,5.00817933232308)
--(axis cs:35,5.03040728429634)
--(axis cs:36,5.04353642284461)
--(axis cs:37,5.01439090708935)
--(axis cs:38,5.06502722903118)
--(axis cs:39,5.06900197916724)
--(axis cs:40,5.11141150775025)
--(axis cs:41,5.12857338295175)
--(axis cs:42,5.12735920712675)
--(axis cs:43,5.14657298825936)
--(axis cs:44,5.13213535107829)
--(axis cs:45,5.13425206233691)
--(axis cs:46,5.14122101690133)
--(axis cs:47,5.16998234130399)
--(axis cs:48,5.18773281905329)
--(axis cs:49,5.19298944949428)
--(axis cs:49,5.3097883282835)
--(axis cs:49,5.3097883282835)
--(axis cs:48,5.28726718094671)
--(axis cs:47,5.26335099202934)
--(axis cs:46,5.242112316432)
--(axis cs:45,5.22408127099642)
--(axis cs:44,5.19842020447726)
--(axis cs:43,5.2089825672962)
--(axis cs:42,5.2170852373177)
--(axis cs:41,5.22420439482603)
--(axis cs:40,5.20525515891642)
--(axis cs:39,5.17266468749943)
--(axis cs:38,5.17386165985771)
--(axis cs:37,5.12449798179954)
--(axis cs:36,5.13979691048872)
--(axis cs:35,5.13625938237033)
--(axis cs:34,5.13348733434359)
--(axis cs:33,5.09538010799811)
--(axis cs:32,5.04308992968698)
--(axis cs:31,5.06846964358549)
--(axis cs:30,5.06439672019227)
--(axis cs:29,5.08236139368788)
--(axis cs:28,5.08224918976501)
--(axis cs:27,5.12409935969874)
--(axis cs:26,5.16049828841574)
--(axis cs:25,5.16804304338603)
--(axis cs:24,5.15909539867794)
--(axis cs:23,5.15894726927989)
--(axis cs:22,5.19724118644072)
--(axis cs:21,5.33112673899415)
--(axis cs:20,5.38526978311029)
--(axis cs:19,5.54088974655038)
--(axis cs:18,5.51636924790395)
--(axis cs:17,5.50649436081304)
--(axis cs:16,5.48649687672238)
--(axis cs:15,5.4883851683478)
--(axis cs:14,5.48653189777983)
--(axis cs:13,5.48472735586039)
--(axis cs:12,5.40281668909855)
--(axis cs:11,5.40191221943298)
--(axis cs:10,5.42582773742827)
--(axis cs:9,5.44480790882147)
--(axis cs:8,5.39849846327757)
--(axis cs:7,5.34314910421249)
--(axis cs:6,5.23300352202692)
--(axis cs:5,5.10785132716288)
--(axis cs:4,5.03292432404472)
--(axis cs:3,4.91533560534654)
--(axis cs:2,4.83452889066083)
--(axis cs:1,4.52622393754521)
--(axis cs:0,4.02144002530418)
--cycle;

\path [fill=color2, fill opacity=0.2]
(axis cs:0,4.24087844844843)
--(axis cs:0,4.00912155155157)
--(axis cs:1,4.39664658927802)
--(axis cs:2,4.69457603362129)
--(axis cs:3,4.8398361182501)
--(axis cs:4,4.9810160825015)
--(axis cs:5,4.76974724306153)
--(axis cs:6,4.70706526963987)
--(axis cs:7,4.70306611862484)
--(axis cs:8,4.61826996233264)
--(axis cs:9,4.57842005468969)
--(axis cs:10,4.533876854686)
--(axis cs:11,4.53273000775084)
--(axis cs:12,4.46548210236148)
--(axis cs:13,4.4471439910229)
--(axis cs:14,4.40615976275748)
--(axis cs:15,4.39195374093677)
--(axis cs:16,4.38265214976844)
--(axis cs:17,4.31318482898994)
--(axis cs:18,4.30144981422366)
--(axis cs:19,4.25984746387729)
--(axis cs:20,4.17288167635361)
--(axis cs:21,4.12638888888889)
--(axis cs:22,4.13916275848978)
--(axis cs:23,4.15859457204603)
--(axis cs:24,4.14919739582668)
--(axis cs:25,4.14483817125207)
--(axis cs:26,4.12857229512812)
--(axis cs:27,4.11094015031334)
--(axis cs:28,4.1158662063287)
--(axis cs:29,4.12093159435245)
--(axis cs:30,4.13049794409376)
--(axis cs:31,4.12229607157669)
--(axis cs:32,4.12818298888121)
--(axis cs:33,4.1202500770966)
--(axis cs:34,4.1276677067534)
--(axis cs:35,4.07173121639785)
--(axis cs:36,4.05497413191581)
--(axis cs:37,4.07378318805621)
--(axis cs:38,4.11564600045173)
--(axis cs:39,4.06695788818006)
--(axis cs:40,4.07168594141567)
--(axis cs:41,4.05446382357017)
--(axis cs:42,4.06863984742259)
--(axis cs:43,4.06087378101711)
--(axis cs:44,4.04388421435052)
--(axis cs:45,4.05068969110124)
--(axis cs:46,4.02313881938989)
--(axis cs:47,4.0424936786701)
--(axis cs:48,4.03347936199844)
--(axis cs:49,3.98091201357798)
--(axis cs:49,4.1718657641998)
--(axis cs:49,4.1718657641998)
--(axis cs:48,4.23874286022378)
--(axis cs:47,4.24639521021879)
--(axis cs:46,4.23519451394345)
--(axis cs:45,4.26319919778765)
--(axis cs:44,4.24500467453837)
--(axis cs:43,4.275237330094)
--(axis cs:42,4.26191570813296)
--(axis cs:41,4.2483139542076)
--(axis cs:40,4.26998072525099)
--(axis cs:39,4.26915322293105)
--(axis cs:38,4.32324288843716)
--(axis cs:37,4.28177236749935)
--(axis cs:36,4.26447031252863)
--(axis cs:35,4.27826878360214)
--(axis cs:34,4.3223322932466)
--(axis cs:33,4.32141658957007)
--(axis cs:32,4.32181701111879)
--(axis cs:31,4.32770392842331)
--(axis cs:30,4.34727983368402)
--(axis cs:29,4.34573507231421)
--(axis cs:28,4.31191157144908)
--(axis cs:27,4.30572651635332)
--(axis cs:26,4.29642770487188)
--(axis cs:25,4.31349516208126)
--(axis cs:24,4.28969149306221)
--(axis cs:23,4.30529431684286)
--(axis cs:22,4.26639279706578)
--(axis cs:21,4.2625)
--(axis cs:20,4.29934054586861)
--(axis cs:19,4.36793031390048)
--(axis cs:18,4.41799463022078)
--(axis cs:17,4.43403739323228)
--(axis cs:16,4.52290340578711)
--(axis cs:15,4.54415737017434)
--(axis cs:14,4.55772912613141)
--(axis cs:13,4.62507823119932)
--(axis cs:12,4.67618456430518)
--(axis cs:11,4.69782554780471)
--(axis cs:10,4.69390092309178)
--(axis cs:9,4.73269105642142)
--(axis cs:8,4.76784114877847)
--(axis cs:7,4.87193388137516)
--(axis cs:6,4.90404584147124)
--(axis cs:5,4.94414164582736)
--(axis cs:4,5.10787280638739)
--(axis cs:3,5.04627499286101)
--(axis cs:2,4.85820174415649)
--(axis cs:1,4.58390896627753)
--(axis cs:0,4.24087844844843)
--cycle;

\path [fill=color3, fill opacity=0.2]
(axis cs:0,5.21253311706404)
--(axis cs:0,5.13746688293596)
--(axis cs:1,4.72893456071066)
--(axis cs:2,3.90471330648731)
--(axis cs:3,3.91576529526087)
--(axis cs:4,3.85162773886304)
--(axis cs:5,3.89979574214402)
--(axis cs:6,3.81556800989022)
--(axis cs:7,3.78457369372148)
--(axis cs:8,3.80638886188272)
--(axis cs:9,3.7408084527492)
--(axis cs:10,3.76555269149845)
--(axis cs:11,3.77314190918548)
--(axis cs:12,3.80930242887266)
--(axis cs:13,3.75812631504421)
--(axis cs:14,3.71718867538162)
--(axis cs:15,3.70364155586826)
--(axis cs:16,3.6951273429351)
--(axis cs:17,3.6702694471994)
--(axis cs:18,3.67992016060015)
--(axis cs:19,3.71836603289784)
--(axis cs:20,3.7578805766269)
--(axis cs:21,3.79649976958275)
--(axis cs:22,3.76763492019619)
--(axis cs:23,3.76392601152334)
--(axis cs:24,3.80399383161162)
--(axis cs:25,3.75714542493075)
--(axis cs:26,3.73985567526182)
--(axis cs:27,3.77806427819132)
--(axis cs:28,3.78234239268618)
--(axis cs:29,3.73598595163651)
--(axis cs:30,3.75704462354397)
--(axis cs:31,3.73818713708074)
--(axis cs:32,3.73118105925737)
--(axis cs:33,3.78689706505025)
--(axis cs:34,3.78288190111802)
--(axis cs:35,3.78204713107696)
--(axis cs:36,3.82071291610273)
--(axis cs:37,3.80459567312043)
--(axis cs:38,3.73005792944817)
--(axis cs:39,3.7739529791332)
--(axis cs:40,3.76325546345967)
--(axis cs:41,3.724358053507)
--(axis cs:42,3.72982964565793)
--(axis cs:43,3.73612878749962)
--(axis cs:44,3.75177686842248)
--(axis cs:45,3.72443197406833)
--(axis cs:46,3.74826327019409)
--(axis cs:47,3.75731632110563)
--(axis cs:48,3.77410303187232)
--(axis cs:49,3.78779382645193)
--(axis cs:49,3.95665061799252)
--(axis cs:49,3.95665061799252)
--(axis cs:48,3.94534141257212)
--(axis cs:47,3.93157256778326)
--(axis cs:46,3.92118117425035)
--(axis cs:45,3.87556802593167)
--(axis cs:44,3.8760009093553)
--(axis cs:43,3.85553787916705)
--(axis cs:42,3.85905924323096)
--(axis cs:41,3.84786416871522)
--(axis cs:40,3.90618898098477)
--(axis cs:39,3.9260470208668)
--(axis cs:38,3.88383095944071)
--(axis cs:37,3.96207099354624)
--(axis cs:36,3.96817597278616)
--(axis cs:35,3.9235084244786)
--(axis cs:34,3.91711809888198)
--(axis cs:33,3.90476960161642)
--(axis cs:32,3.86881894074263)
--(axis cs:31,3.90070175180815)
--(axis cs:30,3.91239982090048)
--(axis cs:29,3.89734738169683)
--(axis cs:28,3.93154649620271)
--(axis cs:27,3.92471349958646)
--(axis cs:26,3.88514432473818)
--(axis cs:25,3.8734101306248)
--(axis cs:24,3.88767283505504)
--(axis cs:23,3.84996287736555)
--(axis cs:22,3.85736507980381)
--(axis cs:21,3.91738911930614)
--(axis cs:20,3.88934164559532)
--(axis cs:19,3.84274507821328)
--(axis cs:18,3.81730206162207)
--(axis cs:17,3.80750833057838)
--(axis cs:16,3.82153932373157)
--(axis cs:15,3.85469177746507)
--(axis cs:14,3.8800335468406)
--(axis cs:13,3.90020701828912)
--(axis cs:12,3.92958646001623)
--(axis cs:11,3.89630253525897)
--(axis cs:10,3.90944730850155)
--(axis cs:9,3.89530265836191)
--(axis cs:8,3.94638891589506)
--(axis cs:7,3.90153741738963)
--(axis cs:6,3.89832087899867)
--(axis cs:5,4.03075981341154)
--(axis cs:4,3.97337226113696)
--(axis cs:3,4.07034581585024)
--(axis cs:2,4.03973113795714)
--(axis cs:1,4.80717655040045)
--(axis cs:0,5.21253311706404)
--cycle;

\path [fill=color4, fill opacity=0.2]
(axis cs:0,5.124683081402)
--(axis cs:0,5.05865025193134)
--(axis cs:1,4.64209880800342)
--(axis cs:2,3.81601422838429)
--(axis cs:3,3.95429866036635)
--(axis cs:4,3.9607532309166)
--(axis cs:5,4.05128194404635)
--(axis cs:6,3.87955318616882)
--(axis cs:7,3.73038550533993)
--(axis cs:8,3.82273410794434)
--(axis cs:9,3.80328505495078)
--(axis cs:10,3.80664084810921)
--(axis cs:11,3.88351175029415)
--(axis cs:12,3.91541372807388)
--(axis cs:13,3.81000849569959)
--(axis cs:14,3.83460538090512)
--(axis cs:15,3.87063942672411)
--(axis cs:16,3.87426599373344)
--(axis cs:17,3.95884019303376)
--(axis cs:18,3.94390093846083)
--(axis cs:19,3.94261384109483)
--(axis cs:20,3.97525962795273)
--(axis cs:21,3.96431217568028)
--(axis cs:22,3.98259280337944)
--(axis cs:23,3.99351252660499)
--(axis cs:24,3.99744749949846)
--(axis cs:25,4.00783443067062)
--(axis cs:26,4.01041221262383)
--(axis cs:27,3.95520302144469)
--(axis cs:28,3.98714201107731)
--(axis cs:29,3.91815902386518)
--(axis cs:30,3.91705090873414)
--(axis cs:31,3.89900219994161)
--(axis cs:32,3.91711369360989)
--(axis cs:33,3.9305861216623)
--(axis cs:34,3.94010297019105)
--(axis cs:35,3.99668221156104)
--(axis cs:36,4.00034450249933)
--(axis cs:37,4.0150779940738)
--(axis cs:38,3.94948625296183)
--(axis cs:39,3.9202030968634)
--(axis cs:40,3.93443137619858)
--(axis cs:41,3.96504870272609)
--(axis cs:42,3.95276042517198)
--(axis cs:43,3.96340164481112)
--(axis cs:44,3.9907944624263)
--(axis cs:45,4.02021258168303)
--(axis cs:46,4.01941586711109)
--(axis cs:47,3.99781724914202)
--(axis cs:48,4.00698145540293)
--(axis cs:49,4.02453334644531)
--(axis cs:49,4.15046665355469)
--(axis cs:49,4.15046665355469)
--(axis cs:48,4.13190743348596)
--(axis cs:47,4.12440497308021)
--(axis cs:46,4.15558413288891)
--(axis cs:45,4.16589852942808)
--(axis cs:44,4.15642775979592)
--(axis cs:43,4.13104279963333)
--(axis cs:42,4.13335068593914)
--(axis cs:41,4.14328463060724)
--(axis cs:40,4.11556862380142)
--(axis cs:39,4.09646356980327)
--(axis cs:38,4.12273596926039)
--(axis cs:37,4.20714422814842)
--(axis cs:36,4.18854438638956)
--(axis cs:35,4.18387334399452)
--(axis cs:34,4.12934147425339)
--(axis cs:33,4.13608054500437)
--(axis cs:32,4.11621963972345)
--(axis cs:31,4.1121089111695)
--(axis cs:30,4.13572686904364)
--(axis cs:29,4.15961875391259)
--(axis cs:28,4.21008021114491)
--(axis cs:27,4.16701920077753)
--(axis cs:26,4.21736556515395)
--(axis cs:25,4.20605445821827)
--(axis cs:24,4.1942191671682)
--(axis cs:23,4.16759858450612)
--(axis cs:22,4.15907386328722)
--(axis cs:21,4.15235449098638)
--(axis cs:20,4.14696259426949)
--(axis cs:19,4.09905282557184)
--(axis cs:18,4.09498795042806)
--(axis cs:17,4.09393758474402)
--(axis cs:16,4.00628956182211)
--(axis cs:15,3.99602723994256)
--(axis cs:14,3.95150573020599)
--(axis cs:13,3.95110261541152)
--(axis cs:12,4.05403071637057)
--(axis cs:11,4.02759936081696)
--(axis cs:10,3.94058137411301)
--(axis cs:9,3.93004827838256)
--(axis cs:8,3.94671033650011)
--(axis cs:7,3.88628116132673)
--(axis cs:6,3.99822459160895)
--(axis cs:5,4.20149583373143)
--(axis cs:4,4.06702454686118)
--(axis cs:3,4.07903467296699)
--(axis cs:2,3.96454132717126)
--(axis cs:1,4.77456785866325)
--(axis cs:0,5.124683081402)
--cycle;

\path [fill=color5, fill opacity=0.2]
(axis cs:0,4.10900572621907)
--(axis cs:0,3.96043871822538)
--(axis cs:1,4.05632761427204)
--(axis cs:2,4.29666137302128)
--(axis cs:3,4.17434236277378)
--(axis cs:4,4.22599504138135)
--(axis cs:5,4.17096920785056)
--(axis cs:6,4.2087767592418)
--(axis cs:7,4.14747182163195)
--(axis cs:8,4.08402247051399)
--(axis cs:9,4.09246440481158)
--(axis cs:10,4.04645954265983)
--(axis cs:11,4.0314948679491)
--(axis cs:12,4.02290151624877)
--(axis cs:13,3.96900957573834)
--(axis cs:14,3.97404536177954)
--(axis cs:15,3.95163911068912)
--(axis cs:16,3.91961355957431)
--(axis cs:17,3.93816164846286)
--(axis cs:18,3.97878887895948)
--(axis cs:19,3.97061258847745)
--(axis cs:20,3.97981293716523)
--(axis cs:21,3.96921731633638)
--(axis cs:22,3.99025625492988)
--(axis cs:23,3.98521116199865)
--(axis cs:24,3.97634116693154)
--(axis cs:25,4.02101282206134)
--(axis cs:26,3.97325783546649)
--(axis cs:27,4.03668213737902)
--(axis cs:28,4.05938352174376)
--(axis cs:29,4.03577154883397)
--(axis cs:30,4.0176413969873)
--(axis cs:31,3.99393186383329)
--(axis cs:32,4.01072472164955)
--(axis cs:33,3.99189783730541)
--(axis cs:34,3.9689968921274)
--(axis cs:35,3.9826498231479)
--(axis cs:36,3.94278700844991)
--(axis cs:37,3.95765373662797)
--(axis cs:38,3.96126106985874)
--(axis cs:39,3.99346996901728)
--(axis cs:40,4.00495413084494)
--(axis cs:41,4.00236687938699)
--(axis cs:42,3.96828821349513)
--(axis cs:43,3.93923842768452)
--(axis cs:44,3.95206171431602)
--(axis cs:45,3.95824637744814)
--(axis cs:46,3.96695687937663)
--(axis cs:47,3.95459069395949)
--(axis cs:48,3.93753406161601)
--(axis cs:49,3.94911108128835)
--(axis cs:49,4.07588891871165)
--(axis cs:49,4.07588891871165)
--(axis cs:48,4.06524371616177)
--(axis cs:47,4.08152041715162)
--(axis cs:46,4.10804312062337)
--(axis cs:45,4.10008695588519)
--(axis cs:44,4.09793828568398)
--(axis cs:43,4.08853935009326)
--(axis cs:42,4.1400451198382)
--(axis cs:41,4.17263312061301)
--(axis cs:40,4.16171253582172)
--(axis cs:39,4.16764114209383)
--(axis cs:38,4.13596115236348)
--(axis cs:37,4.12290181892759)
--(axis cs:36,4.10443521377231)
--(axis cs:35,4.14512795462988)
--(axis cs:34,4.15878088565038)
--(axis cs:33,4.17754660713904)
--(axis cs:32,4.16149750057267)
--(axis cs:31,4.13940146950004)
--(axis cs:30,4.15180304745714)
--(axis cs:29,4.14756178449936)
--(axis cs:28,4.16561647825624)
--(axis cs:27,4.14387341817654)
--(axis cs:26,4.09896438675574)
--(axis cs:25,4.15398717793866)
--(axis cs:24,4.12643661084624)
--(axis cs:23,4.16201106022358)
--(axis cs:22,4.16252152284789)
--(axis cs:21,4.11689379477473)
--(axis cs:20,4.10629817394588)
--(axis cs:19,4.09883185596699)
--(axis cs:18,4.12954445437385)
--(axis cs:17,4.08683835153714)
--(axis cs:16,4.04705310709236)
--(axis cs:15,4.05391644486644)
--(axis cs:14,4.07873241599824)
--(axis cs:13,4.10321264648388)
--(axis cs:12,4.1937651504179)
--(axis cs:11,4.24072735427312)
--(axis cs:10,4.24798490178461)
--(axis cs:9,4.30198003963287)
--(axis cs:8,4.31597752948601)
--(axis cs:7,4.35530595614583)
--(axis cs:6,4.44400101853598)
--(axis cs:5,4.36791968103833)
--(axis cs:4,4.41567162528532)
--(axis cs:3,4.40899097055955)
--(axis cs:2,4.58667196031205)
--(axis cs:1,4.30200571906129)
--(axis cs:0,4.10900572621907)
--cycle;

\path [fill=color6, fill opacity=0.2]
(axis cs:0,4.53484085817247)
--(axis cs:0,4.34015914182753)
--(axis cs:1,4.41329671744301)
--(axis cs:2,4.59497995958809)
--(axis cs:3,4.57814841221861)
--(axis cs:4,4.39498923471212)
--(axis cs:5,4.39465610702807)
--(axis cs:6,4.37559220095033)
--(axis cs:7,4.452867049735)
--(axis cs:8,4.40494505285801)
--(axis cs:9,4.3816101036635)
--(axis cs:10,4.42766220016567)
--(axis cs:11,4.39620416721787)
--(axis cs:12,4.44837810249874)
--(axis cs:13,4.53931752867872)
--(axis cs:14,4.53664287062737)
--(axis cs:15,4.54849610410591)
--(axis cs:16,4.52565007796312)
--(axis cs:17,4.47721414887626)
--(axis cs:18,4.45106641872807)
--(axis cs:19,4.46521132332918)
--(axis cs:20,4.44145249920033)
--(axis cs:21,4.42096921949571)
--(axis cs:22,4.44680688164543)
--(axis cs:23,4.41886252815675)
--(axis cs:24,4.36335271955009)
--(axis cs:25,4.35993265145191)
--(axis cs:26,4.3811472368508)
--(axis cs:27,4.38529831559332)
--(axis cs:28,4.40309608032203)
--(axis cs:29,4.44089962479865)
--(axis cs:30,4.48671962805822)
--(axis cs:31,4.51023481274119)
--(axis cs:32,4.51131235628258)
--(axis cs:33,4.44774231333818)
--(axis cs:34,4.43632232552463)
--(axis cs:35,4.43056778532761)
--(axis cs:36,4.44573907314077)
--(axis cs:37,4.41399729924169)
--(axis cs:38,4.43436745186748)
--(axis cs:39,4.44028956408465)
--(axis cs:40,4.39940020866798)
--(axis cs:41,4.36747351388603)
--(axis cs:42,4.37513630514895)
--(axis cs:43,4.41891367318433)
--(axis cs:44,4.39752516328474)
--(axis cs:45,4.39149317899779)
--(axis cs:46,4.41276071769496)
--(axis cs:47,4.40843049518566)
--(axis cs:48,4.3810379739167)
--(axis cs:49,4.3769643584821)
--(axis cs:49,4.58414675262901)
--(axis cs:49,4.58414675262901)
--(axis cs:48,4.5689620260833)
--(axis cs:47,4.61101394925879)
--(axis cs:46,4.6177948378606)
--(axis cs:45,4.61128459877999)
--(axis cs:44,4.61636372560415)
--(axis cs:43,4.63386410459344)
--(axis cs:42,4.60541925040661)
--(axis cs:41,4.59363759722508)
--(axis cs:40,4.59782201355424)
--(axis cs:39,4.63748821369313)
--(axis cs:38,4.62396588146586)
--(axis cs:37,4.60544714520276)
--(axis cs:36,4.62648314908146)
--(axis cs:35,4.61387665911684)
--(axis cs:34,4.6220110078087)
--(axis cs:33,4.64114657555071)
--(axis cs:32,4.68868764371742)
--(axis cs:31,4.66198740948103)
--(axis cs:30,4.64939148305289)
--(axis cs:29,4.61187815297913)
--(axis cs:28,4.5802372530113)
--(axis cs:27,4.57025723996223)
--(axis cs:26,4.54940831870476)
--(axis cs:25,4.51228957077032)
--(axis cs:24,4.51720283600547)
--(axis cs:23,4.54224858295436)
--(axis cs:22,4.58930422946568)
--(axis cs:21,4.58180855828207)
--(axis cs:20,4.59188083413301)
--(axis cs:19,4.57923312111526)
--(axis cs:18,4.55171135904971)
--(axis cs:17,4.59500807334596)
--(axis cs:16,4.65490547759243)
--(axis cs:15,4.65428167367187)
--(axis cs:14,4.63835712937263)
--(axis cs:13,4.63846024909906)
--(axis cs:12,4.60439967527904)
--(axis cs:11,4.56768472167101)
--(axis cs:10,4.57789335538988)
--(axis cs:9,4.54061211855872)
--(axis cs:8,4.56727716936421)
--(axis cs:7,4.66102183915389)
--(axis cs:6,4.57718557682745)
--(axis cs:5,4.55256611519415)
--(axis cs:4,4.53278854306566)
--(axis cs:3,4.71074047667028)
--(axis cs:2,4.72446448485636)
--(axis cs:1,4.5978143936681)
--(axis cs:0,4.53484085817247)
--cycle;

\path [fill=color7, fill opacity=0.2]
(axis cs:0,4.19436407275023)
--(axis cs:0,4.03619148280533)
--(axis cs:1,4.41359642215037)
--(axis cs:2,4.69939176538031)
--(axis cs:3,4.85625443611335)
--(axis cs:4,4.84182239661796)
--(axis cs:5,4.88449616864072)
--(axis cs:6,4.91938902630182)
--(axis cs:7,4.95608495933549)
--(axis cs:8,5.02188831048465)
--(axis cs:9,5.14709204611005)
--(axis cs:10,5.19310855143629)
--(axis cs:11,5.19828974285383)
--(axis cs:12,5.20475048907309)
--(axis cs:13,5.29423487339353)
--(axis cs:14,5.36139761152465)
--(axis cs:15,5.41336123261384)
--(axis cs:16,5.4309055000726)
--(axis cs:17,5.48126042405327)
--(axis cs:18,5.51120722130607)
--(axis cs:19,5.53088837938348)
--(axis cs:20,5.63785247299603)
--(axis cs:21,5.70515581801726)
--(axis cs:22,5.77018567245952)
--(axis cs:23,5.82513146501457)
--(axis cs:24,5.84422565723766)
--(axis cs:25,5.84242632308919)
--(axis cs:26,5.89149130925334)
--(axis cs:27,5.92143850082762)
--(axis cs:28,5.9483573608063)
--(axis cs:29,5.990912291265)
--(axis cs:30,5.98658830615547)
--(axis cs:31,6.02551111188229)
--(axis cs:32,6.06936100839279)
--(axis cs:33,6.07780074474396)
--(axis cs:34,6.09641738499416)
--(axis cs:35,6.07129710150575)
--(axis cs:36,6.10896572542677)
--(axis cs:37,6.12641881849576)
--(axis cs:38,6.15603236203072)
--(axis cs:39,6.16646448063053)
--(axis cs:40,6.18394349923295)
--(axis cs:41,6.19864814605802)
--(axis cs:42,6.2184315373364)
--(axis cs:43,6.22599504138135)
--(axis cs:44,6.23610189750824)
--(axis cs:45,6.20990904683919)
--(axis cs:46,6.15889538215067)
--(axis cs:47,6.16800376360882)
--(axis cs:48,6.17160856900711)
--(axis cs:49,6.16786723711855)
--(axis cs:49,6.36824387399256)
--(axis cs:49,6.36824387399256)
--(axis cs:48,6.37561365321511)
--(axis cs:47,6.37088512528007)
--(axis cs:46,6.36332684007155)
--(axis cs:45,6.40120206427192)
--(axis cs:44,6.42500921360287)
--(axis cs:43,6.41567162528532)
--(axis cs:42,6.40379068488582)
--(axis cs:41,6.37912963171976)
--(axis cs:40,6.37994538965594)
--(axis cs:39,6.3668688527028)
--(axis cs:38,6.3550787490804)
--(axis cs:37,6.32080340372646)
--(axis cs:36,6.29936760790657)
--(axis cs:35,6.26759178738314)
--(axis cs:34,6.25636039278362)
--(axis cs:33,6.24164369970049)
--(axis cs:32,6.22786121382943)
--(axis cs:31,6.18559999922882)
--(axis cs:30,6.13841169384453)
--(axis cs:29,6.12575437540167)
--(axis cs:28,6.07942041697148)
--(axis cs:27,6.04245038806127)
--(axis cs:26,6.02795313519111)
--(axis cs:25,5.97701812135526)
--(axis cs:24,5.98077434276234)
--(axis cs:23,5.97486853498543)
--(axis cs:22,5.92148099420714)
--(axis cs:21,5.84762195976052)
--(axis cs:20,5.74825863811508)
--(axis cs:19,5.65522273172763)
--(axis cs:18,5.61934833424949)
--(axis cs:17,5.57429513150228)
--(axis cs:16,5.53298338881629)
--(axis cs:15,5.52830543405282)
--(axis cs:14,5.50804683291979)
--(axis cs:13,5.45298734882869)
--(axis cs:12,5.3702495109269)
--(axis cs:11,5.39893247936839)
--(axis cs:10,5.35411367078593)
--(axis cs:9,5.30568573166773)
--(axis cs:8,5.19477835618201)
--(axis cs:7,5.16891504066451)
--(axis cs:6,5.13894430703151)
--(axis cs:5,5.09050383135928)
--(axis cs:4,4.99428871449315)
--(axis cs:3,5.02985667499776)
--(axis cs:2,4.88394156795303)
--(axis cs:1,4.67529246673852)
--(axis cs:0,4.19436407275023)
--cycle;

\addplot [line width=2pt, color0]
table {%
0 5.08194444444444
1 4.38194444444444
2 4.73333333333333
3 4.31527777777778
4 4.37222222222222
5 4.275
6 4.39861111111111
7 4.41666666666667
8 4.41944444444445
9 4.40972222222222
10 4.39861111111111
11 4.3375
12 4.31944444444444
13 4.35416666666667
14 4.36666666666667
15 4.31527777777778
16 4.35972222222222
17 4.32638888888889
18 4.27916666666667
19 4.24583333333333
20 4.29444444444445
21 4.30972222222222
22 4.32777777777778
23 4.32638888888889
24 4.34166666666667
25 4.31388888888889
26 4.3
27 4.27361111111111
28 4.20138888888889
29 4.21666666666667
30 4.18333333333333
31 4.17916666666667
32 4.13888888888889
33 4.07916666666667
34 4.05
35 4.04305555555556
36 3.99583333333333
37 3.98472222222222
38 3.98472222222222
39 3.96666666666667
40 3.94722222222222
41 3.975
42 3.96944444444444
43 3.95
44 3.94027777777778
45 3.93472222222222
46 3.94305555555556
47 3.93611111111111
48 3.94722222222222
49 3.95138888888889
};
\addplot [line width=2pt, color1]
table {%
0 3.93888888888889
1 4.42222222222222
2 4.73472222222222
3 4.85277777777778
4 4.95416666666667
5 5.04583333333333
6 5.16805555555556
7 5.27361111111111
8 5.33194444444444
9 5.36527777777778
10 5.35277777777778
11 5.34027777777778
12 5.33333333333333
13 5.40138888888889
14 5.41111111111111
15 5.42916666666667
16 5.43333333333333
17 5.45833333333333
18 5.47222222222222
19 5.48888888888889
20 5.33194444444444
21 5.25972222222222
22 5.14583333333333
23 5.1
24 5.09305555555556
25 5.10138888888889
26 5.1
27 5.05555555555556
28 5.01111111111111
29 5.01805555555556
30 5.00138888888889
31 5.0125
32 4.98472222222222
33 5.03194444444444
34 5.07083333333333
35 5.08333333333333
36 5.09166666666667
37 5.06944444444444
38 5.11944444444444
39 5.12083333333333
40 5.15833333333333
41 5.17638888888889
42 5.17222222222222
43 5.17777777777778
44 5.16527777777778
45 5.17916666666667
46 5.19166666666667
47 5.21666666666667
48 5.2375
49 5.25138888888889
};
\addplot [line width=2pt, color2]
table {%
0 4.125
1 4.49027777777778
2 4.77638888888889
3 4.94305555555556
4 5.04444444444444
5 4.85694444444444
6 4.80555555555556
7 4.7875
8 4.69305555555556
9 4.65555555555556
10 4.61388888888889
11 4.61527777777778
12 4.57083333333333
13 4.53611111111111
14 4.48194444444445
15 4.46805555555556
16 4.45277777777778
17 4.37361111111111
18 4.35972222222222
19 4.31388888888889
20 4.23611111111111
21 4.19444444444444
22 4.20277777777778
23 4.23194444444444
24 4.21944444444444
25 4.22916666666667
26 4.2125
27 4.20833333333333
28 4.21388888888889
29 4.23333333333333
30 4.23888888888889
31 4.225
32 4.225
33 4.22083333333333
34 4.225
35 4.175
36 4.15972222222222
37 4.17777777777778
38 4.21944444444444
39 4.16805555555556
40 4.17083333333333
41 4.15138888888889
42 4.16527777777778
43 4.16805555555556
44 4.14444444444444
45 4.15694444444444
46 4.12916666666667
47 4.14444444444444
48 4.13611111111111
49 4.07638888888889
};
\addplot [line width=2pt, color3]
table {%
0 5.175
1 4.76805555555555
2 3.97222222222222
3 3.99305555555556
4 3.9125
5 3.96527777777778
6 3.85694444444445
7 3.84305555555556
8 3.87638888888889
9 3.81805555555556
10 3.8375
11 3.83472222222222
12 3.86944444444444
13 3.82916666666667
14 3.79861111111111
15 3.77916666666667
16 3.75833333333333
17 3.73888888888889
18 3.74861111111111
19 3.78055555555556
20 3.82361111111111
21 3.85694444444444
22 3.8125
23 3.80694444444444
24 3.84583333333333
25 3.81527777777778
26 3.8125
27 3.85138888888889
28 3.85694444444444
29 3.81666666666667
30 3.83472222222222
31 3.81944444444444
32 3.8
33 3.84583333333333
34 3.85
35 3.85277777777778
36 3.89444444444444
37 3.88333333333333
38 3.80694444444444
39 3.85
40 3.83472222222222
41 3.78611111111111
42 3.79444444444444
43 3.79583333333333
44 3.81388888888889
45 3.8
46 3.83472222222222
47 3.84444444444444
48 3.85972222222222
49 3.87222222222222
};
\addplot [line width=2pt, color4]
table {%
0 5.09166666666667
1 4.70833333333333
2 3.89027777777778
3 4.01666666666667
4 4.01388888888889
5 4.12638888888889
6 3.93888888888889
7 3.80833333333333
8 3.88472222222222
9 3.86666666666667
10 3.87361111111111
11 3.95555555555556
12 3.98472222222222
13 3.88055555555556
14 3.89305555555556
15 3.93333333333333
16 3.94027777777778
17 4.02638888888889
18 4.01944444444444
19 4.02083333333333
20 4.06111111111111
21 4.05833333333333
22 4.07083333333333
23 4.08055555555555
24 4.09583333333333
25 4.10694444444444
26 4.11388888888889
27 4.06111111111111
28 4.09861111111111
29 4.03888888888889
30 4.02638888888889
31 4.00555555555555
32 4.01666666666667
33 4.03333333333333
34 4.03472222222222
35 4.09027777777778
36 4.09444444444444
37 4.11111111111111
38 4.03611111111111
39 4.00833333333333
40 4.025
41 4.05416666666667
42 4.04305555555556
43 4.04722222222222
44 4.07361111111111
45 4.09305555555556
46 4.0875
47 4.06111111111111
48 4.06944444444444
49 4.0875
};
\addplot [line width=2pt, color5]
table {%
0 4.03472222222222
1 4.17916666666667
2 4.44166666666667
3 4.29166666666667
4 4.32083333333333
5 4.26944444444444
6 4.32638888888889
7 4.25138888888889
8 4.2
9 4.19722222222222
10 4.14722222222222
11 4.13611111111111
12 4.10833333333333
13 4.03611111111111
14 4.02638888888889
15 4.00277777777778
16 3.98333333333333
17 4.0125
18 4.05416666666667
19 4.03472222222222
20 4.04305555555556
21 4.04305555555556
22 4.07638888888889
23 4.07361111111111
24 4.05138888888889
25 4.0875
26 4.03611111111111
27 4.09027777777778
28 4.1125
29 4.09166666666667
30 4.08472222222222
31 4.06666666666667
32 4.08611111111111
33 4.08472222222222
34 4.06388888888889
35 4.06388888888889
36 4.02361111111111
37 4.04027777777778
38 4.04861111111111
39 4.08055555555555
40 4.08333333333333
41 4.0875
42 4.05416666666667
43 4.01388888888889
44 4.025
45 4.02916666666667
46 4.0375
47 4.01805555555556
48 4.00138888888889
49 4.0125
};
\addplot [line width=2pt, color6]
table {%
0 4.4375
1 4.50555555555556
2 4.65972222222222
3 4.64444444444444
4 4.46388888888889
5 4.47361111111111
6 4.47638888888889
7 4.55694444444444
8 4.48611111111111
9 4.46111111111111
10 4.50277777777778
11 4.48194444444444
12 4.52638888888889
13 4.58888888888889
14 4.5875
15 4.60138888888889
16 4.59027777777778
17 4.53611111111111
18 4.50138888888889
19 4.52222222222222
20 4.51666666666667
21 4.50138888888889
22 4.51805555555555
23 4.48055555555556
24 4.44027777777778
25 4.43611111111111
26 4.46527777777778
27 4.47777777777778
28 4.49166666666667
29 4.52638888888889
30 4.56805555555556
31 4.58611111111111
32 4.6
33 4.54444444444444
34 4.52916666666667
35 4.52222222222222
36 4.53611111111111
37 4.50972222222222
38 4.52916666666667
39 4.53888888888889
40 4.49861111111111
41 4.48055555555556
42 4.49027777777778
43 4.52638888888889
44 4.50694444444444
45 4.50138888888889
46 4.51527777777778
47 4.50972222222222
48 4.475
49 4.48055555555556
};
\addplot [line width=2pt, color7]
table {%
0 4.11527777777778
1 4.54444444444444
2 4.79166666666667
3 4.94305555555556
4 4.91805555555556
5 4.9875
6 5.02916666666667
7 5.0625
8 5.10833333333333
9 5.22638888888889
10 5.27361111111111
11 5.29861111111111
12 5.2875
13 5.37361111111111
14 5.43472222222222
15 5.47083333333333
16 5.48194444444444
17 5.52777777777778
18 5.56527777777778
19 5.59305555555556
20 5.69305555555556
21 5.77638888888889
22 5.84583333333333
23 5.9
24 5.9125
25 5.90972222222222
26 5.95972222222222
27 5.98194444444445
28 6.01388888888889
29 6.05833333333333
30 6.0625
31 6.10555555555556
32 6.14861111111111
33 6.15972222222222
34 6.17638888888889
35 6.16944444444444
36 6.20416666666667
37 6.22361111111111
38 6.25555555555556
39 6.26666666666667
40 6.28194444444444
41 6.28888888888889
42 6.31111111111111
43 6.32083333333333
44 6.33055555555556
45 6.30555555555556
46 6.26111111111111
47 6.26944444444444
48 6.27361111111111
49 6.26805555555556
};

\end{axis}

\end{tikzpicture}

%% file: icml2023/figures/bayesmark_mean_plot.tex
\begin{tikzpicture}
\colorlet{color0}{colorHEBO}
\colorlet{color1}{colorHyperopt}
\colorlet{color2}{colorPFN_BNN}
\colorlet{color3}{colorPFN_GP_init_at_min}
\colorlet{color4}{colorPFN_GP_init_at_mid}
\colorlet{color5}{colorPFN_GP}
\colorlet{color6}{colorPySOT}
\colorlet{color7}{colorRandom}

\begin{axis}[
legend cell align={left},
legend style={fill opacity=0.8, draw opacity=1, text opacity=1, draw=white!80!black},
log basis y={10},
tick align=outside,
tick pos=left,
x grid style={white!69.0196078431373!black},
xmin=-2.45, xmax=51.45,
xtick style={color=black},
y grid style={white!69.0196078431373!black},
ymin=0.0621347204801858, ymax=0.361896068732605,
ymode=log,
ytick style={color=black},
height=\benchmarkplotheight,
width=\benchmarkplotwidth,
xlabel={Number of trials},
ylabel={Average Regret},
]

\path [fill=color0, fill opacity=0.2]
(axis cs:0,0.332240950134486)
--(axis cs:0,0.327798748593361)
--(axis cs:1,0.19708669593025)
--(axis cs:2,0.191861000139621)
--(axis cs:3,0.157563832777206)
--(axis cs:4,0.146643054374294)
--(axis cs:5,0.135864064127887)
--(axis cs:6,0.128797855390297)
--(axis cs:7,0.123760326621113)
--(axis cs:8,0.1176367498166)
--(axis cs:9,0.113749964652796)
--(axis cs:10,0.111396864293077)
--(axis cs:11,0.105754562598721)
--(axis cs:12,0.102477499312156)
--(axis cs:13,0.0998819365713421)
--(axis cs:14,0.0977285225258446)
--(axis cs:15,0.0947270054115125)
--(axis cs:16,0.0939461996562704)
--(axis cs:17,0.0918408562345314)
--(axis cs:18,0.0897628244954286)
--(axis cs:19,0.0876093234486511)
--(axis cs:20,0.0869824975534464)
--(axis cs:21,0.0854415124065672)
--(axis cs:22,0.0840115215940752)
--(axis cs:23,0.0824174958273076)
--(axis cs:24,0.0813498469772858)
--(axis cs:25,0.0803236199072787)
--(axis cs:26,0.0795812315839099)
--(axis cs:27,0.0786417458782936)
--(axis cs:28,0.077804498866886)
--(axis cs:29,0.0763164733485973)
--(axis cs:30,0.0756019449231067)
--(axis cs:31,0.0748214488114087)
--(axis cs:32,0.0740872847496113)
--(axis cs:33,0.0734690392246162)
--(axis cs:34,0.0729713224803989)
--(axis cs:35,0.0722695555868355)
--(axis cs:36,0.0718212379348545)
--(axis cs:37,0.0714614465338488)
--(axis cs:38,0.0708487107402222)
--(axis cs:39,0.0705794431434848)
--(axis cs:40,0.0703621910362383)
--(axis cs:41,0.0702806094441046)
--(axis cs:42,0.0696112025299502)
--(axis cs:43,0.0694122916736407)
--(axis cs:44,0.0689615564304304)
--(axis cs:45,0.0687492517472403)
--(axis cs:46,0.0685237055222967)
--(axis cs:47,0.0682263843364233)
--(axis cs:48,0.0681407629238495)
--(axis cs:49,0.0679505216418163)
--(axis cs:49,0.0689956531177345)
--(axis cs:49,0.0689956531177345)
--(axis cs:48,0.0692292138102638)
--(axis cs:47,0.0694190390441241)
--(axis cs:46,0.0697461697070332)
--(axis cs:45,0.0703390302580588)
--(axis cs:44,0.0707069725383684)
--(axis cs:43,0.0710515557550055)
--(axis cs:42,0.0715804518665156)
--(axis cs:41,0.072243981483118)
--(axis cs:40,0.0722982678687012)
--(axis cs:39,0.0725261115192257)
--(axis cs:38,0.072774819866505)
--(axis cs:37,0.0733221654512805)
--(axis cs:36,0.0737169679019215)
--(axis cs:35,0.0740885099315012)
--(axis cs:34,0.0746751551276184)
--(axis cs:33,0.0755245681904588)
--(axis cs:32,0.0763310079718247)
--(axis cs:31,0.0778034260120047)
--(axis cs:30,0.078615482978089)
--(axis cs:29,0.0793364717377826)
--(axis cs:28,0.081778848622177)
--(axis cs:27,0.0826317249634418)
--(axis cs:26,0.0837970644476136)
--(axis cs:25,0.0843605276094312)
--(axis cs:24,0.085295009239865)
--(axis cs:23,0.0865139854773163)
--(axis cs:22,0.0880980556420964)
--(axis cs:21,0.0893125234098108)
--(axis cs:20,0.0909448601122243)
--(axis cs:19,0.091716792680363)
--(axis cs:18,0.0939606802929591)
--(axis cs:17,0.0954603971415922)
--(axis cs:16,0.0975677786383533)
--(axis cs:15,0.0985268354765447)
--(axis cs:14,0.100937676017614)
--(axis cs:13,0.103477530378502)
--(axis cs:12,0.106380430567904)
--(axis cs:11,0.109918091934495)
--(axis cs:10,0.115437140321363)
--(axis cs:9,0.117794938562804)
--(axis cs:8,0.121846839784648)
--(axis cs:7,0.128070717092465)
--(axis cs:6,0.133247056475631)
--(axis cs:5,0.140094024864001)
--(axis cs:4,0.150527047306776)
--(axis cs:3,0.15938217286238)
--(axis cs:2,0.194948813010166)
--(axis cs:1,0.20010727348784)
--(axis cs:0,0.332240950134486)
--cycle;

\path [fill=color1, fill opacity=0.2]
(axis cs:0,0.277022859269944)
--(axis cs:0,0.264487319227462)
--(axis cs:1,0.209497206891297)
--(axis cs:2,0.185242669968635)
--(axis cs:3,0.170002134544221)
--(axis cs:4,0.160515371993443)
--(axis cs:5,0.156038628021094)
--(axis cs:6,0.148275280749031)
--(axis cs:7,0.145997640004791)
--(axis cs:8,0.142562416537889)
--(axis cs:9,0.139307959691266)
--(axis cs:10,0.13402516463461)
--(axis cs:11,0.130179168382711)
--(axis cs:12,0.126110586611512)
--(axis cs:13,0.124465138891202)
--(axis cs:14,0.123455746515538)
--(axis cs:15,0.121300280066303)
--(axis cs:16,0.120147456629025)
--(axis cs:17,0.119553624454743)
--(axis cs:18,0.118815789630289)
--(axis cs:19,0.118153630337568)
--(axis cs:20,0.110325179776388)
--(axis cs:21,0.10667241209004)
--(axis cs:22,0.104784466590547)
--(axis cs:23,0.102284225727404)
--(axis cs:24,0.099592019170922)
--(axis cs:25,0.0982067050224823)
--(axis cs:26,0.0974469277025911)
--(axis cs:27,0.0968624937736333)
--(axis cs:28,0.0959727032029502)
--(axis cs:29,0.0942893085314061)
--(axis cs:30,0.0927459685713631)
--(axis cs:31,0.0922124065351273)
--(axis cs:32,0.0915706359973521)
--(axis cs:33,0.0912544391445566)
--(axis cs:34,0.0909168032434333)
--(axis cs:35,0.0908460428370338)
--(axis cs:36,0.0904710193940031)
--(axis cs:37,0.0901485773215814)
--(axis cs:38,0.0895699384929585)
--(axis cs:39,0.0894528870503339)
--(axis cs:40,0.0894303300814675)
--(axis cs:41,0.0893356737092939)
--(axis cs:42,0.0890411015682728)
--(axis cs:43,0.0880896291901639)
--(axis cs:44,0.0879687551393553)
--(axis cs:45,0.0876156814534841)
--(axis cs:46,0.0875000776340904)
--(axis cs:47,0.0874893283021407)
--(axis cs:48,0.0874593733229698)
--(axis cs:49,0.0871942807093512)
--(axis cs:49,0.0901908695051199)
--(axis cs:49,0.0901908695051199)
--(axis cs:48,0.0906142040454456)
--(axis cs:47,0.0907046784913544)
--(axis cs:46,0.0907099971963062)
--(axis cs:45,0.0908471354095115)
--(axis cs:44,0.0912994335328318)
--(axis cs:43,0.0914492978685627)
--(axis cs:42,0.093756483141896)
--(axis cs:41,0.0940665399073856)
--(axis cs:40,0.0941665448226158)
--(axis cs:39,0.0942270644442201)
--(axis cs:38,0.0943401612058818)
--(axis cs:37,0.0947791229840625)
--(axis cs:36,0.0950959436472905)
--(axis cs:35,0.0953615523373926)
--(axis cs:34,0.0954242424523748)
--(axis cs:33,0.0957927756838819)
--(axis cs:32,0.096125731929076)
--(axis cs:31,0.0969251650436589)
--(axis cs:30,0.0974157659894639)
--(axis cs:29,0.0984910934405433)
--(axis cs:28,0.101387840965836)
--(axis cs:27,0.101961786225303)
--(axis cs:26,0.102768608401209)
--(axis cs:25,0.103450501088309)
--(axis cs:24,0.104732595808918)
--(axis cs:23,0.107780384410535)
--(axis cs:22,0.109800912869728)
--(axis cs:21,0.111241787486095)
--(axis cs:20,0.114273116234597)
--(axis cs:19,0.122903303087515)
--(axis cs:18,0.123492508842153)
--(axis cs:17,0.124542726245196)
--(axis cs:16,0.125268263422243)
--(axis cs:15,0.126511158574877)
--(axis cs:14,0.1290550489934)
--(axis cs:13,0.130318539201541)
--(axis cs:12,0.131781501393854)
--(axis cs:11,0.135002663962617)
--(axis cs:10,0.139088563945809)
--(axis cs:9,0.143924289893661)
--(axis cs:8,0.146228604128306)
--(axis cs:7,0.150231067067762)
--(axis cs:6,0.152127964182091)
--(axis cs:5,0.161990423123408)
--(axis cs:4,0.169638777021706)
--(axis cs:3,0.180197695688287)
--(axis cs:2,0.19644791799149)
--(axis cs:1,0.223668174663312)
--(axis cs:0,0.277022859269944)
--cycle;

\path [fill=color2, fill opacity=0.2]
(axis cs:0,0.281830738462571)
--(axis cs:0,0.268654075990826)
--(axis cs:1,0.204682754729142)
--(axis cs:2,0.183978398677988)
--(axis cs:3,0.172366219646213)
--(axis cs:4,0.16378088328148)
--(axis cs:5,0.151519276324965)
--(axis cs:6,0.142505189202569)
--(axis cs:7,0.137536710841933)
--(axis cs:8,0.130985996716173)
--(axis cs:9,0.126783245456127)
--(axis cs:10,0.12441829483906)
--(axis cs:11,0.120717175238644)
--(axis cs:12,0.11707391099597)
--(axis cs:13,0.112531553237864)
--(axis cs:14,0.11040074632939)
--(axis cs:15,0.107996375907972)
--(axis cs:16,0.104303634035263)
--(axis cs:17,0.102981732552981)
--(axis cs:18,0.100781534734401)
--(axis cs:19,0.0993031445196884)
--(axis cs:20,0.0974747886715894)
--(axis cs:21,0.0956838277329197)
--(axis cs:22,0.0949662647908504)
--(axis cs:23,0.0944250153432875)
--(axis cs:24,0.0938717675946295)
--(axis cs:25,0.0933514214833398)
--(axis cs:26,0.0925645827595749)
--(axis cs:27,0.0896167358229079)
--(axis cs:28,0.0880360250902833)
--(axis cs:29,0.0876390940456922)
--(axis cs:30,0.0868499257507457)
--(axis cs:31,0.0865090550477784)
--(axis cs:32,0.0859309042447227)
--(axis cs:33,0.0850334223671802)
--(axis cs:34,0.0845117673978875)
--(axis cs:35,0.083513163908827)
--(axis cs:36,0.0828807318206001)
--(axis cs:37,0.0814292517702635)
--(axis cs:38,0.0812253871395079)
--(axis cs:39,0.0790246718481118)
--(axis cs:40,0.0786822455822604)
--(axis cs:41,0.0783951734003294)
--(axis cs:42,0.0775468118785804)
--(axis cs:43,0.077302537536789)
--(axis cs:44,0.0772141057657516)
--(axis cs:45,0.0769972481812475)
--(axis cs:46,0.0764896055117476)
--(axis cs:47,0.0760917140389588)
--(axis cs:48,0.0757461806904689)
--(axis cs:49,0.0754595059327376)
--(axis cs:49,0.0828049941485027)
--(axis cs:49,0.0828049941485027)
--(axis cs:48,0.0830247389108053)
--(axis cs:47,0.0832558334370054)
--(axis cs:46,0.083695760881041)
--(axis cs:45,0.084139844684276)
--(axis cs:44,0.08438838595361)
--(axis cs:43,0.0844833625562494)
--(axis cs:42,0.084665172969811)
--(axis cs:41,0.0852533977652377)
--(axis cs:40,0.0854967166015027)
--(axis cs:39,0.0857773873705614)
--(axis cs:38,0.0876724579048712)
--(axis cs:37,0.0878551122299856)
--(axis cs:36,0.0895026445418728)
--(axis cs:35,0.0900704148071283)
--(axis cs:34,0.090947646613093)
--(axis cs:33,0.0914083336795484)
--(axis cs:32,0.0921630198636217)
--(axis cs:31,0.0929663760038196)
--(axis cs:30,0.0933261722403928)
--(axis cs:29,0.0945923982183904)
--(axis cs:28,0.0949813610910886)
--(axis cs:27,0.0971579050808155)
--(axis cs:26,0.0998445446937949)
--(axis cs:25,0.100603120824335)
--(axis cs:24,0.101156289061721)
--(axis cs:23,0.101842467196886)
--(axis cs:22,0.102332942351362)
--(axis cs:21,0.103022706653195)
--(axis cs:20,0.104469662262383)
--(axis cs:19,0.106089473166698)
--(axis cs:18,0.107521426393237)
--(axis cs:17,0.110336593068225)
--(axis cs:16,0.111416706373453)
--(axis cs:15,0.113981607382476)
--(axis cs:14,0.115897166893548)
--(axis cs:13,0.117943439884662)
--(axis cs:12,0.123321254757428)
--(axis cs:11,0.125765354872198)
--(axis cs:10,0.129226771660208)
--(axis cs:9,0.131898343752953)
--(axis cs:8,0.136298493637335)
--(axis cs:7,0.141851044896836)
--(axis cs:6,0.147815920191027)
--(axis cs:5,0.156550259957005)
--(axis cs:4,0.169391251269352)
--(axis cs:3,0.178762689246398)
--(axis cs:2,0.18943845267638)
--(axis cs:1,0.214005652371447)
--(axis cs:0,0.281830738462571)
--cycle;

\path [fill=color3, fill opacity=0.2]
(axis cs:0,0.334041016411597)
--(axis cs:0,0.331193340929397)
--(axis cs:1,0.203271736453595)
--(axis cs:2,0.1285351809444)
--(axis cs:3,0.120585677405853)
--(axis cs:4,0.114796060953349)
--(axis cs:5,0.110589761490168)
--(axis cs:6,0.106796236754231)
--(axis cs:7,0.103320366023551)
--(axis cs:8,0.101237619477144)
--(axis cs:9,0.0982345750010617)
--(axis cs:10,0.0957511572295859)
--(axis cs:11,0.0934173690198699)
--(axis cs:12,0.0911832970449107)
--(axis cs:13,0.088880607483475)
--(axis cs:14,0.0868549425186893)
--(axis cs:15,0.0851940523327463)
--(axis cs:16,0.0829878192201246)
--(axis cs:17,0.081011400191831)
--(axis cs:18,0.0800482008160567)
--(axis cs:19,0.0796322478259279)
--(axis cs:20,0.0787305153760433)
--(axis cs:21,0.077275973760059)
--(axis cs:22,0.076288219523941)
--(axis cs:23,0.0756383583929058)
--(axis cs:24,0.0753193467852355)
--(axis cs:25,0.0744201067638166)
--(axis cs:26,0.074103441548386)
--(axis cs:27,0.0736283456421705)
--(axis cs:28,0.0733539717599663)
--(axis cs:29,0.0728753144713763)
--(axis cs:30,0.072644544217263)
--(axis cs:31,0.072192874316855)
--(axis cs:32,0.0718167911952488)
--(axis cs:33,0.0715957184849333)
--(axis cs:34,0.0714409923211525)
--(axis cs:35,0.0710195869753184)
--(axis cs:36,0.0708135949434402)
--(axis cs:37,0.0704113360561497)
--(axis cs:38,0.0700783863940009)
--(axis cs:39,0.0698153094664084)
--(axis cs:40,0.0695109748717099)
--(axis cs:41,0.0688107986740676)
--(axis cs:42,0.0686915338576569)
--(axis cs:43,0.0685161545938808)
--(axis cs:44,0.0681448551236805)
--(axis cs:45,0.0679688943686002)
--(axis cs:46,0.06789797758861)
--(axis cs:47,0.0676912828636428)
--(axis cs:48,0.067574179280659)
--(axis cs:49,0.0673160180002309)
--(axis cs:49,0.0686596955883226)
--(axis cs:49,0.0686596955883226)
--(axis cs:48,0.0689520864369789)
--(axis cs:47,0.0690340509591293)
--(axis cs:46,0.0692932314621798)
--(axis cs:45,0.0693708501838755)
--(axis cs:44,0.0695125151935755)
--(axis cs:43,0.0697672887419524)
--(axis cs:42,0.0700143878741794)
--(axis cs:41,0.0702662716367075)
--(axis cs:40,0.0709936182026196)
--(axis cs:39,0.0713711279196539)
--(axis cs:38,0.0716150588128132)
--(axis cs:37,0.0719926186011866)
--(axis cs:36,0.0723498066631558)
--(axis cs:35,0.0726065143047987)
--(axis cs:34,0.0731526524094244)
--(axis cs:33,0.0732816156629712)
--(axis cs:32,0.0735780241236098)
--(axis cs:31,0.0739540901912147)
--(axis cs:30,0.0743643234167797)
--(axis cs:29,0.07455599611586)
--(axis cs:28,0.0750570858360681)
--(axis cs:27,0.0753204913002534)
--(axis cs:26,0.0757687998528888)
--(axis cs:25,0.0760367583957048)
--(axis cs:24,0.0769083082420409)
--(axis cs:23,0.0773771744219156)
--(axis cs:22,0.0783122499085826)
--(axis cs:21,0.0795208002867675)
--(axis cs:20,0.0808278591593777)
--(axis cs:19,0.0815410720836374)
--(axis cs:18,0.0819874856502612)
--(axis cs:17,0.0829937343373784)
--(axis cs:16,0.0849911648911135)
--(axis cs:15,0.0877184478070281)
--(axis cs:14,0.0894623516811041)
--(axis cs:13,0.0913576359084164)
--(axis cs:12,0.0932681830111565)
--(axis cs:11,0.0955700535536369)
--(axis cs:10,0.097638967108639)
--(axis cs:9,0.10002202696743)
--(axis cs:8,0.103167483163521)
--(axis cs:7,0.105753761036453)
--(axis cs:6,0.108940264113682)
--(axis cs:5,0.112868446951833)
--(axis cs:4,0.117581667522554)
--(axis cs:3,0.123748552915815)
--(axis cs:2,0.132848310083049)
--(axis cs:1,0.20636059018129)
--(axis cs:0,0.334041016411597)
--cycle;

\path [fill=color4, fill opacity=0.2]
(axis cs:0,0.33118041445025)
--(axis cs:0,0.329382002747645)
--(axis cs:1,0.198063583278573)
--(axis cs:2,0.12737732974854)
--(axis cs:3,0.122004935679858)
--(axis cs:4,0.117165615037318)
--(axis cs:5,0.112606670420043)
--(axis cs:6,0.107294278041993)
--(axis cs:7,0.102972639322243)
--(axis cs:8,0.100680143804264)
--(axis cs:9,0.0974601757488282)
--(axis cs:10,0.0953553282750886)
--(axis cs:11,0.09336632119266)
--(axis cs:12,0.0919806236459085)
--(axis cs:13,0.0891731633603904)
--(axis cs:14,0.0882704858299183)
--(axis cs:15,0.0872838656991348)
--(axis cs:16,0.0855015959161754)
--(axis cs:17,0.0844063683850554)
--(axis cs:18,0.0836748879100293)
--(axis cs:19,0.0830505273297814)
--(axis cs:20,0.0819264066134556)
--(axis cs:21,0.0802368621642631)
--(axis cs:22,0.0796869244507635)
--(axis cs:23,0.0791716172018993)
--(axis cs:24,0.0781515119395151)
--(axis cs:25,0.0771500088431258)
--(axis cs:26,0.0767796807899159)
--(axis cs:27,0.0761363512938857)
--(axis cs:28,0.075983417910648)
--(axis cs:29,0.0752002792253987)
--(axis cs:30,0.0742277260786415)
--(axis cs:31,0.0739862582500381)
--(axis cs:32,0.0735755152203173)
--(axis cs:33,0.0733529229852135)
--(axis cs:34,0.0731062256304727)
--(axis cs:35,0.0729290064729419)
--(axis cs:36,0.0724319121698844)
--(axis cs:37,0.0719721939504087)
--(axis cs:38,0.0712764656141925)
--(axis cs:39,0.0708846326252126)
--(axis cs:40,0.0706686730439991)
--(axis cs:41,0.0704225488182095)
--(axis cs:42,0.070172985612183)
--(axis cs:43,0.070081042042666)
--(axis cs:44,0.0697614969840606)
--(axis cs:45,0.0696968453788813)
--(axis cs:46,0.0693290224368255)
--(axis cs:47,0.0691317880677614)
--(axis cs:48,0.0687747224178302)
--(axis cs:49,0.0685195222231164)
--(axis cs:49,0.0705418924567275)
--(axis cs:49,0.0705418924567275)
--(axis cs:48,0.0707427062592564)
--(axis cs:47,0.0711458429468333)
--(axis cs:46,0.0714098012544091)
--(axis cs:45,0.0717856120635039)
--(axis cs:44,0.0718553853717129)
--(axis cs:43,0.0721259960734868)
--(axis cs:42,0.072287066633228)
--(axis cs:41,0.0725310439074254)
--(axis cs:40,0.0727387745200401)
--(axis cs:39,0.0730796789544794)
--(axis cs:38,0.0735034644905904)
--(axis cs:37,0.0742132907229508)
--(axis cs:36,0.0747555509253793)
--(axis cs:35,0.0751291738829501)
--(axis cs:34,0.0752884188069772)
--(axis cs:33,0.0756259765553449)
--(axis cs:32,0.0758712288349037)
--(axis cs:31,0.0763242074967185)
--(axis cs:30,0.0765982331891981)
--(axis cs:29,0.0775460249096775)
--(axis cs:28,0.0784945048346387)
--(axis cs:27,0.0786778265130906)
--(axis cs:26,0.0794830219985787)
--(axis cs:25,0.0800317006789607)
--(axis cs:24,0.0809421510259238)
--(axis cs:23,0.0818482509602627)
--(axis cs:22,0.0824624280682974)
--(axis cs:21,0.0832504241434079)
--(axis cs:20,0.0851197871126336)
--(axis cs:19,0.0859536653620338)
--(axis cs:18,0.0865187713157297)
--(axis cs:17,0.0872186998334386)
--(axis cs:16,0.0881295521846776)
--(axis cs:15,0.0899103626582733)
--(axis cs:14,0.0909794489752799)
--(axis cs:13,0.0921739173116657)
--(axis cs:12,0.0948652185459877)
--(axis cs:11,0.0963812779741815)
--(axis cs:10,0.0979421009573876)
--(axis cs:9,0.100086661567335)
--(axis cs:8,0.103187227377026)
--(axis cs:7,0.105722943057457)
--(axis cs:6,0.109625819403055)
--(axis cs:5,0.115514887348627)
--(axis cs:4,0.120379749405475)
--(axis cs:3,0.125741108446197)
--(axis cs:2,0.131357509821473)
--(axis cs:1,0.201288291286629)
--(axis cs:0,0.33118041445025)
--cycle;

\path [fill=color5, fill opacity=0.2]
(axis cs:0,0.280833441272328)
--(axis cs:0,0.259830937329217)
--(axis cs:1,0.200777506679754)
--(axis cs:2,0.17956603872308)
--(axis cs:3,0.152417340870321)
--(axis cs:4,0.144763956430544)
--(axis cs:5,0.137785305560442)
--(axis cs:6,0.132624514403698)
--(axis cs:7,0.128421430961906)
--(axis cs:8,0.122110322476189)
--(axis cs:9,0.119318119640653)
--(axis cs:10,0.114879230666987)
--(axis cs:11,0.112573198801604)
--(axis cs:12,0.109476914903884)
--(axis cs:13,0.107055881808571)
--(axis cs:14,0.105265244597207)
--(axis cs:15,0.102688752496508)
--(axis cs:16,0.100769608643629)
--(axis cs:17,0.0992480099617816)
--(axis cs:18,0.0968726951592288)
--(axis cs:19,0.0944530506884249)
--(axis cs:20,0.0909783927384198)
--(axis cs:21,0.0892580846789482)
--(axis cs:22,0.0872125283027773)
--(axis cs:23,0.0866576216172464)
--(axis cs:24,0.0858620217710688)
--(axis cs:25,0.0844143212426499)
--(axis cs:26,0.0826211946344434)
--(axis cs:27,0.0820672253936903)
--(axis cs:28,0.0816441508312392)
--(axis cs:29,0.0808742022436401)
--(axis cs:30,0.0797684219552414)
--(axis cs:31,0.0789306746633799)
--(axis cs:32,0.0781146505752387)
--(axis cs:33,0.0774688856662163)
--(axis cs:34,0.0769974530176268)
--(axis cs:35,0.0763857779187652)
--(axis cs:36,0.0756875359128839)
--(axis cs:37,0.0739786585227402)
--(axis cs:38,0.0737332415926437)
--(axis cs:39,0.0736391374729433)
--(axis cs:40,0.0729677302138561)
--(axis cs:41,0.0726389936419154)
--(axis cs:42,0.0718805083159407)
--(axis cs:43,0.0712519261775891)
--(axis cs:44,0.0710561233006776)
--(axis cs:45,0.0708067036365847)
--(axis cs:46,0.0704750719540486)
--(axis cs:47,0.0702101293398831)
--(axis cs:48,0.0696603468168591)
--(axis cs:49,0.069219060818825)
--(axis cs:49,0.0704045235654125)
--(axis cs:49,0.0704045235654125)
--(axis cs:48,0.0709338020359805)
--(axis cs:47,0.0723264517043724)
--(axis cs:46,0.072553142469158)
--(axis cs:45,0.0728075453041816)
--(axis cs:44,0.0732035304648634)
--(axis cs:43,0.0733871211128458)
--(axis cs:42,0.0739508927424543)
--(axis cs:41,0.0746787136484579)
--(axis cs:40,0.0752908934567033)
--(axis cs:39,0.0760654995533225)
--(axis cs:38,0.0761616696214782)
--(axis cs:37,0.0764019710433918)
--(axis cs:36,0.0797079764597801)
--(axis cs:35,0.0803947514832773)
--(axis cs:34,0.0809473564164417)
--(axis cs:33,0.0818003996293165)
--(axis cs:32,0.082564200922841)
--(axis cs:31,0.0833598646081657)
--(axis cs:30,0.0843289528778707)
--(axis cs:29,0.0852779260036486)
--(axis cs:28,0.0859093327604942)
--(axis cs:27,0.0864444212133445)
--(axis cs:26,0.087099826704096)
--(axis cs:25,0.0888978311610885)
--(axis cs:24,0.0913595376991414)
--(axis cs:23,0.0923957432303782)
--(axis cs:22,0.0932154021885817)
--(axis cs:21,0.095685176660269)
--(axis cs:20,0.0988582571647249)
--(axis cs:19,0.102728822391174)
--(axis cs:18,0.105230591227857)
--(axis cs:17,0.107576908563971)
--(axis cs:16,0.109063195750344)
--(axis cs:15,0.110873164743833)
--(axis cs:14,0.113229854282669)
--(axis cs:13,0.114589110046856)
--(axis cs:12,0.117256491000919)
--(axis cs:11,0.119874419863775)
--(axis cs:10,0.122832628405785)
--(axis cs:9,0.127079454733337)
--(axis cs:8,0.13029446672738)
--(axis cs:7,0.136116466579478)
--(axis cs:6,0.140825393432161)
--(axis cs:5,0.147310024650444)
--(axis cs:4,0.153618409569392)
--(axis cs:3,0.160876190508874)
--(axis cs:2,0.190259474186432)
--(axis cs:1,0.213958037902651)
--(axis cs:0,0.280833441272328)
--cycle;

\path [fill=color6, fill opacity=0.2]
(axis cs:0,0.302636698651927)
--(axis cs:0,0.287332127828235)
--(axis cs:1,0.210687929267157)
--(axis cs:2,0.18739451233749)
--(axis cs:3,0.168883246969345)
--(axis cs:4,0.157932360809273)
--(axis cs:5,0.15043197451712)
--(axis cs:6,0.140460345711519)
--(axis cs:7,0.135796383688432)
--(axis cs:8,0.131851204250547)
--(axis cs:9,0.128607547397654)
--(axis cs:10,0.124317770078861)
--(axis cs:11,0.121508437127749)
--(axis cs:12,0.12013101119102)
--(axis cs:13,0.117977744288429)
--(axis cs:14,0.115258557806168)
--(axis cs:15,0.112945785087067)
--(axis cs:16,0.110497383962823)
--(axis cs:17,0.106513358721184)
--(axis cs:18,0.104384673387693)
--(axis cs:19,0.102659495292786)
--(axis cs:20,0.0984551407798155)
--(axis cs:21,0.094360129329161)
--(axis cs:22,0.0931643908244089)
--(axis cs:23,0.0910328625601326)
--(axis cs:24,0.0886863321055475)
--(axis cs:25,0.0875283515694407)
--(axis cs:26,0.0869321328842761)
--(axis cs:27,0.08656470724347)
--(axis cs:28,0.0857779351607057)
--(axis cs:29,0.0847620520726112)
--(axis cs:30,0.0844899865855857)
--(axis cs:31,0.0836448647737782)
--(axis cs:32,0.0828304195678971)
--(axis cs:33,0.0820750032388039)
--(axis cs:34,0.0812988549472723)
--(axis cs:35,0.0808304605321592)
--(axis cs:36,0.0805997356782944)
--(axis cs:37,0.0801165211712552)
--(axis cs:38,0.079816761570283)
--(axis cs:39,0.0796911471928538)
--(axis cs:40,0.0793212819270323)
--(axis cs:41,0.0790471261778756)
--(axis cs:42,0.0785923996105581)
--(axis cs:43,0.0783574391264108)
--(axis cs:44,0.0781302845272085)
--(axis cs:45,0.0778722261721258)
--(axis cs:46,0.0776014175787381)
--(axis cs:47,0.0774086379141683)
--(axis cs:48,0.0772763137925574)
--(axis cs:49,0.0771452296164668)
--(axis cs:49,0.0835163483969649)
--(axis cs:49,0.0835163483969649)
--(axis cs:48,0.0835975403706505)
--(axis cs:47,0.0837816297912907)
--(axis cs:46,0.0841149819817881)
--(axis cs:45,0.0842880087870297)
--(axis cs:44,0.084559726335549)
--(axis cs:43,0.0847619775257363)
--(axis cs:42,0.0849470420392528)
--(axis cs:41,0.085487127600228)
--(axis cs:40,0.0857249378498784)
--(axis cs:39,0.0860184770075364)
--(axis cs:38,0.0862390671373714)
--(axis cs:37,0.0864785920611758)
--(axis cs:36,0.0871321210548238)
--(axis cs:35,0.0873652401709005)
--(axis cs:34,0.0878737379425549)
--(axis cs:33,0.0884707385807674)
--(axis cs:32,0.0891068190039433)
--(axis cs:31,0.0899569557215277)
--(axis cs:30,0.0907618422419895)
--(axis cs:29,0.0910203906278473)
--(axis cs:28,0.0920518665284314)
--(axis cs:27,0.0927091043199056)
--(axis cs:26,0.0930985670069508)
--(axis cs:25,0.0938201196802079)
--(axis cs:24,0.0956635413315315)
--(axis cs:23,0.0975630518937332)
--(axis cs:22,0.099790936281202)
--(axis cs:21,0.100731572379533)
--(axis cs:20,0.104103528444381)
--(axis cs:19,0.107586256840144)
--(axis cs:18,0.109777587388319)
--(axis cs:17,0.112663467409314)
--(axis cs:16,0.116648187015275)
--(axis cs:15,0.118916682456733)
--(axis cs:14,0.121349630079003)
--(axis cs:13,0.123089786679315)
--(axis cs:12,0.125151978926126)
--(axis cs:11,0.12670604822868)
--(axis cs:10,0.12843370442759)
--(axis cs:9,0.131848736913411)
--(axis cs:8,0.136356253338744)
--(axis cs:7,0.1406076294745)
--(axis cs:6,0.148602470742137)
--(axis cs:5,0.15835223207284)
--(axis cs:4,0.165172137878786)
--(axis cs:3,0.175582770807228)
--(axis cs:2,0.197177704968379)
--(axis cs:1,0.224010358112042)
--(axis cs:0,0.302636698651927)
--cycle;

\path [fill=color7, fill opacity=0.2]
(axis cs:0,0.274285619966148)
--(axis cs:0,0.267083947064776)
--(axis cs:1,0.217661592984356)
--(axis cs:2,0.192570608915492)
--(axis cs:3,0.174917699276362)
--(axis cs:4,0.163172949123686)
--(axis cs:5,0.155394963361235)
--(axis cs:6,0.148854307503853)
--(axis cs:7,0.142192705391945)
--(axis cs:8,0.138526673377865)
--(axis cs:9,0.136092434587345)
--(axis cs:10,0.13400827999675)
--(axis cs:11,0.13096586334974)
--(axis cs:12,0.127724843268232)
--(axis cs:13,0.126619176772953)
--(axis cs:14,0.124490145552746)
--(axis cs:15,0.123055228031935)
--(axis cs:16,0.121763811894767)
--(axis cs:17,0.121313943047658)
--(axis cs:18,0.120862276100078)
--(axis cs:19,0.119463522675337)
--(axis cs:20,0.118342938858892)
--(axis cs:21,0.117092379266678)
--(axis cs:22,0.115022697074999)
--(axis cs:23,0.113898749307558)
--(axis cs:24,0.113194762178864)
--(axis cs:25,0.112638179997438)
--(axis cs:26,0.111815348044441)
--(axis cs:27,0.108768100511295)
--(axis cs:28,0.108085239912599)
--(axis cs:29,0.107740507031716)
--(axis cs:30,0.106847662588328)
--(axis cs:31,0.106614719105115)
--(axis cs:32,0.106256751212466)
--(axis cs:33,0.10618375963953)
--(axis cs:34,0.105785493194633)
--(axis cs:35,0.10477049395088)
--(axis cs:36,0.104486632551882)
--(axis cs:37,0.104053084625079)
--(axis cs:38,0.103288402285041)
--(axis cs:39,0.102962169124415)
--(axis cs:40,0.102835875054294)
--(axis cs:41,0.102454339776835)
--(axis cs:42,0.101875340606357)
--(axis cs:43,0.101231362671046)
--(axis cs:44,0.101215758111622)
--(axis cs:45,0.101077807734638)
--(axis cs:46,0.10004635119588)
--(axis cs:47,0.100035873777549)
--(axis cs:48,0.100008784746612)
--(axis cs:49,0.099588071275949)
--(axis cs:49,0.104048307206589)
--(axis cs:49,0.104048307206589)
--(axis cs:48,0.104471890271892)
--(axis cs:47,0.1045312991679)
--(axis cs:46,0.104560517731847)
--(axis cs:45,0.105248468543103)
--(axis cs:44,0.105430598434598)
--(axis cs:43,0.105439413899594)
--(axis cs:42,0.105952767244529)
--(axis cs:41,0.106287683166372)
--(axis cs:40,0.10662913499121)
--(axis cs:39,0.106776513608555)
--(axis cs:38,0.106995469159995)
--(axis cs:37,0.107846718780717)
--(axis cs:36,0.108307857544613)
--(axis cs:35,0.108779480132368)
--(axis cs:34,0.109586823142867)
--(axis cs:33,0.109987192768806)
--(axis cs:32,0.110078756526532)
--(axis cs:31,0.110513525794629)
--(axis cs:30,0.110792591740145)
--(axis cs:29,0.111369491324936)
--(axis cs:28,0.112103664200296)
--(axis cs:27,0.112691891645361)
--(axis cs:26,0.11523512049809)
--(axis cs:25,0.116144507508003)
--(axis cs:24,0.116898962372914)
--(axis cs:23,0.117825654872429)
--(axis cs:22,0.120394176935669)
--(axis cs:21,0.122609285163312)
--(axis cs:20,0.123512160609748)
--(axis cs:19,0.124737639507936)
--(axis cs:18,0.125316800128234)
--(axis cs:17,0.125663378845735)
--(axis cs:16,0.126137565335791)
--(axis cs:15,0.127750284443135)
--(axis cs:14,0.129087683547931)
--(axis cs:13,0.131569565871915)
--(axis cs:12,0.132970385517742)
--(axis cs:11,0.136334980009416)
--(axis cs:10,0.138923165394395)
--(axis cs:9,0.141708098543823)
--(axis cs:8,0.144739959618846)
--(axis cs:7,0.148407006413086)
--(axis cs:6,0.154426814434929)
--(axis cs:5,0.161112038583284)
--(axis cs:4,0.169417503065519)
--(axis cs:3,0.181017541369332)
--(axis cs:2,0.197217736966735)
--(axis cs:1,0.22818005807036)
--(axis cs:0,0.274285619966148)
--cycle;

\addplot [line width=2pt, color0]
table {%
0 0.330019849363924
1 0.198596984709045
2 0.193404906574894
3 0.158473002819793
4 0.148585050840535
5 0.137979044495944
6 0.131022455932964
7 0.125915521856789
8 0.119741794800624
9 0.1157724516078
10 0.11341700230722
11 0.107836327266608
12 0.10442896494003
13 0.101679733474922
14 0.0993330992717294
15 0.0966269204440286
16 0.0957569891473119
17 0.0936506266880618
18 0.0918617523941938
19 0.0896630580645071
20 0.0889636788328353
21 0.087377017908189
22 0.0860547886180858
23 0.084465740652312
24 0.0833224281085754
25 0.0823420737583549
26 0.0816891480157617
27 0.0806367354208677
28 0.0797916737445315
29 0.07782647254319
30 0.0771087139505978
31 0.0763124374117067
32 0.075209146360718
33 0.0744968037075375
34 0.0738232388040087
35 0.0731790327591683
36 0.072769102918388
37 0.0723918059925646
38 0.0718117653033636
39 0.0715527773313552
40 0.0713302294524697
41 0.0712622954636113
42 0.0705958271982329
43 0.0702319237143231
44 0.0698342644843994
45 0.0695441410026495
46 0.069134937614665
47 0.0688227116902737
48 0.0686849883670566
49 0.0684730873797754
};
\addplot [line width=2pt, color1]
table {%
0 0.270755089248703
1 0.216582690777305
2 0.190845293980062
3 0.175099915116254
4 0.165077074507575
5 0.159014525572251
6 0.150201622465561
7 0.148114353536277
8 0.144395510333098
9 0.141616124792463
10 0.13655686429021
11 0.132590916172664
12 0.128946044002683
13 0.127391839046372
14 0.126255397754469
15 0.12390571932059
16 0.122707860025634
17 0.12204817534997
18 0.121154149236221
19 0.120528466712541
20 0.112299148005493
21 0.108957099788068
22 0.107292689730137
23 0.105032305068969
24 0.10216230748992
25 0.100828603055395
26 0.1001077680519
27 0.0994121399994683
28 0.0986802720843932
29 0.0963902009859747
30 0.0950808672804135
31 0.0945687857893931
32 0.0938481839632141
33 0.0935236074142193
34 0.093170522847904
35 0.0931037975872132
36 0.0927834815206468
37 0.0924638501528219
38 0.0919550498494201
39 0.091839975747277
40 0.0917984374520417
41 0.0917011068083398
42 0.0913987923550844
43 0.0897694635293633
44 0.0896340943360936
45 0.0892314084314978
46 0.0891050374151983
47 0.0890970033967476
48 0.0890367886842077
49 0.0886925751072355
};
\addplot [line width=2pt, color2]
table {%
0 0.275242407226699
1 0.209344203550294
2 0.186708425677184
3 0.175564454446305
4 0.166586067275416
5 0.154034768140985
6 0.145160554696798
7 0.139693877869384
8 0.133642245176754
9 0.12934079460454
10 0.126822533249634
11 0.123241265055421
12 0.120197582876699
13 0.115237496561263
14 0.113148956611469
15 0.110988991645224
16 0.107860170204358
17 0.106659162810603
18 0.104151480563819
19 0.102696308843193
20 0.100972225466986
21 0.0993532671930573
22 0.0986496035711062
23 0.0981337412700866
24 0.0975140283281753
25 0.0969772711538376
26 0.0962045637266849
27 0.0933873204518617
28 0.091508693090686
29 0.0911157461320413
30 0.0900880489955693
31 0.089737715525799
32 0.0890469620541722
33 0.0882208780233643
34 0.0877297070054902
35 0.0867917893579777
36 0.0861916881812364
37 0.0846421820001246
38 0.0844489225221896
39 0.0824010296093366
40 0.0820894810918815
41 0.0818242855827836
42 0.0811059924241957
43 0.0808929500465192
44 0.0808012458596808
45 0.0805685464327618
46 0.0800926831963943
47 0.0796737737379821
48 0.0793854598006371
49 0.0791322500406201
};
\addplot [line width=2pt, color3]
table {%
0 0.332617178670497
1 0.204816163317442
2 0.130691745513725
3 0.122167115160834
4 0.116188864237951
5 0.111729104221
6 0.107868250433957
7 0.104537063530002
8 0.102202551320332
9 0.0991283009842461
10 0.0966950621691125
11 0.0944937112867534
12 0.0922257400280336
13 0.0901191216959457
14 0.0881586470998967
15 0.0864562500698872
16 0.083989492055619
17 0.0820025672646047
18 0.0810178432331589
19 0.0805866599547826
20 0.0797791872677105
21 0.0783983870234133
22 0.0773002347162618
23 0.0765077664074107
24 0.0761138275136382
25 0.0752284325797607
26 0.0749361207006374
27 0.074474418471212
28 0.0742055287980172
29 0.0737156552936182
30 0.0735044338170214
31 0.0730734822540348
32 0.0726974076594293
33 0.0724386670739523
34 0.0722968223652884
35 0.0718130506400586
36 0.071581700803298
37 0.0712019773286681
38 0.0708467226034071
39 0.0705932186930312
40 0.0702522965371647
41 0.0695385351553875
42 0.0693529608659181
43 0.0691417216679166
44 0.068828685158628
45 0.0686698722762379
46 0.0685956045253949
47 0.068362666911386
48 0.068263132858819
49 0.0679878567942767
};
\addplot [line width=2pt, color4]
table {%
0 0.330281208598948
1 0.199675937282601
2 0.129367419785006
3 0.123873022063028
4 0.118772682221396
5 0.114060778884335
6 0.108460048722524
7 0.10434779118985
8 0.101933685590645
9 0.0987734186580819
10 0.0966487146162381
11 0.0948737995834207
12 0.0934229210959481
13 0.090673540336028
14 0.0896249674025991
15 0.0885971141787041
16 0.0868155740504265
17 0.085812534109247
18 0.0850968296128795
19 0.0845020963459076
20 0.0835230968630446
21 0.0817436431538355
22 0.0810746762595304
23 0.080509934081081
24 0.0795468314827194
25 0.0785908547610433
26 0.0781313513942473
27 0.0774070889034882
28 0.0772389613726434
29 0.0763731520675381
30 0.0754129796339198
31 0.0751552328733783
32 0.0747233720276105
33 0.0744894497702792
34 0.074197322218725
35 0.074029090177946
36 0.0735937315476318
37 0.0730927423366798
38 0.0723899650523914
39 0.071982155789846
40 0.0717037237820196
41 0.0714767963628175
42 0.0712300261227055
43 0.0711035190580764
44 0.0708084411778867
45 0.0707412287211926
46 0.0703694118456173
47 0.0701388155072974
48 0.0697587143385433
49 0.069530707339922
};
\addplot [line width=2pt, color5]
table {%
0 0.270332189300773
1 0.207367772291202
2 0.184912756454756
3 0.156646765689598
4 0.149191182999968
5 0.142547665105443
6 0.136724953917929
7 0.132268948770692
8 0.126202394601785
9 0.123198787186995
10 0.118855929536386
11 0.116223809332689
12 0.113366702952402
13 0.110822495927714
14 0.109247549439938
15 0.10678095862017
16 0.104916402196986
17 0.103412459262876
18 0.101051643193543
19 0.0985909365397993
20 0.0949183249515724
21 0.0924716306696086
22 0.0902139652456795
23 0.0895266824238123
24 0.0886107797351051
25 0.0866560762018692
26 0.0848605106692697
27 0.0842558233035174
28 0.0837767417958667
29 0.0830760641236443
30 0.082048687416556
31 0.0811452696357728
32 0.0803394257490399
33 0.0796346426477664
34 0.0789724047170343
35 0.0783902647010213
36 0.077697756186332
37 0.075190314783066
38 0.074947455607061
39 0.0748523185131329
40 0.0741293118352797
41 0.0736588536451867
42 0.0729157005291975
43 0.0723195236452175
44 0.0721298268827705
45 0.0718071244703831
46 0.0715141072116033
47 0.0712682905221278
48 0.0702970744264198
49 0.0698117921921188
};
\addplot [line width=2pt, color6]
table {%
0 0.294984413240081
1 0.2173491436896
2 0.192286108652935
3 0.172233008888286
4 0.16155224934403
5 0.15439210329498
6 0.144531408226828
7 0.138202006581466
8 0.134103728794646
9 0.130228142155532
10 0.126375737253225
11 0.124107242678214
12 0.122641495058573
13 0.120533765483872
14 0.118304093942586
15 0.1159312337719
16 0.113572785489049
17 0.109588413065249
18 0.107081130388006
19 0.105122876066465
20 0.101279334612098
21 0.0975458508543471
22 0.0964776635528054
23 0.0942979572269329
24 0.0921749367185395
25 0.0906742356248243
26 0.0900153499456134
27 0.0896369057816878
28 0.0889149008445686
29 0.0878912213502293
30 0.0876259144137876
31 0.0868009102476529
32 0.0859686192859202
33 0.0852728709097857
34 0.0845862964449136
35 0.0840978503515299
36 0.0838659283665591
37 0.0832975566162155
38 0.0830279143538272
39 0.0828548121001951
40 0.0825231098884554
41 0.0822671268890518
42 0.0817697208249055
43 0.0815597083260736
44 0.0813450054313787
45 0.0810801174795778
46 0.0808581997802631
47 0.0805951338527295
48 0.080436927081604
49 0.0803307890067159
};
\addplot [line width=2pt, color7]
table {%
0 0.270684783515462
1 0.222920825527358
2 0.194894172941114
3 0.177967620322847
4 0.166295226094602
5 0.15825350097226
6 0.151640560969391
7 0.145299855902516
8 0.141633316498355
9 0.138900266565584
10 0.136465722695573
11 0.133650421679578
12 0.130347614392987
13 0.129094371322434
14 0.126788914550339
15 0.125402756237535
16 0.123950688615279
17 0.123488660946696
18 0.123089538114156
19 0.122100581091636
20 0.12092754973432
21 0.119850832214995
22 0.117708437005334
23 0.115862202089993
24 0.115046862275889
25 0.114391343752721
26 0.113525234271266
27 0.110729996078328
28 0.110094452056447
29 0.109554999178326
30 0.108820127164236
31 0.108564122449872
32 0.108167753869499
33 0.108085476204168
34 0.10768615816875
35 0.106774987041624
36 0.106397245048247
37 0.105949901702898
38 0.105141935722518
39 0.104869341366485
40 0.104732505022752
41 0.104371011471604
42 0.103914053925443
43 0.10333538828532
44 0.10332317827311
45 0.103163138138871
46 0.102303434463863
47 0.102283586472725
48 0.102240337509252
49 0.101818189241269
};

\end{axis}

\end{tikzpicture}

%% file: icml2023/figures/riemann.tex
\pgfmathdeclarefunction{gauss}{3}{%
  \pgfmathparse{1/(#2*sqrt(2*pi))*#3*exp(-((x-#1)^2)/(2*#2^2))}%
}
\usepgfplotslibrary{fillbetween}

 \begin{tikzpicture}
 \begin{axis}[width=\textwidth,
     height=.98\textwidth,
     axis lines=center,
     axis y line*=left,
     xmin=-4,xmax=4,
     ymin=0,ymax=.5,
     grid=major,
     yminorgrids,
     ylabel near ticks,
     xlabel near ticks,
     major grid style={thick},
     tick style={thick},
     axis line style={thick},
     xlabel=$y_q$,
     ylabel=Density,
     xtick={-3,-1,0,1,3},
     ytick={0,.2,.4},
     ticklabel style={font=\tiny},
     label style={font=\tiny},
     enlarge x limits={upper,value=0.02},
     enlarge y limits={upper,value=0.1},
     ]
      \addplot [name path=A,domain=-10:-1.5] {gauss(-1.5, .75, .1*2)}; 
 \addplot[name path=line, domain=-10:-1.5, gray, no markers, line width=1pt] {0};
 \addplot[BarStyle] fill between [of=A and line];
    \addplot[BarStyle]coordinates{(-1.5,.15)(-.5,.15)};
    \addplot[BarStyle]coordinates{(-.5,.25)(0,.25)};
    \addplot[BarStyle]coordinates{(.0,.35)(.5,.25)};
    \addplot[BarStyle]coordinates{(.5,.20)(2,.15)};
     \addplot [name path=B,domain=2:10] {gauss(2, .5,.05*2)}; 
 \addplot[name path=line, domain=2:10, gray, no markers, line width=1pt] {0};
 \addplot[BarStyle] fill between [of=B and line];
  \end{axis}
 \end{tikzpicture}

%% file: icml2023/figures/hpob_nonmyopic_rank_test.tex
\begin{tikzpicture}

\begin{axis}[
legend cell align={left},
legend style={
  fill opacity=0.8,
  draw opacity=1,
  text opacity=1,
  at={(0.55,-0.15)},
  anchor=south,
  draw=white!80!black
},
tick align=outside,
tick pos=left,
x grid style={white!69.0196078431373!black},
xlabel={Number of trials},
xmin=2.75, xmax=52.25,
xtick style={color=black},
y grid style={white!69.0196078431373!black},
ylabel={Average Rank},
ymin=2.90743197362865, ymax=5.28532559341556,
ytick style={color=black},
height=.6\textwidth,
width=.8\textwidth,
]
\path [fill=colorRandom, fill opacity=0.2]
(axis cs:5,4)
--(axis cs:5,4)
--(axis cs:6,4.16155915109726)
--(axis cs:7,4.38420555034615)
--(axis cs:8,4.48100327902876)
--(axis cs:9,4.415365573135)
--(axis cs:10,4.38498149594354)
--(axis cs:11,4.37270239290935)
--(axis cs:12,4.34773008204993)
--(axis cs:13,4.43407925664706)
--(axis cs:14,4.52388951594511)
--(axis cs:15,4.54823935639246)
--(axis cs:16,4.57666624174489)
--(axis cs:17,4.5881843253352)
--(axis cs:18,4.64512943976353)
--(axis cs:19,4.64063910068535)
--(axis cs:20,4.68179527595437)
--(axis cs:21,4.70852234484819)
--(axis cs:22,4.73100633297473)
--(axis cs:23,4.77532294563437)
--(axis cs:24,4.81459488372769)
--(axis cs:25,4.86943468306064)
--(axis cs:26,4.82252751769557)
--(axis cs:27,4.84288409298926)
--(axis cs:28,4.81359188823285)
--(axis cs:29,4.87037209811011)
--(axis cs:30,4.87171995274422)
--(axis cs:31,4.8893768443235)
--(axis cs:32,4.90623539166014)
--(axis cs:33,4.94419305199581)
--(axis cs:34,4.95048783373275)
--(axis cs:35,4.96833304554446)
--(axis cs:36,4.95850880691364)
--(axis cs:37,4.96873623314693)
--(axis cs:38,4.94018145145583)
--(axis cs:39,4.92013700076322)
--(axis cs:40,4.92013700076322)
--(axis cs:41,4.92991401145052)
--(axis cs:42,4.92868364128028)
--(axis cs:43,4.89229999601334)
--(axis cs:44,4.90317269451846)
--(axis cs:45,4.90170784863217)
--(axis cs:46,4.900547745323)
--(axis cs:47,4.91840178425564)
--(axis cs:48,4.92285038956239)
--(axis cs:49,4.92709461930505)
--(axis cs:50,4.9490762696848)
--(axis cs:50,5.17723951978889)
--(axis cs:50,5.17723951978889)
--(axis cs:49,5.15711590701074)
--(axis cs:48,5.17188645254287)
--(axis cs:47,5.16580874206015)
--(axis cs:46,5.13103120204542)
--(axis cs:45,5.12987109873626)
--(axis cs:44,5.11787993706049)
--(axis cs:43,5.10770000398666)
--(axis cs:42,5.10289530608814)
--(axis cs:41,5.1016649359179)
--(axis cs:40,5.12196826239468)
--(axis cs:39,5.12196826239468)
--(axis cs:38,5.13350275907048)
--(axis cs:37,5.14705324053729)
--(axis cs:36,5.13622803519162)
--(axis cs:35,5.1264037965608)
--(axis cs:34,5.14424900837252)
--(axis cs:33,5.16107010589893)
--(axis cs:32,5.14639618728723)
--(axis cs:31,5.12114947146597)
--(axis cs:30,5.10722741567683)
--(axis cs:29,5.10857527031094)
--(axis cs:28,5.07061863808294)
--(axis cs:27,5.07290538069495)
--(axis cs:26,5.05115669283075)
--(axis cs:25,5.08846005378146)
--(axis cs:24,5.02751037943021)
--(axis cs:23,5.01415073857615)
--(axis cs:22,4.98478314070948)
--(axis cs:21,4.95463554988866)
--(axis cs:20,4.91820472404563)
--(axis cs:19,4.87515037299886)
--(axis cs:18,4.9022389812891)
--(axis cs:17,4.86444725361217)
--(axis cs:16,4.9075442845709)
--(axis cs:15,4.86228695939702)
--(axis cs:14,4.81295258931805)
--(axis cs:13,4.76592074335294)
--(axis cs:12,4.63121728637113)
--(axis cs:11,4.69045550182749)
--(axis cs:10,4.60449218826699)
--(axis cs:9,4.57410811107553)
--(axis cs:8,4.59268093149756)
--(axis cs:7,4.61579444965385)
--(axis cs:6,4.36475663837642)
--(axis cs:5,4)
--cycle;

\path [fill=colorHEBO, fill opacity=0.2]
(axis cs:5,4)
--(axis cs:5,4)
--(axis cs:6,4.08124414534368)
--(axis cs:7,3.88652631578947)
--(axis cs:8,3.70736221620398)
--(axis cs:9,3.71829541147022)
--(axis cs:10,3.85189134439623)
--(axis cs:11,3.7380469093442)
--(axis cs:12,3.64489290309281)
--(axis cs:13,3.5259620063104)
--(axis cs:14,3.53459431522331)
--(axis cs:15,3.48757028541569)
--(axis cs:16,3.48464493935682)
--(axis cs:17,3.45633879796972)
--(axis cs:18,3.44543325756026)
--(axis cs:19,3.35909954092314)
--(axis cs:20,3.24142183798808)
--(axis cs:21,3.17713515702771)
--(axis cs:22,3.13868515520189)
--(axis cs:23,3.12089166182609)
--(axis cs:24,3.1532456490189)
--(axis cs:25,3.18596966469014)
--(axis cs:26,3.21048414616034)
--(axis cs:27,3.1873993154863)
--(axis cs:28,3.17988076341423)
--(axis cs:29,3.22627719305552)
--(axis cs:30,3.19134155047595)
--(axis cs:31,3.20942752306379)
--(axis cs:32,3.14672616530556)
--(axis cs:33,3.14487377843601)
--(axis cs:34,3.15818434553612)
--(axis cs:35,3.19571421255851)
--(axis cs:36,3.18864322171907)
--(axis cs:37,3.12941154317228)
--(axis cs:38,3.10457726987421)
--(axis cs:39,3.09297769347396)
--(axis cs:40,3.0862664369235)
--(axis cs:41,3.05525551304981)
--(axis cs:42,3.06416260851966)
--(axis cs:43,3.07632443427178)
--(axis cs:44,3.09405955257629)
--(axis cs:45,3.08210615902267)
--(axis cs:46,3.04975939812412)
--(axis cs:47,3.01704883534304)
--(axis cs:48,3.0229372362159)
--(axis cs:49,3.01551804725533)
--(axis cs:50,3.03316043291504)
--(axis cs:50,3.14578693550601)
--(axis cs:50,3.14578693550601)
--(axis cs:49,3.12132405800783)
--(axis cs:48,3.12443118483673)
--(axis cs:47,3.11979326992012)
--(axis cs:46,3.17129323345483)
--(axis cs:45,3.18105173571417)
--(axis cs:44,3.17962465795002)
--(axis cs:43,3.16578082888612)
--(axis cs:42,3.15689002305928)
--(axis cs:41,3.15527080273966)
--(axis cs:40,3.19794408939229)
--(axis cs:39,3.2017591486313)
--(axis cs:38,3.20068588802053)
--(axis cs:37,3.20743056209087)
--(axis cs:36,3.23240940985988)
--(axis cs:35,3.25691736638886)
--(axis cs:34,3.22076302288493)
--(axis cs:33,3.20249464261662)
--(axis cs:32,3.26380015048391)
--(axis cs:31,3.33794089798884)
--(axis cs:30,3.31392160741879)
--(axis cs:29,3.39477543852343)
--(axis cs:28,3.31485607869103)
--(axis cs:27,3.32839015819791)
--(axis cs:26,3.38951585383966)
--(axis cs:25,3.36139875636249)
--(axis cs:24,3.33096487729689)
--(axis cs:23,3.30016096975286)
--(axis cs:22,3.28236747637705)
--(axis cs:21,3.2965490534986)
--(axis cs:20,3.36910447780139)
--(axis cs:19,3.47247940644528)
--(axis cs:18,3.50193516349237)
--(axis cs:17,3.51208225466186)
--(axis cs:16,3.57851295538002)
--(axis cs:15,3.57558760932115)
--(axis cs:14,3.66540568477669)
--(axis cs:13,3.69509062526855)
--(axis cs:12,3.80773867585456)
--(axis cs:11,3.85142677486633)
--(axis cs:10,4.10600339244588)
--(axis cs:9,4.00802037800346)
--(axis cs:8,3.98737462590128)
--(axis cs:7,4.11347368421053)
--(axis cs:6,4.26612427570895)
--(axis cs:5,4)
--cycle;

\path [fill=colorGP, fill opacity=0.2]
(axis cs:5,4)
--(axis cs:5,4)
--(axis cs:6,3.63808450265565)
--(axis cs:7,3.69679275271297)
--(axis cs:8,3.91791602680927)
--(axis cs:9,3.8423877444195)
--(axis cs:10,3.92251237724555)
--(axis cs:11,3.99725975273666)
--(axis cs:12,4.09884254541451)
--(axis cs:13,4.14524818036358)
--(axis cs:14,4.10472543249483)
--(axis cs:15,4.19979794646277)
--(axis cs:16,4.22143637979417)
--(axis cs:17,4.21695949658296)
--(axis cs:18,4.16946008377846)
--(axis cs:19,4.23193675092023)
--(axis cs:20,4.17840964302115)
--(axis cs:21,4.24274762807603)
--(axis cs:22,4.26526315789474)
--(axis cs:23,4.21473242303283)
--(axis cs:24,4.08262191515819)
--(axis cs:25,4.12605664889601)
--(axis cs:26,4.13383010371806)
--(axis cs:27,4.16518245837768)
--(axis cs:28,4.16655510241943)
--(axis cs:29,4.19221509916109)
--(axis cs:30,4.22595301217391)
--(axis cs:31,4.25002375144066)
--(axis cs:32,4.23982295435819)
--(axis cs:33,4.25462290801341)
--(axis cs:34,4.26485297096455)
--(axis cs:35,4.26206290995206)
--(axis cs:36,4.28616526722228)
--(axis cs:37,4.3100661351105)
--(axis cs:38,4.32740725567427)
--(axis cs:39,4.34454328119227)
--(axis cs:40,4.35383320080931)
--(axis cs:41,4.36813133860093)
--(axis cs:42,4.38083641090929)
--(axis cs:43,4.38657390190697)
--(axis cs:44,4.32870676654517)
--(axis cs:45,4.35635047122076)
--(axis cs:46,4.32036920640913)
--(axis cs:47,4.30627543971516)
--(axis cs:48,4.25861744328887)
--(axis cs:49,4.25957894736842)
--(axis cs:50,4.24440971743753)
--(axis cs:50,4.39769554572036)
--(axis cs:50,4.39769554572036)
--(axis cs:49,4.42463157894737)
--(axis cs:48,4.42559308302692)
--(axis cs:47,4.44109298133748)
--(axis cs:46,4.44805184622244)
--(axis cs:45,4.47522847614766)
--(axis cs:44,4.45024060187589)
--(axis cs:43,4.52921557177724)
--(axis cs:42,4.53495306277492)
--(axis cs:41,4.52660550350433)
--(axis cs:40,4.50932469392753)
--(axis cs:39,4.48703566617615)
--(axis cs:38,4.45154011274678)
--(axis cs:37,4.42677597015266)
--(axis cs:36,4.40857157488298)
--(axis cs:35,4.38004235320583)
--(axis cs:34,4.40883123956177)
--(axis cs:33,4.39800867093396)
--(axis cs:32,4.41280862458918)
--(axis cs:31,4.49734466961197)
--(axis cs:30,4.47931014572083)
--(axis cs:29,4.43936384820733)
--(axis cs:28,4.44397121337004)
--(axis cs:27,4.43481754162232)
--(axis cs:26,4.41353831733457)
--(axis cs:25,4.42131177215662)
--(axis cs:24,4.39106229536812)
--(axis cs:23,4.44842547170401)
--(axis cs:22,4.47157894736842)
--(axis cs:21,4.43093658245029)
--(axis cs:20,4.36895877803148)
--(axis cs:19,4.43122114381661)
--(axis cs:18,4.35685570569523)
--(axis cs:17,4.38304050341704)
--(axis cs:16,4.35751098862688)
--(axis cs:15,4.38967573774776)
--(axis cs:14,4.29527456750517)
--(axis cs:13,4.47580445121537)
--(axis cs:12,4.39589429669076)
--(axis cs:11,4.3395823525265)
--(axis cs:10,4.24590867538603)
--(axis cs:9,4.13655962400156)
--(axis cs:8,4.05050502582231)
--(axis cs:7,3.80847040518177)
--(axis cs:6,3.86717865523909)
--(axis cs:5,4)
--cycle;

\path [fill=colorDNGO, fill opacity=0.2]
(axis cs:5,4)
--(axis cs:5,4)
--(axis cs:6,3.96382891773875)
--(axis cs:7,4.07350944910666)
--(axis cs:8,3.95893376798898)
--(axis cs:9,3.90674871866453)
--(axis cs:10,3.95853916552842)
--(axis cs:11,3.939679292554)
--(axis cs:12,4.00592485560329)
--(axis cs:13,3.93172338408822)
--(axis cs:14,3.94655186986635)
--(axis cs:15,4.01601699366982)
--(axis cs:16,4.04003089443367)
--(axis cs:17,4.04594526249614)
--(axis cs:18,4.08474225770585)
--(axis cs:19,4.11748454991191)
--(axis cs:20,4.19094736842105)
--(axis cs:21,4.18067033433876)
--(axis cs:22,4.19790953662372)
--(axis cs:23,4.18202098251116)
--(axis cs:24,4.2143526376405)
--(axis cs:25,4.17522877956776)
--(axis cs:26,4.19669158301176)
--(axis cs:27,4.22892566252808)
--(axis cs:28,4.26857657745301)
--(axis cs:29,4.19525706841022)
--(axis cs:30,4.18972300420924)
--(axis cs:31,4.21272793778499)
--(axis cs:32,4.24726669302371)
--(axis cs:33,4.26594986000511)
--(axis cs:34,4.25626661893274)
--(axis cs:35,4.28714459434467)
--(axis cs:36,4.31632980968912)
--(axis cs:37,4.29738920065112)
--(axis cs:38,4.22828864641818)
--(axis cs:39,4.24396962295457)
--(axis cs:40,4.24396962295457)
--(axis cs:41,4.25298938249918)
--(axis cs:42,4.25561744280601)
--(axis cs:43,4.24503042766992)
--(axis cs:44,4.25918359421472)
--(axis cs:45,4.28526727192311)
--(axis cs:46,4.2872841404059)
--(axis cs:47,4.23497837999187)
--(axis cs:48,4.23854995389532)
--(axis cs:49,4.24498513409127)
--(axis cs:50,4.25336548671893)
--(axis cs:50,4.43084503959686)
--(axis cs:50,4.43084503959686)
--(axis cs:49,4.47080433959294)
--(axis cs:48,4.46671320399941)
--(axis cs:47,4.45975846211339)
--(axis cs:46,4.50218954380463)
--(axis cs:45,4.5147327280769)
--(axis cs:44,4.48818482683791)
--(axis cs:43,4.46023273022482)
--(axis cs:42,4.4601720308782)
--(axis cs:41,4.45227377539556)
--(axis cs:40,4.44024090336122)
--(axis cs:39,4.44024090336122)
--(axis cs:38,4.42434293252918)
--(axis cs:37,4.44997922040151)
--(axis cs:36,4.45209124294246)
--(axis cs:35,4.41811856355007)
--(axis cs:34,4.40689127580411)
--(axis cs:33,4.41826066631068)
--(axis cs:32,4.41589120171313)
--(axis cs:31,4.37674574642554)
--(axis cs:30,4.34711910105392)
--(axis cs:29,4.35211135264241)
--(axis cs:28,4.39458131728383)
--(axis cs:27,4.36054802168245)
--(axis cs:26,4.31909789067245)
--(axis cs:25,4.33003437832698)
--(axis cs:24,4.43827894130687)
--(axis cs:23,4.39692638590989)
--(axis cs:22,4.41261677916575)
--(axis cs:21,4.36669808671388)
--(axis cs:20,4.33536842105263)
--(axis cs:19,4.26146281850914)
--(axis cs:18,4.2520998475573)
--(axis cs:17,4.20668631645123)
--(axis cs:16,4.18102173714528)
--(axis cs:15,4.15240405896176)
--(axis cs:14,4.13765865644944)
--(axis cs:13,4.13143451064862)
--(axis cs:12,4.09933830229145)
--(axis cs:11,4.03926807586705)
--(axis cs:10,4.07303978184001)
--(axis cs:9,4.10377759712494)
--(axis cs:8,4.14632938990575)
--(axis cs:7,4.27385897194597)
--(axis cs:6,4.17301318752441)
--(axis cs:5,4)
--cycle;

\path [fill=colorDGP, fill opacity=0.2]
(axis cs:5,4)
--(axis cs:5,4)
--(axis cs:6,4.37845825493894)
--(axis cs:7,4.27220238748611)
--(axis cs:8,4.25580855207439)
--(axis cs:9,4.31531140996676)
--(axis cs:10,4.12977384441772)
--(axis cs:11,4.28483615664416)
--(axis cs:12,4.30424265066314)
--(axis cs:13,4.37774839785069)
--(axis cs:14,4.27569113260232)
--(axis cs:15,4.29306264486269)
--(axis cs:16,4.26541946566607)
--(axis cs:17,4.23617044188806)
--(axis cs:18,4.19008066889669)
--(axis cs:19,4.14936414917618)
--(axis cs:20,4.13083871281728)
--(axis cs:21,4.16540905108648)
--(axis cs:22,4.14094694250147)
--(axis cs:23,4.17280962163578)
--(axis cs:24,4.21381529851486)
--(axis cs:25,4.21799633513615)
--(axis cs:26,4.20285301359972)
--(axis cs:27,4.16829148234274)
--(axis cs:28,4.21594499850564)
--(axis cs:29,4.22892566252808)
--(axis cs:30,4.2747756767269)
--(axis cs:31,4.2840291376659)
--(axis cs:32,4.27995965024147)
--(axis cs:33,4.32477175385829)
--(axis cs:34,4.32326695615084)
--(axis cs:35,4.3096080377384)
--(axis cs:36,4.28769302457019)
--(axis cs:37,4.31582064080715)
--(axis cs:38,4.33820514456003)
--(axis cs:39,4.32406969534019)
--(axis cs:40,4.33080700832238)
--(axis cs:41,4.33870776378223)
--(axis cs:42,4.31510358566368)
--(axis cs:43,4.3078937543947)
--(axis cs:44,4.32292049784437)
--(axis cs:45,4.32981979873712)
--(axis cs:46,4.34440874282241)
--(axis cs:47,4.3817845557119)
--(axis cs:48,4.38704771360664)
--(axis cs:49,4.39211390795599)
--(axis cs:50,4.40743005420168)
--(axis cs:50,4.49783310369306)
--(axis cs:50,4.49783310369306)
--(axis cs:49,4.48157030257033)
--(axis cs:48,4.48663649691968)
--(axis cs:47,4.48137333902494)
--(axis cs:46,4.46611757296706)
--(axis cs:45,4.43860125389446)
--(axis cs:44,4.44550055478721)
--(axis cs:43,4.41842203507899)
--(axis cs:42,4.41121220381)
--(axis cs:41,4.4507659204283)
--(axis cs:40,4.4376140443092)
--(axis cs:39,4.43382504150192)
--(axis cs:38,4.44074222386103)
--(axis cs:37,4.42102146445601)
--(axis cs:36,4.38599118595613)
--(axis cs:35,4.45881301489318)
--(axis cs:34,4.46620672805968)
--(axis cs:33,4.46470193035224)
--(axis cs:32,4.41477719186379)
--(axis cs:31,4.41070770443936)
--(axis cs:30,4.39890853379942)
--(axis cs:29,4.36054802168245)
--(axis cs:28,4.34194973833646)
--(axis cs:27,4.31591904397305)
--(axis cs:26,4.36556803903186)
--(axis cs:25,4.35042471749543)
--(axis cs:24,4.38618470148514)
--(axis cs:23,4.41666406257474)
--(axis cs:22,4.39589516276169)
--(axis cs:21,4.44511726470299)
--(axis cs:20,4.38495076086693)
--(axis cs:19,4.38747795608698)
--(axis cs:18,4.44149827847174)
--(axis cs:17,4.51119797916458)
--(axis cs:16,4.53458053433393)
--(axis cs:15,4.51746367092678)
--(axis cs:14,4.57694044634505)
--(axis cs:13,4.60119897057037)
--(axis cs:12,4.59049419144212)
--(axis cs:11,4.50463752756636)
--(axis cs:10,4.47022615558228)
--(axis cs:9,4.58995174792798)
--(axis cs:8,4.59682302687298)
--(axis cs:7,4.54885024409284)
--(axis cs:6,4.5268049029558)
--(axis cs:5,4)
--cycle;

\path [fill=colorPFN_GP, fill opacity=0.2]
(axis cs:5,4)
--(axis cs:5,4)
--(axis cs:6,3.57081848814285)
--(axis cs:7,3.55571793350874)
--(axis cs:8,3.55960258547427)
--(axis cs:9,3.5367528168164)
--(axis cs:10,3.31275855814956)
--(axis cs:11,3.3070356401754)
--(axis cs:12,3.30333871450665)
--(axis cs:13,3.32649739487964)
--(axis cs:14,3.40903698140705)
--(axis cs:15,3.31950875348687)
--(axis cs:16,3.33567762112101)
--(axis cs:17,3.36933517198559)
--(axis cs:18,3.37866775579169)
--(axis cs:19,3.43217376929583)
--(axis cs:20,3.4501603863209)
--(axis cs:21,3.4537377771901)
--(axis cs:22,3.42160684336539)
--(axis cs:23,3.39121852437713)
--(axis cs:24,3.43569401718127)
--(axis cs:25,3.41098175695001)
--(axis cs:26,3.44621708437331)
--(axis cs:27,3.4710006287067)
--(axis cs:28,3.47846550615554)
--(axis cs:29,3.39710469391186)
--(axis cs:30,3.41943110860644)
--(axis cs:31,3.31221052631579)
--(axis cs:32,3.34663795758638)
--(axis cs:33,3.34795652027564)
--(axis cs:34,3.34759594103824)
--(axis cs:35,3.33210100731803)
--(axis cs:36,3.36818202077285)
--(axis cs:37,3.39372719032819)
--(axis cs:38,3.41326368889782)
--(axis cs:39,3.40585797982215)
--(axis cs:40,3.4051335876059)
--(axis cs:41,3.39183705980171)
--(axis cs:42,3.39643191833297)
--(axis cs:43,3.39643191833297)
--(axis cs:44,3.42563236430387)
--(axis cs:45,3.38980919571148)
--(axis cs:46,3.4151172611827)
--(axis cs:47,3.42742681361839)
--(axis cs:48,3.43363485797613)
--(axis cs:49,3.43363485797613)
--(axis cs:50,3.41578698569939)
--(axis cs:50,3.53158143535324)
--(axis cs:50,3.53158143535324)
--(axis cs:49,3.54531251044493)
--(axis cs:48,3.54531251044493)
--(axis cs:47,3.54099423901319)
--(axis cs:46,3.52172484408046)
--(axis cs:45,3.49440133060431)
--(axis cs:44,3.55331500411718)
--(axis cs:43,3.54041018693019)
--(axis cs:42,3.54041018693019)
--(axis cs:41,3.53447872967198)
--(axis cs:40,3.5527611492362)
--(axis cs:39,3.55203675701996)
--(axis cs:38,3.55515736373376)
--(axis cs:37,3.53258859914549)
--(axis cs:36,3.50550218975346)
--(axis cs:35,3.47842530847144)
--(axis cs:34,3.48398300633018)
--(axis cs:33,3.50467505867173)
--(axis cs:32,3.51651993715046)
--(axis cs:31,3.47726315789474)
--(axis cs:30,3.58056889139356)
--(axis cs:29,3.57131635871972)
--(axis cs:28,3.6689029148971)
--(axis cs:27,3.68689410813541)
--(axis cs:26,3.6695723893109)
--(axis cs:25,3.64164982199736)
--(axis cs:24,3.66956914071347)
--(axis cs:23,3.5982551598334)
--(axis cs:22,3.59944578821356)
--(axis cs:21,3.59889380175727)
--(axis cs:20,3.60247119262647)
--(axis cs:19,3.57835254649364)
--(axis cs:18,3.55817434947147)
--(axis cs:17,3.53592798590915)
--(axis cs:16,3.51695395782635)
--(axis cs:15,3.5331228254605)
--(axis cs:14,3.569910387014)
--(axis cs:13,3.44192365775194)
--(axis cs:12,3.48613496970387)
--(axis cs:11,3.54559593877197)
--(axis cs:10,3.6240835471136)
--(axis cs:9,3.74745770949939)
--(axis cs:8,3.78776583557836)
--(axis cs:7,3.80217680333337)
--(axis cs:6,3.81865519606767)
--(axis cs:5,4)
--cycle;

\path [fill=colorHyperopt, fill opacity=0.2]
(axis cs:5,4)
--(axis cs:5,4)
--(axis cs:6,3.48135697658313)
--(axis cs:7,3.38167344701479)
--(axis cs:8,3.39693846410078)
--(axis cs:9,3.41675262026449)
--(axis cs:10,3.54727021906454)
--(axis cs:11,3.59835181091148)
--(axis cs:12,3.57383375046055)
--(axis cs:13,3.49498616210398)
--(axis cs:14,3.47707861833201)
--(axis cs:15,3.46725322920999)
--(axis cs:16,3.39058452964798)
--(axis cs:17,3.43534815664233)
--(axis cs:18,3.4258334734209)
--(axis cs:19,3.4230423303679)
--(axis cs:20,3.46539298524151)
--(axis cs:21,3.38437406387864)
--(axis cs:22,3.39629220584619)
--(axis cs:23,3.38810190816947)
--(axis cs:24,3.36502355325591)
--(axis cs:25,3.36291696360408)
--(axis cs:26,3.32958877150221)
--(axis cs:27,3.31632886926529)
--(axis cs:28,3.28150129742647)
--(axis cs:29,3.28805658389993)
--(axis cs:30,3.23533061280722)
--(axis cs:31,3.24803272165792)
--(axis cs:32,3.27205049564072)
--(axis cs:33,3.25852441767152)
--(axis cs:34,3.26436770907322)
--(axis cs:35,3.25284210526316)
--(axis cs:36,3.21264943450196)
--(axis cs:37,3.16689290441018)
--(axis cs:38,3.19415234375958)
--(axis cs:39,3.16664039299124)
--(axis cs:40,3.15899554783986)
--(axis cs:41,3.18204312211965)
--(axis cs:42,3.18466758599279)
--(axis cs:43,3.19128881237368)
--(axis cs:44,3.1776071149203)
--(axis cs:45,3.14487703357826)
--(axis cs:46,3.15933006401845)
--(axis cs:47,3.18514392130897)
--(axis cs:48,3.19048596603095)
--(axis cs:49,3.19357409046333)
--(axis cs:50,3.19669158301176)
--(axis cs:50,3.31909789067245)
--(axis cs:50,3.31909789067245)
--(axis cs:49,3.33274169901035)
--(axis cs:48,3.32530350765327)
--(axis cs:47,3.32011923658577)
--(axis cs:46,3.28277519913945)
--(axis cs:45,3.27617559800069)
--(axis cs:44,3.25397183244812)
--(axis cs:43,3.28239539815264)
--(axis cs:42,3.26796399295457)
--(axis cs:41,3.26006214103824)
--(axis cs:40,3.24100445216014)
--(axis cs:39,3.26493855437718)
--(axis cs:38,3.24795291939831)
--(axis cs:37,3.23310709558982)
--(axis cs:36,3.26103477602435)
--(axis cs:35,3.27347368421053)
--(axis cs:34,3.30405334355836)
--(axis cs:33,3.30989663496006)
--(axis cs:32,3.39110739909612)
--(axis cs:31,3.37301990992103)
--(axis cs:30,3.36466938719278)
--(axis cs:29,3.37510131083691)
--(axis cs:28,3.36060396573143)
--(axis cs:27,3.42051323599787)
--(axis cs:26,3.44935859691885)
--(axis cs:25,3.45813566797486)
--(axis cs:24,3.47708170990198)
--(axis cs:23,3.58031914446211)
--(axis cs:22,3.56160253099592)
--(axis cs:21,3.59457330454241)
--(axis cs:20,3.68197543581112)
--(axis cs:19,3.64011556436894)
--(axis cs:18,3.64785073710542)
--(axis cs:17,3.63833605388399)
--(axis cs:16,3.60941547035202)
--(axis cs:15,3.63800992868475)
--(axis cs:14,3.67028980272063)
--(axis cs:13,3.65238225894865)
--(axis cs:12,3.71037677585524)
--(axis cs:11,3.79112187329905)
--(axis cs:10,3.76851925461967)
--(axis cs:9,3.68851053763024)
--(axis cs:8,3.56095627274133)
--(axis cs:7,3.58674760561679)
--(axis cs:6,3.7081167076274)
--(axis cs:5,4)
--cycle;

\addplot [line width=2pt, colorRandom]
table {%
5 4
6 4.26315789473684
7 4.5
8 4.53684210526316
9 4.49473684210526
10 4.49473684210526
11 4.53157894736842
12 4.48947368421053
13 4.6
14 4.66842105263158
15 4.70526315789474
16 4.7421052631579
17 4.72631578947368
18 4.77368421052632
19 4.7578947368421
20 4.8
21 4.83157894736842
22 4.85789473684211
23 4.89473684210526
24 4.92105263157895
25 4.97894736842105
26 4.93684210526316
27 4.95789473684211
28 4.94210526315789
29 4.98947368421053
30 4.98947368421053
31 5.00526315789474
32 5.02631578947368
33 5.05263157894737
34 5.04736842105263
35 5.04736842105263
36 5.04736842105263
37 5.05789473684211
38 5.03684210526316
39 5.02105263157895
40 5.02105263157895
41 5.01578947368421
42 5.01578947368421
43 5
44 5.01052631578947
45 5.01578947368421
46 5.01578947368421
47 5.04210526315789
48 5.04736842105263
49 5.04210526315789
50 5.06315789473684
};
\addlegendentry{Random}
\addplot [line width=2pt, colorHEBO]
table {%
5 4
6 4.17368421052632
7 4
8 3.84736842105263
9 3.86315789473684
10 3.97894736842105
11 3.79473684210526
12 3.72631578947368
13 3.61052631578947
14 3.6
15 3.53157894736842
16 3.53157894736842
17 3.48421052631579
18 3.47368421052632
19 3.41578947368421
20 3.30526315789474
21 3.23684210526316
22 3.21052631578947
23 3.21052631578947
24 3.24210526315789
25 3.27368421052632
26 3.3
27 3.25789473684211
28 3.24736842105263
29 3.31052631578947
30 3.25263157894737
31 3.27368421052632
32 3.20526315789474
33 3.17368421052632
34 3.18947368421053
35 3.22631578947368
36 3.21052631578947
37 3.16842105263158
38 3.15263157894737
39 3.14736842105263
40 3.14210526315789
41 3.10526315789474
42 3.11052631578947
43 3.12105263157895
44 3.13684210526316
45 3.13157894736842
46 3.11052631578947
47 3.06842105263158
48 3.07368421052632
49 3.06842105263158
50 3.08947368421053
};
\addlegendentry{HEBO}
\addplot [line width=2pt, colorGP]
table {%
5 4
6 3.75263157894737
7 3.75263157894737
8 3.98421052631579
9 3.98947368421053
10 4.08421052631579
11 4.16842105263158
12 4.24736842105263
13 4.31052631578947
14 4.2
15 4.29473684210526
16 4.28947368421053
17 4.3
18 4.26315789473684
19 4.33157894736842
20 4.27368421052632
21 4.33684210526316
22 4.36842105263158
23 4.33157894736842
24 4.23684210526316
25 4.27368421052632
26 4.27368421052632
27 4.3
28 4.30526315789474
29 4.31578947368421
30 4.35263157894737
31 4.37368421052632
32 4.32631578947368
33 4.32631578947368
34 4.33684210526316
35 4.32105263157895
36 4.34736842105263
37 4.36842105263158
38 4.38947368421053
39 4.41578947368421
40 4.43157894736842
41 4.44736842105263
42 4.45789473684211
43 4.45789473684211
44 4.38947368421053
45 4.41578947368421
46 4.38421052631579
47 4.37368421052632
48 4.34210526315789
49 4.34210526315789
50 4.32105263157895
};
\addlegendentry{GP}
\addplot [line width=2pt, colorDNGO]
table {%
5 4
6 4.06842105263158
7 4.17368421052632
8 4.05263157894737
9 4.00526315789474
10 4.01578947368421
11 3.98947368421053
12 4.05263157894737
13 4.03157894736842
14 4.04210526315789
15 4.08421052631579
16 4.11052631578947
17 4.12631578947368
18 4.16842105263158
19 4.18947368421053
20 4.26315789473684
21 4.27368421052632
22 4.30526315789474
23 4.28947368421053
24 4.32631578947368
25 4.25263157894737
26 4.2578947368421
27 4.29473684210526
28 4.33157894736842
29 4.27368421052632
30 4.26842105263158
31 4.29473684210526
32 4.33157894736842
33 4.34210526315789
34 4.33157894736842
35 4.35263157894737
36 4.38421052631579
37 4.37368421052632
38 4.32631578947368
39 4.34210526315789
40 4.34210526315789
41 4.35263157894737
42 4.35789473684211
43 4.35263157894737
44 4.37368421052632
45 4.4
46 4.39473684210526
47 4.34736842105263
48 4.35263157894737
49 4.35789473684211
50 4.34210526315789
};
\addlegendentry{DNGO}
\addplot [line width=2pt, colorDGP]
table {%
5 4
6 4.45263157894737
7 4.41052631578947
8 4.42631578947368
9 4.45263157894737
10 4.3
11 4.39473684210526
12 4.44736842105263
13 4.48947368421053
14 4.42631578947368
15 4.40526315789474
16 4.4
17 4.37368421052632
18 4.31578947368421
19 4.26842105263158
20 4.2578947368421
21 4.30526315789474
22 4.26842105263158
23 4.29473684210526
24 4.3
25 4.28421052631579
26 4.28421052631579
27 4.2421052631579
28 4.27894736842105
29 4.29473684210526
30 4.33684210526316
31 4.34736842105263
32 4.34736842105263
33 4.39473684210526
34 4.39473684210526
35 4.38421052631579
36 4.33684210526316
37 4.36842105263158
38 4.38947368421053
39 4.37894736842105
40 4.38421052631579
41 4.39473684210526
42 4.36315789473684
43 4.36315789473684
44 4.38421052631579
45 4.38421052631579
46 4.40526315789474
47 4.43157894736842
48 4.43684210526316
49 4.43684210526316
50 4.45263157894737
};
\addlegendentry{DGP}
\addplot [line width=2pt, colorPFN_GP]
table {%
5 4
6 3.69473684210526
7 3.67894736842105
8 3.67368421052632
9 3.64210526315789
10 3.46842105263158
11 3.42631578947368
12 3.39473684210526
13 3.38421052631579
14 3.48947368421053
15 3.42631578947368
16 3.42631578947368
17 3.45263157894737
18 3.46842105263158
19 3.50526315789474
20 3.52631578947368
21 3.52631578947368
22 3.51052631578947
23 3.49473684210526
24 3.55263157894737
25 3.52631578947368
26 3.55789473684211
27 3.57894736842105
28 3.57368421052632
29 3.48421052631579
30 3.5
31 3.39473684210526
32 3.43157894736842
33 3.42631578947368
34 3.41578947368421
35 3.40526315789474
36 3.43684210526316
37 3.46315789473684
38 3.48421052631579
39 3.47894736842105
40 3.47894736842105
41 3.46315789473684
42 3.46842105263158
43 3.46842105263158
44 3.48947368421053
45 3.44210526315789
46 3.46842105263158
47 3.48421052631579
48 3.48947368421053
49 3.48947368421053
50 3.47368421052632
};
\addlegendentry{EI}
\addplot [line width=2pt, colorHyperopt]
table {%
5 4
6 3.59473684210526
7 3.48421052631579
8 3.47894736842105
9 3.55263157894737
10 3.65789473684211
11 3.69473684210526
12 3.64210526315789
13 3.57368421052632
14 3.57368421052632
15 3.55263157894737
16 3.5
17 3.53684210526316
18 3.53684210526316
19 3.53157894736842
20 3.57368421052632
21 3.48947368421053
22 3.47894736842105
23 3.48421052631579
24 3.42105263157895
25 3.41052631578947
26 3.38947368421053
27 3.36842105263158
28 3.32105263157895
29 3.33157894736842
30 3.3
31 3.31052631578947
32 3.33157894736842
33 3.28421052631579
34 3.28421052631579
35 3.26315789473684
36 3.23684210526316
37 3.2
38 3.22105263157895
39 3.21578947368421
40 3.2
41 3.22105263157895
42 3.22631578947368
43 3.23684210526316
44 3.21578947368421
45 3.21052631578947
46 3.22105263157895
47 3.25263157894737
48 3.25789473684211
49 3.26315789473684
50 3.25789473684211
};
\addlegendentry{EI + KG}
\legend{};
\end{axis}

\end{tikzpicture}

%% file: icml2023/figures/hpob_nonmyopic_regret_test.tex
\begin{tikzpicture}

\begin{axis}[
legend cell align={left},
legend style={
  fill opacity=0.8,
  draw opacity=1,
  text opacity=1,
  at={(0.55,-0.15)},
  anchor=south,
  draw=white!80!black
},
log basis y={10},
tick align=outside,
tick pos=left,
x grid style={white!69.0196078431373!black},
xlabel={Number of trials},
xmin=2.75, xmax=52.25,
xtick style={color=black},
y grid style={white!69.0196078431373!black},
ylabel={Average Regret},
ymin=0.00810597829468994, ymax=0.198863070592161,
ymode=log,
ytick style={color=black},
height=.6\textwidth,
width=.8\textwidth,
]
\path [fill=colorRandom, fill opacity=0.2]
(axis cs:5,0.171942701250577)
--(axis cs:5,0.131709969573849)
--(axis cs:6,0.107953049665483)
--(axis cs:7,0.0917237313617074)
--(axis cs:8,0.0799307044298644)
--(axis cs:9,0.0705018233806989)
--(axis cs:10,0.0655681805958182)
--(axis cs:11,0.0525757318310409)
--(axis cs:12,0.0453949884293952)
--(axis cs:13,0.0446194603050026)
--(axis cs:14,0.043085720666871)
--(axis cs:15,0.0401485899759881)
--(axis cs:16,0.0397084851942063)
--(axis cs:17,0.0393344492743204)
--(axis cs:18,0.0389614461726876)
--(axis cs:19,0.0386342011404024)
--(axis cs:20,0.0383673587855373)
--(axis cs:21,0.0383673587855373)
--(axis cs:22,0.0380701506391254)
--(axis cs:23,0.0379403448725462)
--(axis cs:24,0.0376287797906567)
--(axis cs:25,0.0376287797906567)
--(axis cs:26,0.0352386244793612)
--(axis cs:27,0.0351556736588477)
--(axis cs:28,0.0349584016002827)
--(axis cs:29,0.0348054274350595)
--(axis cs:30,0.0345820636666534)
--(axis cs:31,0.0344831542890687)
--(axis cs:32,0.0330670028196475)
--(axis cs:33,0.0325818167209202)
--(axis cs:34,0.0322638491681974)
--(axis cs:35,0.0322241912778449)
--(axis cs:36,0.0295909733179801)
--(axis cs:37,0.029408757056329)
--(axis cs:38,0.0291488412823729)
--(axis cs:39,0.0288925488701289)
--(axis cs:40,0.0288925488701289)
--(axis cs:41,0.0288458058725171)
--(axis cs:42,0.028800809523349)
--(axis cs:43,0.0264830893873049)
--(axis cs:44,0.0264030765645907)
--(axis cs:45,0.0261682012841563)
--(axis cs:46,0.0246841279357637)
--(axis cs:47,0.0246841279357637)
--(axis cs:48,0.0246841279357637)
--(axis cs:49,0.0241318249671857)
--(axis cs:50,0.0241318249671857)
--(axis cs:50,0.0278310367809559)
--(axis cs:50,0.0278310367809559)
--(axis cs:49,0.0278310367809559)
--(axis cs:48,0.0302849095173544)
--(axis cs:47,0.0302849095173544)
--(axis cs:46,0.0302849095173544)
--(axis cs:45,0.0318625738255101)
--(axis cs:44,0.0319242144828073)
--(axis cs:43,0.0320546746199565)
--(axis cs:42,0.0337798258690691)
--(axis cs:41,0.0337848654832496)
--(axis cs:40,0.0339418844002582)
--(axis cs:39,0.0339418844002582)
--(axis cs:38,0.0342824807539718)
--(axis cs:37,0.0346549306204788)
--(axis cs:36,0.0349024341667986)
--(axis cs:35,0.0352754323330664)
--(axis cs:34,0.0354506343466994)
--(axis cs:33,0.0360222146071715)
--(axis cs:32,0.037278588132005)
--(axis cs:31,0.0377764424031695)
--(axis cs:30,0.037828778096616)
--(axis cs:29,0.0381619955316003)
--(axis cs:28,0.03830752160344)
--(axis cs:27,0.0385483974570182)
--(axis cs:26,0.0385916737258615)
--(axis cs:25,0.0412541339057232)
--(axis cs:24,0.0412541339057232)
--(axis cs:23,0.0415977980284952)
--(axis cs:22,0.0416510505907812)
--(axis cs:21,0.0417554835974724)
--(axis cs:20,0.0417554835974724)
--(axis cs:19,0.0421151381454235)
--(axis cs:18,0.0426798080539616)
--(axis cs:17,0.0428895892086099)
--(axis cs:16,0.0438622070907334)
--(axis cs:15,0.0445430661123948)
--(axis cs:14,0.0474939146617861)
--(axis cs:13,0.0491471936443456)
--(axis cs:12,0.0497992839676828)
--(axis cs:11,0.0596066547890235)
--(axis cs:10,0.0783082378297599)
--(axis cs:9,0.0863176719669485)
--(axis cs:8,0.101511976295974)
--(axis cs:7,0.113436951078821)
--(axis cs:6,0.132003820065526)
--(axis cs:5,0.171942701250577)
--cycle;

\path [fill=colorHEBO, fill opacity=0.2]
(axis cs:5,0.171942701250577)
--(axis cs:5,0.131709969573849)
--(axis cs:6,0.117304230566511)
--(axis cs:7,0.0895633313826469)
--(axis cs:8,0.0737585812407267)
--(axis cs:9,0.064452248361371)
--(axis cs:10,0.0620187864283746)
--(axis cs:11,0.0545945886448174)
--(axis cs:12,0.0408909004843866)
--(axis cs:13,0.0360840283502013)
--(axis cs:14,0.0325606823986568)
--(axis cs:15,0.029834347384329)
--(axis cs:16,0.027421182469876)
--(axis cs:17,0.0238252261704918)
--(axis cs:18,0.0236799921958651)
--(axis cs:19,0.0224495237350814)
--(axis cs:20,0.021339116668334)
--(axis cs:21,0.0194292329936008)
--(axis cs:22,0.0161452599369058)
--(axis cs:23,0.0158836711360263)
--(axis cs:24,0.0158389285890272)
--(axis cs:25,0.0158389285890272)
--(axis cs:26,0.0157581671353973)
--(axis cs:27,0.0155828009074436)
--(axis cs:28,0.0148107403630681)
--(axis cs:29,0.0148107403630681)
--(axis cs:30,0.0143246175176749)
--(axis cs:31,0.0143246175176749)
--(axis cs:32,0.0137386554448293)
--(axis cs:33,0.0132485906573577)
--(axis cs:34,0.0130890912653012)
--(axis cs:35,0.0129131070196989)
--(axis cs:36,0.0122878291514073)
--(axis cs:37,0.0121773083884467)
--(axis cs:38,0.012151595563437)
--(axis cs:39,0.01209660973479)
--(axis cs:40,0.0113762191851921)
--(axis cs:41,0.0111505955564567)
--(axis cs:42,0.0111505955564567)
--(axis cs:43,0.0111505955564567)
--(axis cs:44,0.0111505955564567)
--(axis cs:45,0.0108816663562454)
--(axis cs:46,0.010414846392576)
--(axis cs:47,0.0100892944709201)
--(axis cs:48,0.0100892944709201)
--(axis cs:49,0.0100714839728398)
--(axis cs:50,0.0100714839728398)
--(axis cs:50,0.0125444309542262)
--(axis cs:50,0.0125444309542262)
--(axis cs:49,0.0125444309542262)
--(axis cs:48,0.012557379621337)
--(axis cs:47,0.012557379621337)
--(axis cs:46,0.0125968587433458)
--(axis cs:45,0.0130195865928714)
--(axis cs:44,0.0136402052058551)
--(axis cs:43,0.0136402052058551)
--(axis cs:42,0.0136402052058551)
--(axis cs:41,0.0136402052058551)
--(axis cs:40,0.0139531699028844)
--(axis cs:39,0.0151467850938722)
--(axis cs:38,0.0153065009513138)
--(axis cs:37,0.0153487676212164)
--(axis cs:36,0.01546832812694)
--(axis cs:35,0.0186710617398196)
--(axis cs:34,0.0188189641338935)
--(axis cs:33,0.018983351381513)
--(axis cs:32,0.0206179522294514)
--(axis cs:31,0.0212232965316962)
--(axis cs:30,0.0212232965316962)
--(axis cs:29,0.021520132719656)
--(axis cs:28,0.021520132719656)
--(axis cs:27,0.0222306518639387)
--(axis cs:26,0.0225876602571019)
--(axis cs:25,0.0226426812231801)
--(axis cs:24,0.0226426812231801)
--(axis cs:23,0.0229218253158571)
--(axis cs:22,0.0229841231546537)
--(axis cs:21,0.0244143507755578)
--(axis cs:20,0.0255100501405329)
--(axis cs:19,0.026188152613427)
--(axis cs:18,0.0272367726283542)
--(axis cs:17,0.0273760572716707)
--(axis cs:16,0.0306305047206174)
--(axis cs:15,0.0312981999308026)
--(axis cs:14,0.036544218159344)
--(axis cs:13,0.0403233726652711)
--(axis cs:12,0.0487217772453283)
--(axis cs:11,0.0659694614818657)
--(axis cs:10,0.0735432259463067)
--(axis cs:9,0.0746792229354757)
--(axis cs:8,0.0869413181752619)
--(axis cs:7,0.120635204583711)
--(axis cs:6,0.149205940398073)
--(axis cs:5,0.171942701250577)
--cycle;

\path [fill=colorGP, fill opacity=0.2]
(axis cs:5,0.171942701250577)
--(axis cs:5,0.131709969573849)
--(axis cs:6,0.0853714163036909)
--(axis cs:7,0.0733454010375716)
--(axis cs:8,0.0641014948256549)
--(axis cs:9,0.0566292893300475)
--(axis cs:10,0.0538716338867031)
--(axis cs:11,0.0528189533241563)
--(axis cs:12,0.0500373988279434)
--(axis cs:13,0.046018072909621)
--(axis cs:14,0.0419590847268678)
--(axis cs:15,0.0394911371884673)
--(axis cs:16,0.0369709108257468)
--(axis cs:17,0.0364924556597054)
--(axis cs:18,0.0363140805891105)
--(axis cs:19,0.0363140805891105)
--(axis cs:20,0.0355311892494335)
--(axis cs:21,0.0355311892494335)
--(axis cs:22,0.0355148022534749)
--(axis cs:23,0.034933173092053)
--(axis cs:24,0.028673820162404)
--(axis cs:25,0.0286715506027953)
--(axis cs:26,0.0280402080473674)
--(axis cs:27,0.028022420882457)
--(axis cs:28,0.0279734905454329)
--(axis cs:29,0.0279566167204334)
--(axis cs:30,0.0279566167204334)
--(axis cs:31,0.0279566167204334)
--(axis cs:32,0.0270988821997989)
--(axis cs:33,0.0267113566599834)
--(axis cs:34,0.0266418840660875)
--(axis cs:35,0.0265292853349005)
--(axis cs:36,0.0265292853349005)
--(axis cs:37,0.0263033962030898)
--(axis cs:38,0.0263033962030898)
--(axis cs:39,0.0263033962030898)
--(axis cs:40,0.0263033962030898)
--(axis cs:41,0.0263033962030898)
--(axis cs:42,0.0257408925251851)
--(axis cs:43,0.0256111498456616)
--(axis cs:44,0.0248187710270605)
--(axis cs:45,0.0248187710270605)
--(axis cs:46,0.0247279192586504)
--(axis cs:47,0.0231856009604742)
--(axis cs:48,0.0230446328605851)
--(axis cs:49,0.0230348191777943)
--(axis cs:50,0.0227729611052066)
--(axis cs:50,0.0263486835486575)
--(axis cs:50,0.0263486835486575)
--(axis cs:49,0.0269493004415107)
--(axis cs:48,0.0269933152978045)
--(axis cs:47,0.027067048884004)
--(axis cs:46,0.0277024502195201)
--(axis cs:45,0.0282593986778996)
--(axis cs:44,0.0282593986778996)
--(axis cs:43,0.0295426314234201)
--(axis cs:42,0.0297094046816282)
--(axis cs:41,0.0303329647546501)
--(axis cs:40,0.0303329647546501)
--(axis cs:39,0.0303329647546501)
--(axis cs:38,0.0303329647546501)
--(axis cs:37,0.0303329647546501)
--(axis cs:36,0.0306831426560942)
--(axis cs:35,0.0306831426560942)
--(axis cs:34,0.0308238035442906)
--(axis cs:33,0.0309465451253268)
--(axis cs:32,0.0316157593894003)
--(axis cs:31,0.0346859866461778)
--(axis cs:30,0.0346859866461778)
--(axis cs:29,0.0346859866461778)
--(axis cs:28,0.0347191757319643)
--(axis cs:27,0.0349240432729091)
--(axis cs:26,0.0349523948557855)
--(axis cs:25,0.0354517780815155)
--(axis cs:24,0.0354642151988792)
--(axis cs:23,0.0405340389201207)
--(axis cs:22,0.0410875449198743)
--(axis cs:21,0.0411005712778603)
--(axis cs:20,0.0411005712778603)
--(axis cs:19,0.0439942740948728)
--(axis cs:18,0.0439942740948728)
--(axis cs:17,0.0442343895724243)
--(axis cs:16,0.0446119658242574)
--(axis cs:15,0.0464691624635678)
--(axis cs:14,0.0473663569261679)
--(axis cs:13,0.0568280784861891)
--(axis cs:12,0.0595622490905951)
--(axis cs:11,0.0612487945203658)
--(axis cs:10,0.0619256592741816)
--(axis cs:9,0.0652874636480121)
--(axis cs:8,0.0709159239985639)
--(axis cs:7,0.0829594736235985)
--(axis cs:6,0.105216558223194)
--(axis cs:5,0.171942701250577)
--cycle;

\path [fill=colorDNGO, fill opacity=0.2]
(axis cs:5,0.171942701250577)
--(axis cs:5,0.131709969573849)
--(axis cs:6,0.112708789544014)
--(axis cs:7,0.103781599881411)
--(axis cs:8,0.0787297747180652)
--(axis cs:9,0.0622777694902754)
--(axis cs:10,0.0533442638435899)
--(axis cs:11,0.0486196125200434)
--(axis cs:12,0.0477099250159095)
--(axis cs:13,0.0461356251339006)
--(axis cs:14,0.0408018260008672)
--(axis cs:15,0.0393020459429676)
--(axis cs:16,0.0386306981652062)
--(axis cs:17,0.0383314313242266)
--(axis cs:18,0.038284571941595)
--(axis cs:19,0.0382359046311243)
--(axis cs:20,0.0382359046311243)
--(axis cs:21,0.0378523549205242)
--(axis cs:22,0.0377714470629922)
--(axis cs:23,0.0376651248637112)
--(axis cs:24,0.0350862875632218)
--(axis cs:25,0.0344887709206559)
--(axis cs:26,0.0331265884571676)
--(axis cs:27,0.0330797275926471)
--(axis cs:28,0.0329233647470729)
--(axis cs:29,0.0325919036095588)
--(axis cs:30,0.0314818893937393)
--(axis cs:31,0.0314818893937393)
--(axis cs:32,0.0314818893937393)
--(axis cs:33,0.0312040235062621)
--(axis cs:34,0.0310777174400289)
--(axis cs:35,0.0310393834192546)
--(axis cs:36,0.0307357432810844)
--(axis cs:37,0.0295178661309379)
--(axis cs:38,0.0293603929746979)
--(axis cs:39,0.0293524342935235)
--(axis cs:40,0.0293524342935235)
--(axis cs:41,0.0293524342935235)
--(axis cs:42,0.0293524342935235)
--(axis cs:43,0.0292909891669)
--(axis cs:44,0.0292909891669)
--(axis cs:45,0.0292909891669)
--(axis cs:46,0.0290765897966657)
--(axis cs:47,0.0285544517998176)
--(axis cs:48,0.0285544517998176)
--(axis cs:49,0.0285544517998176)
--(axis cs:50,0.0283900360663959)
--(axis cs:50,0.0355967812684306)
--(axis cs:50,0.0355967812684306)
--(axis cs:49,0.0363219133482039)
--(axis cs:48,0.0363219133482039)
--(axis cs:47,0.0363219133482039)
--(axis cs:46,0.036772513818418)
--(axis cs:45,0.0368546573333528)
--(axis cs:44,0.0368546573333528)
--(axis cs:43,0.0368546573333528)
--(axis cs:42,0.0368764028605966)
--(axis cs:41,0.0368764028605966)
--(axis cs:40,0.0368764028605966)
--(axis cs:39,0.0368764028605966)
--(axis cs:38,0.0368831508563946)
--(axis cs:37,0.0369404063336805)
--(axis cs:36,0.0406181206305788)
--(axis cs:35,0.0408519895923686)
--(axis cs:34,0.0410241420051764)
--(axis cs:33,0.0416240108408249)
--(axis cs:32,0.0417765608135581)
--(axis cs:31,0.0417765608135581)
--(axis cs:30,0.0417765608135581)
--(axis cs:29,0.0422387891506534)
--(axis cs:28,0.042581698758612)
--(axis cs:27,0.04263582234662)
--(axis cs:26,0.0426567644068951)
--(axis cs:25,0.0432085876839927)
--(axis cs:24,0.0436330035983363)
--(axis cs:23,0.0454108585409303)
--(axis cs:22,0.0456343363360805)
--(axis cs:21,0.045768130164637)
--(axis cs:20,0.0460208172962137)
--(axis cs:19,0.0460208172962137)
--(axis cs:18,0.0460547000086671)
--(axis cs:17,0.0461254000694371)
--(axis cs:16,0.0464375955424353)
--(axis cs:15,0.0475929348772622)
--(axis cs:14,0.0481590254531049)
--(axis cs:13,0.0517026691817205)
--(axis cs:12,0.0527285769610391)
--(axis cs:11,0.0545847336982583)
--(axis cs:10,0.075791049071923)
--(axis cs:9,0.0884641078227471)
--(axis cs:8,0.0973386754135451)
--(axis cs:7,0.144438335727939)
--(axis cs:6,0.150098172628931)
--(axis cs:5,0.171942701250577)
--cycle;

\path [fill=colorDGP, fill opacity=0.2]
(axis cs:5,0.171942701250577)
--(axis cs:5,0.131709969573849)
--(axis cs:6,0.122225006109484)
--(axis cs:7,0.105401070461368)
--(axis cs:8,0.0856211166807105)
--(axis cs:9,0.0830491138130478)
--(axis cs:10,0.065992578971265)
--(axis cs:11,0.0649616321784654)
--(axis cs:12,0.0615976289297153)
--(axis cs:13,0.057697683814174)
--(axis cs:14,0.0486669113753851)
--(axis cs:15,0.045735951647533)
--(axis cs:16,0.0449250624944319)
--(axis cs:17,0.0415232341622362)
--(axis cs:18,0.0387881657788215)
--(axis cs:19,0.0382557951938519)
--(axis cs:20,0.0377455780655465)
--(axis cs:21,0.0369220709343417)
--(axis cs:22,0.0355452273574677)
--(axis cs:23,0.0355452273574677)
--(axis cs:24,0.0349268974730079)
--(axis cs:25,0.0348610923830908)
--(axis cs:26,0.0346755795363536)
--(axis cs:27,0.0327632986730963)
--(axis cs:28,0.0327632986730963)
--(axis cs:29,0.0298157884651446)
--(axis cs:30,0.0297826272152767)
--(axis cs:31,0.0295759719385922)
--(axis cs:32,0.0291108989948434)
--(axis cs:33,0.0291108989948434)
--(axis cs:34,0.0290652281340872)
--(axis cs:35,0.028525985732269)
--(axis cs:36,0.0283920473342543)
--(axis cs:37,0.0272086589257235)
--(axis cs:38,0.0272086589257235)
--(axis cs:39,0.0270193517186836)
--(axis cs:40,0.0270193517186836)
--(axis cs:41,0.0270193517186836)
--(axis cs:42,0.026895541709255)
--(axis cs:43,0.0268659291956513)
--(axis cs:44,0.0268659291956513)
--(axis cs:45,0.0268171799793645)
--(axis cs:46,0.0267182751966212)
--(axis cs:47,0.0267182751966212)
--(axis cs:48,0.0267182751966212)
--(axis cs:49,0.026713092623393)
--(axis cs:50,0.026707896340493)
--(axis cs:50,0.0320810597931022)
--(axis cs:50,0.0320810597931022)
--(axis cs:49,0.0320905701871745)
--(axis cs:48,0.0321000942909186)
--(axis cs:47,0.0321000942909186)
--(axis cs:46,0.0321000942909186)
--(axis cs:45,0.0321675708159099)
--(axis cs:44,0.0322020392009279)
--(axis cs:43,0.0322020392009279)
--(axis cs:42,0.0322556173411915)
--(axis cs:41,0.0323465090178514)
--(axis cs:40,0.0323465090178514)
--(axis cs:39,0.0323465090178514)
--(axis cs:38,0.032367715191831)
--(axis cs:37,0.032367715191831)
--(axis cs:36,0.0361013265156966)
--(axis cs:35,0.0362500692986829)
--(axis cs:34,0.036926304001736)
--(axis cs:33,0.0371302051025816)
--(axis cs:32,0.0371302051025816)
--(axis cs:31,0.0372689230658362)
--(axis cs:30,0.0373118397507537)
--(axis cs:29,0.0373521648995589)
--(axis cs:28,0.0384431526063606)
--(axis cs:27,0.0384431526063606)
--(axis cs:26,0.039969534551264)
--(axis cs:25,0.0402469593711492)
--(axis cs:24,0.0403611863157853)
--(axis cs:23,0.0410556999005318)
--(axis cs:22,0.0410556999005318)
--(axis cs:21,0.0443939607432685)
--(axis cs:20,0.0457215179607295)
--(axis cs:19,0.0468261936276446)
--(axis cs:18,0.048977127561577)
--(axis cs:17,0.0647772212423803)
--(axis cs:16,0.0671307460074921)
--(axis cs:15,0.0675954492755749)
--(axis cs:14,0.0716423713653905)
--(axis cs:13,0.078098161141527)
--(axis cs:12,0.0802864420659304)
--(axis cs:11,0.0835547666982572)
--(axis cs:10,0.0879746172900576)
--(axis cs:9,0.119598916000186)
--(axis cs:8,0.123708751365111)
--(axis cs:7,0.138111832439353)
--(axis cs:6,0.162739646429015)
--(axis cs:5,0.171942701250577)
--cycle;

\path [fill=colorPFN_GP, fill opacity=0.2]
(axis cs:5,0.171942701250577)
--(axis cs:5,0.131709969573849)
--(axis cs:6,0.0916537411333802)
--(axis cs:7,0.0690146945543044)
--(axis cs:8,0.0570636002500037)
--(axis cs:9,0.0534763195402432)
--(axis cs:10,0.0466969611086196)
--(axis cs:11,0.0451608366206251)
--(axis cs:12,0.0422230639124124)
--(axis cs:13,0.040800440496201)
--(axis cs:14,0.03741396282368)
--(axis cs:15,0.033067558159877)
--(axis cs:16,0.0326641665912554)
--(axis cs:17,0.0325766618009299)
--(axis cs:18,0.0314824376729034)
--(axis cs:19,0.0313841279020239)
--(axis cs:20,0.0313214670905851)
--(axis cs:21,0.031081292404142)
--(axis cs:22,0.0307230253825866)
--(axis cs:23,0.0305650483637355)
--(axis cs:24,0.030440063671113)
--(axis cs:25,0.0295647126064967)
--(axis cs:26,0.0295401704909295)
--(axis cs:27,0.029459124117797)
--(axis cs:28,0.0288853086528379)
--(axis cs:29,0.0247154571659743)
--(axis cs:30,0.0245987680937454)
--(axis cs:31,0.0212321966647904)
--(axis cs:32,0.0212270280874112)
--(axis cs:33,0.0205439566909651)
--(axis cs:34,0.0205417542628925)
--(axis cs:35,0.0191858313759098)
--(axis cs:36,0.0191858313759098)
--(axis cs:37,0.0191858313759098)
--(axis cs:38,0.0191858313759098)
--(axis cs:39,0.0191704217451706)
--(axis cs:40,0.0191646669761936)
--(axis cs:41,0.0191557404463894)
--(axis cs:42,0.0191557404463894)
--(axis cs:43,0.0191557404463894)
--(axis cs:44,0.0191557404463894)
--(axis cs:45,0.0189842705548021)
--(axis cs:46,0.0189842705548021)
--(axis cs:47,0.0189832614111387)
--(axis cs:48,0.0189832614111387)
--(axis cs:49,0.0189832614111387)
--(axis cs:50,0.0189437740765432)
--(axis cs:50,0.0274444216917689)
--(axis cs:50,0.0274444216917689)
--(axis cs:49,0.0274729138520857)
--(axis cs:48,0.0274729138520857)
--(axis cs:47,0.0274729138520857)
--(axis cs:46,0.0274795944997202)
--(axis cs:45,0.0274795944997202)
--(axis cs:44,0.0275908057891337)
--(axis cs:43,0.0275908057891337)
--(axis cs:42,0.0275908057891337)
--(axis cs:41,0.0275908057891337)
--(axis cs:40,0.0276498587542418)
--(axis cs:39,0.0276881770299934)
--(axis cs:38,0.0277260320095212)
--(axis cs:37,0.0277260320095212)
--(axis cs:36,0.0277260320095212)
--(axis cs:35,0.0277260320095212)
--(axis cs:34,0.0287639199937357)
--(axis cs:33,0.0287770971482587)
--(axis cs:32,0.0292800895027392)
--(axis cs:31,0.0292826107166578)
--(axis cs:30,0.0310361300945774)
--(axis cs:29,0.0311727006417319)
--(axis cs:28,0.0351245750331793)
--(axis cs:27,0.0357641940210913)
--(axis cs:26,0.035925467495874)
--(axis cs:25,0.0360624109975607)
--(axis cs:24,0.0371699174739547)
--(axis cs:23,0.0377155894565734)
--(axis cs:22,0.0387667574295364)
--(axis cs:21,0.0390554904515092)
--(axis cs:20,0.0392000247784431)
--(axis cs:19,0.0392718047672424)
--(axis cs:18,0.040109529968103)
--(axis cs:17,0.0410909174041981)
--(axis cs:16,0.0413337880998454)
--(axis cs:15,0.0416094851998808)
--(axis cs:14,0.0475793934566217)
--(axis cs:13,0.050805261434812)
--(axis cs:12,0.0516301996504128)
--(axis cs:11,0.0646177517329314)
--(axis cs:10,0.0674973808533955)
--(axis cs:9,0.0753473372737531)
--(axis cs:8,0.0818238896819726)
--(axis cs:7,0.0984445729810539)
--(axis cs:6,0.135280232185557)
--(axis cs:5,0.171942701250577)
--cycle;

\path [fill=colorHyperopt, fill opacity=0.2]
(axis cs:5,0.171942701250577)
--(axis cs:5,0.131709969573849)
--(axis cs:6,0.0794223211087063)
--(axis cs:7,0.0697643508167897)
--(axis cs:8,0.0520508129606247)
--(axis cs:9,0.0493841285901047)
--(axis cs:10,0.0487869915986627)
--(axis cs:11,0.047933210701804)
--(axis cs:12,0.0444945494398577)
--(axis cs:13,0.042245801569608)
--(axis cs:14,0.0391964909420186)
--(axis cs:15,0.0366534801761898)
--(axis cs:16,0.035244246601818)
--(axis cs:17,0.0350626177764654)
--(axis cs:18,0.0332743312206387)
--(axis cs:19,0.0305419516734588)
--(axis cs:20,0.030322928019035)
--(axis cs:21,0.028351041907213)
--(axis cs:22,0.0265098341005228)
--(axis cs:23,0.0250337653414102)
--(axis cs:24,0.0214874796913636)
--(axis cs:25,0.021465468748125)
--(axis cs:26,0.0204464124937625)
--(axis cs:27,0.0189771329445281)
--(axis cs:28,0.0184725184976733)
--(axis cs:29,0.0177849119160486)
--(axis cs:30,0.0166978135790283)
--(axis cs:31,0.0166978135790283)
--(axis cs:32,0.0165447810847255)
--(axis cs:33,0.0157676625391517)
--(axis cs:34,0.0155903953476305)
--(axis cs:35,0.0155533954569679)
--(axis cs:36,0.0152955108987475)
--(axis cs:37,0.0146572738181555)
--(axis cs:38,0.0118280132480564)
--(axis cs:39,0.0112410989709057)
--(axis cs:40,0.0112387078018237)
--(axis cs:41,0.0112387078018237)
--(axis cs:42,0.010229537551464)
--(axis cs:43,0.010229537551464)
--(axis cs:44,0.0100548893843845)
--(axis cs:45,0.00963150158914354)
--(axis cs:46,0.00950614325327776)
--(axis cs:47,0.00950614325327776)
--(axis cs:48,0.00950614325327776)
--(axis cs:49,0.00950614325327776)
--(axis cs:50,0.00937509834445527)
--(axis cs:50,0.0159967984372119)
--(axis cs:50,0.0159967984372119)
--(axis cs:49,0.0161896401680655)
--(axis cs:48,0.0161896401680655)
--(axis cs:47,0.0161896401680655)
--(axis cs:46,0.0161896401680655)
--(axis cs:45,0.0163881684718758)
--(axis cs:44,0.0168543284898299)
--(axis cs:43,0.0169329399421337)
--(axis cs:42,0.0169329399421337)
--(axis cs:41,0.0173761194955377)
--(axis cs:40,0.0173761194955377)
--(axis cs:39,0.017388435003428)
--(axis cs:38,0.017790087842299)
--(axis cs:37,0.0279213398010147)
--(axis cs:36,0.0294888008597596)
--(axis cs:35,0.0297695046273039)
--(axis cs:34,0.0297934339257756)
--(axis cs:33,0.0299400533739306)
--(axis cs:32,0.0304063361569178)
--(axis cs:31,0.0305771903022912)
--(axis cs:30,0.0305771903022912)
--(axis cs:29,0.0310265001930138)
--(axis cs:28,0.038475648115907)
--(axis cs:27,0.0388751953034741)
--(axis cs:26,0.0397478133711984)
--(axis cs:25,0.0406813207907163)
--(axis cs:24,0.040877513019795)
--(axis cs:23,0.0437202446350336)
--(axis cs:22,0.0443197874400426)
--(axis cs:21,0.0479507766267177)
--(axis cs:20,0.0497652872126483)
--(axis cs:19,0.0498889182437429)
--(axis cs:18,0.0516461172692545)
--(axis cs:17,0.0524580586898548)
--(axis cs:16,0.0525965187142331)
--(axis cs:15,0.05333200027499)
--(axis cs:14,0.0551894458366164)
--(axis cs:13,0.0571915871725679)
--(axis cs:12,0.0588506737202554)
--(axis cs:11,0.0619592820066975)
--(axis cs:10,0.0652123124319801)
--(axis cs:9,0.0672608063876726)
--(axis cs:8,0.0697860469236099)
--(axis cs:7,0.0921531483413473)
--(axis cs:6,0.101247486166889)
--(axis cs:5,0.171942701250577)
--cycle;

\addplot [line width=2pt, colorRandom]
table {%
5 0.151826335412213
6 0.119978434865505
7 0.102580341220264
8 0.0907213403629194
9 0.0784097476738237
10 0.0719382092127891
11 0.0560911933100322
12 0.047597136198539
13 0.0468833269746741
14 0.0452898176643286
15 0.0423458280441914
16 0.0417853461424698
17 0.0411120192414652
18 0.0408206271133246
19 0.0403746696429129
20 0.0400614211915049
21 0.0400614211915049
22 0.0398606006149533
23 0.0397690714505207
24 0.03944145684819
25 0.03944145684819
26 0.0369151491026114
27 0.036852035557933
28 0.0366329616018613
29 0.0364837114833299
30 0.0362054208816347
31 0.0361297983461191
32 0.0351727954758262
33 0.0343020156640459
34 0.0338572417574484
35 0.0337498118054557
36 0.0322467037423894
37 0.0320318438384039
38 0.0317156610181723
39 0.0314172166351936
40 0.0314172166351936
41 0.0313153356778834
42 0.0312903176962091
43 0.0292688820036307
44 0.029163645523699
45 0.0290153875548332
46 0.027484518726559
47 0.027484518726559
48 0.027484518726559
49 0.0259814308740708
50 0.0259814308740708
};
\addplot [line width=2pt, colorHEBO]
table {%
5 0.151826335412213
6 0.133255085482292
7 0.105099267983179
8 0.0803499497079943
9 0.0695657356484233
10 0.0677810061873406
11 0.0602820250633415
12 0.0448063388648574
13 0.0382037005077362
14 0.0345524502790004
15 0.0305662736575658
16 0.0290258435952467
17 0.0256006417210812
18 0.0254583824121097
19 0.0243188381742542
20 0.0234245834044334
21 0.0219217918845793
22 0.0195646915457797
23 0.0194027482259417
24 0.0192408049061036
25 0.0192408049061036
26 0.0191729136962496
27 0.0189067263856912
28 0.018165436541362
29 0.018165436541362
30 0.0177739570246856
31 0.0177739570246856
32 0.0171783038371404
33 0.0161159710194354
34 0.0159540276995973
35 0.0157920843797593
36 0.0138780786391736
37 0.0137630380048315
38 0.0137290482573754
39 0.0136216974143311
40 0.0126646945440382
41 0.0123954003811559
42 0.0123954003811559
43 0.0123954003811559
44 0.0123954003811559
45 0.0119506264745584
46 0.0115058525679609
47 0.0113233370461286
48 0.0113233370461286
49 0.011307957463533
50 0.011307957463533
};
\addplot [line width=2pt, colorGP]
table {%
5 0.151826335412213
6 0.0952939872634423
7 0.0781524373305851
8 0.0675087094121094
9 0.0609583764890298
10 0.0578986465804424
11 0.057033873922261
12 0.0547998239592693
13 0.0514230756979051
14 0.0446627208265179
15 0.0429801498260176
16 0.0407914383250021
17 0.0403634226160648
18 0.0401541773419916
19 0.0401541773419916
20 0.0383158802636469
21 0.0383158802636469
22 0.0383011735866746
23 0.0377336060060869
24 0.0320690176806416
25 0.0320616643421554
26 0.0314963014515765
27 0.0314732320776831
28 0.0313463331386986
29 0.0313213016833056
30 0.0313213016833056
31 0.0313213016833056
32 0.0293573207945996
33 0.0288289508926551
34 0.028732843805189
35 0.0286062139954974
36 0.0286062139954974
37 0.02831818047887
38 0.02831818047887
39 0.02831818047887
40 0.02831818047887
41 0.02831818047887
42 0.0277251486034067
43 0.0275768906345408
44 0.0265390848524801
45 0.0265390848524801
46 0.0262151847390852
47 0.0251263249222391
48 0.0250189740791948
49 0.0249920598096525
50 0.0245608223269321
};
\addplot [line width=2pt, colorDNGO]
table {%
5 0.151826335412213
6 0.131403481086473
7 0.124109967804675
8 0.0880342250658052
9 0.0753709386565112
10 0.0645676564577564
11 0.0516021731091508
12 0.0502192509884743
13 0.0489191471578105
14 0.0444804257269861
15 0.0434474904101149
16 0.0425341468538208
17 0.0422284156968319
18 0.0421696359751311
19 0.042128360963669
20 0.042128360963669
21 0.0418102425425806
22 0.0417028916995363
23 0.0415379917023207
24 0.039359645580779
25 0.0388486793023243
26 0.0378916764320314
27 0.0378577749696335
28 0.0377525317528425
29 0.0374153463801061
30 0.0366292251036487
31 0.0366292251036487
32 0.0366292251036487
33 0.0364140171735435
34 0.0360509297226026
35 0.0359456865058116
36 0.0356769319558316
37 0.0332291362323092
38 0.0331217719155462
39 0.0331144185770601
40 0.0331144185770601
41 0.0331144185770601
42 0.0331144185770601
43 0.0330728232501264
44 0.0330728232501264
45 0.0330728232501264
46 0.0329245518075419
47 0.0324381825740108
48 0.0324381825740108
49 0.0324381825740108
50 0.0319934086674133
};
\addplot [line width=2pt, colorDGP]
table {%
5 0.151826335412213
6 0.14248232626925
7 0.12175645145036
8 0.104664934022911
9 0.101324014906617
10 0.0769835981306613
11 0.0742581994383613
12 0.0709420354978229
13 0.0678979224778505
14 0.0601546413703878
15 0.056665700461554
16 0.056027904250962
17 0.0531502277023083
18 0.0438826466701992
19 0.0425409944107483
20 0.041733548013138
21 0.0406580158388051
22 0.0383004636289998
23 0.0383004636289998
24 0.0376440418943966
25 0.03755402587712
26 0.0373225570438088
27 0.0356032256397285
28 0.0356032256397285
29 0.0335839766823517
30 0.0335472334830152
31 0.0334224475022142
32 0.0331205520487125
33 0.0331205520487125
34 0.0329957660679116
35 0.032388027515476
36 0.0322466869249755
37 0.0297881870587773
38 0.0297881870587773
39 0.0296829303682675
40 0.0296829303682675
41 0.0296829303682675
42 0.0295755795252232
43 0.0295339841982896
44 0.0295339841982896
45 0.0294923753976372
46 0.0294091847437699
47 0.0294091847437699
48 0.0294091847437699
49 0.0294018314052837
50 0.0293944780667976
};
\addplot [line width=2pt, colorPFN_GP] 
table {%
5 0.151826335412213
6 0.113466986659469
7 0.0837296337676792
8 0.0694437449659881
9 0.0644118284069981
10 0.0570971709810076
11 0.0548892941767783
12 0.0469266317814126
13 0.0458028509655065
14 0.0424966781401509
15 0.0373385216798789
16 0.0369989773455504
17 0.036833789602564
18 0.0357959838205032
19 0.0353279663346331
20 0.0352607459345141
21 0.0350683914278256
22 0.0347448914060615
23 0.0341403189101544
24 0.0338049905725338
25 0.0328135618020287
26 0.0327328189934017
27 0.0326116590694441
28 0.0320049418430086
29 0.0279440789038531
30 0.0278174490941614
31 0.0252574036907241
32 0.0252535587950752
33 0.0246605269196119
34 0.0246528371283141
35 0.0234559316927155
36 0.0234559316927155
37 0.0234559316927155
38 0.0234559316927155
39 0.023429299387582
40 0.0234072628652177
41 0.0233732731177616
42 0.0233732731177616
43 0.0233732731177616
44 0.0233732731177616
45 0.0232319325272611
46 0.0232319325272611
47 0.0232280876316122
48 0.0232280876316122
49 0.0232280876316122
50 0.023194097884156
};
\addplot [line width=2pt, colorHyperopt] 
table {%
5 0.151826335412213
6 0.0903349036377975
7 0.0809587495790685
8 0.0609184299421173
9 0.0583224674888887
10 0.0569996520153214
11 0.0549462463542508
12 0.0516726115800566
13 0.0497186943710879
14 0.0471929683893175
15 0.0449927402255899
16 0.0439203826580256
17 0.0437603382331601
18 0.0424602242449466
19 0.0402154349586009
20 0.0400441076158417
21 0.0381509092669654
22 0.0354148107702827
23 0.0343770049882219
24 0.0311824963555793
25 0.0310733947694206
26 0.0300971129324804
27 0.0289261641240011
28 0.0284740833067902
29 0.0244057060545312
30 0.0236375019406597
31 0.0236375019406597
32 0.0234755586208217
33 0.0228538579565411
34 0.0226919146367031
35 0.0226614500421359
36 0.0223921558792535
37 0.0212893068095851
38 0.0148090505451777
39 0.0143147669871669
40 0.0143074136486807
41 0.0143074136486807
42 0.0135812387467988
43 0.0135812387467988
44 0.0134546089371072
45 0.0130098350305097
46 0.0128478917106716
47 0.0128478917106716
48 0.0128478917106716
49 0.0128478917106716
50 0.0126859483908336
};
\end{axis}

\end{tikzpicture}

%% file: MetaLearnWorkshop/figures/ll_ranks_hebo_warp_obsnoise.tex
\begin{tikzpicture}

\begin{axis}[
colorbar,
colorbar style={ytick={0.2,0.4,0.6,0.8},yticklabels={0.2,0.4,0.6,0.8},ylabel={}},
colormap={mymap}{[1pt]
 rgb(0pt)=(0.969088811995386,0.966474432910419,0.964936562860438);
  rgb(1pt)=(0.969088811995386,0.966474432910419,0.964936562860438);
  rgb(2pt)=(0.970011534025375,0.962168396770473,0.957554786620531);
  rgb(3pt)=(0.970011534025375,0.962168396770473,0.957554786620531);
  rgb(4pt)=(0.970934256055363,0.957862360630527,0.950173010380623);
  rgb(5pt)=(0.970934256055363,0.957862360630527,0.950173010380623);
  rgb(6pt)=(0.971856978085352,0.953556324490581,0.942791234140715);
  rgb(7pt)=(0.971856978085352,0.953556324490581,0.942791234140715);
  rgb(8pt)=(0.97277970011534,0.949250288350635,0.935409457900808);
  rgb(9pt)=(0.97277970011534,0.949250288350635,0.935409457900808);
  rgb(10pt)=(0.973702422145329,0.944944252210688,0.9280276816609);
  rgb(11pt)=(0.973702422145329,0.944944252210688,0.9280276816609);
  rgb(12pt)=(0.974625144175317,0.940638216070742,0.920645905420992);
  rgb(13pt)=(0.974625144175317,0.940638216070742,0.920645905420992);
  rgb(14pt)=(0.975547866205306,0.936332179930796,0.913264129181084);
  rgb(15pt)=(0.975547866205306,0.936332179930796,0.913264129181084);
  rgb(16pt)=(0.976470588235294,0.93202614379085,0.905882352941176);
  rgb(17pt)=(0.976470588235294,0.93202614379085,0.905882352941176);
  rgb(18pt)=(0.977393310265283,0.927720107650904,0.898500576701269);
  rgb(19pt)=(0.977393310265283,0.927720107650904,0.898500576701269);
  rgb(20pt)=(0.978316032295271,0.923414071510957,0.891118800461361);
  rgb(21pt)=(0.978316032295271,0.923414071510957,0.891118800461361);
  rgb(22pt)=(0.97923875432526,0.919108035371011,0.883737024221453);
  rgb(23pt)=(0.97923875432526,0.919108035371011,0.883737024221453);
  rgb(24pt)=(0.980161476355248,0.914801999231065,0.876355247981546);
  rgb(25pt)=(0.980161476355248,0.914801999231065,0.876355247981546);
  rgb(26pt)=(0.981084198385236,0.910495963091119,0.868973471741638);
  rgb(27pt)=(0.981084198385236,0.910495963091119,0.868973471741638);
  rgb(28pt)=(0.982006920415225,0.906189926951173,0.86159169550173);
  rgb(29pt)=(0.982006920415225,0.906189926951173,0.86159169550173);
  rgb(30pt)=(0.982929642445213,0.901883890811226,0.854209919261822);
  rgb(31pt)=(0.982929642445213,0.901883890811226,0.854209919261822);
  rgb(32pt)=(0.983852364475202,0.89757785467128,0.846828143021915);
  rgb(33pt)=(0.983852364475202,0.89757785467128,0.846828143021915);
  rgb(34pt)=(0.98477508650519,0.893271818531334,0.839446366782007);
  rgb(35pt)=(0.98477508650519,0.893271818531334,0.839446366782007);
  rgb(36pt)=(0.985697808535179,0.888965782391388,0.832064590542099);
  rgb(37pt)=(0.985697808535179,0.888965782391388,0.832064590542099);
  rgb(38pt)=(0.986620530565167,0.884659746251442,0.824682814302191);
  rgb(39pt)=(0.986620530565167,0.884659746251442,0.824682814302191);
  rgb(40pt)=(0.987543252595156,0.880353710111496,0.817301038062284);
  rgb(41pt)=(0.987543252595156,0.880353710111496,0.817301038062284);
  rgb(42pt)=(0.988465974625144,0.876047673971549,0.809919261822376);
  rgb(43pt)=(0.988465974625144,0.876047673971549,0.809919261822376);
  rgb(44pt)=(0.989388696655133,0.871741637831603,0.802537485582468);
  rgb(45pt)=(0.989388696655133,0.871741637831603,0.802537485582468);
  rgb(46pt)=(0.990311418685121,0.867435601691657,0.795155709342561);
  rgb(47pt)=(0.990311418685121,0.867435601691657,0.795155709342561);
  rgb(48pt)=(0.99123414071511,0.863129565551711,0.787773933102653);
  rgb(49pt)=(0.99123414071511,0.863129565551711,0.787773933102653);
  rgb(50pt)=(0.992156862745098,0.858823529411765,0.780392156862745);
  rgb(51pt)=(0.992156862745098,0.858823529411765,0.780392156862745);
  rgb(52pt)=(0.990772779700115,0.850519031141868,0.769780853517878);
  rgb(53pt)=(0.990772779700115,0.850519031141868,0.769780853517878);
  rgb(54pt)=(0.989388696655133,0.842214532871972,0.75916955017301);
  rgb(55pt)=(0.989388696655133,0.842214532871972,0.75916955017301);
  rgb(56pt)=(0.98800461361015,0.833910034602076,0.748558246828143);
  rgb(57pt)=(0.98800461361015,0.833910034602076,0.748558246828143);
  rgb(58pt)=(0.986620530565167,0.82560553633218,0.737946943483276);
  rgb(59pt)=(0.986620530565167,0.82560553633218,0.737946943483276);
  rgb(60pt)=(0.985236447520185,0.817301038062284,0.727335640138408);
  rgb(61pt)=(0.985236447520185,0.817301038062284,0.727335640138408);
  rgb(62pt)=(0.983852364475202,0.808996539792388,0.716724336793541);
  rgb(63pt)=(0.983852364475202,0.808996539792388,0.716724336793541);
  rgb(64pt)=(0.982468281430219,0.800692041522491,0.706113033448674);
  rgb(65pt)=(0.982468281430219,0.800692041522491,0.706113033448674);
  rgb(66pt)=(0.981084198385237,0.792387543252595,0.695501730103806);
  rgb(67pt)=(0.981084198385237,0.792387543252595,0.695501730103806);
  rgb(68pt)=(0.979700115340254,0.784083044982699,0.684890426758939);
  rgb(69pt)=(0.979700115340254,0.784083044982699,0.684890426758939);
  rgb(70pt)=(0.978316032295271,0.775778546712803,0.674279123414071);
  rgb(71pt)=(0.978316032295271,0.775778546712803,0.674279123414071);
  rgb(72pt)=(0.976931949250288,0.767474048442907,0.663667820069204);
  rgb(73pt)=(0.976931949250288,0.767474048442907,0.663667820069204);
  rgb(74pt)=(0.975547866205306,0.75916955017301,0.653056516724337);
  rgb(75pt)=(0.975547866205306,0.75916955017301,0.653056516724337);
  rgb(76pt)=(0.974163783160323,0.750865051903114,0.642445213379469);
  rgb(77pt)=(0.974163783160323,0.750865051903114,0.642445213379469);
  rgb(78pt)=(0.97277970011534,0.742560553633218,0.631833910034602);
  rgb(79pt)=(0.97277970011534,0.742560553633218,0.631833910034602);
  rgb(80pt)=(0.971395617070358,0.734256055363322,0.621222606689735);
  rgb(81pt)=(0.971395617070358,0.734256055363322,0.621222606689735);
  rgb(82pt)=(0.970011534025375,0.725951557093426,0.610611303344867);
  rgb(83pt)=(0.970011534025375,0.725951557093426,0.610611303344867);
  rgb(84pt)=(0.968627450980392,0.717647058823529,0.6);
  rgb(85pt)=(0.968627450980392,0.717647058823529,0.6);
  rgb(86pt)=(0.96724336793541,0.709342560553633,0.589388696655133);
  rgb(87pt)=(0.96724336793541,0.709342560553633,0.589388696655133);
  rgb(88pt)=(0.965859284890427,0.701038062283737,0.578777393310265);
  rgb(89pt)=(0.965859284890427,0.701038062283737,0.578777393310265);
  rgb(90pt)=(0.964475201845444,0.692733564013841,0.568166089965398);
  rgb(91pt)=(0.964475201845444,0.692733564013841,0.568166089965398);
  rgb(92pt)=(0.963091118800461,0.684429065743945,0.557554786620531);
  rgb(93pt)=(0.963091118800461,0.684429065743945,0.557554786620531);
  rgb(94pt)=(0.961707035755479,0.676124567474048,0.546943483275663);
  rgb(95pt)=(0.961707035755479,0.676124567474048,0.546943483275663);
  rgb(96pt)=(0.960322952710496,0.667820069204152,0.536332179930796);
  rgb(97pt)=(0.960322952710496,0.667820069204152,0.536332179930796);
  rgb(98pt)=(0.958938869665513,0.659515570934256,0.525720876585928);
  rgb(99pt)=(0.958938869665513,0.659515570934256,0.525720876585928);
  rgb(100pt)=(0.957554786620531,0.65121107266436,0.515109573241061);
  rgb(101pt)=(0.957554786620531,0.65121107266436,0.515109573241061);
  rgb(102pt)=(0.954555940023068,0.641753171856978,0.505728565936178);
  rgb(103pt)=(0.954555940023068,0.641753171856978,0.505728565936178);
  rgb(104pt)=(0.949942329873126,0.631141868512111,0.49757785467128);
  rgb(105pt)=(0.949942329873126,0.631141868512111,0.49757785467128);
  rgb(106pt)=(0.945328719723183,0.620530565167243,0.489427143406382);
  rgb(107pt)=(0.945328719723183,0.620530565167243,0.489427143406382);
  rgb(108pt)=(0.940715109573241,0.609919261822376,0.481276432141484);
  rgb(109pt)=(0.940715109573241,0.609919261822376,0.481276432141484);
  rgb(110pt)=(0.936101499423299,0.599307958477509,0.473125720876586);
  rgb(111pt)=(0.936101499423299,0.599307958477509,0.473125720876586);
  rgb(112pt)=(0.931487889273356,0.588696655132641,0.464975009611688);
  rgb(113pt)=(0.931487889273356,0.588696655132641,0.464975009611688);
  rgb(114pt)=(0.926874279123414,0.578085351787774,0.45682429834679);
  rgb(115pt)=(0.926874279123414,0.578085351787774,0.45682429834679);
  rgb(116pt)=(0.922260668973472,0.567474048442907,0.448673587081892);
  rgb(117pt)=(0.922260668973472,0.567474048442907,0.448673587081892);
  rgb(118pt)=(0.917647058823529,0.556862745098039,0.440522875816993);
  rgb(119pt)=(0.917647058823529,0.556862745098039,0.440522875816993);
  rgb(120pt)=(0.913033448673587,0.546251441753172,0.432372164552095);
  rgb(121pt)=(0.913033448673587,0.546251441753172,0.432372164552095);
  rgb(122pt)=(0.908419838523645,0.535640138408304,0.424221453287197);
  rgb(123pt)=(0.908419838523645,0.535640138408304,0.424221453287197);
  rgb(124pt)=(0.903806228373703,0.525028835063437,0.416070742022299);
  rgb(125pt)=(0.903806228373703,0.525028835063437,0.416070742022299);
  rgb(126pt)=(0.89919261822376,0.51441753171857,0.407920030757401);
  rgb(127pt)=(0.89919261822376,0.51441753171857,0.407920030757401);
  rgb(128pt)=(0.894579008073818,0.503806228373702,0.399769319492503);
  rgb(129pt)=(0.894579008073818,0.503806228373702,0.399769319492503);
  rgb(130pt)=(0.889965397923876,0.493194925028835,0.391618608227605);
  rgb(131pt)=(0.889965397923876,0.493194925028835,0.391618608227605);
  rgb(132pt)=(0.885351787773933,0.482583621683968,0.383467896962707);
  rgb(133pt)=(0.885351787773933,0.482583621683968,0.383467896962707);
  rgb(134pt)=(0.880738177623991,0.4719723183391,0.375317185697808);
  rgb(135pt)=(0.880738177623991,0.4719723183391,0.375317185697808);
  rgb(136pt)=(0.876124567474049,0.461361014994233,0.367166474432911);
  rgb(137pt)=(0.876124567474049,0.461361014994233,0.367166474432911);
  rgb(138pt)=(0.871510957324106,0.450749711649366,0.359015763168012);
  rgb(139pt)=(0.871510957324106,0.450749711649366,0.359015763168012);
  rgb(140pt)=(0.866897347174164,0.440138408304498,0.350865051903114);
  rgb(141pt)=(0.866897347174164,0.440138408304498,0.350865051903114);
  rgb(142pt)=(0.862283737024221,0.429527104959631,0.342714340638216);
  rgb(143pt)=(0.862283737024221,0.429527104959631,0.342714340638216);
  rgb(144pt)=(0.857670126874279,0.418915801614764,0.334563629373318);
  rgb(145pt)=(0.857670126874279,0.418915801614764,0.334563629373318);
  rgb(146pt)=(0.853056516724337,0.408304498269896,0.32641291810842);
  rgb(147pt)=(0.853056516724337,0.408304498269896,0.32641291810842);
  rgb(148pt)=(0.848442906574394,0.397693194925029,0.318262206843522);
  rgb(149pt)=(0.848442906574394,0.397693194925029,0.318262206843522);
  rgb(150pt)=(0.843829296424452,0.387081891580161,0.310111495578624);
  rgb(151pt)=(0.843829296424452,0.387081891580161,0.310111495578624);
  rgb(152pt)=(0.83921568627451,0.376470588235294,0.301960784313725);
  rgb(153pt)=(0.83921568627451,0.376470588235294,0.301960784313725);
  rgb(154pt)=(0.833679354094579,0.365397923875433,0.296732026143791);
  rgb(155pt)=(0.833679354094579,0.365397923875433,0.296732026143791);
  rgb(156pt)=(0.828143021914648,0.354325259515571,0.291503267973856);
  rgb(157pt)=(0.828143021914648,0.354325259515571,0.291503267973856);
  rgb(158pt)=(0.822606689734717,0.343252595155709,0.286274509803922);
  rgb(159pt)=(0.822606689734717,0.343252595155709,0.286274509803922);
  rgb(160pt)=(0.817070357554787,0.332179930795848,0.281045751633987);
  rgb(161pt)=(0.817070357554787,0.332179930795848,0.281045751633987);
  rgb(162pt)=(0.811534025374856,0.321107266435986,0.275816993464052);
  rgb(163pt)=(0.811534025374856,0.321107266435986,0.275816993464052);
  rgb(164pt)=(0.805997693194925,0.310034602076125,0.270588235294118);
  rgb(165pt)=(0.805997693194925,0.310034602076125,0.270588235294118);
  rgb(166pt)=(0.800461361014994,0.298961937716263,0.265359477124183);
  rgb(167pt)=(0.800461361014994,0.298961937716263,0.265359477124183);
  rgb(168pt)=(0.794925028835064,0.287889273356402,0.260130718954248);
  rgb(169pt)=(0.794925028835064,0.287889273356402,0.260130718954248);
  rgb(170pt)=(0.789388696655133,0.27681660899654,0.254901960784314);
  rgb(171pt)=(0.789388696655133,0.27681660899654,0.254901960784314);
  rgb(172pt)=(0.783852364475202,0.265743944636678,0.249673202614379);
  rgb(173pt)=(0.783852364475202,0.265743944636678,0.249673202614379);
  rgb(174pt)=(0.778316032295271,0.254671280276817,0.244444444444444);
  rgb(175pt)=(0.778316032295271,0.254671280276817,0.244444444444444);
  rgb(176pt)=(0.77277970011534,0.243598615916955,0.23921568627451);
  rgb(177pt)=(0.77277970011534,0.243598615916955,0.23921568627451);
  rgb(178pt)=(0.767243367935409,0.232525951557093,0.233986928104575);
  rgb(179pt)=(0.767243367935409,0.232525951557093,0.233986928104575);
  rgb(180pt)=(0.761707035755479,0.221453287197232,0.228758169934641);
  rgb(181pt)=(0.761707035755479,0.221453287197232,0.228758169934641);
  rgb(182pt)=(0.756170703575548,0.21038062283737,0.223529411764706);
  rgb(183pt)=(0.756170703575548,0.21038062283737,0.223529411764706);
  rgb(184pt)=(0.750634371395617,0.199307958477509,0.218300653594771);
  rgb(185pt)=(0.750634371395617,0.199307958477509,0.218300653594771);
  rgb(186pt)=(0.745098039215686,0.188235294117647,0.213071895424837);
  rgb(187pt)=(0.745098039215686,0.188235294117647,0.213071895424837);
  rgb(188pt)=(0.739561707035755,0.177162629757785,0.207843137254902);
  rgb(189pt)=(0.739561707035755,0.177162629757785,0.207843137254902);
  rgb(190pt)=(0.734025374855825,0.166089965397924,0.202614379084967);
  rgb(191pt)=(0.734025374855825,0.166089965397924,0.202614379084967);
  rgb(192pt)=(0.728489042675894,0.155017301038062,0.197385620915033);
  rgb(193pt)=(0.728489042675894,0.155017301038062,0.197385620915033);
  rgb(194pt)=(0.722952710495963,0.143944636678201,0.192156862745098);
  rgb(195pt)=(0.722952710495963,0.143944636678201,0.192156862745098);
  rgb(196pt)=(0.717416378316032,0.132871972318339,0.186928104575163);
  rgb(197pt)=(0.717416378316032,0.132871972318339,0.186928104575163);
  rgb(198pt)=(0.711880046136101,0.121799307958477,0.181699346405229);
  rgb(199pt)=(0.711880046136101,0.121799307958477,0.181699346405229);
  rgb(200pt)=(0.706343713956171,0.110726643598616,0.176470588235294);
  rgb(201pt)=(0.706343713956171,0.110726643598616,0.176470588235294);
  rgb(202pt)=(0.70080738177624,0.0996539792387544,0.17124183006536);
  rgb(203pt)=(0.70080738177624,0.0996539792387544,0.17124183006536);
  rgb(204pt)=(0.692272202998847,0.0922722029988466,0.167704728950404);
  rgb(205pt)=(0.692272202998847,0.0922722029988466,0.167704728950404);
  rgb(206pt)=(0.680738177623991,0.0885813148788927,0.165859284890427);
  rgb(207pt)=(0.680738177623991,0.0885813148788927,0.165859284890427);
  rgb(208pt)=(0.669204152249135,0.0848904267589389,0.16401384083045);
  rgb(209pt)=(0.669204152249135,0.0848904267589389,0.16401384083045);
  rgb(210pt)=(0.657670126874279,0.081199538638985,0.162168396770473);
  rgb(211pt)=(0.657670126874279,0.081199538638985,0.162168396770473);
  rgb(212pt)=(0.646136101499423,0.0775086505190311,0.160322952710496);
  rgb(213pt)=(0.646136101499423,0.0775086505190311,0.160322952710496);
  rgb(214pt)=(0.634602076124567,0.0738177623990773,0.158477508650519);
  rgb(215pt)=(0.634602076124567,0.0738177623990773,0.158477508650519);
  rgb(216pt)=(0.623068050749712,0.0701268742791234,0.156632064590542);
  rgb(217pt)=(0.623068050749712,0.0701268742791234,0.156632064590542);
  rgb(218pt)=(0.611534025374856,0.0664359861591695,0.154786620530565);
  rgb(219pt)=(0.611534025374856,0.0664359861591695,0.154786620530565);
  rgb(220pt)=(0.6,0.0627450980392157,0.152941176470588);
  rgb(221pt)=(0.6,0.0627450980392157,0.152941176470588);
  rgb(222pt)=(0.588465974625144,0.0590542099192618,0.151095732410611);
  rgb(223pt)=(0.588465974625144,0.0590542099192618,0.151095732410611);
  rgb(224pt)=(0.576931949250288,0.055363321799308,0.149250288350634);
  rgb(225pt)=(0.576931949250288,0.055363321799308,0.149250288350634);
  rgb(226pt)=(0.565397923875433,0.0516724336793541,0.147404844290657);
  rgb(227pt)=(0.565397923875433,0.0516724336793541,0.147404844290657);
  rgb(228pt)=(0.553863898500577,0.0479815455594002,0.145559400230681);
  rgb(229pt)=(0.553863898500577,0.0479815455594002,0.145559400230681);
  rgb(230pt)=(0.542329873125721,0.0442906574394464,0.143713956170704);
  rgb(231pt)=(0.542329873125721,0.0442906574394464,0.143713956170704);
  rgb(232pt)=(0.530795847750865,0.0405997693194926,0.141868512110727);
  rgb(233pt)=(0.530795847750865,0.0405997693194926,0.141868512110727);
  rgb(234pt)=(0.519261822376009,0.0369088811995386,0.14002306805075);
  rgb(235pt)=(0.519261822376009,0.0369088811995386,0.14002306805075);
  rgb(236pt)=(0.507727797001153,0.0332179930795848,0.138177623990773);
  rgb(237pt)=(0.507727797001153,0.0332179930795848,0.138177623990773);
  rgb(238pt)=(0.496193771626298,0.0295271049596309,0.136332179930796);
  rgb(239pt)=(0.496193771626298,0.0295271049596309,0.136332179930796);
  rgb(240pt)=(0.484659746251442,0.025836216839677,0.134486735870819);
  rgb(241pt)=(0.484659746251442,0.025836216839677,0.134486735870819);
  rgb(242pt)=(0.473125720876586,0.0221453287197232,0.132641291810842);
  rgb(243pt)=(0.473125720876586,0.0221453287197232,0.132641291810842);
  rgb(244pt)=(0.46159169550173,0.0184544405997693,0.130795847750865);
  rgb(245pt)=(0.46159169550173,0.0184544405997693,0.130795847750865);
  rgb(246pt)=(0.450057670126874,0.0147635524798155,0.128950403690888);
  rgb(247pt)=(0.450057670126874,0.0147635524798155,0.128950403690888);
  rgb(248pt)=(0.438523644752018,0.0110726643598616,0.127104959630911);
  rgb(249pt)=(0.438523644752018,0.0110726643598616,0.127104959630911);
  rgb(250pt)=(0.426989619377163,0.00738177623990774,0.125259515570934);
  rgb(251pt)=(0.426989619377163,0.00738177623990774,0.125259515570934);
  rgb(252pt)=(0.415455594002307,0.00369088811995386,0.123414071510957);
  rgb(253pt)=(0.415455594002307,0.00369088811995386,0.123414071510957);
  rgb(254pt)=(0.403921568627451,0,0.12156862745098);
  rgb(255pt)=(0.403921568627451,0,0.12156862745098)
},
point meta max=0.937,
point meta min=0.001,
tick align=outside,
tick pos=left,
x grid style={white!69.0196078431373!black},
xlabel={\# Features},
xmin=0, xmax=3,
xtick style={color=black},
xtick={0.5,1.5,2.5},
xticklabels={1,2,10},
y dir=reverse,
y grid style={white!69.0196078431373!black},
ylabel={\# Training Examples},
ymin=0, ymax=4,
ytick style={color=black},
ytick={0.5,1.5,2.5,3.5},
yticklabel style={rotate=90.0},
yticklabels={10,20,30,40},
font=\small,width=.75\textwidth, height=.55\textwidth,
]
\addplot graphics [includegraphics cmd=\pgfimage,xmin=0, xmax=3, ymin=4, ymax=0] {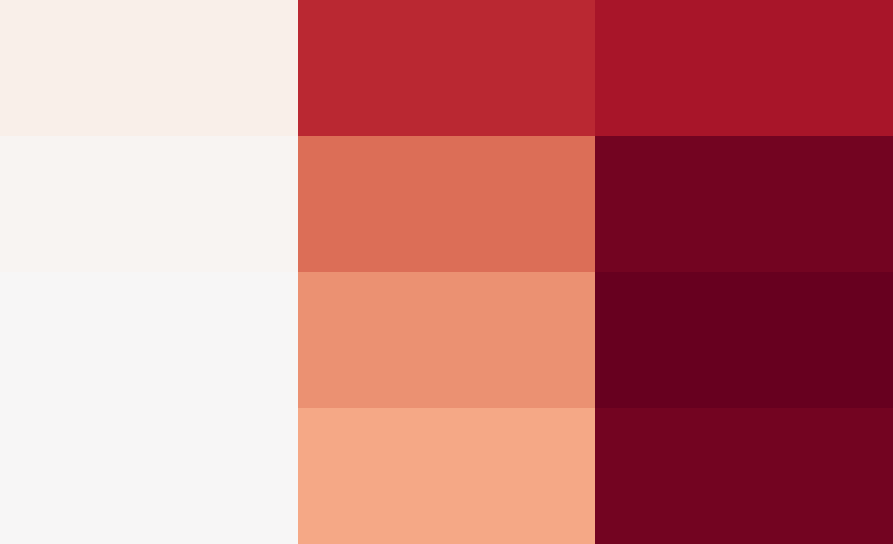};
\draw (axis cs:0.5,0.5) node[
  text=white!15!black,
  rotate=0.0
]{0.053};
\draw (axis cs:1.5,0.5) node[
  text=white,
  rotate=0.0
]{0.71};
\draw (axis cs:2.5,0.5) node[
  text=white,
  rotate=0.0
]{0.77};
\draw (axis cs:0.5,1.5) node[
  text=white!15!black,
  rotate=0.0
]{0.02};
\draw (axis cs:1.5,1.5) node[
  text=white,
  rotate=0.0
]{0.53};
\draw (axis cs:2.5,1.5) node[
  text=white,
  rotate=0.0
]{0.9};
\draw (axis cs:0.5,2.5) node[
  text=white!15!black,
  rotate=0.0
]{0.005};
\draw (axis cs:1.5,2.5) node[
  text=white,
  rotate=0.0
]{0.43};
\draw (axis cs:2.5,2.5) node[
  text=white,
  rotate=0.0
]{0.94};
\draw (axis cs:0.5,3.5) node[
  text=white!15!black,
  rotate=0.0
]{0.001};
\draw (axis cs:1.5,3.5) node[
  text=white!15!black,
  rotate=0.0
]{0.36};
\draw (axis cs:2.5,3.5) node[
  text=white,
  rotate=0.0
]{0.91};
\end{axis}

\end{tikzpicture}

%% file: icml2023/figures/hpob_diffbaselines_rank_test.tex
\begin{tikzpicture}

\begin{axis}[
legend cell align={left},
legend style={
  fill opacity=0.8,
  draw opacity=1,
  text opacity=1,
  at={(0.55,-0.15)},
  anchor=south,
  draw=white!80!black
},
  height=.6\textwidth,
    width=.8\textwidth,
tick align=outside,
tick pos=left,
x grid style={white!69.0196078431373!black},
xlabel={Number of trials},
xmin=2.75, xmax=52.25,
xtick style={color=black},
y grid style={white!69.0196078431373!black},
ylabel={Average Rank},
ymin=2.89841728742347, ymax=5.72187797803099,
ytick style={color=black}
]
\path [fill=colorRandom, fill opacity=0.2]
(axis cs:5,4.23095960690422)
--(axis cs:5,4.09845215780166)
--(axis cs:6,4.5090373408398)
--(axis cs:7,4.59937109639382)
--(axis cs:8,4.6073771403292)
--(axis cs:9,4.57588832991028)
--(axis cs:10,4.61874998443501)
--(axis cs:11,4.76489377407815)
--(axis cs:12,4.80152350432528)
--(axis cs:13,4.89336899841156)
--(axis cs:14,4.90414750148562)
--(axis cs:15,4.99360332747644)
--(axis cs:16,5.01969683762245)
--(axis cs:17,5.07026240806039)
--(axis cs:18,5.09380937374877)
--(axis cs:19,5.09821187935203)
--(axis cs:20,5.16322075899301)
--(axis cs:21,5.21393863561823)
--(axis cs:22,5.2135464455046)
--(axis cs:23,5.23744106070679)
--(axis cs:24,5.22666214976116)
--(axis cs:25,5.27440849052642)
--(axis cs:26,5.29023496505714)
--(axis cs:27,5.32982750307457)
--(axis cs:28,5.32473960148193)
--(axis cs:29,5.35556229732025)
--(axis cs:30,5.35386107920493)
--(axis cs:31,5.35680292723363)
--(axis cs:32,5.3643100792185)
--(axis cs:33,5.35936538253295)
--(axis cs:34,5.36741672216956)
--(axis cs:35,5.36734799203838)
--(axis cs:36,5.37961792727632)
--(axis cs:37,5.35575881601266)
--(axis cs:38,5.36104096627775)
--(axis cs:39,5.36375065225555)
--(axis cs:40,5.37236028590365)
--(axis cs:41,5.37791659030838)
--(axis cs:42,5.36869645225233)
--(axis cs:43,5.37162140426904)
--(axis cs:44,5.36085083441908)
--(axis cs:45,5.39132970085693)
--(axis cs:46,5.41292191508067)
--(axis cs:47,5.42518322520462)
--(axis cs:48,5.41130532351595)
--(axis cs:49,5.42474336731045)
--(axis cs:50,5.43391212466151)
--(axis cs:50,5.59353885573065)
--(axis cs:50,5.59353885573065)
--(axis cs:49,5.579178201317)
--(axis cs:48,5.56516526471935)
--(axis cs:47,5.56697363754048)
--(axis cs:46,5.55962710452717)
--(axis cs:45,5.56161147561366)
--(axis cs:44,5.54110994989465)
--(axis cs:43,5.55386879180939)
--(axis cs:42,5.54502903794375)
--(axis cs:41,5.53973046851515)
--(axis cs:40,5.52960049841008)
--(axis cs:39,5.52252385754837)
--(axis cs:38,5.50954726901637)
--(axis cs:37,5.51482941928146)
--(axis cs:36,5.51057815115506)
--(axis cs:35,5.51108338051064)
--(axis cs:34,5.51885778763436)
--(axis cs:33,5.49945814687882)
--(axis cs:32,5.49843501882072)
--(axis cs:31,5.48633432766834)
--(axis cs:30,5.46182519530488)
--(axis cs:29,5.44443770267975)
--(axis cs:28,5.43604471224357)
--(axis cs:27,5.42311367339602)
--(axis cs:26,5.39211797611933)
--(axis cs:25,5.3883366075128)
--(axis cs:24,5.35373000710159)
--(axis cs:23,5.3468726647834)
--(axis cs:22,5.32763002508363)
--(axis cs:21,5.31547312908765)
--(axis cs:20,5.26423022139915)
--(axis cs:19,5.20767047358914)
--(axis cs:18,5.21599454781986)
--(axis cs:17,5.16111014095922)
--(axis cs:16,5.14500904473049)
--(axis cs:15,5.12404373134709)
--(axis cs:14,5.02526426322027)
--(axis cs:13,4.99290551139236)
--(axis cs:12,4.9043588486159)
--(axis cs:11,4.92138073572577)
--(axis cs:10,4.78517158419244)
--(axis cs:9,4.73391559165835)
--(axis cs:8,4.71026991849433)
--(axis cs:7,4.70651125654735)
--(axis cs:6,4.61645285523863)
--(axis cs:5,4.23095960690422)
--cycle;

\path [fill=colorHEBO, fill opacity=0.2]
(axis cs:5,4.23095960690422)
--(axis cs:5,4.09845215780166)
--(axis cs:6,4.40493100148895)
--(axis cs:7,4.52324194574595)
--(axis cs:8,4.4989444246479)
--(axis cs:9,4.59167981381391)
--(axis cs:10,4.60032080305248)
--(axis cs:11,4.50302115392715)
--(axis cs:12,4.36912500521407)
--(axis cs:13,4.22972853055087)
--(axis cs:14,4.21864482058694)
--(axis cs:15,4.14472120389765)
--(axis cs:16,4.05105466332413)
--(axis cs:17,4.04082398822861)
--(axis cs:18,4.01222372959759)
--(axis cs:19,4.01086389651949)
--(axis cs:20,3.91906171590823)
--(axis cs:21,3.80265379276382)
--(axis cs:22,3.74714740704144)
--(axis cs:23,3.67905474796521)
--(axis cs:24,3.67208648423792)
--(axis cs:25,3.64608409854608)
--(axis cs:26,3.60443725617496)
--(axis cs:27,3.58638359515056)
--(axis cs:28,3.5560161659234)
--(axis cs:29,3.56408259866647)
--(axis cs:30,3.48954367321228)
--(axis cs:31,3.44449545721896)
--(axis cs:32,3.44346172100895)
--(axis cs:33,3.44178162337744)
--(axis cs:34,3.41799192422801)
--(axis cs:35,3.40081015502723)
--(axis cs:36,3.40446414090913)
--(axis cs:37,3.38068095035502)
--(axis cs:38,3.38585001182014)
--(axis cs:39,3.39149089840417)
--(axis cs:40,3.36344056881845)
--(axis cs:41,3.3741212008766)
--(axis cs:42,3.34858823014094)
--(axis cs:43,3.33632905247724)
--(axis cs:44,3.34150677146809)
--(axis cs:45,3.29871718765408)
--(axis cs:46,3.24755195305821)
--(axis cs:47,3.24773435131488)
--(axis cs:48,3.24497179155689)
--(axis cs:49,3.24858823014094)
--(axis cs:50,3.22659838968283)
--(axis cs:50,3.30673494365051)
--(axis cs:50,3.30673494365051)
--(axis cs:49,3.3082745149571)
--(axis cs:48,3.31189095354115)
--(axis cs:47,3.30912839378316)
--(axis cs:46,3.32107549792219)
--(axis cs:45,3.36402791038513)
--(axis cs:44,3.39574813049269)
--(axis cs:43,3.4126905553659)
--(axis cs:42,3.4082745149571)
--(axis cs:41,3.43764350500575)
--(axis cs:40,3.46401041157371)
--(axis cs:39,3.4830189055174)
--(axis cs:38,3.50042449798378)
--(axis cs:37,3.48598571631165)
--(axis cs:36,3.4974966434046)
--(axis cs:35,3.48938592340414)
--(axis cs:34,3.49573356596806)
--(axis cs:33,3.56213994525001)
--(axis cs:32,3.5604598476185)
--(axis cs:31,3.57119081729084)
--(axis cs:30,3.63594652286615)
--(axis cs:29,3.71434877388255)
--(axis cs:28,3.69888579486091)
--(axis cs:27,3.73910660092787)
--(axis cs:26,3.75242548892308)
--(axis cs:25,3.78528845047353)
--(axis cs:24,3.7985017510562)
--(axis cs:23,3.83074917360341)
--(axis cs:22,3.87638200472327)
--(axis cs:21,3.92675797194206)
--(axis cs:20,4.02603632330746)
--(axis cs:19,4.08325375053933)
--(axis cs:18,4.08973705471614)
--(axis cs:17,4.14741130588903)
--(axis cs:16,4.14894533667587)
--(axis cs:15,4.23174938433764)
--(axis cs:14,4.29900223823658)
--(axis cs:13,4.29184009690011)
--(axis cs:12,4.43479656341338)
--(axis cs:11,4.64207688528854)
--(axis cs:10,4.74085566753576)
--(axis cs:9,4.69851626461746)
--(axis cs:8,4.63046734005799)
--(axis cs:7,4.65322864248934)
--(axis cs:6,4.53232390047184)
--(axis cs:5,4.23095960690422)
--cycle;

\path [fill=colorOPT_BEI, fill opacity=0.2]
(axis cs:5,3.75386960549584)
--(axis cs:5,3.42260098273945)
--(axis cs:6,3.2530543574121)
--(axis cs:7,3.37810142927477)
--(axis cs:8,3.43493078972016)
--(axis cs:9,3.49829586823853)
--(axis cs:10,3.55974307136895)
--(axis cs:11,3.64906573683572)
--(axis cs:12,3.71091741861831)
--(axis cs:13,3.72510318054808)
--(axis cs:14,3.77336314978964)
--(axis cs:15,3.79745647081234)
--(axis cs:16,3.8498091146679)
--(axis cs:17,3.87002211226583)
--(axis cs:18,3.86868962816659)
--(axis cs:19,3.90744896276975)
--(axis cs:20,3.96184529888496)
--(axis cs:21,4.01214878962564)
--(axis cs:22,4.04622469324524)
--(axis cs:23,4.08093951294165)
--(axis cs:24,4.11575368473584)
--(axis cs:25,4.13381249132902)
--(axis cs:26,4.19303806729231)
--(axis cs:27,4.21234202776965)
--(axis cs:28,4.21848693277502)
--(axis cs:29,4.22639977891716)
--(axis cs:30,4.26838589399128)
--(axis cs:31,4.29611082856485)
--(axis cs:32,4.30402858350668)
--(axis cs:33,4.33082633162787)
--(axis cs:34,4.3391098112969)
--(axis cs:35,4.35064176850475)
--(axis cs:36,4.38198850367999)
--(axis cs:37,4.41768399254911)
--(axis cs:38,4.43379437430384)
--(axis cs:39,4.40392787308792)
--(axis cs:40,4.42981775001914)
--(axis cs:41,4.43722073833964)
--(axis cs:42,4.43724018249137)
--(axis cs:43,4.43078200627989)
--(axis cs:44,4.47177551450276)
--(axis cs:45,4.50554083884109)
--(axis cs:46,4.5339660896812)
--(axis cs:47,4.52600588362594)
--(axis cs:48,4.53814203443957)
--(axis cs:49,4.55006281401198)
--(axis cs:50,4.55218099869404)
--(axis cs:50,4.71840723660008)
--(axis cs:50,4.71840723660008)
--(axis cs:49,4.70091757814488)
--(axis cs:48,4.68146580869768)
--(axis cs:47,4.67399411637406)
--(axis cs:46,4.68956332208351)
--(axis cs:45,4.67092974939421)
--(axis cs:44,4.65371468157567)
--(axis cs:43,4.62019838587698)
--(axis cs:42,4.61374020966549)
--(axis cs:41,4.60591651656232)
--(axis cs:40,4.57802538723576)
--(axis cs:39,4.54901330338267)
--(axis cs:38,4.53483307667655)
--(axis cs:37,4.52349247803913)
--(axis cs:36,4.49252130024158)
--(axis cs:35,4.45720136875015)
--(axis cs:34,4.44128234556585)
--(axis cs:33,4.41819327621526)
--(axis cs:32,4.38616749492469)
--(axis cs:31,4.38232054398417)
--(axis cs:30,4.36690822365578)
--(axis cs:29,4.33046296618088)
--(axis cs:28,4.33053267506812)
--(axis cs:27,4.30138346242643)
--(axis cs:26,4.29715801113906)
--(axis cs:25,4.25834437141608)
--(axis cs:24,4.26071690349946)
--(axis cs:23,4.20925656548972)
--(axis cs:22,4.14201060087241)
--(axis cs:21,4.10549826919789)
--(axis cs:20,4.04991940699739)
--(axis cs:19,4.01019809605378)
--(axis cs:18,4.00582017575498)
--(axis cs:17,3.98487984851849)
--(axis cs:16,3.96979872846936)
--(axis cs:15,3.93195529389354)
--(axis cs:14,3.89330351687702)
--(axis cs:13,3.87489681945192)
--(axis cs:12,3.84986689510718)
--(axis cs:11,3.78230681218389)
--(axis cs:10,3.68339418353301)
--(axis cs:9,3.64680217097715)
--(axis cs:8,3.58859862204454)
--(axis cs:7,3.57091817856837)
--(axis cs:6,3.50380838768594)
--(axis cs:5,3.75386960549584)
--cycle;

\path [fill=colorOPT_REI, fill opacity=0.2]
(axis cs:5,3.75386960549584)
--(axis cs:5,3.42260098273945)
--(axis cs:6,3.18437760394158)
--(axis cs:7,3.36617775030756)
--(axis cs:8,3.38440230900222)
--(axis cs:9,3.53824478326505)
--(axis cs:10,3.55442731908691)
--(axis cs:11,3.54944976047507)
--(axis cs:12,3.63819710582868)
--(axis cs:13,3.64763827276623)
--(axis cs:14,3.68228783398091)
--(axis cs:15,3.6919950369464)
--(axis cs:16,3.71679872835558)
--(axis cs:17,3.70467957357041)
--(axis cs:18,3.7355504394621)
--(axis cs:19,3.75020981742996)
--(axis cs:20,3.75928457038227)
--(axis cs:21,3.79045375777317)
--(axis cs:22,3.84541480506067)
--(axis cs:23,3.86223404422)
--(axis cs:24,3.87581544863688)
--(axis cs:25,3.87417561354975)
--(axis cs:26,3.92479503815626)
--(axis cs:27,3.95808463030357)
--(axis cs:28,3.95031054977737)
--(axis cs:29,3.98090925992554)
--(axis cs:30,4.00161253002942)
--(axis cs:31,4.00460711663509)
--(axis cs:32,4.04632855987201)
--(axis cs:33,4.04241483830288)
--(axis cs:34,4.04972091517081)
--(axis cs:35,4.05746831013973)
--(axis cs:36,4.09550806429603)
--(axis cs:37,4.11814994169722)
--(axis cs:38,4.14291999848778)
--(axis cs:39,4.16051503775358)
--(axis cs:40,4.10716227211411)
--(axis cs:41,4.13118966142429)
--(axis cs:42,4.1214219004258)
--(axis cs:43,4.11259737632357)
--(axis cs:44,4.11678311735339)
--(axis cs:45,4.15890829175934)
--(axis cs:46,4.14515375017158)
--(axis cs:47,4.15317169695584)
--(axis cs:48,4.13658135628311)
--(axis cs:49,4.14376853376067)
--(axis cs:50,4.12106398035096)
--(axis cs:50,4.18481837259021)
--(axis cs:50,4.18481837259021)
--(axis cs:49,4.18956479957266)
--(axis cs:48,4.20459511430512)
--(axis cs:47,4.2036910481422)
--(axis cs:46,4.19602272041666)
--(axis cs:45,4.20971915922105)
--(axis cs:44,4.15380511794073)
--(axis cs:43,4.16975556485291)
--(axis cs:42,4.17269574663303)
--(axis cs:41,4.19037896602669)
--(axis cs:40,4.21048478670942)
--(axis cs:39,4.26693594263857)
--(axis cs:38,4.25315843288477)
--(axis cs:37,4.24263437202827)
--(axis cs:36,4.23390370040985)
--(axis cs:35,4.20527678789948)
--(axis cs:34,4.20125947698605)
--(axis cs:33,4.18503614208928)
--(axis cs:32,4.18504398914759)
--(axis cs:31,4.15225562846295)
--(axis cs:30,4.15917178369607)
--(axis cs:29,4.12497309301564)
--(axis cs:28,4.07321886198734)
--(axis cs:27,4.07720948734349)
--(axis cs:26,4.02814613831433)
--(axis cs:25,3.98856948448947)
--(axis cs:24,3.94771396312783)
--(axis cs:23,3.93384438715255)
--(axis cs:22,3.90752637140991)
--(axis cs:21,3.8840560461484)
--(axis cs:20,3.86032327275499)
--(axis cs:19,3.90077057472691)
--(axis cs:18,3.8722926977928)
--(axis cs:17,3.80120277937077)
--(axis cs:16,3.83222087948756)
--(axis cs:15,3.77467162972026)
--(axis cs:14,3.7569278522936)
--(axis cs:13,3.70922447233181)
--(axis cs:12,3.68729309024975)
--(axis cs:11,3.61525612187787)
--(axis cs:10,3.61027856326603)
--(axis cs:9,3.59901011869574)
--(axis cs:8,3.51363690668405)
--(axis cs:7,3.48088107322185)
--(axis cs:6,3.43915180782313)
--(axis cs:5,3.75386960549584)
--cycle;

\path [fill=colorBOHAMIANN, fill opacity=0.2]
(axis cs:5,4.23095960690422)
--(axis cs:5,4.09845215780166)
--(axis cs:6,4.10502643852251)
--(axis cs:7,4.0371096699239)
--(axis cs:8,3.99947696023058)
--(axis cs:9,4.0337692213253)
--(axis cs:10,4.07475979288306)
--(axis cs:11,3.98505643114512)
--(axis cs:12,4.05132624785386)
--(axis cs:13,4.02388186233609)
--(axis cs:14,3.94543429849292)
--(axis cs:15,3.92908396374313)
--(axis cs:16,3.94470840400732)
--(axis cs:17,3.98539137737889)
--(axis cs:18,3.970083924597)
--(axis cs:19,3.9763087624366)
--(axis cs:20,3.96524006561052)
--(axis cs:21,3.96873268909746)
--(axis cs:22,3.9920105724987)
--(axis cs:23,3.95708667120922)
--(axis cs:24,3.93378208982121)
--(axis cs:25,3.95927867454378)
--(axis cs:26,3.93917984550312)
--(axis cs:27,3.93270962637527)
--(axis cs:28,3.95112878789184)
--(axis cs:29,3.92927721778668)
--(axis cs:30,3.95214031705382)
--(axis cs:31,3.97430996612676)
--(axis cs:32,3.95820841298381)
--(axis cs:33,3.96218197854884)
--(axis cs:34,3.96142057105904)
--(axis cs:35,3.96286324199976)
--(axis cs:36,3.96005237484772)
--(axis cs:37,3.96184529888496)
--(axis cs:38,3.97876414462122)
--(axis cs:39,3.96071190679938)
--(axis cs:40,4.00329791419889)
--(axis cs:41,4.02269181173906)
--(axis cs:42,4.03753079506535)
--(axis cs:43,4.02392730776444)
--(axis cs:44,4.0167537028759)
--(axis cs:45,4.03104672178588)
--(axis cs:46,4.03504965151474)
--(axis cs:47,4.05042462172746)
--(axis cs:48,4.05619678075711)
--(axis cs:49,3.99402623112246)
--(axis cs:50,3.97858114173543)
--(axis cs:50,4.07632081904888)
--(axis cs:50,4.07632081904888)
--(axis cs:49,4.09224827868146)
--(axis cs:48,4.14772478787034)
--(axis cs:47,4.14565380964508)
--(axis cs:46,4.12573466221075)
--(axis cs:45,4.12189445468471)
--(axis cs:44,4.14403061084959)
--(axis cs:43,4.15254328047085)
--(axis cs:42,4.1507044990523)
--(axis cs:41,4.12240622747662)
--(axis cs:40,4.10650600736974)
--(axis cs:39,4.04320966182807)
--(axis cs:38,4.05260840439839)
--(axis cs:37,4.04991940699739)
--(axis cs:36,4.03602605652483)
--(axis cs:35,4.04890146388259)
--(axis cs:34,4.04642256619586)
--(axis cs:33,4.04958272733351)
--(axis cs:32,4.0653209987809)
--(axis cs:31,4.08059199465755)
--(axis cs:30,4.07138909471088)
--(axis cs:29,4.05503650770351)
--(axis cs:28,4.08808689838267)
--(axis cs:27,4.07121194225218)
--(axis cs:26,4.09219270351648)
--(axis cs:25,4.13483897251504)
--(axis cs:24,4.10151202782585)
--(axis cs:23,4.09389372094764)
--(axis cs:22,4.11779334906993)
--(axis cs:21,4.07832613443195)
--(axis cs:20,4.06613248340909)
--(axis cs:19,4.05898535521046)
--(axis cs:18,4.06128862442261)
--(axis cs:17,4.09696156379758)
--(axis cs:16,4.09058571363973)
--(axis cs:15,4.07091603625687)
--(axis cs:14,4.09378138778159)
--(axis cs:13,4.11729460825214)
--(axis cs:12,4.10945806587163)
--(axis cs:11,4.08161023552155)
--(axis cs:10,4.15269118750909)
--(axis cs:9,4.1819170531845)
--(axis cs:8,4.16915049074981)
--(axis cs:7,4.23347856537022)
--(axis cs:6,4.31458140461474)
--(axis cs:5,4.23095960690422)
--cycle;

\path [fill=colorPFN_GP, fill opacity=0.2]
(axis cs:5,4.23095960690422)
--(axis cs:5,4.09845215780166)
--(axis cs:6,3.96550696719471)
--(axis cs:7,3.82582954439413)
--(axis cs:8,3.8365492919393)
--(axis cs:9,3.71822189671434)
--(axis cs:10,3.68082654969785)
--(axis cs:11,3.65763088715845)
--(axis cs:12,3.63896081741859)
--(axis cs:13,3.63078094494604)
--(axis cs:14,3.62451327999838)
--(axis cs:15,3.56972847275224)
--(axis cs:16,3.58379247279526)
--(axis cs:17,3.57471657451804)
--(axis cs:18,3.52851231422207)
--(axis cs:19,3.52995355400818)
--(axis cs:20,3.49578679555586)
--(axis cs:21,3.44707419730843)
--(axis cs:22,3.37038371341957)
--(axis cs:23,3.34593975484542)
--(axis cs:24,3.38128579581351)
--(axis cs:25,3.33838445644912)
--(axis cs:26,3.32408503138009)
--(axis cs:27,3.32041258192728)
--(axis cs:28,3.30800663972425)
--(axis cs:29,3.26652785994391)
--(axis cs:30,3.25295019975226)
--(axis cs:31,3.24049560544744)
--(axis cs:32,3.22081633789159)
--(axis cs:33,3.17154247163605)
--(axis cs:34,3.19463177884383)
--(axis cs:35,3.18382205574421)
--(axis cs:36,3.14781507736165)
--(axis cs:37,3.12953859562998)
--(axis cs:38,3.06096931559742)
--(axis cs:39,3.08969515593962)
--(axis cs:40,3.04574089010089)
--(axis cs:41,3.03902339046553)
--(axis cs:42,3.05164295487617)
--(axis cs:43,3.06291506537821)
--(axis cs:44,3.06343072696479)
--(axis cs:45,3.02675640972381)
--(axis cs:46,3.0564969699074)
--(axis cs:47,3.0609704724715)
--(axis cs:48,3.07639363274324)
--(axis cs:49,3.09615270253981)
--(axis cs:50,3.10859255553015)
--(axis cs:50,3.20121136603848)
--(axis cs:50,3.20121136603848)
--(axis cs:49,3.19404337589156)
--(axis cs:48,3.17850832804108)
--(axis cs:47,3.17040207654811)
--(axis cs:46,3.1748755791122)
--(axis cs:45,3.14187104125658)
--(axis cs:44,3.19147123381952)
--(axis cs:43,3.19983003266101)
--(axis cs:42,3.17972959414344)
--(axis cs:41,3.18058445267173)
--(axis cs:40,3.19739636480107)
--(axis cs:39,3.23579504013881)
--(axis cs:38,3.22922676283396)
--(axis cs:37,3.26653983574257)
--(axis cs:36,3.30316531479521)
--(axis cs:35,3.31421715994207)
--(axis cs:34,3.31125057409735)
--(axis cs:33,3.31081046954042)
--(axis cs:32,3.32428170132409)
--(axis cs:31,3.34381812004276)
--(axis cs:30,3.35881450613009)
--(axis cs:29,3.38837410084041)
--(axis cs:28,3.46846394851104)
--(axis cs:27,3.46390114356291)
--(axis cs:26,3.47199339999246)
--(axis cs:25,3.50475279845284)
--(axis cs:24,3.59126322379433)
--(axis cs:23,3.54817789221341)
--(axis cs:22,3.57471432579611)
--(axis cs:21,3.63920031249549)
--(axis cs:20,3.68852692993433)
--(axis cs:19,3.68965428912908)
--(axis cs:18,3.71462494067989)
--(axis cs:17,3.73116577842313)
--(axis cs:16,3.77699184093023)
--(axis cs:15,3.75576172332619)
--(axis cs:14,3.75980044549182)
--(axis cs:13,3.78882689819122)
--(axis cs:12,3.78064702571867)
--(axis cs:11,3.85609460303763)
--(axis cs:10,3.9074087444198)
--(axis cs:9,3.91707222093272)
--(axis cs:8,4.08501933551168)
--(axis cs:7,3.99769986737058)
--(axis cs:6,4.09331656221706)
--(axis cs:5,4.23095960690422)
--cycle;

\path [fill=colorPFN_BNN, fill opacity=0.2]
(axis cs:5,4.23095960690422)
--(axis cs:5,4.09845215780166)
--(axis cs:6,3.94161210511987)
--(axis cs:7,3.76005899904748)
--(axis cs:8,3.69943309129529)
--(axis cs:9,3.59465296349087)
--(axis cs:10,3.47988110304353)
--(axis cs:11,3.47563380806791)
--(axis cs:12,3.46791250575974)
--(axis cs:13,3.50262399146809)
--(axis cs:14,3.47024814952066)
--(axis cs:15,3.42085656749299)
--(axis cs:16,3.36275140249968)
--(axis cs:17,3.3648027889355)
--(axis cs:18,3.37615610022522)
--(axis cs:19,3.3324369644404)
--(axis cs:20,3.34132623977465)
--(axis cs:21,3.35820860672896)
--(axis cs:22,3.3727436246027)
--(axis cs:23,3.3893619481472)
--(axis cs:24,3.30187233934553)
--(axis cs:25,3.27677228105996)
--(axis cs:26,3.26305808495815)
--(axis cs:27,3.20022632045408)
--(axis cs:28,3.20155427175201)
--(axis cs:29,3.21509573177008)
--(axis cs:30,3.21619760008125)
--(axis cs:31,3.22493914760277)
--(axis cs:32,3.21861127447363)
--(axis cs:33,3.23546529453583)
--(axis cs:34,3.22352271423728)
--(axis cs:35,3.21379431909533)
--(axis cs:36,3.17218409147021)
--(axis cs:37,3.18294722920066)
--(axis cs:38,3.18951443080114)
--(axis cs:39,3.1781286061212)
--(axis cs:40,3.20271313119523)
--(axis cs:41,3.18953538635335)
--(axis cs:42,3.20668118802314)
--(axis cs:43,3.20287676755626)
--(axis cs:44,3.20204857457935)
--(axis cs:45,3.18850356861959)
--(axis cs:46,3.18454856587418)
--(axis cs:47,3.16054670811649)
--(axis cs:48,3.13107819391626)
--(axis cs:49,3.15180648848323)
--(axis cs:50,3.15764502905562)
--(axis cs:50,3.34039418663066)
--(axis cs:50,3.34039418663066)
--(axis cs:49,3.32662488406579)
--(axis cs:48,3.31598062961315)
--(axis cs:47,3.30611995855018)
--(axis cs:46,3.31741221843955)
--(axis cs:45,3.32914349020394)
--(axis cs:44,3.34697103326379)
--(axis cs:43,3.35006440891433)
--(axis cs:42,3.3580246943298)
--(axis cs:41,3.35164108423488)
--(axis cs:40,3.38944373154987)
--(axis cs:39,3.35128315858469)
--(axis cs:38,3.3673483142969)
--(axis cs:37,3.36999394726993)
--(axis cs:36,3.38467865362783)
--(axis cs:35,3.43718607306154)
--(axis cs:34,3.43137924654704)
--(axis cs:33,3.43120137213084)
--(axis cs:32,3.42452598042833)
--(axis cs:31,3.4417275190639)
--(axis cs:30,3.41125338031091)
--(axis cs:29,3.40451211136717)
--(axis cs:28,3.39452415962054)
--(axis cs:27,3.38408740503612)
--(axis cs:26,3.42713799347322)
--(axis cs:25,3.43695320913612)
--(axis cs:24,3.4393041312427)
--(axis cs:23,3.48514785577437)
--(axis cs:22,3.46647206167181)
--(axis cs:21,3.45747766778084)
--(axis cs:20,3.4390659170881)
--(axis cs:19,3.4440336237949)
--(axis cs:18,3.45521644879439)
--(axis cs:17,3.46656976008411)
--(axis cs:16,3.50783683279444)
--(axis cs:15,3.5634571579972)
--(axis cs:14,3.55328126224405)
--(axis cs:13,3.57188581245348)
--(axis cs:12,3.55561690600497)
--(axis cs:11,3.51652305467718)
--(axis cs:10,3.55149144597608)
--(axis cs:9,3.67201370317579)
--(axis cs:8,3.84174337929294)
--(axis cs:7,3.86739198134468)
--(axis cs:6,4.13681926742915)
--(axis cs:5,4.23095960690422)
--cycle;

\addplot [line width=2pt, colorRandom]
table {%
5 4.16470588235294
6 4.56274509803922
7 4.65294117647059
8 4.65882352941177
9 4.65490196078431
10 4.70196078431373
11 4.84313725490196
12 4.85294117647059
13 4.94313725490196
14 4.96470588235294
15 5.05882352941176
16 5.08235294117647
17 5.1156862745098
18 5.15490196078431
19 5.15294117647059
20 5.21372549019608
21 5.26470588235294
22 5.27058823529412
23 5.2921568627451
24 5.29019607843137
25 5.33137254901961
26 5.34117647058823
27 5.37647058823529
28 5.38039215686275
29 5.4
30 5.4078431372549
31 5.42156862745098
32 5.43137254901961
33 5.42941176470588
34 5.44313725490196
35 5.43921568627451
36 5.44509803921569
37 5.43529411764706
38 5.43529411764706
39 5.44313725490196
40 5.45098039215686
41 5.45882352941176
42 5.45686274509804
43 5.46274509803922
44 5.45098039215686
45 5.47647058823529
46 5.48627450980392
47 5.49607843137255
48 5.48823529411765
49 5.50196078431373
50 5.51372549019608
};
\addplot [line width=2pt, colorHEBO]
table {%
5 4.16470588235294
6 4.46862745098039
7 4.58823529411765
8 4.56470588235294
9 4.64509803921569
10 4.67058823529412
11 4.57254901960784
12 4.40196078431373
13 4.26078431372549
14 4.25882352941176
15 4.18823529411765
16 4.1
17 4.09411764705882
18 4.05098039215686
19 4.04705882352941
20 3.97254901960784
21 3.86470588235294
22 3.81176470588235
23 3.75490196078431
24 3.73529411764706
25 3.7156862745098
26 3.67843137254902
27 3.66274509803922
28 3.62745098039216
29 3.63921568627451
30 3.56274509803922
31 3.5078431372549
32 3.50196078431373
33 3.50196078431373
34 3.45686274509804
35 3.44509803921569
36 3.45098039215686
37 3.43333333333333
38 3.44313725490196
39 3.43725490196078
40 3.41372549019608
41 3.40588235294118
42 3.37843137254902
43 3.37450980392157
44 3.36862745098039
45 3.33137254901961
46 3.2843137254902
47 3.27843137254902
48 3.27843137254902
49 3.27843137254902
50 3.26666666666667
};
\addplot [line width=2pt, colorOPT_BEI]
table {%
5 3.58823529411765
6 3.37843137254902
7 3.47450980392157
8 3.51176470588235
9 3.57254901960784
10 3.62156862745098
11 3.7156862745098
12 3.78039215686274
13 3.8
14 3.83333333333333
15 3.86470588235294
16 3.90980392156863
17 3.92745098039216
18 3.93725490196078
19 3.95882352941176
20 4.00588235294118
21 4.05882352941176
22 4.09411764705882
23 4.14509803921569
24 4.18823529411765
25 4.19607843137255
26 4.24509803921569
27 4.25686274509804
28 4.27450980392157
29 4.27843137254902
30 4.31764705882353
31 4.33921568627451
32 4.34509803921569
33 4.37450980392157
34 4.39019607843137
35 4.40392156862745
36 4.43725490196078
37 4.47058823529412
38 4.4843137254902
39 4.47647058823529
40 4.50392156862745
41 4.52156862745098
42 4.52549019607843
43 4.52549019607843
44 4.56274509803922
45 4.58823529411765
46 4.61176470588235
47 4.6
48 4.60980392156863
49 4.62549019607843
50 4.63529411764706
};
\addplot [line width=2pt, colorOPT_REI]
table {%
5 3.58823529411765
6 3.31176470588235
7 3.42352941176471
8 3.44901960784314
9 3.56862745098039
10 3.58235294117647
11 3.58235294117647
12 3.66274509803922
13 3.67843137254902
14 3.71960784313726
15 3.73333333333333
16 3.77450980392157
17 3.75294117647059
18 3.80392156862745
19 3.82549019607843
20 3.80980392156863
21 3.83725490196078
22 3.87647058823529
23 3.89803921568627
24 3.91176470588235
25 3.93137254901961
26 3.97647058823529
27 4.01764705882353
28 4.01176470588235
29 4.05294117647059
30 4.08039215686274
31 4.07843137254902
32 4.1156862745098
33 4.11372549019608
34 4.12549019607843
35 4.13137254901961
36 4.16470588235294
37 4.18039215686275
38 4.19803921568627
39 4.21372549019608
40 4.15882352941177
41 4.16078431372549
42 4.14705882352941
43 4.14117647058824
44 4.13529411764706
45 4.1843137254902
46 4.17058823529412
47 4.17843137254902
48 4.17058823529412
49 4.16666666666667
50 4.15294117647059
};
\addplot [line width=2pt, colorBOHAMIANN]
table {%
5 4.16470588235294
6 4.20980392156863
7 4.13529411764706
8 4.0843137254902
9 4.1078431372549
10 4.11372549019608
11 4.03333333333333
12 4.08039215686274
13 4.07058823529412
14 4.01960784313725
15 4
16 4.01764705882353
17 4.04117647058824
18 4.0156862745098
19 4.01764705882353
20 4.0156862745098
21 4.02352941176471
22 4.05490196078431
23 4.02549019607843
24 4.01764705882353
25 4.04705882352941
26 4.0156862745098
27 4.00196078431373
28 4.01960784313725
29 3.9921568627451
30 4.01176470588235
31 4.02745098039216
32 4.01176470588235
33 4.00588235294118
34 4.00392156862745
35 4.00588235294118
36 3.99803921568627
37 4.00588235294118
38 4.0156862745098
39 4.00196078431373
40 4.05490196078431
41 4.07254901960784
42 4.09411764705882
43 4.08823529411765
44 4.08039215686274
45 4.07647058823529
46 4.08039215686274
47 4.09803921568627
48 4.10196078431373
49 4.04313725490196
50 4.02745098039216
};
\addplot [line width=2pt, colorPFN_GP]
table {%
5 4.16470588235294
6 4.02941176470588
7 3.91176470588235
8 3.96078431372549
9 3.81764705882353
10 3.79411764705882
11 3.75686274509804
12 3.70980392156863
13 3.70980392156863
14 3.6921568627451
15 3.66274509803922
16 3.68039215686274
17 3.65294117647059
18 3.62156862745098
19 3.60980392156863
20 3.5921568627451
21 3.54313725490196
22 3.47254901960784
23 3.44705882352941
24 3.48627450980392
25 3.42156862745098
26 3.39803921568627
27 3.3921568627451
28 3.38823529411765
29 3.32745098039216
30 3.30588235294118
31 3.2921568627451
32 3.27254901960784
33 3.24117647058824
34 3.25294117647059
35 3.24901960784314
36 3.22549019607843
37 3.19803921568627
38 3.14509803921569
39 3.16274509803922
40 3.12156862745098
41 3.10980392156863
42 3.1156862745098
43 3.13137254901961
44 3.12745098039216
45 3.0843137254902
46 3.1156862745098
47 3.1156862745098
48 3.12745098039216
49 3.14509803921569
50 3.15490196078431
};
\addplot [line width=2pt, colorPFN_BNN]
table {%
5 4.16470588235294
6 4.03921568627451
7 3.81372549019608
8 3.77058823529412
9 3.63333333333333
10 3.5156862745098
11 3.49607843137255
12 3.51176470588235
13 3.53725490196078
14 3.51176470588235
15 3.4921568627451
16 3.43529411764706
17 3.4156862745098
18 3.4156862745098
19 3.38823529411765
20 3.39019607843137
21 3.4078431372549
22 3.41960784313725
23 3.43725490196078
24 3.37058823529412
25 3.35686274509804
26 3.34509803921569
27 3.2921568627451
28 3.29803921568627
29 3.30980392156863
30 3.31372549019608
31 3.33333333333333
32 3.32156862745098
33 3.33333333333333
34 3.32745098039216
35 3.32549019607843
36 3.27843137254902
37 3.27647058823529
38 3.27843137254902
39 3.26470588235294
40 3.29607843137255
41 3.27058823529412
42 3.28235294117647
43 3.27647058823529
44 3.27450980392157
45 3.25882352941176
46 3.25098039215686
47 3.23333333333333
48 3.22352941176471
49 3.23921568627451
50 3.24901960784314
};
\legend{};
\end{axis}

\end{tikzpicture}

%% file: icml2023/figures/hpob_diffbaselines_regret_test.tex
\begin{tikzpicture}

\begin{axis}[
legend cell align={left},
legend style={
  fill opacity=0.8,
  draw opacity=1,
  text opacity=1,
  at={(0.55,-0.15)},
  anchor=south,
  draw=white!80!black
},
log basis y={10},
tick align=outside,
tick pos=left,
height=.6\textwidth,
width=.8\textwidth,
x grid style={white!69.0196078431373!black},
xlabel={Number of trials},
xmin=2.75, xmax=52.25,
xtick style={color=black},
y grid style={white!69.0196078431373!black},
ylabel={Average Regret},
ymin=0.0153171731609242, ymax=0.261636933606157,
ymode=log,
ytick style={color=black}
]
\path [fill=colorRandom, fill opacity=0.2]
(axis cs:5,0.229972229042153)
--(axis cs:5,0.21710238749041)
--(axis cs:6,0.210551931890934)
--(axis cs:7,0.191336856479292)
--(axis cs:8,0.18398611450507)
--(axis cs:9,0.173174426009038)
--(axis cs:10,0.161566672085494)
--(axis cs:11,0.153221170289856)
--(axis cs:12,0.147103225495095)
--(axis cs:13,0.144964010471113)
--(axis cs:14,0.133513953193397)
--(axis cs:15,0.130869269073569)
--(axis cs:16,0.129588704326338)
--(axis cs:17,0.129056064670601)
--(axis cs:18,0.128240983502576)
--(axis cs:19,0.119655546102882)
--(axis cs:20,0.119261341744938)
--(axis cs:21,0.118952916746962)
--(axis cs:22,0.115740984667637)
--(axis cs:23,0.113822815372513)
--(axis cs:24,0.111239996489407)
--(axis cs:25,0.111109586652343)
--(axis cs:26,0.110677655794738)
--(axis cs:27,0.110054406629553)
--(axis cs:28,0.10869636607477)
--(axis cs:29,0.105425670492834)
--(axis cs:30,0.104148807039482)
--(axis cs:31,0.103036857102853)
--(axis cs:32,0.102326505972465)
--(axis cs:33,0.101512284262623)
--(axis cs:34,0.101273140087157)
--(axis cs:35,0.101095535190606)
--(axis cs:36,0.100374825444411)
--(axis cs:37,0.0987592666410219)
--(axis cs:38,0.0985667843139294)
--(axis cs:39,0.0983883405881294)
--(axis cs:40,0.0982740935933329)
--(axis cs:41,0.097971176092753)
--(axis cs:42,0.0978147028919637)
--(axis cs:43,0.0970845745081627)
--(axis cs:44,0.0967919127817516)
--(axis cs:45,0.0967251933978113)
--(axis cs:46,0.0964900529287663)
--(axis cs:47,0.0948039725845666)
--(axis cs:48,0.0914377093499991)
--(axis cs:49,0.0913116376456255)
--(axis cs:50,0.0909194074778135)
--(axis cs:50,0.102760161966289)
--(axis cs:50,0.102760161966289)
--(axis cs:49,0.10355391050267)
--(axis cs:48,0.103618894177447)
--(axis cs:47,0.105118030426644)
--(axis cs:46,0.105890409431554)
--(axis cs:45,0.106714359948635)
--(axis cs:44,0.106830096253732)
--(axis cs:43,0.107401874324678)
--(axis cs:42,0.109019289222573)
--(axis cs:41,0.109232186712493)
--(axis cs:40,0.110249516016526)
--(axis cs:39,0.110511645921156)
--(axis cs:38,0.110593493901512)
--(axis cs:37,0.110726131211684)
--(axis cs:36,0.112007148271516)
--(axis cs:35,0.112514679946367)
--(axis cs:34,0.112607841546255)
--(axis cs:33,0.112700097536489)
--(axis cs:32,0.113557523551764)
--(axis cs:31,0.115252610748441)
--(axis cs:30,0.115924337046734)
--(axis cs:29,0.116760957112317)
--(axis cs:28,0.118213697427976)
--(axis cs:27,0.120825416295935)
--(axis cs:26,0.121896892724957)
--(axis cs:25,0.122429135447138)
--(axis cs:24,0.122801304845827)
--(axis cs:23,0.124100798337417)
--(axis cs:22,0.129089778757046)
--(axis cs:21,0.133831452125081)
--(axis cs:20,0.133983446483229)
--(axis cs:19,0.134736088401152)
--(axis cs:18,0.139933467453242)
--(axis cs:17,0.14085611822347)
--(axis cs:16,0.141347747166577)
--(axis cs:15,0.142075058190751)
--(axis cs:14,0.145751665683148)
--(axis cs:13,0.152858007901119)
--(axis cs:12,0.15497155554015)
--(axis cs:11,0.164415684662798)
--(axis cs:10,0.170072803621721)
--(axis cs:9,0.180017514871002)
--(axis cs:8,0.186814074102229)
--(axis cs:7,0.196881831709089)
--(axis cs:6,0.219660583406081)
--(axis cs:5,0.229972229042153)
--cycle;

\path [fill=colorHEBO, fill opacity=0.2]
(axis cs:5,0.229972229042153)
--(axis cs:5,0.21710238749041)
--(axis cs:6,0.206767140505162)
--(axis cs:7,0.200269236542652)
--(axis cs:8,0.191893881430571)
--(axis cs:9,0.178401295834877)
--(axis cs:10,0.172228709564284)
--(axis cs:11,0.160242156579528)
--(axis cs:12,0.148288228185233)
--(axis cs:13,0.13706877128598)
--(axis cs:14,0.131755403377912)
--(axis cs:15,0.124634753528544)
--(axis cs:16,0.122314300804427)
--(axis cs:17,0.117386177238596)
--(axis cs:18,0.109443737179489)
--(axis cs:19,0.107964035869148)
--(axis cs:20,0.101478655615687)
--(axis cs:21,0.0894026065920785)
--(axis cs:22,0.0834904358748244)
--(axis cs:23,0.0810612376894862)
--(axis cs:24,0.0792019900308486)
--(axis cs:25,0.077272045739271)
--(axis cs:26,0.0743406880548449)
--(axis cs:27,0.0723679242034906)
--(axis cs:28,0.0692501331132998)
--(axis cs:29,0.0679576455343032)
--(axis cs:30,0.0642696604071299)
--(axis cs:31,0.0619732061149486)
--(axis cs:32,0.0588216188244668)
--(axis cs:33,0.0575674376163326)
--(axis cs:34,0.0543787145588783)
--(axis cs:35,0.0542253016853686)
--(axis cs:36,0.052423616783675)
--(axis cs:37,0.0507972753430667)
--(axis cs:38,0.0447286066464856)
--(axis cs:39,0.0427350322893294)
--(axis cs:40,0.0396577344033624)
--(axis cs:41,0.0388375831557457)
--(axis cs:42,0.0353897302858142)
--(axis cs:43,0.0352365203911538)
--(axis cs:44,0.0342058190248711)
--(axis cs:45,0.0333112179082447)
--(axis cs:46,0.0297218006534051)
--(axis cs:47,0.02925041183829)
--(axis cs:48,0.0289366390239416)
--(axis cs:49,0.0287281894276406)
--(axis cs:50,0.027704668884422)
--(axis cs:50,0.0379314778610262)
--(axis cs:50,0.0379314778610262)
--(axis cs:49,0.0385168128715617)
--(axis cs:48,0.0393725565609319)
--(axis cs:47,0.0395320769678211)
--(axis cs:46,0.039810107941024)
--(axis cs:45,0.0419644805299724)
--(axis cs:44,0.0441549254466072)
--(axis cs:43,0.047486456652137)
--(axis cs:42,0.0489356418855603)
--(axis cs:41,0.0519426310609125)
--(axis cs:40,0.0524295108410772)
--(axis cs:39,0.0560439699596165)
--(axis cs:38,0.0601576956040919)
--(axis cs:37,0.0653047631088551)
--(axis cs:36,0.0665071595620215)
--(axis cs:35,0.0679698668069191)
--(axis cs:34,0.0681561666136415)
--(axis cs:33,0.0698212639436179)
--(axis cs:32,0.0711194692498258)
--(axis cs:31,0.0735111490708296)
--(axis cs:30,0.0756630863393933)
--(axis cs:29,0.0796081418458316)
--(axis cs:28,0.0812864229024938)
--(axis cs:27,0.0834186826816302)
--(axis cs:26,0.0858884492261974)
--(axis cs:25,0.0903980956630372)
--(axis cs:24,0.0918888972060752)
--(axis cs:23,0.0939932055801463)
--(axis cs:22,0.0972281217787287)
--(axis cs:21,0.101010725685203)
--(axis cs:20,0.110119388570072)
--(axis cs:19,0.115691538281368)
--(axis cs:18,0.117482158326457)
--(axis cs:17,0.120996473127716)
--(axis cs:16,0.126426727255947)
--(axis cs:15,0.128761250870552)
--(axis cs:14,0.135914421443853)
--(axis cs:13,0.142248332662099)
--(axis cs:12,0.152950954979235)
--(axis cs:11,0.166301876686783)
--(axis cs:10,0.175840787690108)
--(axis cs:9,0.183015318333832)
--(axis cs:8,0.198153077650713)
--(axis cs:7,0.205652428193418)
--(axis cs:6,0.211559225297995)
--(axis cs:5,0.229972229042153)
--cycle;

\path [fill=colorOPT_BEI, fill opacity=0.2]
(axis cs:5,0.189439949135977)
--(axis cs:5,0.189439949135977)
--(axis cs:6,0.11391716511058)
--(axis cs:7,0.0958814307263215)
--(axis cs:8,0.0885134608338407)
--(axis cs:9,0.0828140217675857)
--(axis cs:10,0.0803804895098994)
--(axis cs:11,0.0783368703321028)
--(axis cs:12,0.076127434048452)
--(axis cs:13,0.0722266472152937)
--(axis cs:14,0.0713345989216402)
--(axis cs:15,0.0682342859988593)
--(axis cs:16,0.066638856091403)
--(axis cs:17,0.0649541353536289)
--(axis cs:18,0.0631102467311284)
--(axis cs:19,0.0622115104499682)
--(axis cs:20,0.0613440726736537)
--(axis cs:21,0.0606037530175092)
--(axis cs:22,0.0601883950111331)
--(axis cs:23,0.0588002159427384)
--(axis cs:24,0.0581139487982263)
--(axis cs:25,0.0575265301345649)
--(axis cs:26,0.0563355120891425)
--(axis cs:27,0.0555375632380128)
--(axis cs:28,0.0551545039350805)
--(axis cs:29,0.0546045227860134)
--(axis cs:30,0.0544181208133627)
--(axis cs:31,0.0541968282515756)
--(axis cs:32,0.0538097508709512)
--(axis cs:33,0.0532441676422011)
--(axis cs:34,0.0527334226542564)
--(axis cs:35,0.052130769418789)
--(axis cs:36,0.0513347359113986)
--(axis cs:37,0.0503654841279909)
--(axis cs:38,0.0498883509210866)
--(axis cs:39,0.0494581080060895)
--(axis cs:40,0.0492524418808456)
--(axis cs:41,0.0488738462613362)
--(axis cs:42,0.0483461769534117)
--(axis cs:43,0.0474526466457785)
--(axis cs:44,0.0471691629787914)
--(axis cs:45,0.0459675373238928)
--(axis cs:46,0.044419719871272)
--(axis cs:47,0.0440478881441445)
--(axis cs:48,0.043876655412969)
--(axis cs:49,0.043786236600278)
--(axis cs:50,0.0435362816871891)
--(axis cs:50,0.0488178990992764)
--(axis cs:50,0.0488178990992764)
--(axis cs:49,0.0490193397711489)
--(axis cs:48,0.0490898618889277)
--(axis cs:47,0.0493401768667377)
--(axis cs:46,0.0495892640573568)
--(axis cs:45,0.051085578490267)
--(axis cs:44,0.0524108994548425)
--(axis cs:43,0.052686480243535)
--(axis cs:42,0.0533811820045455)
--(axis cs:41,0.0539454849864869)
--(axis cs:40,0.0541798200979281)
--(axis cs:39,0.0544865790270586)
--(axis cs:38,0.0549124061852839)
--(axis cs:37,0.0553322118534851)
--(axis cs:36,0.0559086854968861)
--(axis cs:35,0.0567158637344447)
--(axis cs:34,0.0571733032318605)
--(axis cs:33,0.0578379200175894)
--(axis cs:32,0.0582334934883091)
--(axis cs:31,0.0589645318479317)
--(axis cs:30,0.0591990768471818)
--(axis cs:29,0.0594379950697145)
--(axis cs:28,0.059840768622284)
--(axis cs:27,0.0600728351978269)
--(axis cs:26,0.0607174811021406)
--(axis cs:25,0.0613035152813663)
--(axis cs:24,0.0619807459547081)
--(axis cs:23,0.0632241677842049)
--(axis cs:22,0.0638786891142101)
--(axis cs:21,0.0643628271540229)
--(axis cs:20,0.0655043776317339)
--(axis cs:19,0.066148374680021)
--(axis cs:18,0.0672324150936229)
--(axis cs:17,0.0683778743462828)
--(axis cs:16,0.0694713599082017)
--(axis cs:15,0.0713175587125908)
--(axis cs:14,0.0739570534848816)
--(axis cs:13,0.0752394708926404)
--(axis cs:12,0.078996043282903)
--(axis cs:11,0.0812592763791696)
--(axis cs:10,0.0831673898639383)
--(axis cs:9,0.0854890036551166)
--(axis cs:8,0.0917397280673817)
--(axis cs:7,0.10022661192539)
--(axis cs:6,0.11391716511058)
--(axis cs:5,0.189439949135977)
--cycle;

\path [fill=colorOPT_REI, fill opacity=0.2]
(axis cs:5,0.189439949135977)
--(axis cs:5,0.189439949135977)
--(axis cs:6,0.10662630444577)
--(axis cs:7,0.0973468681118798)
--(axis cs:8,0.0902238167304435)
--(axis cs:9,0.0841129369333837)
--(axis cs:10,0.075566982552995)
--(axis cs:11,0.0680874714149858)
--(axis cs:12,0.0664492353052708)
--(axis cs:13,0.0631484759180334)
--(axis cs:14,0.0605495163438003)
--(axis cs:15,0.059488847053429)
--(axis cs:16,0.057592896361635)
--(axis cs:17,0.0546092249155445)
--(axis cs:18,0.0528207846785914)
--(axis cs:19,0.0511387746473232)
--(axis cs:20,0.0499794657639762)
--(axis cs:21,0.048855922439964)
--(axis cs:22,0.0487023559924626)
--(axis cs:23,0.0480918712892826)
--(axis cs:24,0.0471215062717954)
--(axis cs:25,0.0461134824234532)
--(axis cs:26,0.045931350356749)
--(axis cs:27,0.044507644566413)
--(axis cs:28,0.0440044955489367)
--(axis cs:29,0.0438094305902367)
--(axis cs:30,0.0427384853692034)
--(axis cs:31,0.0421553075050688)
--(axis cs:32,0.0419025424770654)
--(axis cs:33,0.0411762209114611)
--(axis cs:34,0.0407343345208621)
--(axis cs:35,0.039604639081469)
--(axis cs:36,0.0393646089172242)
--(axis cs:37,0.0391183545036277)
--(axis cs:38,0.0381194232917359)
--(axis cs:39,0.0380233522570014)
--(axis cs:40,0.0367655591976589)
--(axis cs:41,0.0363361203322567)
--(axis cs:42,0.0359705881317438)
--(axis cs:43,0.0358414067364278)
--(axis cs:44,0.0357051680048242)
--(axis cs:45,0.0356097480432182)
--(axis cs:46,0.0351366985293083)
--(axis cs:47,0.0348793273789554)
--(axis cs:48,0.0343735073711842)
--(axis cs:49,0.034132751486408)
--(axis cs:50,0.0329524287081989)
--(axis cs:50,0.0360124267735108)
--(axis cs:50,0.0360124267735108)
--(axis cs:49,0.0364786599770962)
--(axis cs:48,0.0366308045514828)
--(axis cs:47,0.0378049816193878)
--(axis cs:46,0.0380601365889418)
--(axis cs:45,0.0383107862587264)
--(axis cs:44,0.0383805601101324)
--(axis cs:43,0.0384823717035323)
--(axis cs:42,0.0385712395140671)
--(axis cs:41,0.0388802383501453)
--(axis cs:40,0.0392742312527116)
--(axis cs:39,0.0401658957407394)
--(axis cs:38,0.040316271056413)
--(axis cs:37,0.0406935060567429)
--(axis cs:36,0.0409151161199053)
--(axis cs:35,0.0411138056389144)
--(axis cs:34,0.0421915192768972)
--(axis cs:33,0.0425956844859081)
--(axis cs:32,0.0445512004314925)
--(axis cs:31,0.0447519299237112)
--(axis cs:30,0.0451885932096386)
--(axis cs:29,0.0461263861594604)
--(axis cs:28,0.0463185308781511)
--(axis cs:27,0.0478171776304085)
--(axis cs:26,0.0489657837699832)
--(axis cs:25,0.0490590250002563)
--(axis cs:24,0.0499121264279578)
--(axis cs:23,0.0513880430483191)
--(axis cs:22,0.0522375861398504)
--(axis cs:21,0.0524796156734707)
--(axis cs:20,0.0542220635440508)
--(axis cs:19,0.0555162760099655)
--(axis cs:18,0.0564388758748901)
--(axis cs:17,0.0582316674189874)
--(axis cs:16,0.0601994830251131)
--(axis cs:15,0.0619832750276972)
--(axis cs:14,0.0642718506946292)
--(axis cs:13,0.0661980214539539)
--(axis cs:12,0.0688549983782885)
--(axis cs:11,0.0715357780406711)
--(axis cs:10,0.0802680078084322)
--(axis cs:9,0.0886879865382148)
--(axis cs:8,0.09420275983209)
--(axis cs:7,0.100458352009531)
--(axis cs:6,0.112200511723085)
--(axis cs:5,0.189439949135977)
--cycle;

\path [fill=colorBOHAMIANN, fill opacity=0.2]
(axis cs:5,0.229972229042153)
--(axis cs:5,0.21710238749041)
--(axis cs:6,0.177687276520144)
--(axis cs:7,0.153565548454098)
--(axis cs:8,0.137438036391217)
--(axis cs:9,0.118177344353962)
--(axis cs:10,0.111167352270213)
--(axis cs:11,0.0947020035036127)
--(axis cs:12,0.0872941799053517)
--(axis cs:13,0.0791116724847716)
--(axis cs:14,0.0697527250607514)
--(axis cs:15,0.062482156988987)
--(axis cs:16,0.0602296832319325)
--(axis cs:17,0.0591160718877381)
--(axis cs:18,0.0556297650861536)
--(axis cs:19,0.0532121290692355)
--(axis cs:20,0.0511390178111483)
--(axis cs:21,0.0484319901174184)
--(axis cs:22,0.0472373392477794)
--(axis cs:23,0.0445094255412033)
--(axis cs:24,0.0426612310261246)
--(axis cs:25,0.0426004356193128)
--(axis cs:26,0.0402307007324025)
--(axis cs:27,0.0398321354667507)
--(axis cs:28,0.0389842304794989)
--(axis cs:29,0.0371957782693215)
--(axis cs:30,0.0370255889785422)
--(axis cs:31,0.0355774641051242)
--(axis cs:32,0.0339227528718047)
--(axis cs:33,0.0337545868010225)
--(axis cs:34,0.0331952912374969)
--(axis cs:35,0.0318458960485431)
--(axis cs:36,0.0314090842215674)
--(axis cs:37,0.0306010945096677)
--(axis cs:38,0.0296228586983223)
--(axis cs:39,0.0286116849605075)
--(axis cs:40,0.0268770114151158)
--(axis cs:41,0.0268023561677497)
--(axis cs:42,0.0267268406593893)
--(axis cs:43,0.0258659218556026)
--(axis cs:44,0.0251738608911883)
--(axis cs:45,0.0247583388787159)
--(axis cs:46,0.0241128866954101)
--(axis cs:47,0.0241082316471053)
--(axis cs:48,0.0240521888773482)
--(axis cs:49,0.0237310032509895)
--(axis cs:50,0.0226599201112285)
--(axis cs:50,0.0252937859820833)
--(axis cs:50,0.0252937859820833)
--(axis cs:49,0.0262407636310776)
--(axis cs:48,0.0265627697502868)
--(axis cs:47,0.0265858197725222)
--(axis cs:46,0.0265893088998177)
--(axis cs:45,0.0275883957822513)
--(axis cs:44,0.0280629810714331)
--(axis cs:43,0.0287584400082819)
--(axis cs:42,0.0291252283904728)
--(axis cs:41,0.0292205258147758)
--(axis cs:40,0.0293653614499084)
--(axis cs:39,0.0300606140076752)
--(axis cs:38,0.0312218947119284)
--(axis cs:37,0.0325785189826713)
--(axis cs:36,0.0348491115952494)
--(axis cs:35,0.0364033668285655)
--(axis cs:34,0.037220926525751)
--(axis cs:33,0.037576816600039)
--(axis cs:32,0.0380813478086645)
--(axis cs:31,0.0399907001879624)
--(axis cs:30,0.042632047070722)
--(axis cs:29,0.0431290517073818)
--(axis cs:28,0.0439888235627246)
--(axis cs:27,0.044521639276184)
--(axis cs:26,0.0451497830800701)
--(axis cs:25,0.0463055107439702)
--(axis cs:24,0.0464111283257172)
--(axis cs:23,0.0475582792462603)
--(axis cs:22,0.0512683734307935)
--(axis cs:21,0.0526998597535191)
--(axis cs:20,0.0558899029482968)
--(axis cs:19,0.0587053111912838)
--(axis cs:18,0.0601903737019777)
--(axis cs:17,0.0659712824498536)
--(axis cs:16,0.0673634626586001)
--(axis cs:15,0.0713694407939708)
--(axis cs:14,0.0807637352977721)
--(axis cs:13,0.0873279730980333)
--(axis cs:12,0.097687596535291)
--(axis cs:11,0.107152180370112)
--(axis cs:10,0.121952669068145)
--(axis cs:9,0.136584974898696)
--(axis cs:8,0.154260474997262)
--(axis cs:7,0.171361650307632)
--(axis cs:6,0.191082038624974)
--(axis cs:5,0.229972229042153)
--cycle;

\path [fill=colorPFN_GP, fill opacity=0.2]
(axis cs:5,0.229972229042153)
--(axis cs:5,0.21710238749041)
--(axis cs:6,0.160278831014592)
--(axis cs:7,0.130782163847823)
--(axis cs:8,0.116476611076302)
--(axis cs:9,0.0964332893365131)
--(axis cs:10,0.0909970940271918)
--(axis cs:11,0.0848937200910018)
--(axis cs:12,0.0799838492545874)
--(axis cs:13,0.0753719536792764)
--(axis cs:14,0.0725192375026625)
--(axis cs:15,0.0662305828369612)
--(axis cs:16,0.0631898957810735)
--(axis cs:17,0.0597904014311965)
--(axis cs:18,0.0566694875673723)
--(axis cs:19,0.0549615438692448)
--(axis cs:20,0.0524587204684884)
--(axis cs:21,0.0502854282954943)
--(axis cs:22,0.0485177375759072)
--(axis cs:23,0.0454456648919411)
--(axis cs:24,0.0444372309565432)
--(axis cs:25,0.0407273515835956)
--(axis cs:26,0.0384756222783664)
--(axis cs:27,0.0367077372183817)
--(axis cs:28,0.0357680803128359)
--(axis cs:29,0.0338661626496755)
--(axis cs:30,0.0328506051745369)
--(axis cs:31,0.0317215049957481)
--(axis cs:32,0.0288762770262921)
--(axis cs:33,0.0282302502599518)
--(axis cs:34,0.0276126265046207)
--(axis cs:35,0.027171861652372)
--(axis cs:36,0.0259207942838157)
--(axis cs:37,0.0244240034290502)
--(axis cs:38,0.0226805049232498)
--(axis cs:39,0.0225551593726975)
--(axis cs:40,0.0216690437706978)
--(axis cs:41,0.0211328043231009)
--(axis cs:42,0.020691153889118)
--(axis cs:43,0.0195500586840891)
--(axis cs:44,0.019347499671286)
--(axis cs:45,0.0182153763867652)
--(axis cs:46,0.0182153763867652)
--(axis cs:47,0.0181595108243169)
--(axis cs:48,0.0180503633167558)
--(axis cs:49,0.0176131802485801)
--(axis cs:50,0.0174261833006113)
--(axis cs:50,0.0190491751478515)
--(axis cs:50,0.0190491751478515)
--(axis cs:49,0.0192911309590044)
--(axis cs:48,0.0197975457074311)
--(axis cs:47,0.019864637828914)
--(axis cs:46,0.020095003759592)
--(axis cs:45,0.020095003759592)
--(axis cs:44,0.0211877494332459)
--(axis cs:43,0.0214391028494531)
--(axis cs:42,0.0255267584376508)
--(axis cs:41,0.0268282171956197)
--(axis cs:40,0.0272343979720654)
--(axis cs:39,0.0277372755738238)
--(axis cs:38,0.027855769914311)
--(axis cs:37,0.0294444760602151)
--(axis cs:36,0.0305886549822034)
--(axis cs:35,0.0320135843531685)
--(axis cs:34,0.0322725951091343)
--(axis cs:33,0.0326056884009755)
--(axis cs:32,0.0335400814715294)
--(axis cs:31,0.0358461008734134)
--(axis cs:30,0.0368527717228782)
--(axis cs:29,0.0374051929806618)
--(axis cs:28,0.0393245500594882)
--(axis cs:27,0.0404841037423493)
--(axis cs:26,0.0439548761617103)
--(axis cs:25,0.0456213709484763)
--(axis cs:24,0.0480627903424171)
--(axis cs:23,0.0492098431367699)
--(axis cs:22,0.054460064143617)
--(axis cs:21,0.0567772872821245)
--(axis cs:20,0.0588415103598554)
--(axis cs:19,0.0613160108730878)
--(axis cs:18,0.0625221929845047)
--(axis cs:17,0.0652598812884552)
--(axis cs:16,0.0694508371730795)
--(axis cs:15,0.0740021478936112)
--(axis cs:14,0.0787037147583354)
--(axis cs:13,0.0841041113492363)
--(axis cs:12,0.0885640030903338)
--(axis cs:11,0.0949998034039603)
--(axis cs:10,0.100816834998825)
--(axis cs:9,0.108115914008586)
--(axis cs:8,0.133084280281661)
--(axis cs:7,0.145019877547662)
--(axis cs:6,0.175050770556326)
--(axis cs:5,0.229972229042153)
--cycle;

\path [fill=colorPFN_BNN, fill opacity=0.2]
(axis cs:5,0.229972229042153)
--(axis cs:5,0.21710238749041)
--(axis cs:6,0.183594967492045)
--(axis cs:7,0.144762447909839)
--(axis cs:8,0.133849891657266)
--(axis cs:9,0.11462367856701)
--(axis cs:10,0.100715892464588)
--(axis cs:11,0.0926872358859926)
--(axis cs:12,0.0889921381211661)
--(axis cs:13,0.0861495072458914)
--(axis cs:14,0.0793350105495573)
--(axis cs:15,0.0742156476452009)
--(axis cs:16,0.0661353879138803)
--(axis cs:17,0.0616343097238552)
--(axis cs:18,0.0579970783460914)
--(axis cs:19,0.0564389749899866)
--(axis cs:20,0.0532713786660544)
--(axis cs:21,0.051352854979348)
--(axis cs:22,0.0502335523502818)
--(axis cs:23,0.0480985102273837)
--(axis cs:24,0.0447505973773237)
--(axis cs:25,0.0440795334630975)
--(axis cs:26,0.0424967914435655)
--(axis cs:27,0.0402867393694132)
--(axis cs:28,0.03753282371629)
--(axis cs:29,0.0365076825310318)
--(axis cs:30,0.0360734312457983)
--(axis cs:31,0.0336938069260269)
--(axis cs:32,0.0327469639836636)
--(axis cs:33,0.0316070492430778)
--(axis cs:34,0.0298969099962349)
--(axis cs:35,0.0294530965083396)
--(axis cs:36,0.0282636705310643)
--(axis cs:37,0.0279108359802767)
--(axis cs:38,0.0276635479891179)
--(axis cs:39,0.0273891226297093)
--(axis cs:40,0.0271182534142053)
--(axis cs:41,0.0266494209042222)
--(axis cs:42,0.0260714939404244)
--(axis cs:43,0.0254125569830256)
--(axis cs:44,0.0252551900182939)
--(axis cs:45,0.0242681928573477)
--(axis cs:46,0.0240153468174058)
--(axis cs:47,0.0237422661506803)
--(axis cs:48,0.0234220815226007)
--(axis cs:49,0.0231110108017196)
--(axis cs:50,0.0226515868825414)
--(axis cs:50,0.0287594111470262)
--(axis cs:50,0.0287594111470262)
--(axis cs:49,0.0290843987232345)
--(axis cs:48,0.0293239019762014)
--(axis cs:47,0.0294701117731199)
--(axis cs:46,0.0295670690004524)
--(axis cs:45,0.0297311055903109)
--(axis cs:44,0.0302775056948107)
--(axis cs:43,0.0303977180892438)
--(axis cs:42,0.030959132179896)
--(axis cs:41,0.0313764433924255)
--(axis cs:40,0.03188812808349)
--(axis cs:39,0.0322496083433775)
--(axis cs:38,0.0328556033655389)
--(axis cs:37,0.0330652085291479)
--(axis cs:36,0.0334618898824823)
--(axis cs:35,0.0345975069817636)
--(axis cs:34,0.0355643080299272)
--(axis cs:33,0.0370100657106305)
--(axis cs:32,0.0380777137789999)
--(axis cs:31,0.0396342344038521)
--(axis cs:30,0.0400867399381067)
--(axis cs:29,0.0408714224779201)
--(axis cs:28,0.0430712454033133)
--(axis cs:27,0.0466118090440831)
--(axis cs:26,0.0478926530683019)
--(axis cs:25,0.048669258382749)
--(axis cs:24,0.0494693973336476)
--(axis cs:23,0.0582870080837376)
--(axis cs:22,0.0612775534728455)
--(axis cs:21,0.062178627040015)
--(axis cs:20,0.0639110246020631)
--(axis cs:19,0.0690984449767098)
--(axis cs:18,0.0714178860155361)
--(axis cs:17,0.075396850561149)
--(axis cs:16,0.0785996259794174)
--(axis cs:15,0.0849003956145668)
--(axis cs:14,0.0884687144954101)
--(axis cs:13,0.0931491585352917)
--(axis cs:12,0.0968654627409627)
--(axis cs:11,0.10495921660506)
--(axis cs:10,0.114553056517067)
--(axis cs:9,0.127804744396634)
--(axis cs:8,0.15088390448601)
--(axis cs:7,0.164539810221328)
--(axis cs:6,0.201746124399505)
--(axis cs:5,0.229972229042153)
--cycle;

\addplot [line width=2pt, colorRandom]
table {%
5 0.223537308266282
6 0.215106257648508
7 0.19410934409419
8 0.185400094303649
9 0.17659597044002
10 0.165819737853608
11 0.158818427476327
12 0.151037390517623
13 0.148911009186116
14 0.139632809438272
15 0.13647216363216
16 0.135468225746457
17 0.134956091447036
18 0.134087225477909
19 0.127195817252017
20 0.126622394114084
21 0.126392184436021
22 0.122415381712342
23 0.118961806854965
24 0.117020650667617
25 0.11676936104974
26 0.116287274259848
27 0.115439911462744
28 0.113455031751373
29 0.111093313802576
30 0.110036572043108
31 0.109144733925647
32 0.107942014762115
33 0.107106190899556
34 0.106940490816706
35 0.106805107568487
36 0.106190986857964
37 0.104742698926353
38 0.104580139107721
39 0.104449993254643
40 0.104261804804929
41 0.103601681402623
42 0.103416996057268
43 0.10224322441642
44 0.101811004517742
45 0.101719776673223
46 0.10119023118016
47 0.0999610015056053
48 0.0975283017637233
49 0.0974327740741479
50 0.0968397847220513
};
\addplot [line width=2pt, colorHEBO]
table {%
5 0.223537308266282
6 0.209163182901579
7 0.202960832368035
8 0.195023479540642
9 0.180708307084355
10 0.174034748627196
11 0.163272016633155
12 0.150619591582234
13 0.139658551974039
14 0.133834912410882
15 0.126698002199548
16 0.124370514030187
17 0.119191325183156
18 0.113462947752973
19 0.111827787075258
20 0.105799022092879
21 0.0952066661386408
22 0.0903592788267765
23 0.0875272216348163
24 0.0855454436184619
25 0.0838350707011541
26 0.0801145686405212
27 0.0778933034425604
28 0.0752682780078968
29 0.0737828936900674
30 0.0699663733732616
31 0.0677421775928891
32 0.0649705440371463
33 0.0636943507799752
34 0.0612674405862599
35 0.0610975842461438
36 0.0594653881728482
37 0.0580510192259609
38 0.0524431511252887
39 0.049389501124473
40 0.0460436226222198
41 0.0453901071083291
42 0.0421626860856872
43 0.0413614885216454
44 0.0391803722357391
45 0.0376378492191086
46 0.0347659542972145
47 0.0343912444030556
48 0.0341545977924367
49 0.0336225011496011
50 0.0328180733727241
};
\addplot [line width=2pt, colorOPT_BEI]
table {%
5 0.189439949135977
6 0.11391716511058
7 0.0980540213258558
8 0.0901265944506112
9 0.0841515127113511
10 0.0817739396869189
11 0.0797980733556362
12 0.0775617386656775
13 0.0737330590539671
14 0.0726458262032609
15 0.069775922355725
16 0.0680551079998024
17 0.0666660048499558
18 0.0651713309123756
19 0.0641799425649946
20 0.0634242251526938
21 0.0624832900857661
22 0.0620335420626716
23 0.0610121918634716
24 0.0600473473764672
25 0.0594150227079656
26 0.0585264965956415
27 0.0578051992179198
28 0.0574976362786823
29 0.057021258927864
30 0.0568085988302722
31 0.0565806800497537
32 0.0560216221796302
33 0.0555410438298952
34 0.0549533629430585
35 0.0544233165766168
36 0.0536217107041424
37 0.052848847990738
38 0.0524003785531853
39 0.0519723435165741
40 0.0517161309893868
41 0.0514096656239115
42 0.0508636794789786
43 0.0500695634446567
44 0.049790031216817
45 0.0485265579070799
46 0.0470044919643144
47 0.0466940325054411
48 0.0464832586509483
49 0.0464027881857135
50 0.0461770903932327
};
\addplot [line width=2pt, colorOPT_REI]
table {%
5 0.189439949135977
6 0.109413408084427
7 0.0989026100607053
8 0.0922132882812668
9 0.0864004617357993
10 0.0779174951807136
11 0.0698116247278285
12 0.0676521168417796
13 0.0646732486859936
14 0.0624106835192147
15 0.0607360610405631
16 0.058896189693374
17 0.056420446167266
18 0.0546298302767408
19 0.0533275253286443
20 0.0521007646540135
21 0.0506677690567174
22 0.0504699710661565
23 0.0497399571688008
24 0.0485168163498766
25 0.0475862537118547
26 0.0474485670633661
27 0.0461624110984107
28 0.0451615132135439
29 0.0449679083748485
30 0.043963539289421
31 0.04345361871439
32 0.0432268714542789
33 0.0418859526986846
34 0.0414629268988797
35 0.0403592223601917
36 0.0401398625185648
37 0.0399059302801853
38 0.0392178471740744
39 0.0390946239988704
40 0.0380198952251852
41 0.037608179341201
42 0.0372709138229055
43 0.03716188921998
44 0.0370428640574783
45 0.0369602671509723
46 0.0365984175591251
47 0.0363421544991716
48 0.0355021559613335
49 0.0353057057317521
50 0.0344824277408548
};
\addplot [line width=2pt, colorBOHAMIANN]
table {%
5 0.223537308266282
6 0.184384657572559
7 0.162463599380865
8 0.14584925569424
9 0.127381159626329
10 0.116560010669179
11 0.100927091936862
12 0.0924908882203214
13 0.0832198227914024
14 0.0752582301792618
15 0.0669257988914789
16 0.0637965729452663
17 0.0625436771687959
18 0.0579100693940657
19 0.0559587201302596
20 0.0535144603797225
21 0.0505659249354687
22 0.0492528563392865
23 0.0460338523937318
24 0.0445361796759209
25 0.0444529731816415
26 0.0426902419062363
27 0.0421768873714674
28 0.0414865270211117
29 0.0401624149883517
30 0.0398288180246321
31 0.0377840821465433
32 0.0360020503402346
33 0.0356657017005308
34 0.035208108881624
35 0.0341246314385543
36 0.0331290979084084
37 0.0315898067461695
38 0.0304223767051254
39 0.0293361494840913
40 0.0281211864325121
41 0.0280114409912628
42 0.0279260345249311
43 0.0273121809319423
44 0.0266184209813107
45 0.0261733673304836
46 0.0253510977976139
47 0.0253470257098138
48 0.0253074793138175
49 0.0249858834410336
50 0.0239768530466559
};
\addplot [line width=2pt, colorPFN_GP]
table {%
5 0.223537308266282
6 0.167664800785459
7 0.137901020697743
8 0.124780445678982
9 0.10227460167255
10 0.0959069645130084
11 0.089946761747481
12 0.0842739261724606
13 0.0797380325142564
14 0.0756114761304989
15 0.0701163653652862
16 0.0663203664770765
17 0.0625251413598259
18 0.0595958402759385
19 0.0581387773711663
20 0.0556501154141719
21 0.0535313577888094
22 0.0514889008597621
23 0.0473277540143555
24 0.0462500106494802
25 0.0431743612660359
26 0.0412152492200384
27 0.0385959204803655
28 0.0375463151861621
29 0.0356356778151686
30 0.0348516884487076
31 0.0337838029345808
32 0.0312081792489108
33 0.0304179693304636
34 0.0299426108068775
35 0.0295927230027702
36 0.0282547246330095
37 0.0269342397446327
38 0.0252681374187804
39 0.0251462174732606
40 0.0244517208713816
41 0.0239805107593603
42 0.0231089561633844
43 0.0204945807667711
44 0.020267624552266
45 0.0191551900731786
46 0.0191551900731786
47 0.0190120743266155
48 0.0189239545120935
49 0.0184521556037923
50 0.0182376792242314
};
\addplot [line width=2pt, colorPFN_BNN]
table {%
5 0.223537308266282
6 0.192670545945775
7 0.154651129065583
8 0.142366898071638
9 0.121214211481822
10 0.107634474490827
11 0.0988232262455261
12 0.0929288004310644
13 0.0896493328905916
14 0.0839018625224837
15 0.0795580216298839
16 0.0723675069466488
17 0.0685155801425021
18 0.0647074821808137
19 0.0627687099833482
20 0.0585912016340588
21 0.0567657410096815
22 0.0557555529115636
23 0.0531927591555606
24 0.0471099973554856
25 0.0463743959229232
26 0.0451947222559337
27 0.0434492742067481
28 0.0403020345598016
29 0.038689552504476
30 0.0380800855919525
31 0.0366640206649395
32 0.0354123388813317
33 0.0343085574768542
34 0.0327306090130811
35 0.0320253017450516
36 0.0308627802067733
37 0.0304880222547123
38 0.0302595756773284
39 0.0298193654865434
40 0.0295031907488476
41 0.0290129321483239
42 0.0285153130601602
43 0.0279051375361347
44 0.0277663478565523
45 0.0269996492238293
46 0.0267912079089291
47 0.0266061889619001
48 0.026372991749401
49 0.0260977047624771
50 0.0257054990147838
};
\legend{};
\end{axis}

\end{tikzpicture}

%% file: MetaLearnWorkshop/figures/hebo_on_hebo_example_1.tex
 
\begin{tikzpicture}

\begin{axis}[
tick align=outside,
tick pos=left,
x grid style={white!69.0196078431373!black},
xmin=-0.0498972654342651, xmax=1.0497177362442,
xtick style={color=black},
y grid style={white!69.0196078431373!black},
ymin=0.0207632094621658, ymax=1.16667629182339,
ytick style={color=black}, width=\textwidth, height=\textwidth,
]
\addplot [semithick, white!50.1960784313725!black, forget plot]
table {%
8.52346420288086e-05 0.291133195161819
0.000220537185668945 0.190055876970291
0.00083845853805542 0.358708500862122
0.00273597240447998 0.231334820389748
0.00465720891952515 0.342697232961655
0.00606405735015869 0.243281468749046
0.00633800029754639 0.221985459327698
0.00695621967315674 0.225422620773315
0.00904583930969238 0.195103511214256
0.00968903303146362 0.0799800157546997
0.0126655697822571 0.341727048158646
0.0145461559295654 0.228800982236862
0.0147328972816467 0.196432158350945
0.0160070061683655 0.215339928865433
0.0165477991104126 0.314940094947815
0.0169711112976074 0.0981291979551315
0.017436146736145 0.163241147994995
0.0178589820861816 0.23307204246521
0.019224226474762 0.243650585412979
0.0193036198616028 0.172765001654625
0.0195304751396179 0.24303674697876
0.0200313329696655 0.177136361598969
0.0209006071090698 0.272572994232178
0.0209113359451294 0.267353594303131
0.021028459072113 0.083729475736618
0.0212217569351196 0.168169885873795
0.0227402448654175 0.275801718235016
0.0244792103767395 0.133950889110565
0.0258387923240662 0.201274007558823
0.0302267074584961 0.201298400759697
0.0311180353164673 0.274504482746124
0.0311998128890991 0.243612319231033
0.031236469745636 0.182980209589005
0.0346618890762329 0.131531849503517
0.0358260273933411 0.176485121250153
0.0363011360168457 0.153287023305893
0.0367897152900696 0.206430092453957
0.0391088128089905 0.24840684235096
0.0394134521484375 0.205404967069626
0.0401457548141479 0.310555458068848
0.0404354929924011 0.112533777952194
0.0420945882797241 0.148366823792458
0.0441054105758667 0.234818920493126
0.0504013895988464 0.271951705217361
0.0517642498016357 0.239942818880081
0.0518956780433655 0.260831713676453
0.0542507171630859 0.252398282289505
0.0562487840652466 0.284987062215805
0.0575753450393677 0.331315875053406
0.0580314993858337 0.239810973405838
0.0586116909980774 0.356246501207352
0.0591593980789185 0.185854062438011
0.0595703125 0.175635382533073
0.0596156716346741 0.244612783193588
0.0599780678749084 0.303186625242233
0.0607773661613464 0.28662446141243
0.0609022378921509 0.115519218146801
0.0619069933891296 0.241463989019394
0.0646771192550659 0.0873922258615494
0.0656799077987671 0.227952033281326
0.0668550133705139 0.21548043191433
0.0682503581047058 0.120884962379932
0.0696618556976318 0.202284902334213
0.0718380808830261 0.204328835010529
0.0720627307891846 0.22459988296032
0.0722064971923828 0.235124230384827
0.0728914737701416 0.190991625189781
0.0735968351364136 0.0728501677513123
0.0749292969703674 0.185408011078835
0.0750841498374939 0.262158155441284
0.0781050324440002 0.310679376125336
0.0794161558151245 0.259737223386765
0.0809480547904968 0.233866542577744
0.0839001536369324 0.23348605632782
0.0850949287414551 0.192785888910294
0.085742175579071 0.236104115843773
0.0903812646865845 0.210035234689713
0.0908154845237732 0.092038631439209
0.0917133688926697 0.285357803106308
0.0930115580558777 0.19328835606575
0.095221221446991 0.268175840377808
0.0975136160850525 0.284065842628479
0.0991360545158386 0.323743283748627
0.0996325612068176 0.289512395858765
0.100641429424286 0.133524268865585
0.101235389709473 0.250800997018814
0.101402640342712 0.267165750265121
0.101477980613708 0.155039370059967
0.103624582290649 0.288202941417694
0.103650450706482 0.379768967628479
0.105658233165741 0.180945545434952
0.106590628623962 0.285069733858109
0.10856419801712 0.205134004354477
0.109183371067047 0.334570586681366
0.10925018787384 0.258425861597061
0.109856188297272 0.266067147254944
0.110911428928375 0.322535991668701
0.111746490001678 0.20901420712471
0.112940549850464 0.242062196135521
0.112941026687622 0.227190732955933
0.113887965679169 0.16431836783886
0.113894164562225 0.194839850068092
0.115043222904205 0.168287202715874
0.115292072296143 0.220416337251663
0.11549037694931 0.344513982534409
0.11875581741333 0.328302502632141
0.120871305465698 0.192998051643372
0.121737241744995 0.273002833127975
0.123547732830048 0.215301275253296
0.124441802501678 0.277540981769562
0.12463116645813 0.236010774970055
0.124758362770081 0.303928405046463
0.126496195793152 0.23900830745697
0.126850187778473 0.319934844970703
0.127949297428131 0.282477647066116
0.127951264381409 0.305340170860291
0.129734933376312 0.324549555778503
0.130945801734924 0.282463610172272
0.132102489471436 0.312464267015457
0.132719337940216 0.360142052173615
0.132850527763367 0.306219756603241
0.133407592773438 0.243596583604813
0.1335089802742 0.287124544382095
0.133756995201111 0.345638543367386
0.134546101093292 0.24272595345974
0.135038614273071 0.270624935626984
0.139540016651154 0.261428952217102
0.139686346054077 0.240289837121964
0.141815543174744 0.323465913534164
0.141994774341583 0.209015905857086
0.142177820205688 0.236370176076889
0.143692076206207 0.397538691759109
0.144455075263977 0.219475477933884
0.144704639911652 0.332656979560852
0.145006895065308 0.272773921489716
0.145786583423615 0.329662382602692
0.146015048027039 0.366026878356934
0.14610767364502 0.422746449708939
0.147069454193115 0.121693730354309
0.147522389888763 0.262944221496582
0.148501217365265 0.288618743419647
0.149721145629883 0.290038019418716
0.14974045753479 0.176395028829575
0.149909198284149 0.256807804107666
0.152248382568359 0.290464848279953
0.154292047023773 0.239956870675087
0.155668616294861 0.125346973538399
0.15629118680954 0.307911247014999
0.156395256519318 0.318084508180618
0.156431615352631 0.321368604898453
0.15730893611908 0.312297314405441
0.158216655254364 0.288377523422241
0.158496677875519 0.187543854117393
0.159429848194122 0.332594245672226
0.160326182842255 0.267446398735046
0.16283243894577 0.335321605205536
0.168377220630646 0.351962983608246
0.168931365013123 0.327784866094589
0.16923850774765 0.396057486534119
0.169637978076935 0.41697832942009
0.172175705432892 0.41791707277298
0.172467768192291 0.328460991382599
0.175769805908203 0.33720126748085
0.17658919095993 0.467761754989624
0.176974475383759 0.303816437721252
0.177221238613129 0.441556602716446
0.178355038166046 0.320109844207764
0.178724348545074 0.3260817527771
0.179067671298981 0.238560348749161
0.180233895778656 0.355582803487778
0.180384278297424 0.312322616577148
0.180542290210724 0.366732269525528
0.1815225481987 0.299874067306519
0.18307763338089 0.327446430921555
0.184250295162201 0.201307505369186
0.185195446014404 0.33823025226593
0.185393631458282 0.267426729202271
0.187025547027588 0.303171217441559
0.190416693687439 0.354278415441513
0.190554857254028 0.298597902059555
0.192553043365479 0.451272368431091
0.192851543426514 0.395838469266891
0.193986058235168 0.32294112443924
0.195306301116943 0.424321562051773
0.196540474891663 0.30873304605484
0.196930885314941 0.337005108594894
0.19761073589325 0.296745091676712
0.198796927928925 0.309718519449234
0.19958233833313 0.310901403427124
0.20103394985199 0.415004044771194
0.201160252094269 0.418995320796967
0.202246963977814 0.284566283226013
0.202638745307922 0.331345230340958
0.202902555465698 0.26447269320488
0.20469331741333 0.293086796998978
0.206405997276306 0.36784565448761
0.208176672458649 0.335948437452316
0.209192514419556 0.219927430152893
0.210101842880249 0.311298161745071
0.211088955402374 0.206988900899887
0.211771965026855 0.336538970470428
0.212001442909241 0.33319279551506
0.213316261768341 0.465543508529663
0.213838994503021 0.499068588018417
0.214916527271271 0.414554625749588
0.215855956077576 0.399649918079376
0.216652274131775 0.474465698003769
0.216834664344788 0.333421379327774
0.217937529087067 0.406087726354599
0.218256235122681 0.380474954843521
0.21925562620163 0.449257582426071
0.220674395561218 0.333047837018967
0.220855057239532 0.395762115716934
0.222877502441406 0.303713798522949
0.223563134670258 0.404504209756851
0.225062787532806 0.493910610675812
0.227192461490631 0.553197205066681
0.227225422859192 0.41915574669838
0.228492796421051 0.539320826530457
0.229353725910187 0.453099936246872
0.229825556278229 0.476109653711319
0.230618238449097 0.379064440727234
0.230863153934479 0.376566588878632
0.231529712677002 0.52615362405777
0.232293546199799 0.405907809734344
0.234022915363312 0.42080420255661
0.237444281578064 0.555793166160583
0.240157127380371 0.462974518537521
0.240422606468201 0.428715109825134
0.240529775619507 0.438666313886642
0.24145370721817 0.326404392719269
0.241626024246216 0.520409166812897
0.241816520690918 0.404221832752228
0.243636012077332 0.473127663135529
0.244035422801971 0.378127157688141
0.244789958000183 0.42659193277359
0.244807362556458 0.487075418233871
0.245333135128021 0.477597892284393
0.248321950435638 0.508722186088562
0.250741302967072 0.452316403388977
0.250891089439392 0.474141001701355
0.250952899456024 0.525581061840057
0.2521733045578 0.438483357429504
0.254678964614868 0.568719148635864
0.255166947841644 0.451238572597504
0.255353808403015 0.560390472412109
0.255556464195251 0.413677752017975
0.257813394069672 0.486448168754578
0.258561193943024 0.356967985630035
0.259115517139435 0.469578862190247
0.260087251663208 0.487933069467545
0.26069962978363 0.580492496490479
0.262701094150543 0.422850221395493
0.264399349689484 0.466352045536041
0.265197336673737 0.424793154001236
0.266014635562897 0.446687549352646
0.269402027130127 0.584706902503967
0.270610988140106 0.523523449897766
0.270729005336761 0.391487866640091
0.271528601646423 0.360863626003265
0.271792411804199 0.418308466672897
0.275497734546661 0.535677134990692
0.275709927082062 0.588598787784576
0.275847315788269 0.548684358596802
0.276239037513733 0.52736109495163
0.277277231216431 0.569197654724121
0.279167830944061 0.575997412204742
0.279838800430298 0.371896266937256
0.282211899757385 0.528735101222992
0.283117949962616 0.460930615663528
0.283523142337799 0.660939931869507
0.283561825752258 0.507692992687225
0.284918665885925 0.358126997947693
0.285549759864807 0.475669264793396
0.28619509935379 0.365298867225647
0.286877632141113 0.460558950901031
0.28694087266922 0.579129159450531
0.287096619606018 0.598691761493683
0.287719309329987 0.625876247882843
0.288107216358185 0.708247482776642
0.289463043212891 0.558381855487823
0.289723873138428 0.549471378326416
0.290348827838898 0.503929018974304
0.292164325714111 0.553253769874573
0.293014466762543 0.440975874662399
0.293560326099396 0.536233425140381
0.294159650802612 0.535838603973389
0.294864058494568 0.639894127845764
0.296637356281281 0.581543684005737
0.299957871437073 0.473068177700043
0.300500512123108 0.575784862041473
0.300996243953705 0.50904780626297
0.304957509040833 0.658901393413544
0.305148601531982 0.637697517871857
0.305178046226501 0.557204723358154
0.306709825992584 0.497022747993469
0.309660255908966 0.563747942447662
0.309926807880402 0.696410953998566
0.310880899429321 0.603282928466797
0.312766194343567 0.551250278949738
0.31489360332489 0.534651935100555
0.317708671092987 0.507068514823914
0.31870824098587 0.555736184120178
0.318920075893402 0.591190993785858
0.319026052951813 0.699483633041382
0.319164037704468 0.471582770347595
0.319267153739929 0.555910527706146
0.319553256034851 0.685565054416656
0.322031736373901 0.659510850906372
0.323155343532562 0.663906633853912
0.323583841323853 0.698763847351074
0.324535012245178 0.580229222774506
0.324745059013367 0.737359404563904
0.33219301700592 0.674508571624756
0.332882523536682 0.628504931926727
0.332949697971344 0.52847546339035
0.333221077919006 0.769786417484283
0.333616554737091 0.696820676326752
0.334072649478912 0.69120866060257
0.337227046489716 0.71652364730835
0.340588808059692 0.698621988296509
0.341925203800201 0.816605448722839
0.343591451644897 0.610135972499847
0.346196174621582 0.675577700138092
0.34884250164032 0.736766874790192
0.348873317241669 0.663905084133148
0.350650310516357 0.748929142951965
0.351543188095093 0.781536459922791
0.354589283466339 0.710460782051086
0.354619204998016 0.68024480342865
0.355625629425049 0.808079719543457
0.355847239494324 0.706413984298706
0.356755256652832 0.816284894943237
0.35686206817627 0.819777429103851
0.35706615447998 0.655704855918884
0.358181357383728 0.707185685634613
0.359050393104553 0.763970136642456
0.36026930809021 0.843882441520691
0.360967695713043 0.8165243268013
0.363143384456635 0.713334262371063
0.363173186779022 0.861489951610565
0.363255977630615 0.920578956604004
0.363328754901886 0.815284132957458
0.36375617980957 0.735824108123779
0.368817627429962 0.854432761669159
0.368886470794678 0.688206553459167
0.370804667472839 0.765926122665405
0.371111929416656 0.905165076255798
0.371653914451599 0.867407858371735
0.373301565647125 0.803149521350861
0.373560428619385 0.878521144390106
0.374024152755737 0.720384359359741
0.374306201934814 0.72947496175766
0.378445088863373 0.804676830768585
0.378772616386414 0.774274051189423
0.379658281803131 0.87795227766037
0.379996836185455 0.806143403053284
0.380324423313141 0.898428797721863
0.381900250911713 0.864015877246857
0.382390201091766 0.825169503688812
0.382779538631439 0.887982845306396
0.384195983409882 0.931018888950348
0.38507217168808 0.884385704994202
0.386103749275208 0.968976080417633
0.387248039245605 0.839808702468872
0.388104796409607 0.876639544963837
0.388121843338013 0.944481134414673
0.388507425785065 0.718878865242004
0.390447676181793 0.831897258758545
0.392211973667145 0.973649024963379
0.392251670360565 0.988537073135376
0.392294824123383 0.887535989284515
0.392300426959991 0.919017314910889
0.39358389377594 1.00822579860687
0.394498646259308 0.910851418972015
0.394821882247925 0.847213983535767
0.395173013210297 0.86043918132782
0.396247148513794 0.838404893875122
0.399033010005951 1.01056170463562
0.400146126747131 0.93982595205307
0.400431036949158 0.860292971134186
0.400593876838684 0.995643019676208
0.401180863380432 0.88323962688446
0.402402818202972 0.882872045040131
0.403287053108215 0.948605597019196
0.403721392154694 0.872640013694763
0.404075264930725 0.910784184932709
0.40414160490036 1.01386201381683
0.405068576335907 0.865586161613464
0.405868530273438 0.955960214138031
0.405909717082977 1.05585932731628
0.405936062335968 0.799963533878326
0.408304989337921 0.946087539196014
0.410912811756134 0.980945229530334
0.411191642284393 0.985248684883118
0.411810994148254 0.899020135402679
0.411881983280182 0.82808917760849
0.412677407264709 0.877067685127258
0.413959383964539 0.897685766220093
0.41544246673584 0.954910635948181
0.41815447807312 0.992578208446503
0.418670833110809 1.00450873374939
0.41948789358139 0.971788942813873
0.420748293399811 0.950726747512817
0.420809626579285 0.919983685016632
0.422456562519073 1.03662419319153
0.422540485858917 1.02879762649536
0.42456990480423 0.961735606193542
0.426298975944519 1.00107753276825
0.426385343074799 0.981194853782654
0.428110182285309 0.886980652809143
0.428433060646057 0.993816018104553
0.42849326133728 0.87762326002121
0.431678175926208 1.05809557437897
0.432132124900818 1.03649067878723
0.432397544384003 0.848028838634491
0.43413233757019 0.926334857940674
0.436286687850952 1.04958319664001
0.438493728637695 1.03790986537933
0.439012289047241 1.00434947013855
0.439452111721039 0.976928174495697
0.43986189365387 1.04589283466339
0.440347790718079 0.997910499572754
0.440456330776215 1.01600015163422
0.441112518310547 1.09926307201385
0.441308796405792 0.873661279678345
0.443403244018555 0.958616852760315
0.444002151489258 1.04563498497009
0.444093108177185 1.03072226047516
0.444178462028503 0.969828128814697
0.445305645465851 1.03837513923645
0.445680379867554 1.03463208675385
0.446228325366974 0.994756519794464
0.44915646314621 1.04873096942902
0.449272394180298 1.03458666801453
0.450604677200317 0.919307947158813
0.451099336147308 1.00055754184723
0.453749060630798 0.971377015113831
0.455396115779877 0.92872679233551
0.456639885902405 1.08679604530334
0.456826210021973 1.05102586746216
0.457016289234161 0.859536528587341
0.458004355430603 1.06897020339966
0.461130976676941 1.04302132129669
0.468546807765961 0.943203210830688
0.469612121582031 0.871877014636993
0.470028281211853 1.05911529064178
0.47035950422287 0.991664826869965
0.470392644405365 1.0235550403595
0.470628678798676 0.977064847946167
0.470661401748657 0.979399561882019
0.47107595205307 0.999050378799438
0.472313642501831 0.913072347640991
0.47395646572113 0.995404481887817
0.474145710468292 0.897049844264984
0.474332928657532 0.894760966300964
0.474968552589417 1.08991396427155
0.475312173366547 0.829710721969604
0.47618967294693 0.94978940486908
0.476969301700592 0.921866238117218
0.479645967483521 0.938471138477325
0.480032622814178 0.895776510238647
0.480459153652191 0.941013693809509
0.480654895305634 0.95587432384491
0.481215178966522 1.10172009468079
0.484621345996857 0.897808492183685
0.485950767993927 0.994762301445007
0.489820241928101 0.806196689605713
0.490106344223022 0.929807305335999
0.490889728069305 1.00430357456207
0.491379737854004 0.872326195240021
0.492365419864655 0.914669513702393
0.49249678850174 1.11458933353424
0.495196282863617 1.04238259792328
0.496517539024353 1.10674095153809
0.497141122817993 0.843351364135742
0.499180257320404 0.906763195991516
0.49930477142334 0.742680191993713
0.499354958534241 0.907268583774567
0.49959272146225 0.975563228130341
0.499758183956146 0.883891224861145
0.500489711761475 1.07022821903229
0.501936197280884 0.879975736141205
0.502381205558777 0.973018288612366
0.502390205860138 0.910817444324493
0.502769351005554 1.01499962806702
0.50390088558197 0.942584455013275
0.50562858581543 0.99922114610672
0.505705058574677 0.765812039375305
0.506111204624176 0.943741023540497
0.506956517696381 1.08585524559021
0.507725358009338 1.06148493289948
0.507761478424072 1.0208181142807
0.509940147399902 0.919509708881378
0.51062673330307 0.88261079788208
0.512452483177185 0.966949701309204
0.512654840946198 0.913566887378693
0.512847721576691 0.980358064174652
0.512902438640594 0.963046610355377
0.513400793075562 0.858234941959381
0.513982534408569 0.886656522750854
0.515919148921967 0.982561111450195
0.518131017684937 0.84987473487854
0.519934058189392 0.95713996887207
0.51997697353363 0.986162185668945
0.520343065261841 0.9204261302948
0.521902918815613 0.890732765197754
0.522745668888092 1.00376045703888
0.522795617580414 0.820377826690674
0.523518562316895 0.870790839195251
0.524305582046509 0.897067487239838
0.524829387664795 0.939477860927582
0.526197910308838 0.986505746841431
0.52657276391983 0.922242820262909
0.530030131340027 0.927573442459106
0.533802688121796 0.906903326511383
0.534465074539185 0.87169736623764
0.535061359405518 0.831379890441895
0.535104155540466 0.938214778900146
0.535816371440887 0.841621875762939
0.535879969596863 0.968203783035278
0.536599338054657 0.975346624851227
0.536784052848816 0.947804987430573
0.536804020404816 0.847392082214355
0.537530124187469 0.948131620883942
0.538085103034973 0.960679352283478
0.538210391998291 0.905161201953888
0.54078996181488 0.968436777591705
0.540870785713196 0.776616215705872
0.540996015071869 0.940863072872162
0.541006445884705 0.750708699226379
0.54118549823761 0.92101263999939
0.541267275810242 0.713598370552063
0.542739570140839 0.999610543251038
0.544417858123779 0.963367938995361
0.544990599155426 0.862401962280273
0.545052230358124 0.847818851470947
0.545093536376953 0.893571257591248
0.545098304748535 0.931350529193878
0.546921730041504 0.863667547702789
0.547452330589294 0.815664172172546
0.548009276390076 0.85366815328598
0.548233270645142 0.932733535766602
0.549476385116577 0.800658583641052
0.550529778003693 0.882820427417755
0.551612019538879 0.849600315093994
0.553167760372162 1.00570821762085
0.554103553295135 0.90585857629776
0.554401516914368 0.826975226402283
0.556763231754303 0.852346658706665
0.556899547576904 0.814224898815155
0.557880878448486 0.878675580024719
0.558060348033905 0.908758044242859
0.558150768280029 0.824082612991333
0.558262228965759 0.809279024600983
0.55834823846817 0.801167249679565
0.558591306209564 0.882307112216949
0.559162497520447 0.87387490272522
0.559988915920258 0.853778541088104
0.560512840747833 0.892246305942535
0.562030911445618 0.833308935165405
0.562895476818085 0.825841426849365
0.562931835651398 0.823157012462616
0.566300928592682 0.928120851516724
0.566549837589264 0.90599513053894
0.567325472831726 0.773273289203644
0.567661464214325 0.900383055210114
0.568280875682831 0.892396569252014
0.569289207458496 0.770938873291016
0.569811582565308 0.655525147914886
0.570483505725861 0.781142115592957
0.572854995727539 0.777050197124481
0.575360357761383 0.812016487121582
0.57647031545639 0.800162613391876
0.577002584934235 0.80826860666275
0.577011823654175 0.738610029220581
0.581000685691833 0.743575215339661
0.581099152565002 0.759232044219971
0.581897616386414 0.864210605621338
0.581961691379547 0.87911319732666
0.582343399524689 0.754026174545288
0.58418333530426 0.804990351200104
0.585272908210754 0.819870591163635
0.588422775268555 0.763289511203766
0.589165568351746 0.937043130397797
0.589285910129547 0.723585188388824
0.590005934238434 0.839179873466492
0.593346655368805 0.797218799591064
0.595482230186462 0.815044581890106
0.595876276493073 0.836595594882965
0.597435772418976 0.830663442611694
0.598685503005981 0.826224863529205
0.599067628383636 0.828388452529907
0.601053059101105 0.752611935138702
0.601823925971985 0.871835947036743
0.602961242198944 0.882820606231689
0.605050086975098 0.932015180587769
0.611085712909698 0.863713324069977
0.611132621765137 0.65439772605896
0.611394107341766 0.676062285900116
0.611597120761871 0.733205378055573
0.614727854728699 0.787245810031891
0.615956366062164 0.775803744792938
0.617069482803345 0.744324207305908
0.617742776870728 0.686392724514008
0.619130313396454 0.707504272460938
0.621442198753357 0.897154927253723
0.622425019741058 1.00288057327271
0.622567117214203 0.795681536197662
0.622647345066071 0.885241329669952
0.622946560382843 0.691779136657715
0.623782098293304 0.824056506156921
0.625869393348694 0.730927348136902
0.625899195671082 0.773500382900238
0.628299474716187 0.802800118923187
0.631012320518494 0.773403882980347
0.632265865802765 0.759315490722656
0.632640242576599 0.686627686023712
0.632669746875763 0.681142568588257
0.634849905967712 0.777040481567383
0.637507438659668 0.755177557468414
0.638323485851288 0.796857118606567
0.638843774795532 0.788782894611359
0.639834225177765 0.734907448291779
0.639883458614349 0.865713655948639
0.641278147697449 0.7649906873703
0.642634630203247 0.734972894191742
0.643203377723694 0.701884508132935
0.645859360694885 0.597112655639648
0.647501528263092 0.669337272644043
0.647805213928223 0.693761885166168
0.648837506771088 0.675831317901611
0.649549961090088 0.696913242340088
0.649925529956818 0.574736297130585
0.650067389011383 0.662048518657684
0.650239884853363 0.824533462524414
0.650544226169586 0.704847037792206
0.652437627315521 0.748818159103394
0.65274840593338 0.707355976104736
0.654480934143066 0.652798771858215
0.654526352882385 0.716390132904053
0.655307352542877 0.736777067184448
0.656225025653839 0.819793879985809
0.656365036964417 0.622492134571075
0.658714771270752 0.668840289115906
0.658925354480743 0.794818878173828
0.659018695354462 0.574976861476898
0.659281730651855 0.712123334407806
0.659536421298981 0.713902413845062
0.659565031528473 0.826691091060638
0.661026895046234 0.728485882282257
0.662578463554382 0.69739043712616
0.662862360477448 0.651555120944977
0.663303136825562 0.591562747955322
0.663590312004089 0.800294876098633
0.665575444698334 0.697277247905731
0.666386842727661 0.639361023902893
0.666478753089905 0.694150328636169
0.66671884059906 0.5272176861763
0.667573928833008 0.76781553030014
0.667755007743835 0.74107152223587
0.668736577033997 0.703277051448822
0.66919219493866 0.616458475589752
0.669954299926758 0.641982793807983
0.672591149806976 0.616837918758392
0.673006176948547 0.670163929462433
0.673557221889496 0.781705796718597
0.673694968223572 0.679193019866943
0.67452085018158 0.623165249824524
0.677889764308929 0.618856191635132
0.679241478443146 0.662560999393463
0.679985523223877 0.841254830360413
0.680927872657776 0.670216739177704
0.682246446609497 0.617480397224426
0.682728171348572 0.636235117912292
0.683104932308197 0.672212958335876
0.684252440929413 0.62712299823761
0.684777855873108 0.535127878189087
0.685512721538544 0.703313231468201
0.68716686964035 0.628434240818024
0.688127040863037 0.700554668903351
0.689122974872589 0.620566606521606
0.690440952777863 0.655558049678802
0.692194700241089 0.617797374725342
0.692817866802216 0.60010027885437
0.696113049983978 0.620199024677277
0.696246981620789 0.699909687042236
0.696474313735962 0.702060461044312
0.697291135787964 0.633392453193665
0.697302341461182 0.667158186435699
0.697517335414886 0.609405338764191
0.698112368583679 0.703733265399933
0.698269724845886 0.637637794017792
0.69835501909256 0.594149589538574
0.698971569538116 0.716010451316833
0.699137628078461 0.664977312088013
0.69913923740387 0.670449495315552
0.700214087963104 0.681978106498718
0.703082323074341 0.56821745634079
0.704374253749847 0.729952275753021
0.705483675003052 0.586257755756378
0.706195652484894 0.638895034790039
0.708456695079803 0.567189157009125
0.708650231361389 0.565611124038696
0.710206270217896 0.762919902801514
0.71100240945816 0.633706212043762
0.711526095867157 0.680246651172638
0.712998867034912 0.741112172603607
0.715877532958984 0.588097035884857
0.716496706008911 0.570589900016785
0.717505931854248 0.679842591285706
0.719501256942749 0.481424152851105
0.720014691352844 0.525447309017181
0.7201327085495 0.684595584869385
0.72138649225235 0.614733099937439
0.72193306684494 0.579581022262573
0.722175180912018 0.648967921733856
0.722286343574524 0.708336055278778
0.722844421863556 0.718070685863495
0.724134862422943 0.638176620006561
0.724979221820831 0.700418591499329
0.725767076015472 0.630586385726929
0.7265904545784 0.547821938991547
0.726683020591736 0.599741220474243
0.727060914039612 0.647916436195374
0.727489769458771 0.568997025489807
0.729349493980408 0.58421778678894
0.731857061386108 0.579410552978516
0.733681678771973 0.596387207508087
0.734320998191833 0.521370589733124
0.735029637813568 0.592935383319855
0.735053062438965 0.708371460437775
0.735960364341736 0.636051595211029
0.737233877182007 0.632313847541809
0.737927854061127 0.516477942466736
0.738638758659363 0.603987753391266
0.739779651165009 0.546169877052307
0.740231931209564 0.57080614566803
0.740990281105042 0.61369401216507
0.741952002048492 0.570578038692474
0.743195235729218 0.597872912883759
0.743206739425659 0.430045455694199
0.743236064910889 0.547782182693481
0.744196176528931 0.684131324291229
0.744572579860687 0.574510753154755
0.745292484760284 0.616999208927155
0.745376646518707 0.44008731842041
0.745495736598969 0.492301136255264
0.746768355369568 0.435265153646469
0.747916221618652 0.493111729621887
0.749733328819275 0.550040602684021
0.750327706336975 0.582709670066833
0.750944077968597 0.533451616764069
0.751571476459503 0.602015137672424
0.752250552177429 0.49835467338562
0.752313494682312 0.577032208442688
0.754422545433044 0.426008492708206
0.755354046821594 0.46905979514122
0.756339490413666 0.584897756576538
0.757269024848938 0.544463992118835
0.758492588996887 0.451216846704483
0.758496820926666 0.570398330688477
0.759116053581238 0.624641954898834
0.759288787841797 0.506462275981903
0.759818136692047 0.521568715572357
0.760750234127045 0.534923553466797
0.761180520057678 0.460539400577545
0.765182912349701 0.398102819919586
0.765520751476288 0.488896012306213
0.766566574573517 0.437449306249619
0.767966032028198 0.417973160743713
0.76958441734314 0.508031666278839
0.769771099090576 0.54340124130249
0.770561397075653 0.478657841682434
0.771108567714691 0.427851289510727
0.772491037845612 0.426260858774185
0.773934543132782 0.42018324136734
0.774413645267487 0.38590407371521
0.775593519210815 0.385071903467178
0.775602996349335 0.441393256187439
0.775866627693176 0.43598398566246
0.776708960533142 0.473157495260239
0.77766215801239 0.3999302983284
0.778743922710419 0.508148789405823
0.779879152774811 0.4276964366436
0.780417859554291 0.448066025972366
0.780419170856476 0.453856378793716
0.781103312969208 0.518841624259949
0.781149625778198 0.347348004579544
0.781194150447845 0.498339056968689
0.781676054000854 0.471745997667313
0.782239377498627 0.402174443006516
0.783648550510406 0.451345890760422
0.784381687641144 0.481939315795898
0.785483598709106 0.401180922985077
0.787930369377136 0.417968899011612
0.788056433200836 0.538596153259277
0.78807520866394 0.466346442699432
0.790058851242065 0.490572959184647
0.790267825126648 0.462493985891342
0.791922986507416 0.369175970554352
0.795758724212646 0.528627336025238
0.795789420604706 0.450943112373352
0.79665219783783 0.493756204843521
0.797267496585846 0.434871077537537
0.797861933708191 0.431111752986908
0.801692187786102 0.510977029800415
0.803759157657623 0.367748498916626
0.805622160434723 0.452275156974792
0.805865943431854 0.433191657066345
0.806981325149536 0.472896754741669
0.807988524436951 0.433773100376129
0.809130847454071 0.522603750228882
0.810865521430969 0.449628502130508
0.811038136482239 0.634457647800446
0.811396837234497 0.523616671562195
0.812509179115295 0.492984771728516
0.813900709152222 0.387739896774292
0.813984394073486 0.41796001791954
0.814246296882629 0.454816460609436
0.815562307834625 0.46849524974823
0.815814197063446 0.552738130092621
0.81609970331192 0.529974520206451
0.817032039165497 0.476048707962036
0.819423794746399 0.521797299385071
0.820094645023346 0.481008172035217
0.820098161697388 0.502634942531586
0.820338010787964 0.483696728944778
0.820343852043152 0.52307915687561
0.82218199968338 0.454779207706451
0.822318732738495 0.536388397216797
0.823667168617249 0.523280322551727
0.823674321174622 0.574835538864136
0.827399075031281 0.407162576913834
0.827441692352295 0.551086485385895
0.828736186027527 0.465060710906982
0.829689562320709 0.539921462535858
0.831529676914215 0.542847812175751
0.832857429981232 0.529125571250916
0.833512187004089 0.349432289600372
0.833581745624542 0.404031425714493
0.833932220935822 0.493073403835297
0.835229456424713 0.530414044857025
0.836114764213562 0.555651843547821
0.836464464664459 0.560935854911804
0.83811753988266 0.46147608757019
0.839439451694489 0.474531769752502
0.839691162109375 0.490872651338577
0.840052902698517 0.485127151012421
0.841350853443146 0.44624787569046
0.842326760292053 0.519717752933502
0.843134641647339 0.551135361194611
0.843337893486023 0.440773248672485
0.844574809074402 0.50723797082901
0.845124185085297 0.607496380805969
0.846815466880798 0.51130998134613
0.848071217536926 0.611072063446045
0.84828382730484 0.635140240192413
0.848414361476898 0.526213586330414
0.849238514900208 0.613093376159668
0.849566996097565 0.473238229751587
0.849775910377502 0.669857442378998
0.849871456623077 0.547387421131134
0.852007329463959 0.586960792541504
0.853645503520966 0.518757522106171
0.854109168052673 0.500399351119995
0.855393707752228 0.52078914642334
0.855556786060333 0.460366666316986
0.856502950191498 0.406446635723114
0.857758104801178 0.426400125026703
0.857791841030121 0.432245135307312
0.861385107040405 0.463213235139847
0.861647427082062 0.65341454744339
0.861714959144592 0.567756593227386
0.865676164627075 0.567544996738434
0.866383671760559 0.584519565105438
0.867403090000153 0.519476711750031
0.868175029754639 0.592077672481537
0.868794202804565 0.491250872612
0.869205176830292 0.584866046905518
0.869648039340973 0.633928298950195
0.870958864688873 0.524990200996399
0.87123316526413 0.47611141204834
0.871411919593811 0.499679207801819
0.872288286685944 0.542415916919708
0.872929573059082 0.461805015802383
0.875264048576355 0.502415835857391
0.875558197498322 0.457399010658264
0.876104056835175 0.563918650150299
0.876227676868439 0.563479423522949
0.876926481723785 0.594160676002502
0.87721985578537 0.481668591499329
0.878821432590485 0.599969685077667
0.879399180412292 0.567369282245636
0.880362272262573 0.517484486103058
0.880517244338989 0.603677809238434
0.88086074590683 0.553404271602631
0.884024560451508 0.562549710273743
0.885680675506592 0.547514379024506
0.887285709381104 0.492359399795532
0.887566268444061 0.422043293714523
0.88915491104126 0.503682971000671
0.889345169067383 0.532938838005066
0.890521109104156 0.531809449195862
0.891411304473877 0.554557025432587
0.893763780593872 0.538321256637573
0.893934547901154 0.517853796482086
0.896422028541565 0.528525650501251
0.898175239562988 0.519107699394226
0.89886212348938 0.45486706495285
0.89984405040741 0.555767476558685
0.900085091590881 0.556621432304382
0.901403069496155 0.520281970500946
0.902433931827545 0.507114708423615
0.902540922164917 0.449361473321915
0.902693688869476 0.663316071033478
0.903219819068909 0.597478032112122
0.904124617576599 0.535703837871552
0.90525496006012 0.626158535480499
0.905770659446716 0.49345874786377
0.906245350837708 0.630500078201294
0.909311652183533 0.477698683738708
0.910041689872742 0.623989462852478
0.910264253616333 0.412902891635895
0.910971879959106 0.422514915466309
0.912175059318542 0.502094328403473
0.915790438652039 0.524363040924072
0.916977345943451 0.537147045135498
0.918918371200562 0.526193618774414
0.920561075210571 0.560707688331604
0.920840263366699 0.490170061588287
0.920883238315582 0.552538931369781
0.921666085720062 0.591804623603821
0.922611832618713 0.447463929653168
0.924246907234192 0.487885773181915
0.925482094287872 0.408708244562149
0.925991356372833 0.448851674795151
0.929618120193481 0.378801494836807
0.930388927459717 0.587993323802948
0.931363701820374 0.369258105754852
0.932113468647003 0.357322096824646
0.932557284832001 0.597187399864197
0.93282824754715 0.453138768672943
0.935172438621521 0.488484889268875
0.937557339668274 0.561141610145569
0.939596235752106 0.444053173065186
0.939790546894073 0.536981046199799
0.94224750995636 0.40333154797554
0.946286380290985 0.425106585025787
0.94671219587326 0.412225693464279
0.947061061859131 0.315797418355942
0.948757350444794 0.343373477458954
0.948972523212433 0.520504534244537
0.950052797794342 0.268607556819916
0.952594041824341 0.511331140995026
0.955855309963226 0.438464790582657
0.956705808639526 0.412334680557251
0.956948816776276 0.41785055398941
0.957008957862854 0.378464102745056
0.957778871059418 0.412868201732635
0.958848237991333 0.363123834133148
0.9596306681633 0.333796203136444
0.960404574871063 0.340954273939133
0.961005032062531 0.481251418590546
0.96245002746582 0.386128157377243
0.962753117084503 0.399103105068207
0.964278399944305 0.412356585264206
0.965688228607178 0.374429911375046
0.966017663478851 0.391700506210327
0.966664969921112 0.413322895765305
0.967511475086212 0.520873844623566
0.967783212661743 0.39738991856575
0.968621909618378 0.52184671163559
0.970402956008911 0.39834052324295
0.971209347248077 0.319878786802292
0.972378492355347 0.456798046827316
0.974245369434357 0.337647438049316
0.975155293941498 0.458090662956238
0.97547847032547 0.387804925441742
0.975614190101624 0.389824539422989
0.97569066286087 0.401550322771072
0.976303517818451 0.378676235675812
0.977004826068878 0.355845659971237
0.979181408882141 0.390399992465973
0.980850160121918 0.354745000600815
0.981413662433624 0.39481720328331
0.981556057929993 0.334054887294769
0.982095241546631 0.388748407363892
0.985821127891541 0.285376846790314
0.986221075057983 0.223657533526421
0.987662255764008 0.32119545340538
0.989157676696777 0.36841681599617
0.990050256252289 0.340312391519547
0.990136861801147 0.430326014757156
0.990395128726959 0.473032712936401
0.99216902256012 0.290315866470337
0.993385791778564 0.476230561733246
0.998277604579926 0.265586137771606
0.998482644557953 0.451301693916321
0.999735236167908 0.220510095357895
};
\addplot [draw=blue, fill=blue, mark=*, only marks, opacity=0.5]
table{%
x  y
0.44915646314621 1.04873096942902
0.453749060630798 0.971377015113831
0.999735236167908 0.220510095357895
0.431678175926208 1.05809557437897
0.470028281211853 1.05911529064178
0.455396115779877 0.92872679233551
0.456639885902405 1.08679604530334
0.472313642501831 0.913072347640991
0.468546807765961 0.943203210830688
0.450604677200317 0.919307947158813
0.461130976676941 1.04302132129669
8.52346420288086e-05 0.291133195161819
0.474145710468292 0.897049844264984
0.50390088558197 0.942584455013275
0.445680379867554 1.03463208675385
0.469612121582031 0.871877014636993
0.213838994503021 0.499068588018417
0.661026895046234 0.728485882282257
0.47107595205307 0.999050378799438
0.458004355430603 1.06897020339966
0.456826210021973 1.05102586746216
0.457016289234161 0.859536528587341
0.451099336147308 1.00055754184723
0.449272394180298 1.03458666801453
0.47395646572113 0.995404481887817
0.470661401748657 0.979399561882019
0.446228325366974 0.994756519794464
0.470628678798676 0.977064847946167
0.432397544384003 0.848028838634491
0.470392644405365 1.0235550403595
0.47035950422287 0.991664826869965
};
\addlegendentry{PFN}
\addplot [draw=red, fill=red, mark=*, only marks, opacity=0.5]
table{%
x  y
0.44915646314621 1.04873096942902
0.444178462028503 0.969828128814697
0.453749060630798 0.971377015113831
0.444002151489258 1.04563498497009
0.443403244018555 0.958616852760315
0.939596235752106 0.444053173065186
0.455396115779877 0.92872679233551
0.456639885902405 1.08679604530334
0.58418333530426 0.804990351200104
0.590005934238434 0.839179873466492
0.441308796405792 0.873661279678345
0.468546807765961 0.943203210830688
0.229353725910187 0.453099936246872
0.450604677200317 0.919307947158813
0.461130976676941 1.04302132129669
0.445305645465851 1.03837513923645
0.445680379867554 1.03463208675385
0.469612121582031 0.871877014636993
0.213838994503021 0.499068588018417
0.444093108177185 1.03072226047516
0.577011823654175 0.738610029220581
0.458004355430603 1.06897020339966
0.456826210021973 1.05102586746216
0.413959383964539 0.897685766220093
0.698269724845886 0.637637794017792
0.457016289234161 0.859536528587341
0.451099336147308 1.00055754184723
0.449272394180298 1.03458666801453
0.135038614273071 0.270624935626984
0.446228325366974 0.994756519794464
0.530030131340027 0.927573442459106
};
\addlegendentry{GP}
\legend{};
\end{axis}

\end{tikzpicture}

%% file: MetaLearnWorkshop/figures/hebo_on_hebo_example_2.tex
\begin{tikzpicture}

\begin{axis}[
legend cell align={left},
legend style={
  fill opacity=0.8,
  draw opacity=1,
  text opacity=1,
  at={(0.03,0.03)},
  anchor=south west,
  draw=white!80!black
},
tick align=outside,
tick pos=left,
x grid style={white!69.0196078431373!black},
xmin=-0.0489058196544647, xmax=1.04744836688042,
xtick style={color=black},
y grid style={white!69.0196078431373!black},
ymin=-0.730771313607693, ymax=0.433544616401196,
ytick style={color=black}, width=\textwidth, height=\textwidth,
]
\addplot [semithick, white!50.1960784313725!black, forget plot]
table {%
0.00092846155166626 -0.202616214752197
0.00257432460784912 -0.165530353784561
0.00589418411254883 -0.279467433691025
0.0063556432723999 -0.0314770191907883
0.00679057836532593 -0.22300212085247
0.00728720426559448 -0.184675693511963
0.00964581966400146 -0.263805508613586
0.0104939341545105 -0.0114532709121704
0.0107897520065308 0.0514030084013939
0.0112658739089966 0.0481321811676025
0.0114412307739258 -0.221071571111679
0.0122391581535339 0.0367782786488533
0.014276921749115 -0.451548337936401
0.016072690486908 0.0355637893080711
0.0160778760910034 -0.0622736364603043
0.0164305567741394 0.0126902088522911
0.0187850594520569 -0.0164866000413895
0.0212677121162415 0.0760011747479439
0.0221312642097473 -0.151822254061699
0.0222833752632141 -0.318360805511475
0.0230520367622375 -0.0295409075915813
0.0235278606414795 0.00162540376186371
0.023862361907959 -0.0744325891137123
0.0239171981811523 -0.285241365432739
0.0239870548248291 -0.276188462972641
0.0251039266586304 -0.0192468725144863
0.0258756279945374 -0.0920842513442039
0.0259672403335571 -0.151209950447083
0.0263203978538513 0.0213649123907089
0.0270591378211975 -0.151377514004707
0.0276182293891907 -0.151685416698456
0.0288788080215454 0.0639764219522476
0.0289638638496399 -0.129740431904793
0.0291405320167542 0.0449600592255592
0.0291650891304016 -0.0135651677846909
0.0304962992668152 -0.00271373242139816
0.0305755734443665 -0.255924880504608
0.0320157408714294 -0.176588773727417
0.0340774655342102 -0.115180976688862
0.0353922247886658 0.0516872555017471
0.0363065004348755 0.036669634282589
0.0375193953514099 0.12771712243557
0.0387241840362549 -0.088435024023056
0.0400042533874512 -0.160485059022903
0.0410518050193787 -0.174683511257172
0.0417539477348328 -0.0713227093219757
0.0419920682907104 -0.0262116268277168
0.0442166924476624 -0.0782307535409927
0.0447625517845154 -0.396120488643646
0.0448413491249084 0.0559295117855072
0.0451632142066956 -0.152072995901108
0.0453979969024658 0.105597481131554
0.0455912947654724 -0.233971729874611
0.0456152558326721 -0.221630036830902
0.0459954142570496 -0.0300640650093555
0.046363353729248 -0.199593067169189
0.0466258525848389 0.119031809270382
0.0466635823249817 -0.125877916812897
0.0477652549743652 -0.028057087212801
0.048056423664093 -0.105867825448513
0.048327624797821 -0.0801836177706718
0.0489953756332397 -0.140693679451942
0.0495092868804932 -0.055616732686758
0.0513243675231934 -0.27395087480545
0.0559583306312561 -0.229700654745102
0.0590564608573914 -0.0300328135490417
0.0600242614746094 -0.1641586124897
0.0632046461105347 0.0654576420783997
0.0636166930198669 -0.113190539181232
0.064532458782196 0.267953991889954
0.0670799016952515 -0.0711456015706062
0.0677050948143005 -0.259108871221542
0.0681868195533752 -0.123638808727264
0.0683515667915344 -0.151590630412102
0.0692979693412781 -0.123926930129528
0.0695959329605103 -0.084236815571785
0.0711385011672974 -0.0735002458095551
0.0715187788009644 -0.0186069086194038
0.0717700123786926 -0.00530913472175598
0.0730509757995605 -0.0143588036298752
0.0736969113349915 -0.109141424298286
0.0737826824188232 -0.052206352353096
0.0739925503730774 -0.0065082386136055
0.0764964818954468 0.0238832458853722
0.0775805115699768 0.0663033872842789
0.0791187286376953 -0.0922619998455048
0.0798352956771851 -0.107476763427258
0.0802133083343506 0.148982793092728
0.0803200006484985 -0.0414520129561424
0.080586850643158 -0.0559350252151489
0.0822451114654541 -0.0460336767137051
0.0822886824607849 -0.124273873865604
0.0828485488891602 -0.127006262540817
0.0835652947425842 -0.261655867099762
0.0844007730484009 -0.328388094902039
0.0859438180923462 0.116894781589508
0.0861801505088806 -0.161920964717865
0.0863577723503113 0.0748872831463814
0.087590217590332 -0.13120698928833
0.0878874063491821 -0.23109433054924
0.0881634950637817 -0.0303104482591152
0.0889030694961548 -0.0523105710744858
0.0911379456520081 -0.17208394408226
0.092924177646637 -0.159419432282448
0.094292938709259 -0.186478406190872
0.0964751839637756 -0.268509447574615
0.0972840189933777 -0.0426964648067951
0.100539207458496 0.058416374027729
0.101997077465057 -0.185826003551483
0.102087199687958 0.0320552438497543
0.102850794792175 0.106206506490707
0.104059457778931 -0.174655497074127
0.104672074317932 -0.217498496174812
0.105636239051819 -0.132205814123154
0.107144117355347 0.0651131272315979
0.10869437456131 -0.0161384753882885
0.108860909938812 -0.222055673599243
0.109214782714844 -0.0990531146526337
0.109250485897064 -0.179685220122337
0.111354470252991 -0.0453298091888428
0.111513435840607 0.0353116095066071
0.113950669765472 0.185417100787163
0.115558266639709 -0.0113875344395638
0.116531193256378 0.178845539689064
0.117037713527679 0.0397236943244934
0.117187201976776 -0.132990568876266
0.118281304836273 0.0502079203724861
0.119206547737122 -0.2126774340868
0.119339227676392 0.0910086631774902
0.119742691516876 -0.199267849326134
0.121207118034363 -0.105263866484165
0.123034477233887 0.0124762058258057
0.123293578624725 0.211999148130417
0.124092817306519 -0.0717714056372643
0.124265074729919 -0.0685413852334023
0.12440150976181 -0.15027517080307
0.125593066215515 -0.0789089351892471
0.126041948795319 0.072950467467308
0.126150846481323 -0.0220889896154404
0.127145707607269 -0.161133974790573
0.128730297088623 -0.137361943721771
0.129713773727417 -0.0190822221338749
0.130961835384369 0.0291521847248077
0.131067156791687 -0.111388206481934
0.134514093399048 -0.0790471732616425
0.135306119918823 0.0229599233716726
0.136551916599274 0.0151746012270451
0.137236058712006 -0.0200124550610781
0.139781951904297 -0.0425809770822525
0.140080451965332 -0.288468688726425
0.141112864017487 -0.150050833821297
0.141848385334015 -0.0920110195875168
0.142446160316467 -0.151434615254402
0.142733991146088 -0.0816661864519119
0.143089652061462 -0.15395799279213
0.143768489360809 0.0557813048362732
0.144136309623718 -0.229148834943771
0.145042657852173 0.0239839050918818
0.150119423866272 0.00270748231559992
0.151843726634979 0.0980680882930756
0.152107954025269 -0.00516506098210812
0.152158498764038 -0.253243774175644
0.155070066452026 0.112191133201122
0.155118107795715 0.121464237570763
0.155360221862793 0.146080762147903
0.160015165805817 0.0929853618144989
0.162416458129883 0.0970035493373871
0.163224756717682 0.0431322790682316
0.163813412189484 -0.106652639806271
0.164154350757599 -0.0688489899039268
0.165421426296234 -0.0778728574514389
0.167545139789581 0.0818995982408524
0.167811393737793 0.00462069269269705
0.167851269245148 0.109483554959297
0.169511318206787 -0.0719846487045288
0.170672059059143 0.0381217896938324
0.172257304191589 0.0457076281309128
0.172457635402679 -0.024538055062294
0.173421204090118 0.151155859231949
0.173622906208038 -0.0835586786270142
0.1739462018013 0.0525536797940731
0.176334857940674 -0.0538414902985096
0.176541268825531 0.0630232244729996
0.178555905818939 0.198258653283119
0.178997993469238 -0.0204114019870758
0.179110944271088 -0.0807857662439346
0.179623365402222 -0.166345655918121
0.181875109672546 -0.0558607466518879
0.182038366794586 0.0963219478726387
0.182807385921478 0.0590850748121738
0.185605466365814 0.0443370305001736
0.185830295085907 0.108501300215721
0.186591327190399 0.152221783995628
0.188310325145721 0.185427188873291
0.191090524196625 -0.135234370827675
0.191875159740448 0.0947580188512802
0.194044470787048 0.00982263591140509
0.195048570632935 -0.00861820392310619
0.196854174137115 0.0666161328554153
0.196879804134369 0.140948742628098
0.197501063346863 0.116830222308636
0.198099553585052 0.137658953666687
0.203383505344391 -0.0357897318899632
0.209720730781555 0.127094686031342
0.211825370788574 0.103507958352566
0.21237176656723 0.071953073143959
0.21259605884552 0.0894219279289246
0.212820708751678 0.0637220218777657
0.214050233364105 -0.029586736112833
0.214988589286804 0.189351543784142
0.215640127658844 0.0643238425254822
0.217170715332031 -0.100535839796066
0.217518210411072 -0.0410603657364845
0.217960715293884 0.220738276839256
0.218441188335419 0.213144734501839
0.218538641929626 -0.128374963998795
0.218637585639954 0.0585593283176422
0.220242917537689 -0.0542074292898178
0.221825540065765 -0.0718820914626122
0.221908748149872 0.00783794280141592
0.223562777042389 -0.0370888970792294
0.223583281040192 -0.21496108174324
0.224663436412811 0.04893808811903
0.225398480892181 -0.121966660022736
0.227014183998108 0.111658364534378
0.22879821062088 -0.1329295784235
0.23099559545517 -0.00151634402573109
0.232033967971802 0.124788358807564
0.232459425926208 0.107418194413185
0.233624339103699 0.0373077839612961
0.234163820743561 0.0606120824813843
0.235299110412598 0.10160505771637
0.236962258815765 -0.189192593097687
0.237625062465668 -0.0160589516162872
0.239240050315857 -0.120778761804104
0.239303231239319 -0.0553366430103779
0.239346027374268 -0.0285791102796793
0.239483892917633 -0.109698556363583
0.239660918712616 0.00926803983747959
0.240252792835236 0.014792762696743
0.241577327251434 -0.058590579777956
0.243947565555573 -0.0825380086898804
0.244090855121613 0.0653250738978386
0.246447443962097 -0.142790824174881
0.246529042720795 -0.104491710662842
0.246777892112732 0.131956160068512
0.248116135597229 0.00981787592172623
0.248735010623932 0.0120512768626213
0.248915195465088 0.0323902443051338
0.249655961990356 -0.00654731504619122
0.250291049480438 0.0458019822835922
0.251165807247162 -0.156740590929985
0.251270353794098 0.0472500026226044
0.252190589904785 -0.0845130309462547
0.252580940723419 -0.139565497636795
0.253803610801697 -0.119272567331791
0.254267930984497 0.00245456770062447
0.254932761192322 -0.2026347219944
0.255572378635406 -0.146896675229073
0.255648910999298 0.000123727484606206
0.256034851074219 0.110605597496033
0.256331861019135 -0.0934749245643616
0.256362855434418 -0.194400995969772
0.257544338703156 0.112602546811104
0.257941067218781 -0.175110504031181
0.258711278438568 0.118470393121243
0.259100198745728 0.000632722396403551
0.261327683925629 -0.0217186771333218
0.26228940486908 0.105893984436989
0.263210892677307 -0.177927955985069
0.263304173946381 0.0824229419231415
0.263767957687378 0.12839587032795
0.265148758888245 -0.0494514517486095
0.266320586204529 -0.0350280590355396
0.266809105873108 -0.101337417960167
0.266947150230408 0.0636922419071198
0.267576396465302 0.0236340761184692
0.267954230308533 -0.0720308348536491
0.269720137119293 -0.0850820690393448
0.270157098770142 0.0621283948421478
0.270401298999786 0.00915352813899517
0.271250069141388 0.0990232154726982
0.27251136302948 -0.0102019067853689
0.272619366645813 -0.0694758370518684
0.273886919021606 -0.20197831094265
0.275389552116394 0.0911761671304703
0.277250111103058 0.111461713910103
0.278018593788147 -0.106654688715935
0.278896331787109 -0.104028016328812
0.280175149440765 -0.0155598744750023
0.280499517917633 -0.114858150482178
0.281648457050323 0.0842287763953209
0.282632827758789 -0.0649998635053635
0.283860504627228 -0.136869058012962
0.28521203994751 0.0445119813084602
0.286320209503174 -0.000515960156917572
0.287659883499146 0.00968045741319656
0.289305925369263 0.0557753555476665
0.289709568023682 -0.0500231981277466
0.29078209400177 -0.235426843166351
0.291069209575653 -0.0959499031305313
0.292207598686218 0.157271593809128
0.292283952236176 -0.0175091214478016
0.294197082519531 -0.036129042506218
0.295168936252594 0.0833960622549057
0.296064376831055 -0.0882878303527832
0.296705722808838 0.124091237783432
0.297545909881592 -0.0484836176037788
0.298480927944183 0.0039464645087719
0.298593401908875 -0.00303765013813972
0.30056893825531 -0.157568454742432
0.301549732685089 -0.231673523783684
0.301718473434448 -0.196938335895538
0.302456498146057 0.0556124746799469
0.302664220333099 -0.0560546815395355
0.303058862686157 0.0349535793066025
0.303061306476593 -0.10118006169796
0.304725348949432 -0.097360834479332
0.305242717266083 0.0473610460758209
0.307378172874451 0.0735165551304817
0.309487164020538 0.0175314247608185
0.310400009155273 -0.208177879452705
0.310584962368011 0.00235695391893387
0.311017513275146 0.000936232507228851
0.311720609664917 0.0409543812274933
0.312570810317993 -0.0436478927731514
0.314203023910522 -0.133061408996582
0.31428474187851 -0.143751755356789
0.314744174480438 0.057888574898243
0.316275656223297 -0.118167541921139
0.317826569080353 0.00467319041490555
0.320747971534729 -0.127482578158379
0.321011662483215 -0.00718978047370911
0.321969985961914 -0.156222432851791
0.323122441768646 0.159602135419846
0.323171436786652 -0.11655093729496
0.324699878692627 -0.0413679704070091
0.324765563011169 -0.279356896877289
0.326466739177704 -0.0067678689956665
0.326522946357727 -0.0480200760066509
0.326841652393341 -0.0232469439506531
0.327114164829254 0.0369529947638512
0.332080543041229 -0.11236697435379
0.332459807395935 -0.246322765946388
0.332547008991241 -0.153077349066734
0.335393369197845 -0.0321051254868507
0.335488200187683 0.0570100024342537
0.33571583032608 -0.0837614685297012
0.335764706134796 0.04427869617939
0.335817456245422 -0.161684080958366
0.336524188518524 0.0701538920402527
0.336871147155762 -0.11124973744154
0.340639114379883 -0.0259377658367157
0.340729355812073 -0.230804234743118
0.343158364295959 -0.173277705907822
0.343641459941864 -0.198294252157211
0.344507813453674 -0.159449458122253
0.344977736473083 -0.230316549539566
0.345470428466797 -0.0666123777627945
0.34564083814621 -0.11555427312851
0.346384763717651 -0.100786536931992
0.347466111183167 -0.412128090858459
0.347527027130127 -0.161786332726479
0.349535763263702 -0.227761656045914
0.349600315093994 -0.147173032164574
0.349622547626495 -0.023718997836113
0.350798904895782 -0.0547289997339249
0.351008117198944 -0.158954814076424
0.352101624011993 -0.0725385919213295
0.352968811988831 -0.231618970632553
0.354760825634003 -0.248592108488083
0.354762673377991 -0.0381947308778763
0.355374157428741 -0.269818007946014
0.356362223625183 -0.226970672607422
0.357526659965515 -0.105373851954937
0.363303601741791 -0.204733535647392
0.364983260631561 -0.284574270248413
0.365545153617859 -0.135791718959808
0.366953015327454 -0.392126113176346
0.367371499538422 -0.34565669298172
0.368208110332489 -0.262652337551117
0.369657099246979 -0.250547468662262
0.371260583400726 -0.367365181446075
0.372505366802216 -0.120810389518738
0.37386828660965 -0.346094161272049
0.374388098716736 -0.178918182849884
0.374439537525177 -0.197451815009117
0.374542474746704 -0.283219486474991
0.375138998031616 -0.27657812833786
0.375311434268951 -0.197310909628868
0.375629425048828 -0.109276682138443
0.375661015510559 -0.375571072101593
0.377108514308929 -0.222846522927284
0.377328157424927 -0.445245385169983
0.380472362041473 -0.23379223048687
0.381282925605774 -0.215882197022438
0.381424009799957 -0.258029967546463
0.382105946540833 -0.313632786273956
0.382156789302826 -0.348841190338135
0.382582426071167 -0.252305239439011
0.383217632770538 -0.204087108373642
0.384503722190857 -0.289914548397064
0.386863827705383 -0.240544587373734
0.387317836284637 -0.288033753633499
0.387437045574188 -0.297588527202606
0.389028489589691 -0.0451639294624329
0.389258861541748 -0.291029393672943
0.393831789493561 -0.291773229837418
0.394003212451935 -0.475390434265137
0.394229292869568 -0.218855023384094
0.396678149700165 -0.281090825796127
0.39722090959549 -0.136794120073318
0.397243618965149 -0.144984632730484
0.397262930870056 -0.424379765987396
0.399177193641663 -0.677847862243652
0.40240353345871 -0.428837478160858
0.403810858726501 -0.231393843889236
0.404462695121765 -0.247391924262047
0.405092179775238 -0.0834430307149887
0.406755447387695 -0.331745624542236
0.412612557411194 -0.164285972714424
0.414450347423553 -0.368451952934265
0.416091740131378 -0.588584542274475
0.416590392589569 -0.277958005666733
0.418941915035248 -0.416150271892548
0.419247984886169 -0.260497719049454
0.422137677669525 -0.227260410785675
0.42242419719696 -0.285558074712753
0.422842264175415 -0.316706478595734
0.424679577350616 -0.266011655330658
0.425080358982086 -0.398766577243805
0.42706024646759 -0.285598874092102
0.427446305751801 -0.266830712556839
0.42834085226059 -0.380875676870346
0.428719878196716 -0.291363775730133
0.428893864154816 -0.29912456870079
0.431493818759918 -0.303588777780533
0.432411730289459 -0.521839797496796
0.432644844055176 -0.370252758264542
0.435224235057831 -0.15094818174839
0.435293734073639 -0.226388797163963
0.437908709049225 -0.496109902858734
0.437966644763947 -0.149860188364983
0.438612639904022 -0.406605124473572
0.439629256725311 -0.263494938611984
0.43971848487854 -0.323885589838028
0.440348088741302 -0.216986358165741
0.440488994121552 -0.160367757081985
0.441463232040405 -0.298786789178848
0.442157983779907 -0.384137749671936
0.442176282405853 -0.425224840641022
0.44442743062973 -0.495893567800522
0.445964097976685 -0.347939372062683
0.446509420871735 -0.433631598949432
0.446607768535614 -0.184368580579758
0.447093605995178 -0.128192543983459
0.447213351726532 -0.187120839953423
0.448138058185577 -0.180780917406082
0.448326110839844 -0.264333933591843
0.450220882892609 -0.12701079249382
0.450871527194977 -0.20220322906971
0.453795492649078 -0.351696848869324
0.458356499671936 -0.252881765365601
0.458406388759613 -0.39438670873642
0.46025824546814 -0.2695152759552
0.461087346076965 -0.161252558231354
0.461532354354858 -0.227546840906143
0.462877750396729 -0.330843031406403
0.463284015655518 -0.198911681771278
0.463893294334412 -0.29594612121582
0.46615594625473 -0.0180779099464417
0.466971039772034 -0.1555045992136
0.469623267650604 0.162628918886185
0.469737350940704 -0.331307590007782
0.470275282859802 -0.250627726316452
0.471850275993347 -0.329770386219025
0.474522054195404 -0.414229035377502
0.475365936756134 -0.130342155694962
0.476164937019348 -0.230517864227295
0.477377533912659 -0.168521165847778
0.477473437786102 -0.00731219351291656
0.479865312576294 -0.293932020664215
0.479888677597046 0.0285922735929489
0.480873584747314 -0.133400067687035
0.482982635498047 -0.167650073766708
0.484197854995728 -0.218803152441978
0.484357059001923 -0.0449100732803345
0.485254645347595 -0.153078496456146
0.48653519153595 -0.352449089288712
0.48817253112793 -0.201576337218285
0.489183247089386 -0.0167380571365356
0.490688264369965 -0.342879235744476
0.491963922977448 -0.238293513655663
0.493155598640442 -0.294871062040329
0.494678974151611 -0.216745615005493
0.494960606098175 0.0803244411945343
0.495229601860046 -0.133359909057617
0.495906829833984 -0.138498038053513
0.496153295040131 -0.0481662079691887
0.496901333332062 -0.290998548269272
0.497200965881348 -0.21967725455761
0.49739009141922 -0.116724565625191
0.498344659805298 -0.0504092127084732
0.498415529727936 -0.0889063328504562
0.499149441719055 -0.157191157341003
0.499790787696838 0.155073463916779
0.500741720199585 -0.111810997128487
0.502298712730408 -0.00822284817695618
0.504996478557587 -0.15020577609539
0.506202757358551 -0.193476438522339
0.50667804479599 -0.345737606287003
0.506702840328217 -0.222432494163513
0.509410440921783 -0.283090204000473
0.509415090084076 -0.0720508843660355
0.509686648845673 -0.182328253984451
0.511009037494659 -0.31615674495697
0.514650762081146 -0.228933215141296
0.516343891620636 -0.0907626077532768
0.517160415649414 -0.119353346526623
0.517968893051147 -0.145353943109512
0.518354296684265 -0.123282574117184
0.51879870891571 -0.163225397467613
0.519201934337616 0.0763005763292313
0.520697355270386 -0.189918577671051
0.521209001541138 -0.201482385396957
0.521258592605591 -0.101124480366707
0.521602511405945 -0.325580775737762
0.523371398448944 -0.240227594971657
0.524607539176941 -0.0881249010562897
0.525187015533447 -0.294967293739319
0.526157259941101 -0.192589640617371
0.526889383792877 -0.251821607351303
0.529257893562317 -0.0667702704668045
0.529807806015015 0.0202995836734772
0.530944168567657 -0.0464619919657707
0.531639695167542 -0.109113104641438
0.532077312469482 -0.177248269319534
0.532392978668213 -0.111625209450722
0.532628118991852 -0.0947629734873772
0.533356964588165 -0.171077162027359
0.53362649679184 -0.250055849552155
0.533629834651947 -0.178279891610146
0.53476345539093 -0.0842978432774544
0.534786462783813 -0.273721098899841
0.536394357681274 -0.25201228260994
0.536470293998718 -0.232944324612617
0.536643743515015 -0.0155675560235977
0.538148045539856 -0.0467406511306763
0.538256108760834 0.050380751490593
0.540016233921051 -0.105572670698166
0.541216015815735 -0.177525073289871
0.542743921279907 -0.0703559070825577
0.543114364147186 -0.13125641644001
0.543142199516296 -0.00638057291507721
0.543426752090454 -0.307046413421631
0.543698906898499 -0.176927447319031
0.544632732868195 -0.108179613947868
0.54466700553894 0.0168361812829971
0.547326803207397 -0.10585480183363
0.547641754150391 -0.180253639817238
0.54777866601944 0.0119324922561646
0.548544228076935 -0.2977614402771
0.548582911491394 -0.183018490672112
0.548852622509003 -0.370148748159409
0.549278318881989 -0.184826910495758
0.549941778182983 -0.162673622369766
0.554930150508881 -0.0993743613362312
0.554966509342194 -0.302590608596802
0.555148482322693 -0.0760888978838921
0.555404841899872 -0.184509083628654
0.559185385704041 -0.132383778691292
0.560130655765533 -0.231901884078979
0.561016023159027 -0.035066619515419
0.561105072498322 -0.17276793718338
0.563429474830627 -0.264358103275299
0.564282476902008 -0.176384478807449
0.56655752658844 -0.183234080672264
0.566916167736053 -0.222217425704002
0.567676544189453 -0.100605145096779
0.568000495433807 -0.270798891782761
0.56843900680542 -0.166410654783249
0.568960785865784 -0.302012860774994
0.569160401821136 -0.207553505897522
0.569237589836121 -0.258498549461365
0.569598853588104 -0.0964314267039299
0.571213722229004 -0.278140872716904
0.571384489536285 -0.156769886612892
0.573099136352539 -0.310493528842926
0.573448717594147 0.0464151203632355
0.57471626996994 -0.118355348706245
0.577921867370605 -0.136493965983391
0.578072309494019 -0.119272619485855
0.579895615577698 -0.113386929035187
0.580683708190918 -0.191003113985062
0.581475377082825 -0.297353029251099
0.582679569721222 -0.156928867101669
0.583037137985229 -0.11184187233448
0.583255290985107 0.102424547076225
0.583369255065918 -0.366889238357544
0.584141194820404 -0.13872392475605
0.58447939157486 -0.288515329360962
0.585948944091797 -0.00903694331645966
0.58686751127243 -0.184593364596367
0.58840024471283 -0.186203062534332
0.589263260364532 -0.113924361765385
0.589611887931824 -0.0914716124534607
0.589694857597351 -0.283148348331451
0.592982530593872 -0.170163571834564
0.593685209751129 -0.143703415989876
0.5938840508461 -0.0360805317759514
0.596662759780884 -0.217671260237694
0.59682697057724 -0.0944903492927551
0.600017130374908 -0.136647135019302
0.600483894348145 0.184720993041992
0.600576639175415 -0.122703105211258
0.602540016174316 -0.136929288506508
0.603696584701538 -0.152684092521667
0.604700326919556 -0.237905770540237
0.605463862419128 0.205703034996986
0.606118679046631 0.00195527821779251
0.607135415077209 0.147855684161186
0.608395576477051 0.0582891851663589
0.609287798404694 0.0549431145191193
0.609342515468597 0.122639864683151
0.611238121986389 0.157241195440292
0.611392557621002 -0.219704434275627
0.611448109149933 -0.0182504504919052
0.611702919006348 -0.0769062936306
0.614982903003693 0.0912832841277122
0.615996778011322 0.154820665717125
0.617995798587799 0.0696122348308563
0.618065893650055 -0.107528582215309
0.620265007019043 0.0372903123497963
0.621298968791962 -0.0265187900513411
0.622179090976715 0.0189879015088081
0.623098850250244 0.0677294954657555
0.623728811740875 -0.00571704842150211
0.624083995819092 0.205474749207497
0.627203106880188 -0.00824000500142574
0.627209544181824 0.039284884929657
0.627270817756653 0.10959792137146
0.628006339073181 0.00885689444839954
0.628556311130524 0.0682473033666611
0.629070341587067 -0.0336415395140648
0.629483878612518 -0.0786045864224434
0.629497230052948 -0.0756427124142647
0.630395472049713 0.153358101844788
0.631299912929535 0.222693055868149
0.632957756519318 -0.0865523591637611
0.635052025318146 -0.0542895048856735
0.635848581790924 -0.183732435107231
0.637620329856873 0.195750743150711
0.638547420501709 -0.111425451934338
0.639510869979858 -0.00145541131496429
0.642798244953156 0.090482085943222
0.643111646175385 0.0397664122283459
0.643257975578308 0.0295111909508705
0.646967947483063 0.181920409202576
0.647090315818787 0.229876846075058
0.649307310581207 0.242646843194962
0.650205373764038 0.0825276076793671
0.651152014732361 -0.02624287083745
0.653026938438416 -0.0324180535972118
0.657317280769348 0.0508731342852116
0.657345592975616 -0.00595012679696083
0.658673107624054 0.0586850456893444
0.65925794839859 -0.0491213984787464
0.661451399326324 -0.0791514664888382
0.661636829376221 0.180148929357529
0.662343502044678 0.0985780730843544
0.662406742572784 0.09925577044487
0.663185000419617 0.0417996644973755
0.663884341716766 -0.104787349700928
0.664242327213287 0.0847544819116592
0.664372324943542 -0.261060923337936
0.664803683757782 0.00360344210639596
0.66552859544754 0.0358174368739128
0.667357385158539 0.0365262851119041
0.667377233505249 0.0631132647395134
0.668696761131287 0.0557298734784126
0.669117629528046 -0.0861928761005402
0.669268250465393 -0.181439787149429
0.670652091503143 -0.0753918215632439
0.672045588493347 0.0484357848763466
0.672606647014618 -0.0720972567796707
0.674961984157562 -0.0140908919274807
0.676169097423553 -0.0492563545703888
0.676279962062836 0.0738588497042656
0.677945077419281 0.0948577374219894
0.678681492805481 0.137128382921219
0.678841769695282 0.105314128100872
0.681132376194 0.168840900063515
0.682716488838196 -0.160810381174088
0.685561001300812 0.185295298695564
0.686879634857178 0.0122874155640602
0.687134504318237 0.276968210935593
0.687504172325134 0.0273356586694717
0.687834203243256 -0.0496908947825432
0.690232336521149 -0.0383113026618958
0.690828025341034 -0.187220066785812
0.691793441772461 0.0910508334636688
0.692329525947571 -0.0323925390839577
0.692811071872711 0.154967576265335
0.692927718162537 -0.0261839143931866
0.693780183792114 0.106804735958576
0.695235252380371 0.075992576777935
0.69555139541626 -0.126782983541489
0.695965886116028 -0.109904885292053
0.69712907075882 -0.107744909822941
0.701892077922821 0.0692362040281296
0.7022305727005 -0.0851027965545654
0.703263282775879 -0.104214020073414
0.703818619251251 -0.154820263385773
0.704034626483917 0.10168644040823
0.704550981521606 -0.232203900814056
0.704586803913116 -0.191085427999496
0.704950451850891 -0.0238640904426575
0.705589234828949 -0.10121151059866
0.705763638019562 -0.0975021570920944
0.706259191036224 -0.12290009111166
0.707353711128235 -0.14692510664463
0.70835143327713 -0.083916649222374
0.709781348705292 -0.113785944879055
0.709917783737183 -0.212213188409805
0.712600648403168 -0.290630161762238
0.712848484516144 -0.194073304533958
0.713130414485931 -0.0798391252756119
0.714075028896332 -0.149329423904419
0.714850604534149 -0.301685154438019
0.71502959728241 -0.160780116915703
0.715675234794617 -0.0708584114909172
0.715733408927917 -0.0918483734130859
0.716383993625641 -0.217317894101143
0.716779172420502 -0.0928714126348495
0.716839253902435 -0.0868164449930191
0.71817946434021 -0.259514182806015
0.718269169330597 -0.218137815594673
0.719146311283112 -0.039598673582077
0.719301402568817 -0.160358816385269
0.719663083553314 -0.0820512101054192
0.719783961772919 0.0121884644031525
0.721017837524414 -0.184132978320122
0.722213566303253 -0.247120663523674
0.723699867725372 0.132501721382141
0.724256753921509 0.106436029076576
0.725065112113953 -0.0789830982685089
0.72655326128006 -0.324760913848877
0.726656198501587 -0.234011709690094
0.72819447517395 0.0518103986978531
0.72912186384201 -0.353101342916489
0.72965133190155 -0.0706338733434677
0.731324255466461 -0.227524757385254
0.731336116790771 -0.231311067938805
0.732370853424072 0.0209803953766823
0.734109044075012 0.11405374109745
0.734839618206024 -0.0211465135216713
0.737467050552368 -0.0198265500366688
0.737612009048462 -0.240652829408646
0.739625871181488 -0.315923362970352
0.740008890628815 -0.0896397829055786
0.740334093570709 0.0422227568924427
0.741185188293457 -0.0417848378419876
0.742149710655212 0.0636891350150108
0.742704451084137 -0.113709062337875
0.744124948978424 0.0630278140306473
0.745761215686798 -0.024590402841568
0.746435046195984 -0.0216419659554958
0.746834218502045 0.31927964091301
0.751363754272461 0.190749675035477
0.751953959465027 0.083337813615799
0.753550171852112 0.0726500004529953
0.755561828613281 0.0486574545502663
0.755683422088623 0.0645854473114014
0.755926251411438 0.0768747553229332
0.755959630012512 0.0134584493935108
0.756112992763519 -0.00951257348060608
0.757652580738068 -0.0912797898054123
0.757949411869049 0.134255215525627
0.758042633533478 0.0855315700173378
0.759805619716644 -0.0375943332910538
0.760032474994659 -0.0171690285205841
0.762265145778656 -0.117565371096134
0.763808965682983 -0.115045815706253
0.765364766120911 0.0950833857059479
0.765566885471344 0.0681227818131447
0.765744864940643 0.0210709162056446
0.767688155174255 -0.054040402173996
0.767880141735077 0.0757198706269264
0.769467651844025 0.076357789337635
0.769901216030121 0.274438261985779
0.77111941576004 -0.0238509029150009
0.771126091480255 0.00381463766098022
0.77229380607605 0.26980447769165
0.772660255432129 -0.0546405389904976
0.773585915565491 0.114869818091393
0.774224281311035 0.114133387804031
0.774264454841614 -0.0210211500525475
0.775710463523865 0.283788412809372
0.776117324829102 0.0123801454901695
0.776192009449005 -0.0476378723978996
0.776491224765778 0.380621165037155
0.776552259922028 0.210329845547676
0.776834726333618 0.00307083129882812
0.776989758014679 0.200451880693436
0.777871131896973 0.169579029083252
0.777998864650726 -0.0698502659797668
0.77806943655014 0.188337028026581
0.781219780445099 -0.0925766304135323
0.781424224376678 -0.0932901576161385
0.782429814338684 0.11739207804203
0.782959222793579 0.186055839061737
0.783080101013184 0.0899242460727692
0.78570967912674 0.0355177149176598
0.786341190338135 -0.123487018048763
0.788268983364105 -0.148769170045853
0.788321852684021 0.0113519001752138
0.788463056087494 0.0707869529724121
0.788946866989136 0.0300989244133234
0.790223956108093 0.133556559681892
0.791987597942352 0.0648794770240784
0.79369592666626 -8.46367329359055e-05
0.796709895133972 -0.075419194996357
0.797364771366119 -0.221100836992264
0.798034131526947 -0.0341412648558617
0.798335015773773 -0.0764971822500229
0.800311326980591 -0.116513691842556
0.800660908222198 -0.133436962962151
0.801533937454224 -0.310227334499359
0.801606059074402 -0.0167274922132492
0.802084028720856 0.13478410243988
0.802627801895142 0.00198228657245636
0.803198575973511 -0.181006163358688
0.80336332321167 -0.122961893677711
0.806694090366364 -0.306732892990112
0.807184934616089 -0.117490217089653
0.807325661182404 -0.046871542930603
0.808991968631744 -0.164276003837585
0.809589385986328 -0.0541869252920151
0.813106060028076 -0.0986080318689346
0.814676403999329 -0.378612816333771
0.815713942050934 -0.214111134409904
0.816798567771912 -0.275957345962524
0.819327414035797 -0.0412418395280838
0.821330547332764 -0.218488052487373
0.8223996758461 -0.151774778962135
0.823560476303101 0.0222169309854507
0.823899984359741 -0.162841156125069
0.824554443359375 -0.213273823261261
0.825075566768646 -0.109764605760574
0.826074361801147 -0.500460267066956
0.826352417469025 -0.0572832897305489
0.828324615955353 -0.194813251495361
0.829666078090668 -0.238209515810013
0.832796812057495 -0.0618775188922882
0.832877457141876 -0.239016473293304
0.833528101444244 -0.297483861446381
0.833710789680481 -0.375449538230896
0.836322605609894 -0.344379246234894
0.836503624916077 -0.0730126798152924
0.837686419487 -0.115500286221504
0.83800083398819 -0.238486886024475
0.839037954807281 -0.282514631748199
0.839476823806763 -0.129085302352905
0.840176641941071 -0.310683786869049
0.847348690032959 -0.271724700927734
0.848069906234741 -0.0260626822710037
0.848271071910858 -0.266024589538574
0.849345624446869 -0.0512340664863586
0.850594818592072 -0.327375918626785
0.85144704580307 -0.147992372512817
0.853228747844696 -0.226195752620697
0.853862345218658 -0.0656448602676392
0.855116248130798 -0.0956370979547501
0.855592966079712 -0.189602732658386
0.856595098972321 -0.0940464362502098
0.857012689113617 -0.294825732707977
0.863369882106781 -0.17867873609066
0.863778531551361 -0.199362084269524
0.868441045284271 -0.205731987953186
0.871688902378082 -0.440330445766449
0.873691558837891 -0.319149732589722
0.874708652496338 -0.184775084257126
0.875350475311279 -0.354873955249786
0.875974535942078 -0.291502714157104
0.876228392124176 -0.29601326584816
0.879694223403931 -0.329485237598419
0.880945801734924 -0.15063413977623
0.883417129516602 -0.287787735462189
0.885231196880341 -0.245709806680679
0.886090874671936 -0.354403972625732
0.886336266994476 -0.132141694426537
0.886885762214661 -0.138755410909653
0.887837469577789 -0.24248531460762
0.887969732284546 -0.090302512049675
0.888222336769104 -0.211629316210747
0.889349222183228 -0.485541760921478
0.893344879150391 -0.321502327919006
0.893395721912384 -0.31648114323616
0.894942939281464 -0.208423227071762
0.895601570606232 -0.426013827323914
0.896244764328003 -0.280782669782639
0.897523283958435 -0.178609400987625
0.898642182350159 -0.0395382195711136
0.898830592632294 -0.302259296178818
0.899360179901123 -0.236781358718872
0.902750432491302 -0.218829780817032
0.904362678527832 0.0396959185600281
0.908016383647919 -0.337041676044464
0.908308207988739 -0.024037778377533
0.90897011756897 -0.0986391752958298
0.909333288669586 -0.0786584913730621
0.909805357456207 -0.0282237380743027
0.910936176776886 -0.179400950670242
0.911090195178986 -0.139744758605957
0.912229299545288 -0.0557987242937088
0.914561867713928 -0.189098626375198
0.914898157119751 -0.219884842634201
0.915383756160736 -0.171259626746178
0.915829658508301 -0.126068949699402
0.916840314865112 -0.0754516199231148
0.916842460632324 -0.188357576727867
0.917477786540985 -0.446573734283447
0.917917847633362 -0.0780095905065536
0.92144900560379 -0.155238419771194
0.921781778335571 -0.119695611298084
0.922143697738647 -0.160021871328354
0.922898650169373 -0.285574823617935
0.92338627576828 -0.21330252289772
0.924791276454926 -0.241086453199387
0.925277292728424 -0.220423266291618
0.926160454750061 -0.245824977755547
0.926836133003235 0.00881460309028625
0.927080690860748 -0.379820168018341
0.927291035652161 -0.251408487558365
0.92767333984375 -0.198533162474632
0.927787601947784 -0.152717113494873
0.929244995117188 0.00788013637065887
0.930286288261414 -0.111325711011887
0.93401426076889 -0.10343673825264
0.934347629547119 -0.144584074616432
0.934767425060272 -0.104622676968575
0.935693144798279 -0.0267810113728046
0.935742199420929 -0.23298454284668
0.937634229660034 0.150924146175385
0.938472211360931 -0.106974333524704
0.939562797546387 -0.0710161030292511
0.940811038017273 -0.0904157981276512
0.941045939922333 -0.255653470754623
0.94183760881424 -0.261149704456329
0.941950440406799 0.0897286310791969
0.942649602890015 -0.311492919921875
0.942933022975922 -0.0161594301462173
0.943146765232086 -0.204472780227661
0.943453907966614 0.111866362392902
0.945283770561218 -0.0743495002388954
0.946651816368103 -0.135404691100121
0.94723105430603 -0.000476956367492676
0.947386205196381 -0.225117087364197
0.947484135627747 -0.181882411241531
0.947816014289856 -0.0469265207648277
0.948593974113464 -0.0858260989189148
0.9515380859375 0.204644381999969
0.953351318836212 -0.000621102750301361
0.956686198711395 -0.0206398293375969
0.957379043102264 0.262155652046204
0.958347737789154 0.0977236703038216
0.959111511707306 0.182937681674957
0.959280729293823 0.112531997263432
0.959829747676849 0.245938450098038
0.96056067943573 -0.0613225474953651
0.966829538345337 0.0534843802452087
0.967267334461212 -0.0219967663288116
0.967601299285889 -0.219368100166321
0.969941318035126 0.0522764511406422
0.971402049064636 -0.0211997739970684
0.975026607513428 -0.0801417678594589
0.975906431674957 0.0554131269454956
0.976352035999298 -0.00354718416929245
0.976420760154724 -0.0219154823571444
0.977560937404633 -0.183712124824524
0.977567851543427 -0.0305781997740269
0.977963209152222 -0.111693032085896
0.977972269058228 -0.225083142518997
0.978969037532806 -0.104195579886436
0.97898805141449 -0.22681200504303
0.979085683822632 -0.123478084802628
0.979259133338928 -0.108102425932884
0.979533851146698 -0.172923505306244
0.981106638908386 -0.320946455001831
0.982022225856781 -0.218985915184021
0.982273995876312 -0.364784747362137
0.983185708522797 -0.270714789628983
0.983955025672913 -0.33967649936676
0.990147650241852 -0.309531569480896
0.990295886993408 -0.246107131242752
0.99135559797287 -0.108024545013905
0.991639852523804 -0.203696966171265
0.993355631828308 0.045138992369175
0.994155883789062 -0.169834703207016
0.997614085674286 -0.200280711054802
};
\addplot [draw=blue, fill=blue, mark=*, only marks, opacity=0.5]
table{%
x  y
0.246447443962097 -0.142790824174881
0.243947565555573 -0.0825380086898804
0.261327683925629 -0.0217186771333218
0.237625062465668 -0.0160589516162872
0.239483892917633 -0.109698556363583
0.224663436412811 0.04893808811903
0.236962258815765 -0.189192593097687
0.246529042720795 -0.104491710662842
0.0911379456520081 -0.17208394408226
0.22879821062088 -0.1329295784235
0.00092846155166626 -0.202616214752197
0.235299110412598 0.10160505771637
0.239660918712616 0.00926803983747959
0.593685209751129 -0.143703415989876
0.223562777042389 -0.0370888970792294
0.240252792835236 0.014792762696743
0.246777892112732 0.131956160068512
0.239346027374268 -0.0285791102796793
0.239303231239319 -0.0553366430103779
0.244090855121613 0.0653250738978386
0.191875159740448 0.0947580188512802
0.191090524196625 -0.135234370827675
0.357526659965515 -0.105373851954937
0.997614085674286 -0.200280711054802
0.225398480892181 -0.121966660022736
0.239240050315857 -0.120778761804104
0.227014183998108 0.111658364534378
0.234163820743561 0.0606120824813843
0.209720730781555 0.127094686031342
0.241577327251434 -0.058590579777956
0.233624339103699 0.0373077839612961
};
\addlegendentry{PFN}
\addplot [draw=red, fill=red, mark=*, only marks, opacity=0.5]
table{%
x  y
0.256034851074219 0.110605597496033
0.243947565555573 -0.0825380086898804
0.215640127658844 0.0643238425254822
0.0670799016952515 -0.0711456015706062
0.0221312642097473 -0.151822254061699
0.446509420871735 -0.433631598949432
0.886885762214661 -0.138755410909653
0.682716488838196 -0.160810381174088
0.416590392589569 -0.277958005666733
0.340729355812073 -0.230804234743118
0.00092846155166626 -0.202616214752197
0.235299110412598 0.10160505771637
0.092924177646637 -0.159419432282448
0.271250069141388 0.0990232154726982
0.484197854995728 -0.218803152441978
0.064532458782196 0.267953991889954
0.016072690486908 0.0355637893080711
0.623098850250244 0.0677294954657555
0.593685209751129 -0.143703415989876
0.0683515667915344 -0.151590630412102
0.746435046195984 -0.0216419659554958
0.244090855121613 0.0653250738978386
0.807184934616089 -0.117490217089653
0.254932761192322 -0.2026347219944
0.997614085674286 -0.200280711054802
0.165421426296234 -0.0778728574514389
0.573448717594147 0.0464151203632355
0.69712907075882 -0.107744909822941
0.234163820743561 0.0606120824813843
0.836322605609894 -0.344379246234894
0.233624339103699 0.0373077839612961
};
\addlegendentry{GP}
\legend{};
\end{axis}

\end{tikzpicture}

%% file: MetaLearnWorkshop/figures/hebo_on_hebo_example_3.tex
\begin{tikzpicture}

\begin{axis}[
legend cell align={left},
legend style={
  fill opacity=0.8,
  draw opacity=1,
  text opacity=1,
  at={(0.03,0.97)},
  anchor=north west,
  draw=white!80!black
},
tick align=outside,
tick pos=left,
x grid style={white!69.0196078431373!black},
xmin=-0.0496456444263458, xmax=1.04694877266884,
xtick style={color=black},
y grid style={white!69.0196078431373!black},
ymin=-0.716917736828327, ymax=0.469170464575291,
ytick style={color=black}, width=\textwidth, height=\textwidth,
]
\addplot [semithick, white!50.1960784313725!black, forget plot]
table {%
0.000199556350708008 0.145666301250458
0.00023043155670166 0.0628557577729225
0.00027167797088623 0.191887632012367
0.00175052881240845 -0.0372040420770645
0.00517368316650391 -0.174245923757553
0.00954639911651611 -0.135704979300499
0.00989758968353271 -0.218096971511841
0.00994342565536499 -0.127345860004425
0.0113877058029175 -0.146843016147614
0.0122083425521851 -0.026267234236002
0.0140360593795776 -0.14351661503315
0.0149008631706238 -0.0437562763690948
0.0161068439483643 -0.191642999649048
0.0181476473808289 -0.318060338497162
0.019612193107605 -0.0724106356501579
0.0196447372436523 -0.126617699861526
0.0196864008903503 -0.133030533790588
0.0197702050209045 -0.209563657641411
0.0201102495193481 -0.152610331773758
0.0203125476837158 -0.0302870273590088
0.0222485661506653 -0.131944298744202
0.022491991519928 -0.183766782283783
0.0267469882965088 -0.134952455759048
0.0276458859443665 -0.010757252573967
0.0282864570617676 -0.0437518209218979
0.0304391980171204 0.0493149608373642
0.0313344597816467 0.109698623418808
0.035020649433136 0.0501433275640011
0.035807192325592 0.0636945068836212
0.0363283157348633 -0.0232269316911697
0.0367077589035034 -0.0154756307601929
0.0373607873916626 0.0478631593286991
0.0385848879814148 0.0988617986440659
0.0403876900672913 0.12606580555439
0.0404689908027649 0.172991737723351
0.0408816337585449 0.126337289810181
0.041027843952179 0.120125435292721
0.0414511561393738 0.0990162864327431
0.0430284738540649 0.155959933996201
0.043190062046051 0.0530696660280228
0.0448529124259949 0.174211144447327
0.0465652346611023 0.0677418410778046
0.0468344688415527 -0.0572954192757607
0.0497320294380188 -0.105548143386841
0.0507627725601196 -0.101678684353828
0.0508604049682617 -0.150600463151932
0.0527128577232361 -0.0369237959384918
0.0561195015907288 -0.118777059018612
0.0598086714744568 0.0697462931275368
0.0598151087760925 -0.133721381425858
0.0615026950836182 -0.0350483767688274
0.0623730421066284 -0.0129133984446526
0.0634879469871521 -0.115062236785889
0.0641909241676331 0.0584390461444855
0.0656154751777649 -0.0332506150007248
0.0657403469085693 -0.00999992154538631
0.066061794757843 -0.0644572079181671
0.0680695772171021 -0.00604055821895599
0.0693674087524414 0.220157593488693
0.0727055072784424 0.175936669111252
0.0738995671272278 0.0811002478003502
0.0759580731391907 0.0552171766757965
0.0761812329292297 0.161093324422836
0.0765529274940491 0.148418784141541
0.0777890086174011 0.17903670668602
0.0786131620407104 0.165024027228355
0.0792798399925232 0.0738420933485031
0.0802623629570007 0.19438698887825
0.0803787708282471 0.237890750169754
0.0808637738227844 0.125722602009773
0.0811832547187805 0.0631096512079239
0.0814422965049744 0.0359803587198257
0.0828582048416138 0.088337279856205
0.0831218361854553 0.182579278945923
0.0837119221687317 0.0489066690206528
0.0844428539276123 0.0287363901734352
0.0848714709281921 0.0385721549391747
0.0865344405174255 -0.00598935410380363
0.0877798795700073 -0.0278609171509743
0.0887836217880249 -0.142925947904587
0.0890009999275208 0.0894091576337814
0.0893362164497375 0.0913800895214081
0.0904621481895447 0.0178713649511337
0.0924857258796692 0.0779731422662735
0.092795729637146 0.0106068029999733
0.0930639505386353 0.0812702178955078
0.0932570695877075 0.00492999702692032
0.0943295359611511 0.145507663488388
0.0950967669487 -0.0285648033022881
0.0951609015464783 0.110098019242287
0.098792552947998 0.027407955378294
0.09971022605896 -0.0166661441326141
0.0998502373695374 0.0444637648761272
0.100142359733582 -0.015014722943306
0.100846529006958 -0.0372994504868984
0.10098659992218 0.0362991541624069
0.101592063903809 -0.14774888753891
0.101963102817535 0.0107802804559469
0.102078914642334 -0.0728696137666702
0.102587699890137 -0.0157484151422977
0.103609085083008 -0.0709621533751488
0.103995323181152 0.0354047678411007
0.105622112751007 0.0209528990089893
0.105710566043854 0.0484153144061565
0.10574072599411 0.0748045518994331
0.105981290340424 -0.042408999055624
0.106539785861969 0.00891519617289305
0.107134997844696 0.106967553496361
0.109054505825043 0.0765919387340546
0.109800934791565 0.0336511917412281
0.111052215099335 0.0285616889595985
0.111548900604248 0.00718408450484276
0.112822949886322 0.0205075796693563
0.114022612571716 -0.00856020674109459
0.114789724349976 -0.101312845945358
0.117478728294373 -0.11617186665535
0.119935989379883 -0.0835607796907425
0.123427033424377 -0.187452971935272
0.124620497226715 -0.084995336830616
0.124975621700287 -0.172874793410301
0.125634014606476 -0.188828974962234
0.125696241855621 -0.169796466827393
0.126127898693085 -0.112152270972729
0.126274764537811 -0.067999467253685
0.1266028881073 -0.0669586658477783
0.126658260822296 -0.144606053829193
0.126995205879211 -0.0754901617765427
0.128673195838928 -0.0762945711612701
0.128796279430389 -0.0696187913417816
0.12884521484375 -0.0808150991797447
0.129053711891174 -0.0903413593769073
0.131194770336151 -0.111958265304565
0.132387459278107 -0.0867666602134705
0.134108126163483 0.0335772410035133
0.1350958943367 0.0599241405725479
0.138791382312775 -0.0719892680644989
0.139918327331543 0.0687835514545441
0.140721380710602 0.0139329079538584
0.140917181968689 0.0168761555105448
0.141993343830109 0.0344672687351704
0.14440780878067 0.0674231573939323
0.144574284553528 0.10807041823864
0.145911633968353 0.188986986875534
0.147887349128723 0.194910734891891
0.149474680423737 0.0624717958271503
0.150143504142761 0.219835862517357
0.151575863361359 0.0823746472597122
0.151757061481476 0.14762108027935
0.152442276477814 0.174045860767365
0.155065596103668 0.053071454167366
0.155396640300751 0.0674473196268082
0.156430721282959 0.181694179773331
0.157435953617096 -0.0186610966920853
0.159417688846588 0.0649052411317825
0.161822915077209 0.154145359992981
0.162460565567017 0.0695557445287704
0.163102865219116 -0.0280029252171516
0.163825929164886 0.107840076088905
0.165288686752319 0.0406824611127377
0.166110813617706 0.0155406072735786
0.167189240455627 0.110630020499229
0.167263984680176 0.0422388054430485
0.167371451854706 0.0064469650387764
0.168284058570862 0.0670295655727386
0.170334994792938 0.0332461446523666
0.170415878295898 0.0673829317092896
0.171342611312866 0.045255471020937
0.172349333763123 0.0884244441986084
0.172621428966522 0.275365650653839
0.174542188644409 0.0556398183107376
0.174935936927795 0.0796668529510498
0.174939334392548 0.104049168527126
0.175145447254181 0.197625190019608
0.175598561763763 0.0211717486381531
0.176317453384399 0.116683974862099
0.176364302635193 0.0767549872398376
0.177976310253143 0.108689606189728
0.17897367477417 0.0615320801734924
0.179502189159393 -0.0194078683853149
0.179844081401825 0.116580538451672
0.181054830551147 0.0838816612958908
0.181183099746704 0.06023108959198
0.182633817195892 0.164334744215012
0.183551669120789 0.0800414085388184
0.183814644813538 0.0625999122858047
0.184000909328461 0.0824480280280113
0.184984862804413 0.123309955000877
0.184999704360962 0.111242666840553
0.185442328453064 0.0379232950508595
0.186193346977234 0.101287923753262
0.186799645423889 0.122859492897987
0.187065184116364 0.125193476676941
0.187268555164337 0.0474456623196602
0.187768518924713 0.146390244364738
0.189625680446625 0.0470676496624947
0.191118061542511 0.0539610534906387
0.194252967834473 0.0534028708934784
0.196675658226013 0.106492042541504
0.197432160377502 0.10979925096035
0.198406338691711 0.0018073245882988
0.198634743690491 0.189471215009689
0.199058294296265 0.0599605552852154
0.199577033519745 0.124654017388821
0.202345907688141 0.29265969991684
0.204398036003113 0.172628581523895
0.206859290599823 0.192682549357414
0.207098484039307 0.151407584547997
0.207242071628571 0.286000460386276
0.208481967449188 0.248721644282341
0.208665430545807 0.27168396115303
0.209347784519196 0.290922284126282
0.209558308124542 0.218788921833038
0.211481034755707 0.249547839164734
0.211761236190796 0.26262429356575
0.21306985616684 0.273029297590256
0.213146388530731 0.27036851644516
0.21455329656601 0.28043857216835
0.214790940284729 0.267421126365662
0.214899718761444 0.135813817381859
0.215690195560455 0.119044192135334
0.216829597949982 0.259331822395325
0.216872930526733 0.221147164702415
0.220463991165161 0.163063034415245
0.220773160457611 0.171373620629311
0.221756100654602 0.137951090931892
0.22200870513916 0.2250707000494
0.22229528427124 0.151753753423691
0.223166584968567 0.111542269587517
0.224019229412079 0.146554231643677
0.226947367191315 0.218028530478477
0.228215038776398 0.29000797867775
0.228481352329254 0.340885519981384
0.229117155075073 0.218820869922638
0.229390799999237 0.242621824145317
0.229943096637726 0.296884208917618
0.230271935462952 0.364935785531998
0.231298983097076 0.230329394340515
0.231452941894531 0.213828533887863
0.231509983539581 0.303839892148972
0.231761932373047 0.235466927289963
0.232509851455688 0.283288866281509
0.232933282852173 0.396972179412842
0.234853088855743 0.313709855079651
0.234998226165771 0.300308108329773
0.235889494419098 0.320393115282059
0.238475859165192 0.336776494979858
0.239427804946899 0.28470504283905
0.241911709308624 0.324203222990036
0.243977844715118 0.307858884334564
0.244791269302368 0.311605632305145
0.245544493198395 0.257244855165482
0.245714426040649 0.41525736451149
0.245885014533997 0.254983484745026
0.247446298599243 0.363400220870972
0.247704148292542 0.390428930521011
0.249111473560333 0.319875866174698
0.252592206001282 0.143908530473709
0.252834379673004 0.20366083085537
0.252939641475677 0.206685364246368
0.253772675991058 0.299360901117325
0.254187405109406 0.206943392753601
0.254200339317322 0.111794978380203
0.254565894603729 0.193559512495995
0.25550651550293 0.160746291279793
0.255708873271942 0.169465750455856
0.255944907665253 0.264400631189346
0.256943464279175 0.14062212407589
0.262296378612518 0.104703985154629
0.262506067752838 0.0510039925575256
0.26271641254425 0.0488371551036835
0.263327658176422 0.137952029705048
0.264706134796143 0.0682390406727791
0.266517758369446 0.170819595456123
0.267735123634338 0.0163552202284336
0.269775807857513 0.0813116580247879
0.271143078804016 0.0576582849025726
0.271331548690796 0.1385328322649
0.273321449756622 0.0521742664277554
0.276074409484863 -0.00156304612755775
0.276277363300323 0.0485577583312988
0.276349902153015 0.161948919296265
0.276574969291687 0.0557923465967178
0.276872992515564 0.0771120339632034
0.277681231498718 0.115360602736473
0.277978718280792 -0.0170736983418465
0.278561413288116 -0.10356879234314
0.279308795928955 -0.0822730138897896
0.283709168434143 -0.0809214562177658
0.283984899520874 -0.0100894197821617
0.285681307315826 -0.0668140053749084
0.286686658859253 -0.055385634303093
0.286785244941711 -0.0576142147183418
0.287016153335571 -0.0812441557645798
0.287048816680908 -0.0320978201925755
0.288021266460419 -0.0141908386722207
0.289716184139252 0.114670664072037
0.290133416652679 -0.0471586100757122
0.293999433517456 -0.0791354104876518
0.294180750846863 -0.070175975561142
0.294566810131073 -0.0365830920636654
0.295030534267426 -0.0248157698661089
0.296118080615997 0.0338578522205353
0.296538591384888 -0.0971635952591896
0.297735750675201 -0.00234707444906235
0.298365414142609 -0.000521935522556305
0.298834979534149 0.0286030676215887
0.299490511417389 -0.048931397497654
0.299543142318726 -0.0487390160560608
0.299838423728943 0.133652329444885
0.300996601581573 0.0269393175840378
0.301259636878967 0.083875298500061
0.302822411060333 -0.0327374339103699
0.30335545539856 0.117802575230598
0.305219054222107 0.0386386252939701
0.305256485939026 0.0977231115102768
0.307631134986877 0.0867974460124969
0.3083735704422 0.0940080136060715
0.310537755489349 0.274604499340057
0.31118631362915 0.191821306943893
0.312048852443695 0.141502857208252
0.313268542289734 0.0153904110193253
0.313356459140778 0.0765659809112549
0.314294636249542 0.0544808059930801
0.319569408893585 -0.00293812900781631
0.320163011550903 0.0488894060254097
0.320546209812164 0.17867973446846
0.322108328342438 0.0896576121449471
0.322207987308502 0.0804629847407341
0.323419630527496 0.0719103962182999
0.32566499710083 -0.0114563927054405
0.32679009437561 0.063853532075882
0.327588617801666 0.146978139877319
0.327746391296387 0.0494648665189743
0.327993333339691 0.1407790184021
0.333553910255432 -0.0647581219673157
0.334640622138977 -0.13596847653389
0.334647655487061 0.0338571704924107
0.334699690341949 -0.120630130171776
0.335496723651886 -0.0124375987797976
0.336142122745514 -0.0327495262026787
0.336305022239685 0.00826212484389544
0.336527228355408 -0.0139025310054421
0.338179290294647 0.000326947309076786
0.338498413562775 -0.090066097676754
0.341498553752899 -0.103338293731213
0.343162357807159 0.110454797744751
0.343607485294342 -0.00826353766024113
0.347407042980194 -0.0626059025526047
0.347542703151703 0.0477307662367821
0.348264276981354 0.0530186295509338
0.349500477313995 -0.0416359454393387
0.350126802921295 -0.190076544880867
0.351660668849945 -0.0509893149137497
0.353828430175781 -0.0259224586188793
0.359227240085602 0.0556042343378067
0.359232485294342 -0.00485196895897388
0.360924780368805 -0.00769903510808945
0.363193333148956 -0.0892663970589638
0.364121675491333 -0.00307653099298477
0.365239441394806 -0.15324355661869
0.365472972393036 -0.12350907176733
0.368711650371552 -0.198932707309723
0.370233058929443 -0.106786027550697
0.370289862155914 -0.0421772971749306
0.370813131332397 -0.101872093975544
0.37096506357193 -0.166398346424103
0.373386085033417 -0.0540238432586193
0.373680949211121 0.0272978991270065
0.373772442340851 -0.12211948633194
0.374650359153748 -0.111348405480385
0.374650776386261 -0.0730794370174408
0.376131057739258 0.0630142465233803
0.376307249069214 -0.147603034973145
0.376455366611481 -0.126542121171951
0.376774311065674 -0.127510726451874
0.377071440219879 -0.109857559204102
0.377189159393311 -0.22894161939621
0.377425193786621 -0.0993595495820045
0.377487897872925 0.0347563624382019
0.378710746765137 -0.055316586047411
0.379239618778229 -0.0669040083885193
0.38233882188797 -0.0542687997221947
0.38249135017395 -0.0870746150612831
0.388106167316437 -0.185521334409714
0.388247191905975 -0.137968868017197
0.38857227563858 -0.0431132689118385
0.391678512096405 -0.144054219126701
0.394571959972382 -0.138066843152046
0.395127952098846 -0.0894727110862732
0.395386695861816 -0.103120639920235
0.400814414024353 -0.222692593932152
0.402042925357819 -0.198530241847038
0.402283310890198 -0.148742333054543
0.403690993785858 -0.215571939945221
0.405100464820862 -0.157714262604713
0.406060576438904 -0.176500916481018
0.409991025924683 -0.306372672319412
0.411767065525055 -0.233283787965775
0.414960861206055 -0.339129239320755
0.416520416736603 -0.351097613573074
0.417985916137695 -0.167772322893143
0.418614029884338 -0.134773463010788
0.420255959033966 -0.240891292691231
0.42080157995224 -0.17042264342308
0.421439051628113 -0.133069515228271
0.422002792358398 -0.304643869400024
0.424449265003204 -0.29113644361496
0.426474213600159 -0.200243279337883
0.427091360092163 -0.336537063121796
0.427609145641327 -0.181473046541214
0.428620994091034 -0.197592303156853
0.429496943950653 -0.227639928460121
0.430633723735809 -0.25702977180481
0.430745124816895 -0.272735476493835
0.431290030479431 -0.171141624450684
0.433530688285828 -0.0303056836128235
0.433632433414459 -0.143189653754234
0.434848129749298 -0.243452772498131
0.435583829879761 -0.226192325353622
0.436027765274048 -0.230851709842682
0.436466872692108 -0.147265523672104
0.436576902866364 -0.25645187497139
0.43679678440094 -0.18841364979744
0.437246918678284 -0.202439501881599
0.440787076950073 -0.245201140642166
0.443727195262909 -0.191765874624252
0.443925976753235 -0.251331061124802
0.444081604480743 -0.244555696845055
0.444275856018066 -0.189355760812759
0.44471949338913 -0.262444227933884
0.445734441280365 -0.141643241047859
0.447099208831787 -0.226862281560898
0.450183689594269 -0.292869985103607
0.451472699642181 -0.307980418205261
0.45325231552124 -0.257967084646225
0.453256011009216 -0.436047405004501
0.453407824039459 -0.303317040205002
0.454217135906219 -0.328023761510849
0.456218421459198 -0.449597716331482
0.456997156143188 -0.432309031486511
0.457209706306458 -0.292230844497681
0.457342267036438 -0.47091144323349
0.45838874578476 -0.436905533075333
0.461093485355377 -0.37678787112236
0.461234152317047 -0.420156985521317
0.463763773441315 -0.353401392698288
0.46399313211441 -0.566479325294495
0.464927494525909 -0.42665684223175
0.465263426303864 -0.397250324487686
0.467434346675873 -0.384082078933716
0.474468290805817 -0.38869497179985
0.476126670837402 -0.594505786895752
0.4781773686409 -0.561251282691956
0.478883922100067 -0.457114458084106
0.478991448879242 -0.524401307106018
0.47925728559494 -0.478487968444824
0.479732096195221 -0.663004636764526
0.481912732124329 -0.529048800468445
0.484113931655884 -0.448689728975296
0.484343647956848 -0.575249075889587
0.484353244304657 -0.510564982891083
0.48443990945816 -0.457312732934952
0.485650658607483 -0.550899565219879
0.486126720905304 -0.617772102355957
0.486138880252838 -0.57169497013092
0.486746430397034 -0.513918459415436
0.487311959266663 -0.619764864444733
0.488618433475494 -0.525831401348114
0.489480972290039 -0.503082156181335
0.491634786128998 -0.530839383602142
0.492331087589264 -0.512963473796844
0.494148671627045 -0.500677883625031
0.496833443641663 -0.480132162570953
0.499725341796875 -0.54557740688324
0.499776124954224 -0.470571488142014
0.50114905834198 -0.410610765218735
0.503112971782684 -0.422325909137726
0.505796909332275 -0.492293983697891
0.506165564060211 -0.494646906852722
0.506464183330536 -0.474936813116074
0.507280170917511 -0.45634913444519
0.508601188659668 -0.458667904138565
0.509131669998169 -0.502774894237518
0.509403765201569 -0.472439795732498
0.50993424654007 -0.411661595106125
0.510859489440918 -0.346243858337402
0.511559009552002 -0.308403670787811
0.511924684047699 -0.528565883636475
0.512395441532135 -0.434404581785202
0.514192461967468 -0.352502852678299
0.514497697353363 -0.37424635887146
0.515617728233337 -0.377183556556702
0.518471598625183 -0.328564763069153
0.518606662750244 -0.346133142709732
0.518795490264893 -0.347777009010315
0.519646108150482 -0.304009914398193
0.519679427146912 -0.352538615465164
0.519923806190491 -0.424874633550644
0.522444665431976 -0.281921744346619
0.522495329380035 -0.294201701879501
0.523162662982941 -0.279110163450241
0.523230969905853 -0.380935400724411
0.524542152881622 -0.244900494813919
0.524836003780365 -0.333473354578018
0.524854004383087 -0.382528781890869
0.525085508823395 -0.317962169647217
0.525639474391937 -0.139721676707268
0.526428461074829 -0.374136060476303
0.528150856494904 -0.21749572455883
0.528494358062744 -0.313828676939011
0.528824329376221 -0.346771031618118
0.528960168361664 -0.225872546434402
0.531051218509674 -0.169710591435432
0.531968593597412 -0.292560309171677
0.532198131084442 -0.24075736105442
0.532454073429108 -0.132141381502151
0.532510578632355 -0.272889137268066
0.533720970153809 -0.143485605716705
0.537742733955383 -0.171594962477684
0.538532495498657 -0.133192032575607
0.539005994796753 -0.0602266788482666
0.53913688659668 -0.161935150623322
0.539841055870056 -0.0975859090685844
0.540154218673706 0.0308061540126801
0.540729582309723 -0.105424299836159
0.541330456733704 -0.186933755874634
0.541451990604401 -0.127234369516373
0.544271945953369 0.031861275434494
0.545041561126709 -0.0438166446983814
0.546633839607239 -0.162623405456543
0.548749923706055 -0.122443281114101
0.550680935382843 -0.0181443151086569
0.550929367542267 -0.094449408352375
0.552525281906128 -0.0277306959033012
0.554821074008942 -0.0192739963531494
0.555033385753632 0.0120664574205875
0.555178046226501 -0.156337082386017
0.555708408355713 -0.0744329616427422
0.556606769561768 0.0379991047084332
0.5576251745224 -0.0239697061479092
0.55823427438736 -0.0789172798395157
0.558258831501007 0.0893681719899178
0.559643089771271 -0.0193947814404964
0.561315417289734 0.0118995197117329
0.561466872692108 0.0616328492760658
0.565553963184357 -0.0786903351545334
0.566228747367859 -0.00666414573788643
0.566391289234161 0.0197630450129509
0.566491067409515 0.0326383747160435
0.567344546318054 0.0509247556328773
0.568358063697815 0.178064435720444
0.569336116313934 0.0975593030452728
0.570217192173004 0.039101280272007
0.570298552513123 0.117596596479416
0.571104466915131 0.0256703197956085
0.571298956871033 0.164268210530281
0.573487937450409 0.00780980288982391
0.574067413806915 0.163333341479301
0.574203610420227 0.18489283323288
0.574207007884979 0.0442190617322922
0.575501620769501 0.161040410399437
0.575675547122955 0.170977264642715
0.578653275966644 0.192112281918526
0.579489588737488 0.180536776781082
0.580305457115173 0.212542414665222
0.580802381038666 0.10538138449192
0.581196665763855 0.128066271543503
0.58131206035614 0.224588871002197
0.581929504871368 0.124731414020061
0.582538187503815 0.0877773091197014
0.584879815578461 0.107272446155548
0.584931194782257 0.20191915333271
0.585383474826813 0.105604231357574
0.585858821868896 0.207203239202499
0.591184914112091 0.175408899784088
0.592807114124298 0.178556695580482
0.592863082885742 0.211026668548584
0.593362867832184 0.166961222887039
0.594660878181458 0.218379035592079
0.594948470592499 0.224702149629593
0.595067799091339 0.182371228933334
0.595394551753998 0.196941375732422
0.595650434494019 0.0783470794558525
0.597231268882751 0.0751087069511414
0.599363267421722 0.306135684251785
0.601343810558319 0.222637861967087
0.604617238044739 0.256280511617661
0.605667769908905 0.265570282936096
0.609225869178772 0.229795813560486
0.609244585037231 0.19865694642067
0.609363973140717 0.285490602254868
0.610797464847565 0.215965136885643
0.611442267894745 0.114596910774708
0.611451685428619 0.150617569684982
0.613154470920563 0.301522642374039
0.615905404090881 0.0812671035528183
0.616033375263214 0.267264038324356
0.617058634757996 0.177654340863228
0.61783093214035 0.149618878960609
0.619691014289856 0.180012568831444
0.620198845863342 0.137027934193611
0.620838582515717 0.195145010948181
0.621964454650879 0.214419737458229
0.622164845466614 0.207161754369736
0.623395442962646 0.0825863480567932
0.625522017478943 0.230903387069702
0.625656127929688 0.0747528597712517
0.626032114028931 0.111462309956551
0.62616103887558 0.0782541558146477
0.627038955688477 0.208165347576141
0.62725692987442 0.127827152609825
0.629179835319519 0.0669101476669312
0.629596412181854 0.15329821407795
0.629886746406555 0.179240763187408
0.63009786605835 0.15946164727211
0.63225120306015 0.196040987968445
0.632619559764862 0.108110599219799
0.633637607097626 0.0507570914924145
0.635416924953461 0.0690015256404877
0.635424077510834 0.145813256502151
0.635967433452606 0.0427055545151234
0.63662987947464 0.13786692917347
0.637954890727997 0.143955066800117
0.638030290603638 0.109846711158752
0.638634145259857 0.0863989964127541
0.639043807983398 0.129258424043655
0.639657020568848 0.148768827319145
0.640920221805573 0.100590974092484
0.640979707241058 0.0467716194689274
0.641856074333191 0.078387126326561
0.643027305603027 0.0598834827542305
0.645687639713287 0.141718119382858
0.646506726741791 0.0935161039233208
0.647252321243286 0.136759668588638
0.648058474063873 0.159714937210083
0.648106575012207 0.165694355964661
0.649079144001007 0.132405489683151
0.649994730949402 0.13707223534584
0.650914549827576 0.0730653926730156
0.651254177093506 0.101328231394291
0.651597857475281 0.108813837170601
0.652158915996552 -0.00909660011529922
0.653536438941956 0.0603935644030571
0.653761982917786 0.149532616138458
0.653894066810608 0.00261978060007095
0.654071748256683 0.0348462089896202
0.65625137090683 -0.113659784197807
0.656627595424652 0.169282212853432
0.656760394573212 0.144966781139374
0.658389925956726 0.13746702671051
0.658483028411865 0.141550779342651
0.658983826637268 0.0873368382453918
0.658996760845184 0.0633082464337349
0.659094452857971 -0.0189358741044998
0.659453332424164 0.121252432465553
0.661216080188751 0.132450729608536
0.662003636360168 0.0303339324891567
0.66250604391098 0.0516396835446358
0.665185332298279 0.115423142910004
0.665937185287476 0.0355493202805519
0.666290462017059 -0.022829432040453
0.667081117630005 -0.109237648546696
0.667598843574524 0.0517770200967789
0.667879343032837 0.0294108241796494
0.668041884899139 0.0166228730231524
0.668361842632294 0.0313351191580296
0.669692933559418 0.0592782348394394
0.670147657394409 0.173283830285072
0.671071231365204 0.0532138347625732
0.673041522502899 -0.0858234092593193
0.675426542758942 0.0797265321016312
0.676513969898224 -0.0266809724271297
0.677212178707123 0.0224201567471027
0.679137587547302 -0.0608317330479622
0.679229617118835 0.0173703953623772
0.680384874343872 -0.0688756704330444
0.680494129657745 0.12335167825222
0.682635247707367 -0.0515884757041931
0.683237850666046 -0.039927039295435
0.68434727191925 0.100712783634663
0.684428155422211 -0.047901701182127
0.685863733291626 0.140129908919334
0.687007129192352 0.0193036012351513
0.687663972377777 0.0111726466566324
0.689038455486298 -0.160828098654747
0.690020322799683 -0.0102226808667183
0.690078854560852 -0.151710718870163
0.690378487110138 0.142068982124329
0.691297888755798 0.0388687178492546
0.694601953029633 0.0112331779673696
0.695164620876312 0.00585552863776684
0.698246836662292 0.0902585238218307
0.700952470302582 0.0511845052242279
0.701016962528229 -0.0542285516858101
0.701213657855988 -0.0887948349118233
0.701741874217987 -0.00513056851923466
0.702874600887299 0.061434268951416
0.705142021179199 0.0969233959913254
0.705440104007721 0.114208519458771
0.705975592136383 0.0700719133019447
0.706383049488068 0.0494184866547585
0.709213078022003 0.123748779296875
0.714430391788483 0.0378367900848389
0.716340661048889 0.126433506608009
0.717490434646606 0.0215527545660734
0.718671321868896 0.161428526043892
0.718690514564514 0.0503360703587532
0.718870997428894 0.0391260012984276
0.720744013786316 -0.0146024040877819
0.721688628196716 -0.0326010435819626
0.7240851521492 0.033511895686388
0.724371194839478 -0.0184282287955284
0.724599778652191 0.046661090105772
0.724910795688629 0.141220450401306
0.72744345664978 0.0872521102428436
0.728538453578949 0.0866059958934784
0.731929004192352 -0.0566979832947254
0.732148945331573 0.0519014671444893
0.732575297355652 0.110234491527081
0.733170807361603 0.0391818881034851
0.734317600727081 -0.000203095376491547
0.735908508300781 0.030265910550952
0.737829804420471 0.0750056654214859
0.738919794559479 -0.0496148318052292
0.739495396614075 -0.010012611746788
0.739710450172424 0.0169991888105869
0.74145382642746 -0.0389780774712563
0.741659224033356 0.0150046646595001
0.741834223270416 0.0651395544409752
0.741925597190857 0.108970329165459
0.742133677005768 -0.0266054905951023
0.742466270923615 0.10955074429512
0.743338227272034 0.0366539321839809
0.743549406528473 0.0690498948097229
0.745955646038055 -0.030016615986824
0.746697187423706 -0.0400952473282814
0.746809422969818 -0.00854595750570297
0.747628450393677 0.187863975763321
0.749484837055206 0.0830913186073303
0.74991762638092 0.000750593841075897
0.752272248268127 0.0155381336808205
0.752800464630127 0.026187602430582
0.755310595035553 0.154969573020935
0.757800042629242 0.113080330193043
0.758200109004974 0.0406028777360916
0.758343100547791 -0.0168612748384476
0.758499026298523 0.119931295514107
0.758722066879272 0.103559531271458
0.759126842021942 0.103160575032234
0.759566247463226 0.197415858507156
0.76087498664856 0.0178837701678276
0.761159002780914 0.0436029136180878
0.761166214942932 0.114587813615799
0.764002561569214 0.169249549508095
0.764766573905945 0.043889369815588
0.765248775482178 -0.0178244784474373
0.765255153179169 0.0562059730291367
0.766946911811829 -0.0094287320971489
0.767013609409332 0.142179518938065
0.768154084682465 0.12705297768116
0.768227458000183 -0.106509059667587
0.768964827060699 -0.0187835693359375
0.769193708896637 0.126552864909172
0.775266051292419 0.0450914390385151
0.776522397994995 0.123767390847206
0.778371810913086 0.115742221474648
0.77950781583786 0.119487337768078
0.779924869537354 0.00992073863744736
0.780610382556915 0.159482359886169
0.78199428319931 0.0476607792079449
0.782252132892609 0.145325481891632
0.784106373786926 0.0879757702350616
0.784371137619019 0.0406833216547966
0.785624444484711 0.0536305606365204
0.787657022476196 0.0439961403608322
0.788341045379639 0.0786313489079475
0.790717661380768 0.131337940692902
0.793409705162048 0.170772671699524
0.793606698513031 0.10387334227562
0.793983459472656 -0.0259885266423225
0.794019460678101 0.00836236029863358
0.794210135936737 0.153977185487747
0.794986128807068 0.110465861856937
0.795261979103088 -0.0151809081435204
0.796147763729095 0.164952844381332
0.796878755092621 0.0704360157251358
0.797990024089813 0.140041649341583
0.799138486385345 0.0621447153389454
0.799727380275726 0.12677863240242
0.800387322902679 0.0315960869193077
0.80109316110611 0.0736145824193954
0.803361356258392 0.136525183916092
0.803842067718506 0.0817187279462814
0.808896362781525 0.133457735180855
0.81052029132843 0.0913931578397751
0.810665547847748 0.214543998241425
0.811855137348175 0.134256109595299
0.812057614326477 0.103759489953518
0.813202798366547 0.127117127180099
0.81437736749649 0.226619958877563
0.814534902572632 0.105793111026287
0.814591765403748 0.0610987804830074
0.814602792263031 0.0207540094852448
0.818363845348358 0.0674886852502823
0.81876802444458 0.0348009616136551
0.819408893585205 0.00814211368560791
0.82080602645874 0.117781199514866
0.820865273475647 0.109409295022488
0.820875406265259 0.255052983760834
0.821343183517456 0.098651796579361
0.821401953697205 0.184477657079697
0.821572542190552 0.0874812826514244
0.822344660758972 0.0677005499601364
0.825192511081696 0.0387889817357063
0.825464844703674 0.131178572773933
0.825468361377716 0.0952712520956993
0.825495064258575 0.201848194003105
0.825945496559143 0.102933071553707
0.826761245727539 0.174239218235016
0.827197313308716 0.123024918138981
0.827406883239746 0.0181795954704285
0.827880024909973 0.162986621260643
0.828085362911224 0.168118849396706
0.829447448253632 0.0611468851566315
0.831851720809937 0.0614785775542259
0.832289159297943 0.0729686915874481
0.832723557949066 0.0300619676709175
0.833836555480957 0.140214502811432
0.83471143245697 0.0980288311839104
0.83570408821106 0.0860579758882523
0.837533950805664 0.166341662406921
0.837960004806519 0.191083431243896
0.840779781341553 -0.00150179117918015
0.841040968894958 0.070867232978344
0.842049181461334 0.175001502037048
0.842162430286407 0.180322557687759
0.842233180999756 -0.0943198576569557
0.843340516090393 0.0819031074643135
0.844564199447632 0.0300303846597672
0.845101773738861 0.105354979634285
0.84542578458786 0.0162694975733757
0.845562040805817 0.0584565512835979
0.846405982971191 0.148121118545532
0.846573293209076 0.0474166199564934
0.847252130508423 0.0800949856638908
0.84762167930603 0.176268190145493
0.847878754138947 0.0734383314847946
0.849616229534149 0.168999552726746
0.84985888004303 0.0375388972461224
0.850727438926697 0.120782628655434
0.851645827293396 0.123719267547131
0.852636873722076 0.0184046849608421
0.853705048561096 0.168255507946014
0.853800415992737 0.214871197938919
0.853817403316498 0.204893812537193
0.858226656913757 0.0151581987738609
0.858405649662018 0.114376917481422
0.8591228723526 0.153075098991394
0.860078811645508 0.0716007351875305
0.86164778470993 0.155033126473427
0.861804366111755 0.135590448975563
0.862199306488037 0.0723648369312286
0.862287819385529 0.0825504139065742
0.863248467445374 0.107482105493546
0.864294528961182 -0.0228400155901909
0.866258800029755 0.077262744307518
0.867130994796753 0.128767594695091
0.867473423480988 0.138822242617607
0.869111597537994 0.169055983424187
0.870773136615753 0.0948225483298302
0.871430337429047 0.0597420707345009
0.871457397937775 0.16793592274189
0.87228661775589 0.00631748884916306
0.872827291488647 0.209104135632515
0.874287068843842 0.072811484336853
0.877163350582123 0.108583346009254
0.877687215805054 0.0669432133436203
0.87982976436615 0.0484814494848251
0.879925906658173 0.00634866207838058
0.881817638874054 0.0852461382746696
0.882034599781036 -0.0159368142485619
0.884983718395233 0.132143661379814
0.885252952575684 0.235195130109787
0.887497842311859 0.183371633291245
0.889658153057098 0.0769525766372681
0.889898121356964 -0.0408350080251694
0.890372216701508 0.142504274845123
0.891906142234802 0.0465626865625381
0.892964959144592 0.022831529378891
0.893611490726471 0.00484801828861237
0.894010841846466 0.0809841081500053
0.895351231098175 0.0705184787511826
0.895890533924103 0.0895277783274651
0.897125244140625 0.0140867456793785
0.897319793701172 0.0464585050940514
0.901436567306519 0.176580607891083
0.902322173118591 0.174843668937683
0.904527187347412 0.0649407804012299
0.905417025089264 0.0554683245718479
0.905548453330994 0.130811452865601
0.906889677047729 -0.0133910775184631
0.908006072044373 0.0905334204435349
0.908948719501495 0.181525349617004
0.909656882286072 0.0673612356185913
0.912491917610168 0.170792311429977
0.912668347358704 0.110020071268082
0.915349125862122 0.144217550754547
0.917639791965485 0.0907537415623665
0.918327331542969 0.0918713510036469
0.918346524238586 0.00205683708190918
0.919712245464325 0.0394309125840664
0.919826030731201 0.0877793282270432
0.922673106193542 -0.0186132788658142
0.923919439315796 0.0559635162353516
0.92450886964798 0.106005348265171
0.924819529056549 0.139111906290054
0.925417065620422 0.190269201993942
0.926218271255493 0.0420060940086842
0.927005410194397 0.0359202921390533
0.928315401077271 0.0748796612024307
0.928659558296204 0.122727811336517
0.930364012718201 0.0730736702680588
0.932134926319122 0.127940028905869
0.932627260684967 0.109937608242035
0.93420422077179 0.0167041718959808
0.935517430305481 -0.0171645879745483
0.935672342777252 0.0943638607859612
0.935826778411865 -0.0108591765165329
0.93672513961792 0.0788511484861374
0.937280535697937 0.0598872937262058
0.937365233898163 0.0865434408187866
0.937952518463135 0.23395973443985
0.93852174282074 0.0372557379305363
0.939170897006989 0.14812408387661
0.939444839954376 0.262849986553192
0.942980229854584 0.10410538315773
0.943534195423126 0.0580475255846977
0.944198250770569 0.181987702846527
0.94531112909317 0.111934974789619
0.945452153682709 -0.0500031560659409
0.946478545665741 0.155321329832077
0.947280764579773 0.112904235720634
0.950588464736938 0.0856567919254303
0.950772762298584 0.215809345245361
0.953024387359619 0.0423212349414825
0.9544557929039 0.0350288338959217
0.954927325248718 0.138846650719643
0.955338358879089 0.260108470916748
0.955825626850128 0.0466931648552418
0.955916225910187 0.127036720514297
0.95691043138504 0.196032524108887
0.957509160041809 0.0886376723647118
0.958259105682373 0.0980396121740341
0.958288013935089 0.0220291316509247
0.958702147006989 0.0542555414140224
0.959420442581177 0.190031617879868
0.959894895553589 0.104819469153881
0.961052596569061 0.0836371630430222
0.962590456008911 0.0705686956644058
0.962921977043152 0.129996344447136
0.963911771774292 0.098788395524025
0.964198172092438 0.0182646065950394
0.965870499610901 0.208566635847092
0.965922951698303 -0.018428236246109
0.965931057929993 0.0834427326917648
0.965960025787354 -0.0880495756864548
0.966410636901855 0.107032835483551
0.967171370983124 0.103252880275249
0.967790305614471 0.0566208064556122
0.969921052455902 0.18568816781044
0.971257746219635 0.0182464271783829
0.971977889537811 0.100684300065041
0.972075998783112 0.105246663093567
0.972141683101654 -0.0361154302954674
0.973164141178131 0.0624736472964287
0.975789904594421 0.12889552116394
0.976115226745605 -0.00526641309261322
0.976327836513519 0.262745261192322
0.977026760578156 0.121978178620338
0.977763533592224 0.0883419811725616
0.978739440441132 -0.00780612230300903
0.979627251625061 0.0457439199090004
0.982460498809814 0.00245753675699234
0.982534289360046 0.0579778738319874
0.983276665210724 0.114051848649979
0.983926773071289 0.0750444605946541
0.984570026397705 0.065642349421978
0.984847187995911 0.10375003516674
0.987444996833801 0.144166499376297
0.989892482757568 0.1215560734272
0.990497827529907 0.0907557159662247
0.990667283535004 0.128434911370277
0.990828096866608 0.0694819018244743
0.991193532943726 0.197497934103012
0.991631925106049 0.0223874039947987
0.992101550102234 -0.105672746896744
0.992335140705109 0.0927348583936691
0.992498278617859 0.0847035050392151
0.996789634227753 0.042107492685318
0.997103571891785 0.0991952195763588
};
\addplot [draw=blue, fill=blue, mark=*, only marks, opacity=0.5]
table{%
x  y
0.000199556350708008 0.145666301250458
0.879925906658173 0.00634866207838058
0.992335140705109 0.0927348583936691
0.206859290599823 0.192682549357414
0.626032114028931 0.111462309956551
0.997103571891785 0.0991952195763588
0.992498278617859 0.0847035050392151
0.00027167797088623 0.191887632012367
0.609225869178772 0.229795813560486
0.992101550102234 -0.105672746896744
0.996789634227753 0.042107492685318
0.625656127929688 0.0747528597712517
0.623395442962646 0.0825863480567932
0.574207007884979 0.0442190617322922
0.48443990945816 -0.457312732934952
0.00175052881240845 -0.0372040420770645
0.887497842311859 0.183371633291245
0.853817403316498 0.204893812537193
0.0598086714744568 0.0697462931275368
0.63009786605835 0.15946164727211
0.155065596103668 0.053071454167366
0.959894895553589 0.104819469153881
0.00023043155670166 0.0628557577729225
0.0282864570617676 -0.0437518209218979
0.973164141178131 0.0624736472964287
0.714430391788483 0.0378367900848389
0.349500477313995 -0.0416359454393387
0.243977844715118 0.307858884334564
0.625522017478943 0.230903387069702
0.919712245464325 0.0394309125840664
0.840779781341553 -0.00150179117918015
};
\addlegendentry{PFN}
\addplot [draw=red, fill=red, mark=*, only marks, opacity=0.5]
table{%
x  y
0.214899718761444 0.135813817381859
0.997103571891785 0.0991952195763588
0.522444665431976 -0.281921744346619
0.189625680446625 0.0470676496624947
0.211761236190796 0.26262429356575
0.213146388530731 0.27036851644516
0.395127952098846 -0.0894727110862732
0.207242071628571 0.286000460386276
0.168284058570862 0.0670295655727386
0.617058634757996 0.177654340863228
0.208481967449188 0.248721644282341
0.887497842311859 0.183371633291245
0.3083735704422 0.0940080136060715
0.208665430545807 0.27168396115303
0.595394551753998 0.196941375732422
0.22229528427124 0.151753753423691
0.21306985616684 0.273029297590256
0.565553963184357 -0.0786903351545334
0.211481034755707 0.249547839164734
0.437246918678284 -0.202439501881599
0.209347784519196 0.290922284126282
0.209558308124542 0.218788921833038
0.123427033424377 -0.187452971935272
0.862287819385529 0.0825504139065742
0.573487937450409 0.00780980288982391
0.214790940284729 0.267421126365662
0.928659558296204 0.122727811336517
0.172621428966522 0.275365650653839
0.21455329656601 0.28043857216835
0.847878754138947 0.0734383314847946
0.65625137090683 -0.113659784197807
};
\addlegendentry{GP}
\legend{};
\end{axis}

\end{tikzpicture}

%% file: icml2023/figures/hpob_heboabl_rank_test.tex
\begin{tikzpicture}

\definecolor{color0}{rgb}{0.12156862745098,0.466666666666667,0.705882352941177}
\definecolor{color1}{rgb}{1,0.498039215686275,0.0549019607843137}
\definecolor{color2}{rgb}{0.172549019607843,0.627450980392157,0.172549019607843}
\definecolor{color3}{rgb}{0.83921568627451,0.152941176470588,0.156862745098039}
\definecolor{color4}{rgb}{0.580392156862745,0.403921568627451,0.741176470588235}
\definecolor{color5}{rgb}{0.549019607843137,0.337254901960784,0.294117647058824}
\definecolor{color6}{rgb}{0.890196078431372,0.466666666666667,0.76078431372549}
\definecolor{color7}{rgb}{0.737254901960784,0.741176470588235,0.133333333333333}
\definecolor{color8}{rgb}{0.0901960784313725,0.745098039215686,0.811764705882353}

\begin{axis}[
legend cell align={left},
legend style={
  fill opacity=0.8,
  draw opacity=1,
  text opacity=1,
  at={(0.55,-0.15)},
  anchor=south,
  draw=white!80!black
},
tick align=outside,
tick pos=left,
x grid style={white!69.0196078431373!black},
xlabel={Number of trials},
xmin=2.75, xmax=52.25,
xtick style={color=black},
  height=.6\textwidth,
    width=.8\textwidth,
y grid style={white!69.0196078431373!black},
ylabel={Average Rank},
ymin=4.69749737055732, ymax=9.48718233793648,
ytick style={color=black}
]
\path [fill=color0, fill opacity=0.2]
(axis cs:5,6)
--(axis cs:5,6)
--(axis cs:6,6.58562870582007)
--(axis cs:7,6.79478555852609)
--(axis cs:8,6.85605063045067)
--(axis cs:9,6.88080365391713)
--(axis cs:10,6.90327370723433)
--(axis cs:11,7.14195694636429)
--(axis cs:12,7.30765099931619)
--(axis cs:13,7.42872646804694)
--(axis cs:14,7.47485436414123)
--(axis cs:15,7.64346165604801)
--(axis cs:16,7.7532441847336)
--(axis cs:17,7.86224527200844)
--(axis cs:18,7.93493249139216)
--(axis cs:19,7.9871913638014)
--(axis cs:20,8.14973805963093)
--(axis cs:21,8.23443437102114)
--(axis cs:22,8.2972137823833)
--(axis cs:23,8.39894030034532)
--(axis cs:24,8.3687042138004)
--(axis cs:25,8.47003968465408)
--(axis cs:26,8.4910671342023)
--(axis cs:27,8.60287599601126)
--(axis cs:28,8.56481478536549)
--(axis cs:29,8.63054819488385)
--(axis cs:30,8.66867936707245)
--(axis cs:31,8.6412452954255)
--(axis cs:32,8.67133777835913)
--(axis cs:33,8.68298361715465)
--(axis cs:34,8.71289058968531)
--(axis cs:35,8.72388242117256)
--(axis cs:36,8.77696813182446)
--(axis cs:37,8.8130377441834)
--(axis cs:38,8.84290222258238)
--(axis cs:39,8.8524556990745)
--(axis cs:40,8.87836940987583)
--(axis cs:41,8.89744756385564)
--(axis cs:42,8.85738926664117)
--(axis cs:43,8.88086313750547)
--(axis cs:44,8.83867407973571)
--(axis cs:45,8.8701324384514)
--(axis cs:46,8.89624924223883)
--(axis cs:47,8.92882403353237)
--(axis cs:48,8.94708825615041)
--(axis cs:49,8.98198838513929)
--(axis cs:50,8.99719728179287)
--(axis cs:50,9.26946938487379)
--(axis cs:50,9.26946938487379)
--(axis cs:49,9.24154102662541)
--(axis cs:48,9.20977448894763)
--(axis cs:47,9.19666616254606)
--(axis cs:46,9.16649585580038)
--(axis cs:45,9.17300481645056)
--(axis cs:44,9.12603180261723)
--(axis cs:43,9.12697999974943)
--(axis cs:42,9.12692445884902)
--(axis cs:41,9.14961125967377)
--(axis cs:40,9.12555215875162)
--(axis cs:39,9.10048547739609)
--(axis cs:38,9.0708232676137)
--(axis cs:37,9.04186421660092)
--(axis cs:36,9.01126716229319)
--(axis cs:35,8.98984306902352)
--(axis cs:34,8.99299176325587)
--(axis cs:33,8.95231050049241)
--(axis cs:32,8.98748575105264)
--(axis cs:31,8.97051941045685)
--(axis cs:30,8.96269318194716)
--(axis cs:29,8.91062827570438)
--(axis cs:28,8.87440090090902)
--(axis cs:27,8.88732008242011)
--(axis cs:26,8.80697208148397)
--(axis cs:25,8.78878384475768)
--(axis cs:24,8.70972715874862)
--(axis cs:23,8.69909891534095)
--(axis cs:22,8.62435484506769)
--(axis cs:21,8.59693817799846)
--(axis cs:20,8.51300703840829)
--(axis cs:19,8.36967138129664)
--(axis cs:18,8.36702829292157)
--(axis cs:17,8.26324492406999)
--(axis cs:16,8.19969699173699)
--(axis cs:15,8.05849912826572)
--(axis cs:14,7.88592994958426)
--(axis cs:13,7.80656764960012)
--(axis cs:12,7.63352547127205)
--(axis cs:11,7.58353324971414)
--(axis cs:10,7.33202041041273)
--(axis cs:9,7.15056889510248)
--(axis cs:8,7.0419885852356)
--(axis cs:7,7.02482228461117)
--(axis cs:6,6.70848894123876)
--(axis cs:5,6)
--cycle;

\path [fill=color1, fill opacity=0.2]
(axis cs:5,6)
--(axis cs:5,6)
--(axis cs:6,6.36723704751799)
--(axis cs:7,6.5900445510603)
--(axis cs:8,6.60960119307758)
--(axis cs:9,6.79614635913696)
--(axis cs:10,6.8317134754343)
--(axis cs:11,6.7172803731103)
--(axis cs:12,6.53583664628062)
--(axis cs:13,6.34875576535062)
--(axis cs:14,6.3114691607634)
--(axis cs:15,6.27107873769945)
--(axis cs:16,6.23355871735485)
--(axis cs:17,6.22124989991631)
--(axis cs:18,6.18652046755494)
--(axis cs:19,6.12197167911795)
--(axis cs:20,5.96606927044373)
--(axis cs:21,5.79488542544523)
--(axis cs:22,5.74455627868277)
--(axis cs:23,5.65420438019305)
--(axis cs:24,5.63499906402776)
--(axis cs:25,5.62220132662765)
--(axis cs:26,5.5941167402661)
--(axis cs:27,5.59822964282605)
--(axis cs:28,5.53945331881248)
--(axis cs:29,5.58507613679524)
--(axis cs:30,5.5039903975267)
--(axis cs:31,5.49650472221686)
--(axis cs:32,5.51094913449157)
--(axis cs:33,5.50616793720362)
--(axis cs:34,5.48731298657378)
--(axis cs:35,5.46306187369372)
--(axis cs:36,5.46488395723491)
--(axis cs:37,5.42341418674064)
--(axis cs:38,5.44180144299197)
--(axis cs:39,5.46282437803013)
--(axis cs:40,5.39116535880635)
--(axis cs:41,5.41925151439831)
--(axis cs:42,5.4604606621281)
--(axis cs:43,5.48898525597348)
--(axis cs:44,5.507878640694)
--(axis cs:45,5.43692287789143)
--(axis cs:46,5.36209305180498)
--(axis cs:47,5.40384629378025)
--(axis cs:48,5.41419866104302)
--(axis cs:49,5.41371979403711)
--(axis cs:50,5.39361971558765)
--(axis cs:50,5.46912538245157)
--(axis cs:50,5.46912538245157)
--(axis cs:49,5.48824099027661)
--(axis cs:48,5.4720758487609)
--(axis cs:47,5.47458507876877)
--(axis cs:46,5.48104420309699)
--(axis cs:45,5.59837123975563)
--(axis cs:44,5.6842782220511)
--(axis cs:43,5.67179905775201)
--(axis cs:42,5.62581384767582)
--(axis cs:41,5.60819946599385)
--(axis cs:40,5.5892267980564)
--(axis cs:39,5.62737170040124)
--(axis cs:38,5.63662992955705)
--(axis cs:37,5.61580149953387)
--(axis cs:36,5.62923368982391)
--(axis cs:35,5.6310557733651)
--(axis cs:34,5.62249093499485)
--(axis cs:33,5.70559676867873)
--(axis cs:32,5.66944302237118)
--(axis cs:31,5.69957370915569)
--(axis cs:30,5.72738215149291)
--(axis cs:29,5.84629641222437)
--(axis cs:28,5.80172315177575)
--(axis cs:27,5.89588800423278)
--(axis cs:26,5.88039306365547)
--(axis cs:25,5.91897514396059)
--(axis cs:24,5.93362838695264)
--(axis cs:23,6.01246228647361)
--(axis cs:22,6.05936528994468)
--(axis cs:21,6.07570280984889)
--(axis cs:20,6.1515777883798)
--(axis cs:19,6.28587145813695)
--(axis cs:18,6.31936188538623)
--(axis cs:17,6.36698539420134)
--(axis cs:16,6.33114716499809)
--(axis cs:15,6.42696047798682)
--(axis cs:14,6.50813868237386)
--(axis cs:13,6.5061461954337)
--(axis cs:12,6.74651629489585)
--(axis cs:11,6.94154315630146)
--(axis cs:10,7.06240417162452)
--(axis cs:9,6.94503011145128)
--(axis cs:8,6.75118312064791)
--(axis cs:7,6.68838682148872)
--(axis cs:6,6.49550805052123)
--(axis cs:5,6)
--cycle;

\path [fill=color2, fill opacity=0.2]
(axis cs:5,6)
--(axis cs:5,6)
--(axis cs:6,5.63345616484574)
--(axis cs:7,5.62505636865293)
--(axis cs:8,5.43665648863728)
--(axis cs:9,5.43363197993879)
--(axis cs:10,5.29524161529279)
--(axis cs:11,5.2977315765944)
--(axis cs:12,5.37721062086886)
--(axis cs:13,5.41778787458131)
--(axis cs:14,5.48211319807016)
--(axis cs:15,5.57299978765088)
--(axis cs:16,5.59536986693645)
--(axis cs:17,5.69979067586934)
--(axis cs:18,5.70579172487918)
--(axis cs:19,5.71378349832874)
--(axis cs:20,5.79924131288561)
--(axis cs:21,5.78673777043168)
--(axis cs:22,5.84401668837667)
--(axis cs:23,5.90902809388404)
--(axis cs:24,5.89689194782027)
--(axis cs:25,5.9541506644591)
--(axis cs:26,5.99353536571688)
--(axis cs:27,5.98331806760106)
--(axis cs:28,6.02726360180986)
--(axis cs:29,6.05510352462221)
--(axis cs:30,6.05133474033587)
--(axis cs:31,6.07231516397775)
--(axis cs:32,6.09325064601063)
--(axis cs:33,6.13538069623136)
--(axis cs:34,6.1586725212891)
--(axis cs:35,6.11531778301603)
--(axis cs:36,6.05384640016824)
--(axis cs:37,6.08474807576404)
--(axis cs:38,6.06937999710262)
--(axis cs:39,6.08554587189354)
--(axis cs:40,6.10990623694387)
--(axis cs:41,6.08199360558115)
--(axis cs:42,6.16672619337753)
--(axis cs:43,6.11079504330499)
--(axis cs:44,6.02910476120984)
--(axis cs:45,6.09075414506086)
--(axis cs:46,6.12283369292005)
--(axis cs:47,6.13073256822666)
--(axis cs:48,6.08714863066645)
--(axis cs:49,6.10364230855174)
--(axis cs:50,6.11218880194749)
--(axis cs:50,6.42114453138585)
--(axis cs:50,6.42114453138585)
--(axis cs:49,6.39831847576199)
--(axis cs:48,6.39520431051003)
--(axis cs:47,6.41436547098903)
--(axis cs:46,6.41049964041328)
--(axis cs:45,6.38767722748816)
--(axis cs:44,6.343444258398)
--(axis cs:43,6.43038142728325)
--(axis cs:42,6.44896008113227)
--(axis cs:41,6.38075149245807)
--(axis cs:40,6.39597611599731)
--(axis cs:39,6.37719922614567)
--(axis cs:38,6.35807098328954)
--(axis cs:37,6.38584015953008)
--(axis cs:36,6.36576144296902)
--(axis cs:35,6.35919202090554)
--(axis cs:34,6.39034708655404)
--(axis cs:33,6.37834479396472)
--(axis cs:32,6.33420033438152)
--(axis cs:31,6.3237632673948)
--(axis cs:30,6.29376329887981)
--(axis cs:29,6.30568078910328)
--(axis cs:28,6.27469718250386)
--(axis cs:27,6.22452506965384)
--(axis cs:26,6.15940581075371)
--(axis cs:25,6.07330031593306)
--(axis cs:24,6.0089904051209)
--(axis cs:23,5.96940327866498)
--(axis cs:22,5.95598331162333)
--(axis cs:21,5.87992889623499)
--(axis cs:20,5.91448417731047)
--(axis cs:19,5.86268708990656)
--(axis cs:18,5.85499258884631)
--(axis cs:17,5.8492289319738)
--(axis cs:16,5.74972817227924)
--(axis cs:15,5.76817668293736)
--(axis cs:14,5.69435739016513)
--(axis cs:13,5.71554545875202)
--(axis cs:12,5.65416192815075)
--(axis cs:11,5.55717038418991)
--(axis cs:10,5.51260152196211)
--(axis cs:9,5.61734841221807)
--(axis cs:8,5.68099057018625)
--(axis cs:7,5.6925906901706)
--(axis cs:6,5.75085756064446)
--(axis cs:5,6)
--cycle;

\path [fill=color3, fill opacity=0.2]
(axis cs:5,6)
--(axis cs:5,6)
--(axis cs:6,5.43361045186759)
--(axis cs:7,5.43657765465656)
--(axis cs:8,5.35106697074116)
--(axis cs:9,5.22331483872449)
--(axis cs:10,5.33815501593988)
--(axis cs:11,5.369893638773)
--(axis cs:12,5.29618948823499)
--(axis cs:13,5.1661670824918)
--(axis cs:14,5.26666996082025)
--(axis cs:15,5.37193394119806)
--(axis cs:16,5.41753440284062)
--(axis cs:17,5.44327530060765)
--(axis cs:18,5.45500786393712)
--(axis cs:19,5.51568343536263)
--(axis cs:20,5.60568202387062)
--(axis cs:21,5.57554322004792)
--(axis cs:22,5.58747309535858)
--(axis cs:23,5.54545870816026)
--(axis cs:24,5.57254618046067)
--(axis cs:25,5.56622642529471)
--(axis cs:26,5.61465771322672)
--(axis cs:27,5.59621384475155)
--(axis cs:28,5.61941828352319)
--(axis cs:29,5.69431919562735)
--(axis cs:30,5.75181827355369)
--(axis cs:31,5.80247101616379)
--(axis cs:32,5.8439902421695)
--(axis cs:33,5.87818114501712)
--(axis cs:34,5.83508918605169)
--(axis cs:35,5.84717026828356)
--(axis cs:36,5.89442268800241)
--(axis cs:37,5.89406962744194)
--(axis cs:38,5.84358084252055)
--(axis cs:39,5.90180749385036)
--(axis cs:40,5.95640020972812)
--(axis cs:41,5.96884757233772)
--(axis cs:42,5.98382053561506)
--(axis cs:43,5.97289552680047)
--(axis cs:44,5.94321796873159)
--(axis cs:45,5.94619640638728)
--(axis cs:46,5.98771470920296)
--(axis cs:47,5.92227494785241)
--(axis cs:48,5.95049749470066)
--(axis cs:49,5.91501367475492)
--(axis cs:50,5.92256027681695)
--(axis cs:50,6.14410638984971)
--(axis cs:50,6.14410638984971)
--(axis cs:49,6.1477314232843)
--(axis cs:48,6.16714956412287)
--(axis cs:47,6.12870544430445)
--(axis cs:46,6.18483431040489)
--(axis cs:45,6.1675290838088)
--(axis cs:44,6.18227222734685)
--(axis cs:43,6.16435937516031)
--(axis cs:42,6.14951279771827)
--(axis cs:41,6.13703478060346)
--(axis cs:40,6.08281547654639)
--(axis cs:39,6.01976113360062)
--(axis cs:38,5.96818386336181)
--(axis cs:37,5.99612645098943)
--(axis cs:36,5.98793025317406)
--(axis cs:35,5.97635914348115)
--(axis cs:34,5.94922453943851)
--(axis cs:33,5.98064238439465)
--(axis cs:32,5.97953916959521)
--(axis cs:31,5.9426270230519)
--(axis cs:30,5.90308368723063)
--(axis cs:29,5.81548472594128)
--(axis cs:28,5.78450328510426)
--(axis cs:27,5.71359007681708)
--(axis cs:26,5.72651875736151)
--(axis cs:25,5.71220494725431)
--(axis cs:24,5.69019891757854)
--(axis cs:23,5.67022756634954)
--(axis cs:22,5.7262523948375)
--(axis cs:21,5.7264175642658)
--(axis cs:20,5.67667091730585)
--(axis cs:19,5.6333361724805)
--(axis cs:18,5.6391097831217)
--(axis cs:17,5.72535215037274)
--(axis cs:16,5.6844263814731)
--(axis cs:15,5.62414449017449)
--(axis cs:14,5.56862415682681)
--(axis cs:13,5.52402899593957)
--(axis cs:12,5.59400659019638)
--(axis cs:11,5.57128283181523)
--(axis cs:10,5.61478616053071)
--(axis cs:9,5.54139104362845)
--(axis cs:8,5.59403106847453)
--(axis cs:7,5.64969685514736)
--(axis cs:6,5.66835033244613)
--(axis cs:5,6)
--cycle;

\path [fill=color4, fill opacity=0.2]
(axis cs:5,6)
--(axis cs:5,6)
--(axis cs:6,6.06629164775377)
--(axis cs:7,6.01524053515107)
--(axis cs:8,5.70304034989407)
--(axis cs:9,5.77854974362603)
--(axis cs:10,5.54371326612505)
--(axis cs:11,5.67547971277012)
--(axis cs:12,5.57428315548338)
--(axis cs:13,5.62077693084115)
--(axis cs:14,5.54377984103106)
--(axis cs:15,5.51904284611463)
--(axis cs:16,5.54057015399783)
--(axis cs:17,5.53902483775389)
--(axis cs:18,5.54104547197973)
--(axis cs:19,5.63991251298209)
--(axis cs:20,5.74774982920671)
--(axis cs:21,5.82355165844029)
--(axis cs:22,5.85458185091502)
--(axis cs:23,5.8890667516255)
--(axis cs:24,5.89847356596012)
--(axis cs:25,5.85350687621147)
--(axis cs:26,5.91972785848607)
--(axis cs:27,5.9530850602568)
--(axis cs:28,5.93353124183837)
--(axis cs:29,5.95075575419704)
--(axis cs:30,5.94292297304383)
--(axis cs:31,5.95439211814486)
--(axis cs:32,5.9419854174978)
--(axis cs:33,5.96785313081591)
--(axis cs:34,5.98155114262881)
--(axis cs:35,5.96327923998955)
--(axis cs:36,5.97201755260001)
--(axis cs:37,6.02738635262719)
--(axis cs:38,6.05713905076833)
--(axis cs:39,6.07234811784348)
--(axis cs:40,6.06699646174043)
--(axis cs:41,6.05741442147563)
--(axis cs:42,6.03996177486266)
--(axis cs:43,6.06617632416826)
--(axis cs:44,6.09248677900445)
--(axis cs:45,6.11592301666382)
--(axis cs:46,6.12814302469588)
--(axis cs:47,6.14431036224985)
--(axis cs:48,6.14635593418044)
--(axis cs:49,6.13770202613069)
--(axis cs:50,6.15474678686157)
--(axis cs:50,6.37074340921686)
--(axis cs:50,6.37074340921686)
--(axis cs:49,6.37210189543794)
--(axis cs:48,6.39874210503525)
--(axis cs:47,6.41647395147564)
--(axis cs:46,6.40519030863746)
--(axis cs:45,6.40564561078716)
--(axis cs:44,6.41731714256418)
--(axis cs:43,6.41225504838076)
--(axis cs:42,6.4110186172942)
--(axis cs:41,6.44454636283809)
--(axis cs:40,6.42712118531839)
--(axis cs:39,6.4256910978428)
--(axis cs:38,6.4683511453101)
--(axis cs:37,6.43928031403948)
--(axis cs:36,6.38876676112548)
--(axis cs:35,6.36221095608888)
--(axis cs:34,6.33609591619472)
--(axis cs:33,6.31057824173311)
--(axis cs:32,6.34821066093358)
--(axis cs:31,6.34364709754141)
--(axis cs:30,6.30805741911304)
--(axis cs:29,6.35904816737159)
--(axis cs:28,6.33705699345575)
--(axis cs:27,6.34103258680202)
--(axis cs:26,6.25674272974922)
--(axis cs:25,6.2053166532003)
--(axis cs:24,6.19956564972616)
--(axis cs:23,6.15407050327646)
--(axis cs:22,6.05914363928106)
--(axis cs:21,6.1372326552852)
--(axis cs:20,6.0679364453031)
--(axis cs:19,5.94832278113556)
--(axis cs:18,5.88248393978498)
--(axis cs:17,5.82960261322651)
--(axis cs:16,5.83982200286491)
--(axis cs:15,5.790761075454)
--(axis cs:14,5.76210251191011)
--(axis cs:13,5.86549757896278)
--(axis cs:12,5.79434429549701)
--(axis cs:11,5.8421673460534)
--(axis cs:10,5.78177692995338)
--(axis cs:9,5.9234110406877)
--(axis cs:8,5.93225376775299)
--(axis cs:7,6.34946534720187)
--(axis cs:6,6.40037501891289)
--(axis cs:5,6)
--cycle;

\path [fill=color5, fill opacity=0.2]
(axis cs:5,6)
--(axis cs:5,6)
--(axis cs:6,6.03176095932012)
--(axis cs:7,6.10849572775062)
--(axis cs:8,6.29984710365361)
--(axis cs:9,6.31502981999909)
--(axis cs:10,6.18355354771364)
--(axis cs:11,6.01431009025153)
--(axis cs:12,5.95348073069206)
--(axis cs:13,5.99938481991166)
--(axis cs:14,5.93256578043409)
--(axis cs:15,5.91000704163712)
--(axis cs:16,5.78772168652455)
--(axis cs:17,5.69672615580053)
--(axis cs:18,5.72307279744062)
--(axis cs:19,5.75950485478348)
--(axis cs:20,5.71722533691434)
--(axis cs:21,5.83367825944602)
--(axis cs:22,5.81593607268469)
--(axis cs:23,5.71896273375828)
--(axis cs:24,5.69091957843002)
--(axis cs:25,5.6947166599269)
--(axis cs:26,5.67544723614713)
--(axis cs:27,5.61154911577438)
--(axis cs:28,5.52524910672382)
--(axis cs:29,5.46575620676049)
--(axis cs:30,5.46025853867529)
--(axis cs:31,5.40501628078526)
--(axis cs:32,5.45684565662114)
--(axis cs:33,5.43362441036495)
--(axis cs:34,5.43642981353779)
--(axis cs:35,5.47285760208398)
--(axis cs:36,5.51028395292206)
--(axis cs:37,5.54853671038693)
--(axis cs:38,5.56382681668939)
--(axis cs:39,5.53287935929579)
--(axis cs:40,5.56640517339962)
--(axis cs:41,5.56697711986821)
--(axis cs:42,5.50779144907227)
--(axis cs:43,5.46707960532981)
--(axis cs:44,5.46406548605762)
--(axis cs:45,5.5113164998113)
--(axis cs:46,5.45696226576404)
--(axis cs:47,5.49465241908974)
--(axis cs:48,5.49425506764373)
--(axis cs:49,5.48079663031753)
--(axis cs:50,5.51604594631924)
--(axis cs:50,5.71532660270036)
--(axis cs:50,5.71532660270036)
--(axis cs:49,5.68390925203542)
--(axis cs:48,5.68221552059157)
--(axis cs:47,5.67397503189065)
--(axis cs:46,5.66068479305949)
--(axis cs:45,5.69260506881615)
--(axis cs:44,5.66142471002081)
--(axis cs:43,5.66233215937608)
--(axis cs:42,5.71965953131988)
--(axis cs:41,5.75459150758277)
--(axis cs:40,5.73555561091411)
--(axis cs:39,5.71810103286107)
--(axis cs:38,5.72636926174198)
--(axis cs:37,5.76518877980915)
--(axis cs:36,5.73677487060735)
--(axis cs:35,5.70753455477876)
--(axis cs:34,5.67337410803084)
--(axis cs:33,5.66049323669387)
--(axis cs:32,5.68040924533964)
--(axis cs:31,5.66165038588141)
--(axis cs:30,5.66131008877569)
--(axis cs:29,5.6714986952003)
--(axis cs:28,5.77279010896246)
--(axis cs:27,5.83943127638248)
--(axis cs:26,5.90102335208817)
--(axis cs:25,5.91704804595545)
--(axis cs:24,5.88947257843272)
--(axis cs:23,5.90848824663388)
--(axis cs:22,5.92916196653099)
--(axis cs:21,5.93887076016182)
--(axis cs:20,5.8239511336739)
--(axis cs:19,5.90324024325573)
--(axis cs:18,5.9553585751084)
--(axis cs:17,5.96994051086614)
--(axis cs:16,5.98090576445585)
--(axis cs:15,6.08607138973543)
--(axis cs:14,6.17723814113454)
--(axis cs:13,6.17708576832363)
--(axis cs:12,6.14455848499422)
--(axis cs:11,6.22490559602298)
--(axis cs:10,6.34977978561969)
--(axis cs:9,6.52810743490287)
--(axis cs:8,6.43740779830718)
--(axis cs:7,6.26013172322977)
--(axis cs:6,6.24274884460145)
--(axis cs:5,6)
--cycle;

\path [fill=color6, fill opacity=0.2]
(axis cs:5,6)
--(axis cs:5,6)
--(axis cs:6,6.08792003846985)
--(axis cs:7,6.15353576345618)
--(axis cs:8,6.31531957576978)
--(axis cs:9,6.23229160876867)
--(axis cs:10,6.23680406191121)
--(axis cs:11,6.04677868561079)
--(axis cs:12,6.10725802933276)
--(axis cs:13,6.17909617352671)
--(axis cs:14,6.1422657905145)
--(axis cs:15,6.10057444795793)
--(axis cs:16,6.07001411282969)
--(axis cs:17,5.9708530540542)
--(axis cs:18,5.92673974803844)
--(axis cs:19,5.86698071766294)
--(axis cs:20,5.76646138079411)
--(axis cs:21,5.78231800122597)
--(axis cs:22,5.84548681191022)
--(axis cs:23,5.90430425433758)
--(axis cs:24,5.76942696940549)
--(axis cs:25,5.75556126282153)
--(axis cs:26,5.61013356310078)
--(axis cs:27,5.56710746765851)
--(axis cs:28,5.61315667345164)
--(axis cs:29,5.62808577750302)
--(axis cs:30,5.64714826234987)
--(axis cs:31,5.66947669443532)
--(axis cs:32,5.63792142671098)
--(axis cs:33,5.66972358385317)
--(axis cs:34,5.60753799113785)
--(axis cs:35,5.60160141156851)
--(axis cs:36,5.63005401389939)
--(axis cs:37,5.61742923531659)
--(axis cs:38,5.62857245544344)
--(axis cs:39,5.55751046278207)
--(axis cs:40,5.57953671868408)
--(axis cs:41,5.50422086754934)
--(axis cs:42,5.52060504532425)
--(axis cs:43,5.46306144762755)
--(axis cs:44,5.46899090549608)
--(axis cs:45,5.47870726565358)
--(axis cs:46,5.48558018078128)
--(axis cs:47,5.50451154722515)
--(axis cs:48,5.4766546490107)
--(axis cs:49,5.51640950687548)
--(axis cs:50,5.43865117669274)
--(axis cs:50,5.5495841174249)
--(axis cs:50,5.5495841174249)
--(axis cs:49,5.5816297088108)
--(axis cs:48,5.57824731177361)
--(axis cs:47,5.60137080571602)
--(axis cs:46,5.56932178000303)
--(axis cs:45,5.59188096964053)
--(axis cs:44,5.59767576117059)
--(axis cs:43,5.61144835629402)
--(axis cs:42,5.69900279781301)
--(axis cs:41,5.71146540696046)
--(axis cs:40,5.75379661464925)
--(axis cs:39,5.75229345878656)
--(axis cs:38,5.80672166220362)
--(axis cs:37,5.78649233331086)
--(axis cs:36,5.78563226061041)
--(axis cs:35,5.76702603941188)
--(axis cs:34,5.80814828337195)
--(axis cs:33,5.86753131810761)
--(axis cs:32,5.80913739681844)
--(axis cs:31,5.85993507027056)
--(axis cs:30,5.87442036510111)
--(axis cs:29,5.83465932053619)
--(axis cs:28,5.86135313046993)
--(axis cs:27,5.8564219441062)
--(axis cs:26,5.92712133886)
--(axis cs:25,5.95816422737455)
--(axis cs:24,5.9835142070651)
--(axis cs:23,6.09961731428987)
--(axis cs:22,6.07216024691331)
--(axis cs:21,6.02552513602894)
--(axis cs:20,6.08451901136275)
--(axis cs:19,6.14478398821941)
--(axis cs:18,6.10071123235372)
--(axis cs:17,6.13110773025953)
--(axis cs:16,6.22018196560168)
--(axis cs:15,6.26413143439501)
--(axis cs:14,6.30087146438746)
--(axis cs:13,6.34247245392427)
--(axis cs:12,6.32411451968684)
--(axis cs:11,6.32577033399705)
--(axis cs:10,6.37103907534369)
--(axis cs:9,6.30888486181956)
--(axis cs:8,6.49644513011258)
--(axis cs:7,6.29352306007323)
--(axis cs:6,6.24149172623603)
--(axis cs:5,6)
--cycle;

\path [fill=white!49.8039215686275!black, fill opacity=0.2]
(axis cs:5,6)
--(axis cs:5,6)
--(axis cs:6,6.03740565632728)
--(axis cs:7,5.99518427358397)
--(axis cs:8,5.9548260249951)
--(axis cs:9,5.84512900489187)
--(axis cs:10,6.0051320278732)
--(axis cs:11,6.03344226988931)
--(axis cs:12,6.20274319383821)
--(axis cs:13,6.07113824667243)
--(axis cs:14,5.94317019937321)
--(axis cs:15,6.01573510752953)
--(axis cs:16,6.05947294585328)
--(axis cs:17,6.04818984866122)
--(axis cs:18,6.02247664298772)
--(axis cs:19,5.99648725234935)
--(axis cs:20,5.95264304331357)
--(axis cs:21,5.85669168427338)
--(axis cs:22,5.80488718100496)
--(axis cs:23,5.78350957139018)
--(axis cs:24,5.77938021356219)
--(axis cs:25,5.83983306304207)
--(axis cs:26,5.8603276994014)
--(axis cs:27,5.83703830602429)
--(axis cs:28,5.86105900463015)
--(axis cs:29,5.71966886070897)
--(axis cs:30,5.67085978328829)
--(axis cs:31,5.62824619020486)
--(axis cs:32,5.64966874898184)
--(axis cs:33,5.60576879984825)
--(axis cs:34,5.65321910395189)
--(axis cs:35,5.65578199998324)
--(axis cs:36,5.64023952462094)
--(axis cs:37,5.53007830602574)
--(axis cs:38,5.51175421327369)
--(axis cs:39,5.51169008599383)
--(axis cs:40,5.55157296258541)
--(axis cs:41,5.53388880973679)
--(axis cs:42,5.43666737054259)
--(axis cs:43,5.43705839938531)
--(axis cs:44,5.45576226604818)
--(axis cs:45,5.45300988676304)
--(axis cs:46,5.4828119648917)
--(axis cs:47,5.41593504630439)
--(axis cs:48,5.36462771376649)
--(axis cs:49,5.33054550004278)
--(axis cs:50,5.31846982315649)
--(axis cs:50,5.39525566703958)
--(axis cs:50,5.39525566703958)
--(axis cs:49,5.42239567642781)
--(axis cs:48,5.48635267839038)
--(axis cs:47,5.53308456153875)
--(axis cs:46,5.56032529001026)
--(axis cs:45,5.54699011323696)
--(axis cs:44,5.58737498885379)
--(axis cs:43,5.56686316924214)
--(axis cs:42,5.54372478632016)
--(axis cs:41,5.59552295496909)
--(axis cs:40,5.66019174329694)
--(axis cs:39,5.61772167871205)
--(axis cs:38,5.6255006886871)
--(axis cs:37,5.65031385083701)
--(axis cs:36,5.75976047537906)
--(axis cs:35,5.79519839217362)
--(axis cs:34,5.78991815095007)
--(axis cs:33,5.76285865113214)
--(axis cs:32,5.84837046670444)
--(axis cs:31,5.8384204764618)
--(axis cs:30,5.8977676676921)
--(axis cs:29,6.00582133536946)
--(axis cs:28,6.13109785811495)
--(axis cs:27,6.1159028704463)
--(axis cs:26,6.09261347706919)
--(axis cs:25,6.07389242715401)
--(axis cs:24,6.01669821781036)
--(axis cs:23,6.05962768351178)
--(axis cs:22,6.07354419154406)
--(axis cs:21,6.14330831572662)
--(axis cs:20,6.27088636845114)
--(axis cs:19,6.30939510059182)
--(axis cs:18,6.36968021975738)
--(axis cs:17,6.38318270035839)
--(axis cs:16,6.39935058355849)
--(axis cs:15,6.38034332384302)
--(axis cs:14,6.2725160751366)
--(axis cs:13,6.27395979254326)
--(axis cs:12,6.36588425714218)
--(axis cs:11,6.26067537716951)
--(axis cs:10,6.19878954075425)
--(axis cs:9,6.07251805393166)
--(axis cs:8,6.11184064167157)
--(axis cs:7,6.18520788327878)
--(axis cs:6,6.0998492456335)
--(axis cs:5,6)
--cycle;

\path [fill=color7, fill opacity=0.2]
(axis cs:5,6)
--(axis cs:5,6)
--(axis cs:6,5.94454272219348)
--(axis cs:7,5.86847457840163)
--(axis cs:8,5.84216938115332)
--(axis cs:9,5.89866861009166)
--(axis cs:10,5.78461205015314)
--(axis cs:11,5.74375624561783)
--(axis cs:12,5.74280409385349)
--(axis cs:13,5.67504867521007)
--(axis cs:14,5.67433987818711)
--(axis cs:15,5.56520326573648)
--(axis cs:16,5.63911274640832)
--(axis cs:17,5.69595695052748)
--(axis cs:18,5.68335122130123)
--(axis cs:19,5.63031326178297)
--(axis cs:20,5.49019543137522)
--(axis cs:21,5.47950099400957)
--(axis cs:22,5.51032285398408)
--(axis cs:23,5.46392472891674)
--(axis cs:24,5.50416057569389)
--(axis cs:25,5.36700027222606)
--(axis cs:26,5.32635788837016)
--(axis cs:27,5.26679501556143)
--(axis cs:28,5.33265284994893)
--(axis cs:29,5.34587216906232)
--(axis cs:30,5.32145539871765)
--(axis cs:31,5.31816104522248)
--(axis cs:32,5.2131743466638)
--(axis cs:33,5.1642989551353)
--(axis cs:34,5.14977912478823)
--(axis cs:35,5.09023722709079)
--(axis cs:36,5.00341674973336)
--(axis cs:37,5.03660306685454)
--(axis cs:38,5.05715073468453)
--(axis cs:39,5.06295179938602)
--(axis cs:40,4.99419452518262)
--(axis cs:41,4.99090877685124)
--(axis cs:42,4.91521032362001)
--(axis cs:43,4.94884948313656)
--(axis cs:44,4.96823680780705)
--(axis cs:45,4.95564911781045)
--(axis cs:46,4.96503247865176)
--(axis cs:47,4.94345835699879)
--(axis cs:48,4.97336102863502)
--(axis cs:49,4.98598404573015)
--(axis cs:50,4.98328475194428)
--(axis cs:50,5.36181328727141)
--(axis cs:50,5.36181328727141)
--(axis cs:49,5.37872183662279)
--(axis cs:48,5.37173701058067)
--(axis cs:47,5.33889458417768)
--(axis cs:46,5.37614399193648)
--(axis cs:45,5.40513519591504)
--(axis cs:44,5.41607691768315)
--(axis cs:43,5.38056228156932)
--(axis cs:42,5.35145634304665)
--(axis cs:41,5.3424245564821)
--(axis cs:40,5.37835449442522)
--(axis cs:39,5.4076364359081)
--(axis cs:38,5.42128063786449)
--(axis cs:37,5.39869105079252)
--(axis cs:36,5.37305383850194)
--(axis cs:35,5.42740983173274)
--(axis cs:34,5.50512283599608)
--(axis cs:33,5.52589712329607)
--(axis cs:32,5.52800212392443)
--(axis cs:31,5.55634875869908)
--(axis cs:30,5.52952499343921)
--(axis cs:29,5.5560886152514)
--(axis cs:28,5.53793538534519)
--(axis cs:27,5.49006772953661)
--(axis cs:26,5.57168132731612)
--(axis cs:25,5.58986247287198)
--(axis cs:24,5.73113354195317)
--(axis cs:23,5.70470272206365)
--(axis cs:22,5.66614773425121)
--(axis cs:21,5.69696959422572)
--(axis cs:20,5.64705947058557)
--(axis cs:19,5.73047105194252)
--(axis cs:18,5.79508015124779)
--(axis cs:17,5.7942391279039)
--(axis cs:16,5.8628480379054)
--(axis cs:15,5.88577712642039)
--(axis cs:14,6.01585620024427)
--(axis cs:13,5.9837748542017)
--(axis cs:12,6.11994100418573)
--(axis cs:11,6.09153787202923)
--(axis cs:10,6.11734873416059)
--(axis cs:9,6.12878237030049)
--(axis cs:8,6.04802669727805)
--(axis cs:7,6.05701561767681)
--(axis cs:6,6.09859453270849)
--(axis cs:5,6)
--cycle;

\path [fill=color8, fill opacity=0.2]
(axis cs:5,6)
--(axis cs:5,6)
--(axis cs:6,5.58590073408666)
--(axis cs:7,5.37707108642562)
--(axis cs:8,5.41255224749991)
--(axis cs:9,5.2036476187884)
--(axis cs:10,5.16657465124009)
--(axis cs:11,5.18577446440831)
--(axis cs:12,5.13633984905933)
--(axis cs:13,5.22050735107612)
--(axis cs:14,5.32492365453423)
--(axis cs:15,5.27522730656447)
--(axis cs:16,5.36950597066186)
--(axis cs:17,5.35851641074336)
--(axis cs:18,5.35270217274582)
--(axis cs:19,5.3489474526332)
--(axis cs:20,5.33705870823109)
--(axis cs:21,5.21647810459874)
--(axis cs:22,5.16351180415594)
--(axis cs:23,5.07853252586378)
--(axis cs:24,5.15784921555135)
--(axis cs:25,5.14597442237154)
--(axis cs:26,5.19745774430833)
--(axis cs:27,5.23508681188424)
--(axis cs:28,5.1877919934695)
--(axis cs:29,5.18224681112947)
--(axis cs:30,5.22742861180939)
--(axis cs:31,5.208913770119)
--(axis cs:32,5.17779939374594)
--(axis cs:33,5.17041986149635)
--(axis cs:34,5.19996006681098)
--(axis cs:35,5.22513386733949)
--(axis cs:36,5.18521842472671)
--(axis cs:37,5.16660793727454)
--(axis cs:38,5.07383832671294)
--(axis cs:39,5.11274466893612)
--(axis cs:40,5.01943432743674)
--(axis cs:41,5.03646148083244)
--(axis cs:42,5.0756402514606)
--(axis cs:43,5.11231808207752)
--(axis cs:44,5.13793022018367)
--(axis cs:45,5.06612839954056)
--(axis cs:46,5.11538872960163)
--(axis cs:47,5.12414882493779)
--(axis cs:48,5.14222679499578)
--(axis cs:49,5.18096812028941)
--(axis cs:50,5.20891606552217)
--(axis cs:50,5.37931922859548)
--(axis cs:50,5.37931922859548)
--(axis cs:49,5.36020835029882)
--(axis cs:48,5.34796928343559)
--(axis cs:47,5.32290999859162)
--(axis cs:46,5.34735636843759)
--(axis cs:45,5.27504807104767)
--(axis cs:44,5.37579527001241)
--(axis cs:43,5.3739564277264)
--(axis cs:42,5.33612445442176)
--(axis cs:41,5.27726400936364)
--(axis cs:40,5.28644802550444)
--(axis cs:39,5.3500004291031)
--(axis cs:38,5.33400481054196)
--(axis cs:37,5.38633323919605)
--(axis cs:36,5.41086000664584)
--(axis cs:35,5.41408181893502)
--(axis cs:34,5.38043209005177)
--(axis cs:33,5.37467817771934)
--(axis cs:32,5.33200452782268)
--(axis cs:31,5.36363524948884)
--(axis cs:30,5.41178707446512)
--(axis cs:29,5.39030220847838)
--(axis cs:28,5.4278942810403)
--(axis cs:27,5.42765828615498)
--(axis cs:26,5.38293441255441)
--(axis cs:25,5.37951577370689)
--(axis cs:24,5.453915490331)
--(axis cs:23,5.38029100354798)
--(axis cs:22,5.43648819584406)
--(axis cs:21,5.43842385618557)
--(axis cs:20,5.55313737020029)
--(axis cs:19,5.52164078266092)
--(axis cs:18,5.54141547431301)
--(axis cs:17,5.53560123631547)
--(axis cs:16,5.59519991169108)
--(axis cs:15,5.46594916402377)
--(axis cs:14,5.53389987487753)
--(axis cs:13,5.56772794304153)
--(axis cs:12,5.46758171956812)
--(axis cs:11,5.51226475127796)
--(axis cs:10,5.55499397621089)
--(axis cs:9,5.54145042042728)
--(axis cs:8,5.8266634387746)
--(axis cs:7,5.66998773710379)
--(axis cs:6,5.71998161885452)
--(axis cs:5,6)
--cycle;

\path [fill=color0, fill opacity=0.2]
(axis cs:5,6)
--(axis cs:5,6)
--(axis cs:6,5.2503049939105)
--(axis cs:7,4.96842482419009)
--(axis cs:8,5.02059325380259)
--(axis cs:9,5.2128343410577)
--(axis cs:10,5.33128846051119)
--(axis cs:11,5.30033903859759)
--(axis cs:12,5.32539892819818)
--(axis cs:13,5.40490632737357)
--(axis cs:14,5.42626415829823)
--(axis cs:15,5.35601261868245)
--(axis cs:16,5.16814194451981)
--(axis cs:17,5.18146876923684)
--(axis cs:18,5.21238023504352)
--(axis cs:19,5.25219596014052)
--(axis cs:20,5.27375455899666)
--(axis cs:21,5.37002356175577)
--(axis cs:22,5.35000255943864)
--(axis cs:23,5.43545387003575)
--(axis cs:24,5.46656830580459)
--(axis cs:25,5.47446351627629)
--(axis cs:26,5.39702587179687)
--(axis cs:27,5.37960967321956)
--(axis cs:28,5.3930063786209)
--(axis cs:29,5.42425291176335)
--(axis cs:30,5.49737073022099)
--(axis cs:31,5.52055916504483)
--(axis cs:32,5.55347594850151)
--(axis cs:33,5.52176314691464)
--(axis cs:34,5.5670493523768)
--(axis cs:35,5.61703187873519)
--(axis cs:36,5.60594673692549)
--(axis cs:37,5.57184340958636)
--(axis cs:38,5.60131319500048)
--(axis cs:39,5.57457669931176)
--(axis cs:40,5.56864969765597)
--(axis cs:41,5.6161773124206)
--(axis cs:42,5.65882425942859)
--(axis cs:43,5.66711749873164)
--(axis cs:44,5.69413047699527)
--(axis cs:45,5.77018913816036)
--(axis cs:46,5.76711082801562)
--(axis cs:47,5.8012166951149)
--(axis cs:48,5.80204467077779)
--(axis cs:49,5.78646850513641)
--(axis cs:50,5.79667207281606)
--(axis cs:50,6.08175929973296)
--(axis cs:50,6.08175929973296)
--(axis cs:49,6.09196286741261)
--(axis cs:48,6.09207297628103)
--(axis cs:47,6.08505781468902)
--(axis cs:46,6.06818328963144)
--(axis cs:45,6.06118341085925)
--(axis cs:44,6.00783030731845)
--(axis cs:43,5.98386289342522)
--(axis cs:42,5.96470515233612)
--(axis cs:41,5.92499915816764)
--(axis cs:40,5.88233069450089)
--(axis cs:39,5.8764036928451)
--(axis cs:38,5.89280445205835)
--(axis cs:37,5.82031345315874)
--(axis cs:36,5.81366110621177)
--(axis cs:35,5.79473282714716)
--(axis cs:34,5.76236241232909)
--(axis cs:33,5.74490351975202)
--(axis cs:32,5.73279856130242)
--(axis cs:31,5.72257808985713)
--(axis cs:30,5.6869429952692)
--(axis cs:29,5.62280591176606)
--(axis cs:28,5.5991504841242)
--(axis cs:27,5.57725307187848)
--(axis cs:26,5.61473883408548)
--(axis cs:25,5.63926197391979)
--(axis cs:24,5.64323561576403)
--(axis cs:23,5.5606245613368)
--(axis cs:22,5.57940920526724)
--(axis cs:21,5.58683918334227)
--(axis cs:20,5.49095132335628)
--(axis cs:19,5.45760796142811)
--(axis cs:18,5.43075701985844)
--(axis cs:17,5.43421750527296)
--(axis cs:16,5.50244629077431)
--(axis cs:15,5.647908949945)
--(axis cs:14,5.75804956719197)
--(axis cs:13,5.70489759419506)
--(axis cs:12,5.5961696992528)
--(axis cs:11,5.56240605944163)
--(axis cs:10,5.48439781399861)
--(axis cs:9,5.42245977658936)
--(axis cs:8,5.27744596188369)
--(axis cs:7,5.19628105816285)
--(axis cs:6,5.5496950060895)
--(axis cs:5,6)
--cycle;

\addplot [line width=2pt, color0]
table {%
5 6
6 6.64705882352941
7 6.90980392156863
8 6.94901960784314
9 7.0156862745098
10 7.11764705882353
11 7.36274509803922
12 7.47058823529412
13 7.61764705882353
14 7.68039215686275
15 7.85098039215686
16 7.97647058823529
17 8.06274509803922
18 8.15098039215686
19 8.17843137254902
20 8.33137254901961
21 8.4156862745098
22 8.46078431372549
23 8.54901960784314
24 8.53921568627451
25 8.62941176470588
26 8.64901960784314
27 8.74509803921569
28 8.71960784313725
29 8.77058823529412
30 8.8156862745098
31 8.80588235294118
32 8.82941176470588
33 8.81764705882353
34 8.85294117647059
35 8.85686274509804
36 8.89411764705882
37 8.92745098039216
38 8.95686274509804
39 8.97647058823529
40 9.00196078431373
41 9.02352941176471
42 8.9921568627451
43 9.00392156862745
44 8.98235294117647
45 9.02156862745098
46 9.03137254901961
47 9.06274509803922
48 9.07843137254902
49 9.11176470588235
50 9.13333333333333
};

\addplot [line width=2pt, color1]
table {%
5 6
6 6.43137254901961
7 6.63921568627451
8 6.68039215686275
9 6.87058823529412
10 6.94705882352941
11 6.82941176470588
12 6.64117647058824
13 6.42745098039216
14 6.40980392156863
15 6.34901960784314
16 6.28235294117647
17 6.29411764705882
18 6.25294117647059
19 6.20392156862745
20 6.05882352941176
21 5.93529411764706
22 5.90196078431373
23 5.83333333333333
24 5.7843137254902
25 5.77058823529412
26 5.73725490196078
27 5.74705882352941
28 5.67058823529412
29 5.7156862745098
30 5.6156862745098
31 5.59803921568627
32 5.59019607843137
33 5.60588235294118
34 5.55490196078431
35 5.54705882352941
36 5.54705882352941
37 5.51960784313725
38 5.53921568627451
39 5.54509803921569
40 5.49019607843137
41 5.51372549019608
42 5.54313725490196
43 5.58039215686274
44 5.59607843137255
45 5.51764705882353
46 5.42156862745098
47 5.43921568627451
48 5.44313725490196
49 5.45098039215686
50 5.43137254901961
};
\addplot [line width=2pt, color2]
table {%
5 6
6 5.6921568627451
7 5.65882352941177
8 5.55882352941176
9 5.52549019607843
10 5.40392156862745
11 5.42745098039216
12 5.5156862745098
13 5.56666666666667
14 5.58823529411765
15 5.67058823529412
16 5.67254901960784
17 5.77450980392157
18 5.78039215686274
19 5.78823529411765
20 5.85686274509804
21 5.83333333333333
22 5.9
23 5.93921568627451
24 5.95294117647059
25 6.01372549019608
26 6.07647058823529
27 6.10392156862745
28 6.15098039215686
29 6.18039215686275
30 6.17254901960784
31 6.19803921568627
32 6.21372549019608
33 6.25686274509804
34 6.27450980392157
35 6.23725490196078
36 6.20980392156863
37 6.23529411764706
38 6.21372549019608
39 6.23137254901961
40 6.25294117647059
41 6.23137254901961
42 6.3078431372549
43 6.27058823529412
44 6.18627450980392
45 6.23921568627451
46 6.26666666666667
47 6.27254901960784
48 6.24117647058824
49 6.25098039215686
50 6.26666666666667
};
\addplot [line width=2pt, color3]
table {%
5 6
6 5.55098039215686
7 5.54313725490196
8 5.47254901960784
9 5.38235294117647
10 5.47647058823529
11 5.47058823529412
12 5.44509803921569
13 5.34509803921569
14 5.41764705882353
15 5.49803921568627
16 5.55098039215686
17 5.5843137254902
18 5.54705882352941
19 5.57450980392157
20 5.64117647058824
21 5.65098039215686
22 5.65686274509804
23 5.6078431372549
24 5.63137254901961
25 5.63921568627451
26 5.67058823529412
27 5.65490196078431
28 5.70196078431373
29 5.75490196078431
30 5.82745098039216
31 5.87254901960784
32 5.91176470588235
33 5.92941176470588
34 5.8921568627451
35 5.91176470588235
36 5.94117647058824
37 5.94509803921569
38 5.90588235294118
39 5.96078431372549
40 6.01960784313725
41 6.05294117647059
42 6.06666666666667
43 6.06862745098039
44 6.06274509803922
45 6.05686274509804
46 6.08627450980392
47 6.02549019607843
48 6.05882352941176
49 6.03137254901961
50 6.03333333333333
};
\addplot [line width=2pt, color4]
table {%
5 6
6 6.23333333333333
7 6.18235294117647
8 5.81764705882353
9 5.85098039215686
10 5.66274509803922
11 5.75882352941176
12 5.6843137254902
13 5.74313725490196
14 5.65294117647059
15 5.65490196078431
16 5.69019607843137
17 5.6843137254902
18 5.71176470588235
19 5.79411764705882
20 5.9078431372549
21 5.98039215686275
22 5.95686274509804
23 6.02156862745098
24 6.04901960784314
25 6.02941176470588
26 6.08823529411765
27 6.14705882352941
28 6.13529411764706
29 6.15490196078431
30 6.12549019607843
31 6.14901960784314
32 6.14509803921569
33 6.13921568627451
34 6.15882352941177
35 6.16274509803922
36 6.18039215686275
37 6.23333333333333
38 6.26274509803922
39 6.24901960784314
40 6.24705882352941
41 6.25098039215686
42 6.22549019607843
43 6.23921568627451
44 6.25490196078431
45 6.26078431372549
46 6.26666666666667
47 6.28039215686274
48 6.27254901960784
49 6.25490196078431
50 6.26274509803922
};
\addplot [line width=2pt, color5]
table {%
5 6
6 6.13725490196078
7 6.1843137254902
8 6.36862745098039
9 6.42156862745098
10 6.26666666666667
11 6.11960784313725
12 6.04901960784314
13 6.08823529411765
14 6.05490196078431
15 5.99803921568627
16 5.8843137254902
17 5.83333333333333
18 5.83921568627451
19 5.83137254901961
20 5.77058823529412
21 5.88627450980392
22 5.87254901960784
23 5.81372549019608
24 5.79019607843137
25 5.80588235294118
26 5.78823529411765
27 5.72549019607843
28 5.64901960784314
29 5.56862745098039
30 5.56078431372549
31 5.53333333333333
32 5.56862745098039
33 5.54705882352941
34 5.55490196078431
35 5.59019607843137
36 5.62352941176471
37 5.65686274509804
38 5.64509803921569
39 5.62549019607843
40 5.65098039215686
41 5.66078431372549
42 5.61372549019608
43 5.56470588235294
44 5.56274509803922
45 5.60196078431373
46 5.55882352941176
47 5.5843137254902
48 5.58823529411765
49 5.58235294117647
50 5.6156862745098
};
\addplot [line width=2pt, color6]
table {%
5 6
6 6.16470588235294
7 6.22352941176471
8 6.40588235294118
9 6.27058823529412
10 6.30392156862745
11 6.18627450980392
12 6.2156862745098
13 6.26078431372549
14 6.22156862745098
15 6.18235294117647
16 6.14509803921569
17 6.05098039215686
18 6.01372549019608
19 6.00588235294118
20 5.92549019607843
21 5.90392156862745
22 5.95882352941176
23 6.00196078431373
24 5.87647058823529
25 5.85686274509804
26 5.76862745098039
27 5.71176470588235
28 5.73725490196078
29 5.73137254901961
30 5.76078431372549
31 5.76470588235294
32 5.72352941176471
33 5.76862745098039
34 5.7078431372549
35 5.6843137254902
36 5.7078431372549
37 5.70196078431373
38 5.71764705882353
39 5.65490196078431
40 5.66666666666667
41 5.6078431372549
42 5.60980392156863
43 5.53725490196078
44 5.53333333333333
45 5.53529411764706
46 5.52745098039216
47 5.55294117647059
48 5.52745098039216
49 5.54901960784314
50 5.49411764705882
};
\addplot [line width=2pt, white!49.8039215686275!black]
table {%
5 6
6 6.06862745098039
7 6.09019607843137
8 6.03333333333333
9 5.95882352941176
10 6.10196078431373
11 6.14705882352941
12 6.2843137254902
13 6.17254901960784
14 6.1078431372549
15 6.19803921568627
16 6.22941176470588
17 6.2156862745098
18 6.19607843137255
19 6.15294117647059
20 6.11176470588235
21 6
22 5.93921568627451
23 5.92156862745098
24 5.89803921568627
25 5.95686274509804
26 5.97647058823529
27 5.97647058823529
28 5.99607843137255
29 5.86274509803922
30 5.7843137254902
31 5.73333333333333
32 5.74901960784314
33 5.6843137254902
34 5.72156862745098
35 5.72549019607843
36 5.7
37 5.59019607843137
38 5.56862745098039
39 5.56470588235294
40 5.60588235294118
41 5.56470588235294
42 5.49019607843137
43 5.50196078431373
44 5.52156862745098
45 5.5
46 5.52156862745098
47 5.47450980392157
48 5.42549019607843
49 5.37647058823529
50 5.35686274509804
};
\addplot [line width=2pt, color7]
table {%
5 6
6 6.02156862745098
7 5.96274509803922
8 5.94509803921569
9 6.01372549019608
10 5.95098039215686
11 5.91764705882353
12 5.93137254901961
13 5.82941176470588
14 5.84509803921569
15 5.72549019607843
16 5.75098039215686
17 5.74509803921569
18 5.73921568627451
19 5.68039215686275
20 5.56862745098039
21 5.58823529411765
22 5.58823529411765
23 5.5843137254902
24 5.61764705882353
25 5.47843137254902
26 5.44901960784314
27 5.37843137254902
28 5.43529411764706
29 5.45098039215686
30 5.42549019607843
31 5.43725490196078
32 5.37058823529412
33 5.34509803921569
34 5.32745098039216
35 5.25882352941176
36 5.18823529411765
37 5.21764705882353
38 5.23921568627451
39 5.23529411764706
40 5.18627450980392
41 5.16666666666667
42 5.13333333333333
43 5.16470588235294
44 5.1921568627451
45 5.18039215686275
46 5.17058823529412
47 5.14117647058824
48 5.17254901960784
49 5.18235294117647
50 5.17254901960784
};
\addplot [line width=2pt, color8]
table {%
5 6
6 5.65294117647059
7 5.52352941176471
8 5.61960784313725
9 5.37254901960784
10 5.36078431372549
11 5.34901960784314
12 5.30196078431373
13 5.39411764705882
14 5.42941176470588
15 5.37058823529412
16 5.48235294117647
17 5.44705882352941
18 5.44705882352941
19 5.43529411764706
20 5.44509803921569
21 5.32745098039216
22 5.3
23 5.22941176470588
24 5.30588235294118
25 5.26274509803922
26 5.29019607843137
27 5.33137254901961
28 5.3078431372549
29 5.28627450980392
30 5.31960784313725
31 5.28627450980392
32 5.25490196078431
33 5.27254901960784
34 5.29019607843137
35 5.31960784313725
36 5.29803921568627
37 5.27647058823529
38 5.20392156862745
39 5.23137254901961
40 5.15294117647059
41 5.15686274509804
42 5.20588235294118
43 5.24313725490196
44 5.25686274509804
45 5.17058823529412
46 5.23137254901961
47 5.22352941176471
48 5.24509803921569
49 5.27058823529412
50 5.29411764705882
};
\addplot [line width=2pt, colorHEBO]
table {%
5 6
6 5.4
7 5.08235294117647
8 5.14901960784314
9 5.31764705882353
10 5.4078431372549
11 5.43137254901961
12 5.46078431372549
13 5.55490196078431
14 5.5921568627451
15 5.50196078431373
16 5.33529411764706
17 5.3078431372549
18 5.32156862745098
19 5.35490196078431
20 5.38235294117647
21 5.47843137254902
22 5.46470588235294
23 5.49803921568627
24 5.55490196078431
25 5.55686274509804
26 5.50588235294118
27 5.47843137254902
28 5.49607843137255
29 5.52352941176471
30 5.5921568627451
31 5.62156862745098
32 5.64313725490196
33 5.63333333333333
34 5.66470588235294
35 5.70588235294118
36 5.70980392156863
37 5.69607843137255
38 5.74705882352941
39 5.72549019607843
40 5.72549019607843
41 5.77058823529412
42 5.81176470588235
43 5.82549019607843
44 5.85098039215686
45 5.9156862745098
46 5.91764705882353
47 5.94313725490196
48 5.94705882352941
49 5.93921568627451
50 5.93921568627451
};
\end{axis}

\end{tikzpicture}

%% file: icml2023/figures/hpob_heboabl_regret_test.tex
\begin{tikzpicture}

\begin{axis}[
legend cell align={left},
legend style={
  fill opacity=0.8,
  draw opacity=1,
  text opacity=1,
  at={(0.55,-0.15)},
  anchor=south,
  draw=white!80!black
},
log basis y={10},
tick align=outside,
tick pos=left,
x grid style={white!69.0196078431373!black},
xlabel={Number of trials},
xmin=2.75, xmax=52.25,
xtick style={color=black},
  height=.6\textwidth,
    width=.8\textwidth,
y grid style={white!69.0196078431373!black},
ylabel={Average Regret},
ymin=0.0153171731609242, ymax=0.261636933606157,
ymode=log,
ytick style={color=black}
]
\path [fill=color0, fill opacity=0.2]
(axis cs:5,0.229972229042153)
--(axis cs:5,0.21710238749041)
--(axis cs:6,0.210551931890934)
--(axis cs:7,0.191336856479292)
--(axis cs:8,0.18398611450507)
--(axis cs:9,0.173174426009038)
--(axis cs:10,0.161566672085494)
--(axis cs:11,0.153221170289856)
--(axis cs:12,0.147103225495095)
--(axis cs:13,0.144964010471113)
--(axis cs:14,0.133513953193397)
--(axis cs:15,0.130869269073569)
--(axis cs:16,0.129588704326338)
--(axis cs:17,0.129056064670601)
--(axis cs:18,0.128240983502576)
--(axis cs:19,0.119655546102882)
--(axis cs:20,0.119261341744938)
--(axis cs:21,0.118952916746962)
--(axis cs:22,0.115740984667637)
--(axis cs:23,0.113822815372513)
--(axis cs:24,0.111239996489407)
--(axis cs:25,0.111109586652343)
--(axis cs:26,0.110677655794738)
--(axis cs:27,0.110054406629553)
--(axis cs:28,0.10869636607477)
--(axis cs:29,0.105425670492834)
--(axis cs:30,0.104148807039482)
--(axis cs:31,0.103036857102853)
--(axis cs:32,0.102326505972465)
--(axis cs:33,0.101512284262623)
--(axis cs:34,0.101273140087157)
--(axis cs:35,0.101095535190606)
--(axis cs:36,0.100374825444411)
--(axis cs:37,0.0987592666410219)
--(axis cs:38,0.0985667843139294)
--(axis cs:39,0.0983883405881294)
--(axis cs:40,0.0982740935933329)
--(axis cs:41,0.097971176092753)
--(axis cs:42,0.0978147028919637)
--(axis cs:43,0.0970845745081627)
--(axis cs:44,0.0967919127817516)
--(axis cs:45,0.0967251933978113)
--(axis cs:46,0.0964900529287663)
--(axis cs:47,0.0948039725845666)
--(axis cs:48,0.0914377093499991)
--(axis cs:49,0.0913116376456255)
--(axis cs:50,0.0909194074778135)
--(axis cs:50,0.102760161966289)
--(axis cs:50,0.102760161966289)
--(axis cs:49,0.10355391050267)
--(axis cs:48,0.103618894177447)
--(axis cs:47,0.105118030426644)
--(axis cs:46,0.105890409431554)
--(axis cs:45,0.106714359948635)
--(axis cs:44,0.106830096253732)
--(axis cs:43,0.107401874324678)
--(axis cs:42,0.109019289222573)
--(axis cs:41,0.109232186712493)
--(axis cs:40,0.110249516016526)
--(axis cs:39,0.110511645921156)
--(axis cs:38,0.110593493901512)
--(axis cs:37,0.110726131211684)
--(axis cs:36,0.112007148271516)
--(axis cs:35,0.112514679946367)
--(axis cs:34,0.112607841546255)
--(axis cs:33,0.112700097536489)
--(axis cs:32,0.113557523551764)
--(axis cs:31,0.115252610748441)
--(axis cs:30,0.115924337046734)
--(axis cs:29,0.116760957112317)
--(axis cs:28,0.118213697427976)
--(axis cs:27,0.120825416295935)
--(axis cs:26,0.121896892724957)
--(axis cs:25,0.122429135447138)
--(axis cs:24,0.122801304845827)
--(axis cs:23,0.124100798337417)
--(axis cs:22,0.129089778757046)
--(axis cs:21,0.133831452125081)
--(axis cs:20,0.133983446483229)
--(axis cs:19,0.134736088401152)
--(axis cs:18,0.139933467453242)
--(axis cs:17,0.14085611822347)
--(axis cs:16,0.141347747166577)
--(axis cs:15,0.142075058190751)
--(axis cs:14,0.145751665683148)
--(axis cs:13,0.152858007901119)
--(axis cs:12,0.15497155554015)
--(axis cs:11,0.164415684662798)
--(axis cs:10,0.170072803621721)
--(axis cs:9,0.180017514871002)
--(axis cs:8,0.186814074102229)
--(axis cs:7,0.196881831709089)
--(axis cs:6,0.219660583406081)
--(axis cs:5,0.229972229042153)
--cycle;

\path [fill=color1, fill opacity=0.2]
(axis cs:5,0.229972229042153)
--(axis cs:5,0.21710238749041)
--(axis cs:6,0.206767140505162)
--(axis cs:7,0.200269236542652)
--(axis cs:8,0.191893881430571)
--(axis cs:9,0.178401295834877)
--(axis cs:10,0.172228709564284)
--(axis cs:11,0.160242156579528)
--(axis cs:12,0.148288228185233)
--(axis cs:13,0.13706877128598)
--(axis cs:14,0.131755403377912)
--(axis cs:15,0.124634753528544)
--(axis cs:16,0.122314300804427)
--(axis cs:17,0.117386177238596)
--(axis cs:18,0.109443737179489)
--(axis cs:19,0.107964035869148)
--(axis cs:20,0.101478655615687)
--(axis cs:21,0.0894026065920785)
--(axis cs:22,0.0834904358748244)
--(axis cs:23,0.0810612376894862)
--(axis cs:24,0.0792019900308486)
--(axis cs:25,0.077272045739271)
--(axis cs:26,0.0743406880548449)
--(axis cs:27,0.0723679242034906)
--(axis cs:28,0.0692501331132998)
--(axis cs:29,0.0679576455343032)
--(axis cs:30,0.0642696604071299)
--(axis cs:31,0.0619732061149486)
--(axis cs:32,0.0588216188244668)
--(axis cs:33,0.0575674376163326)
--(axis cs:34,0.0543787145588783)
--(axis cs:35,0.0542253016853686)
--(axis cs:36,0.052423616783675)
--(axis cs:37,0.0507972753430667)
--(axis cs:38,0.0447286066464856)
--(axis cs:39,0.0427350322893294)
--(axis cs:40,0.0396577344033624)
--(axis cs:41,0.0388375831557457)
--(axis cs:42,0.0353897302858142)
--(axis cs:43,0.0352365203911538)
--(axis cs:44,0.0342058190248711)
--(axis cs:45,0.0333112179082447)
--(axis cs:46,0.0297218006534051)
--(axis cs:47,0.02925041183829)
--(axis cs:48,0.0289366390239416)
--(axis cs:49,0.0287281894276406)
--(axis cs:50,0.027704668884422)
--(axis cs:50,0.0379314778610262)
--(axis cs:50,0.0379314778610262)
--(axis cs:49,0.0385168128715617)
--(axis cs:48,0.0393725565609319)
--(axis cs:47,0.0395320769678211)
--(axis cs:46,0.039810107941024)
--(axis cs:45,0.0419644805299724)
--(axis cs:44,0.0441549254466072)
--(axis cs:43,0.047486456652137)
--(axis cs:42,0.0489356418855603)
--(axis cs:41,0.0519426310609125)
--(axis cs:40,0.0524295108410772)
--(axis cs:39,0.0560439699596165)
--(axis cs:38,0.0601576956040919)
--(axis cs:37,0.0653047631088551)
--(axis cs:36,0.0665071595620215)
--(axis cs:35,0.0679698668069191)
--(axis cs:34,0.0681561666136415)
--(axis cs:33,0.0698212639436179)
--(axis cs:32,0.0711194692498258)
--(axis cs:31,0.0735111490708296)
--(axis cs:30,0.0756630863393933)
--(axis cs:29,0.0796081418458316)
--(axis cs:28,0.0812864229024938)
--(axis cs:27,0.0834186826816302)
--(axis cs:26,0.0858884492261974)
--(axis cs:25,0.0903980956630372)
--(axis cs:24,0.0918888972060752)
--(axis cs:23,0.0939932055801463)
--(axis cs:22,0.0972281217787287)
--(axis cs:21,0.101010725685203)
--(axis cs:20,0.110119388570072)
--(axis cs:19,0.115691538281368)
--(axis cs:18,0.117482158326457)
--(axis cs:17,0.120996473127716)
--(axis cs:16,0.126426727255947)
--(axis cs:15,0.128761250870552)
--(axis cs:14,0.135914421443853)
--(axis cs:13,0.142248332662099)
--(axis cs:12,0.152950954979235)
--(axis cs:11,0.166301876686783)
--(axis cs:10,0.175840787690108)
--(axis cs:9,0.183015318333832)
--(axis cs:8,0.198153077650713)
--(axis cs:7,0.205652428193418)
--(axis cs:6,0.211559225297995)
--(axis cs:5,0.229972229042153)
--cycle;

\path [fill=color2, fill opacity=0.2]
(axis cs:5,0.229972229042153)
--(axis cs:5,0.21710238749041)
--(axis cs:6,0.200309700446158)
--(axis cs:7,0.172642371797726)
--(axis cs:8,0.143085183754668)
--(axis cs:9,0.126310835209599)
--(axis cs:10,0.111781544013711)
--(axis cs:11,0.104655985146712)
--(axis cs:12,0.0954930207679072)
--(axis cs:13,0.0887366931882569)
--(axis cs:14,0.0855630412434489)
--(axis cs:15,0.0824404849151502)
--(axis cs:16,0.0784176091392623)
--(axis cs:17,0.0768319952522663)
--(axis cs:18,0.0733269428278585)
--(axis cs:19,0.0704938371835569)
--(axis cs:20,0.0681138170508396)
--(axis cs:21,0.0653267719096829)
--(axis cs:22,0.0641226789496801)
--(axis cs:23,0.0614158250634304)
--(axis cs:24,0.0593493787929961)
--(axis cs:25,0.0582443865917597)
--(axis cs:26,0.0568004506012404)
--(axis cs:27,0.0558423988981941)
--(axis cs:28,0.0553442456490849)
--(axis cs:29,0.0544104545936577)
--(axis cs:30,0.0536323661592069)
--(axis cs:31,0.0517465276050149)
--(axis cs:32,0.0507336042348001)
--(axis cs:33,0.0496351817565472)
--(axis cs:34,0.0484386970533811)
--(axis cs:35,0.0463463660032869)
--(axis cs:36,0.0444048631919866)
--(axis cs:37,0.0441747349819475)
--(axis cs:38,0.0436653520525761)
--(axis cs:39,0.0435353016910894)
--(axis cs:40,0.0432458253624085)
--(axis cs:41,0.0417667708372308)
--(axis cs:42,0.0413897535859784)
--(axis cs:43,0.0410878132291635)
--(axis cs:44,0.0384449215455703)
--(axis cs:45,0.0383730844351158)
--(axis cs:46,0.0381951880110114)
--(axis cs:47,0.0377054313396284)
--(axis cs:48,0.0358720138674735)
--(axis cs:49,0.0357519998850474)
--(axis cs:50,0.035500023917921)
--(axis cs:50,0.0414785838658191)
--(axis cs:50,0.0414785838658191)
--(axis cs:49,0.0415709542449192)
--(axis cs:48,0.0420595451521691)
--(axis cs:47,0.0427076898693663)
--(axis cs:46,0.0430664602143396)
--(axis cs:45,0.0440027892782668)
--(axis cs:44,0.0440341908261541)
--(axis cs:43,0.0464411404990339)
--(axis cs:42,0.0466965208392798)
--(axis cs:41,0.0470079338929005)
--(axis cs:40,0.0481436676301634)
--(axis cs:39,0.0484414747204803)
--(axis cs:38,0.0485096306910097)
--(axis cs:37,0.0496542885881978)
--(axis cs:36,0.0502130159902322)
--(axis cs:35,0.0512639533496915)
--(axis cs:34,0.0527011403688273)
--(axis cs:33,0.0581637371210983)
--(axis cs:32,0.0592435461024564)
--(axis cs:31,0.0599278840365871)
--(axis cs:30,0.0625667481418208)
--(axis cs:29,0.0634059744093781)
--(axis cs:28,0.0649432925605156)
--(axis cs:27,0.0656355828718669)
--(axis cs:26,0.066546855501863)
--(axis cs:25,0.0680904189583013)
--(axis cs:24,0.0692755030002074)
--(axis cs:23,0.0712119580874435)
--(axis cs:22,0.0732718818914506)
--(axis cs:21,0.0742441783889251)
--(axis cs:20,0.0781083754973152)
--(axis cs:19,0.0816074647508167)
--(axis cs:18,0.0853813580601277)
--(axis cs:17,0.0894613136000751)
--(axis cs:16,0.0919571557343814)
--(axis cs:15,0.0984375207764523)
--(axis cs:14,0.101314513339932)
--(axis cs:13,0.105894208072124)
--(axis cs:12,0.112036054961722)
--(axis cs:11,0.119415719534675)
--(axis cs:10,0.125867140901724)
--(axis cs:9,0.144155130987797)
--(axis cs:8,0.159529649430728)
--(axis cs:7,0.195512879499695)
--(axis cs:6,0.216515565909631)
--(axis cs:5,0.229972229042153)
--cycle;

\path [fill=color3, fill opacity=0.2]
(axis cs:5,0.229972229042153)
--(axis cs:5,0.21710238749041)
--(axis cs:6,0.162822433883316)
--(axis cs:7,0.139539209343873)
--(axis cs:8,0.119571311398833)
--(axis cs:9,0.100476845568446)
--(axis cs:10,0.0893018934461284)
--(axis cs:11,0.0803298380565958)
--(axis cs:12,0.0740084907070383)
--(axis cs:13,0.0691975747032489)
--(axis cs:14,0.0667881520659029)
--(axis cs:15,0.0632844731668084)
--(axis cs:16,0.0608492413866825)
--(axis cs:17,0.0595477303379055)
--(axis cs:18,0.0576543564766146)
--(axis cs:19,0.0560464667854456)
--(axis cs:20,0.0531854374906936)
--(axis cs:21,0.0512052405958865)
--(axis cs:22,0.0492029111925153)
--(axis cs:23,0.0471314549058904)
--(axis cs:24,0.0456495261502267)
--(axis cs:25,0.0445806292061675)
--(axis cs:26,0.0429737634933274)
--(axis cs:27,0.0391345630836159)
--(axis cs:28,0.0385586734985247)
--(axis cs:29,0.0382896549072748)
--(axis cs:30,0.0378521453828476)
--(axis cs:31,0.037282788838564)
--(axis cs:32,0.0370896090319186)
--(axis cs:33,0.0359482533891422)
--(axis cs:34,0.0334080363755105)
--(axis cs:35,0.0328944874116617)
--(axis cs:36,0.031585738063792)
--(axis cs:37,0.0297598304211245)
--(axis cs:38,0.0278133937588502)
--(axis cs:39,0.0275146103385441)
--(axis cs:40,0.0268849253014019)
--(axis cs:41,0.0263855668282113)
--(axis cs:42,0.0258795309129917)
--(axis cs:43,0.0255397533143883)
--(axis cs:44,0.0251083782107108)
--(axis cs:45,0.0240934740366982)
--(axis cs:46,0.0240210187995755)
--(axis cs:47,0.0237854981920791)
--(axis cs:48,0.0235908514780228)
--(axis cs:49,0.022900708361541)
--(axis cs:50,0.0226726475809663)
--(axis cs:50,0.0304509290957804)
--(axis cs:50,0.0304509290957804)
--(axis cs:49,0.030586024081862)
--(axis cs:48,0.0309909742704933)
--(axis cs:47,0.0310815325475244)
--(axis cs:46,0.0315222446493229)
--(axis cs:45,0.0317484951879258)
--(axis cs:44,0.0334800878216948)
--(axis cs:43,0.0336703436352408)
--(axis cs:42,0.033977415475616)
--(axis cs:41,0.0344604031528995)
--(axis cs:40,0.0347316896577118)
--(axis cs:39,0.0351477971959214)
--(axis cs:38,0.0358081237048363)
--(axis cs:37,0.0373876359444048)
--(axis cs:36,0.0386093155613666)
--(axis cs:35,0.0395452191217254)
--(axis cs:34,0.0401061779547068)
--(axis cs:33,0.0419654791329775)
--(axis cs:32,0.0426141380932124)
--(axis cs:31,0.0428505553013442)
--(axis cs:30,0.0432421031163497)
--(axis cs:29,0.0436759421273459)
--(axis cs:28,0.0440008281216884)
--(axis cs:27,0.0444384004489777)
--(axis cs:26,0.047499593792225)
--(axis cs:25,0.0489702167312815)
--(axis cs:24,0.0500176989533606)
--(axis cs:23,0.0516316461742082)
--(axis cs:22,0.0535708308479751)
--(axis cs:21,0.0566039244409512)
--(axis cs:20,0.0580441963217054)
--(axis cs:19,0.0599638590031843)
--(axis cs:18,0.0635423323752902)
--(axis cs:17,0.0661238700227293)
--(axis cs:16,0.068964215405761)
--(axis cs:15,0.0712236826674407)
--(axis cs:14,0.073434782842495)
--(axis cs:13,0.0756301511979598)
--(axis cs:12,0.0822357917195428)
--(axis cs:11,0.0877323577035962)
--(axis cs:10,0.0988879735817272)
--(axis cs:9,0.106758356254475)
--(axis cs:8,0.128464580868728)
--(axis cs:7,0.147419012832314)
--(axis cs:6,0.17427990721908)
--(axis cs:5,0.229972229042153)
--cycle;

\path [fill=color4, fill opacity=0.2]
(axis cs:5,0.229972229042153)
--(axis cs:5,0.21710238749041)
--(axis cs:6,0.192438086603204)
--(axis cs:7,0.163135106791521)
--(axis cs:8,0.119625231628889)
--(axis cs:9,0.10954466016472)
--(axis cs:10,0.0937855362746626)
--(axis cs:11,0.0890990037746094)
--(axis cs:12,0.0842994862998966)
--(axis cs:13,0.0799689788924252)
--(axis cs:14,0.0726775602503533)
--(axis cs:15,0.0655934380150659)
--(axis cs:16,0.0621944001395246)
--(axis cs:17,0.0596705456249492)
--(axis cs:18,0.0569088993841469)
--(axis cs:19,0.0549917600015294)
--(axis cs:20,0.0541024777120235)
--(axis cs:21,0.0523562637838312)
--(axis cs:22,0.0512910081960112)
--(axis cs:23,0.051222519473203)
--(axis cs:24,0.0507033876968981)
--(axis cs:25,0.0467973465556716)
--(axis cs:26,0.0438422893914974)
--(axis cs:27,0.0418949328471505)
--(axis cs:28,0.0402381110817726)
--(axis cs:29,0.0384319661379432)
--(axis cs:30,0.0370051257173546)
--(axis cs:31,0.0346885178358028)
--(axis cs:32,0.033225398384863)
--(axis cs:33,0.0318517743847542)
--(axis cs:34,0.0316051412611633)
--(axis cs:35,0.0307112364747519)
--(axis cs:36,0.0295322073598076)
--(axis cs:37,0.0288332784607667)
--(axis cs:38,0.0274153342706747)
--(axis cs:39,0.0269797810369456)
--(axis cs:40,0.0263390858289064)
--(axis cs:41,0.0260036035843466)
--(axis cs:42,0.0254192389036474)
--(axis cs:43,0.0253202532230446)
--(axis cs:44,0.0252477695192634)
--(axis cs:45,0.0249521589147681)
--(axis cs:46,0.0245294272177282)
--(axis cs:47,0.0243543125057462)
--(axis cs:48,0.0232888186061312)
--(axis cs:49,0.0229145814456802)
--(axis cs:50,0.0228942178006145)
--(axis cs:50,0.0254795572847339)
--(axis cs:50,0.0254795572847339)
--(axis cs:49,0.0255265864974801)
--(axis cs:48,0.0261582366865377)
--(axis cs:47,0.0266473738663484)
--(axis cs:46,0.0267822617066741)
--(axis cs:45,0.0270721209743868)
--(axis cs:44,0.0275465811126066)
--(axis cs:43,0.0278234829607847)
--(axis cs:42,0.0281733588256886)
--(axis cs:41,0.0288011766382653)
--(axis cs:40,0.0290497796374978)
--(axis cs:39,0.0293921884948253)
--(axis cs:38,0.03032349596395)
--(axis cs:37,0.031582222422669)
--(axis cs:36,0.0333953496127332)
--(axis cs:35,0.0373155792881315)
--(axis cs:34,0.0380465709692343)
--(axis cs:33,0.0382738921160702)
--(axis cs:32,0.0402838636228005)
--(axis cs:31,0.0417920818360335)
--(axis cs:30,0.0436489653083839)
--(axis cs:29,0.0451033453794043)
--(axis cs:28,0.0491933628936531)
--(axis cs:27,0.0506854029036714)
--(axis cs:26,0.0516807720148754)
--(axis cs:25,0.0534535113869909)
--(axis cs:24,0.05548073340081)
--(axis cs:23,0.0561134128982804)
--(axis cs:22,0.0561513687802884)
--(axis cs:21,0.0577997163810045)
--(axis cs:20,0.0598857271602295)
--(axis cs:19,0.0613799246777615)
--(axis cs:18,0.0648945680269161)
--(axis cs:17,0.0670137810815454)
--(axis cs:16,0.0701719251138507)
--(axis cs:15,0.074197467777458)
--(axis cs:14,0.0782485709745389)
--(axis cs:13,0.0854567139715733)
--(axis cs:12,0.0884150537893443)
--(axis cs:11,0.0929517090876777)
--(axis cs:10,0.100657285931842)
--(axis cs:9,0.121865114036954)
--(axis cs:8,0.135250051873942)
--(axis cs:7,0.183693637788156)
--(axis cs:6,0.205858249564612)
--(axis cs:5,0.229972229042153)
--cycle;

\path [fill=color5, fill opacity=0.2]
(axis cs:5,0.229972229042153)
--(axis cs:5,0.21710238749041)
--(axis cs:6,0.201367112274834)
--(axis cs:7,0.193419176564881)
--(axis cs:8,0.183382007809458)
--(axis cs:9,0.177231572163759)
--(axis cs:10,0.172436763243834)
--(axis cs:11,0.163430350698415)
--(axis cs:12,0.14850104602059)
--(axis cs:13,0.140445885642101)
--(axis cs:14,0.129097951467228)
--(axis cs:15,0.119011932734486)
--(axis cs:16,0.11100724458171)
--(axis cs:17,0.107209749448798)
--(axis cs:18,0.100717728315283)
--(axis cs:19,0.0956815181125684)
--(axis cs:20,0.093052030862558)
--(axis cs:21,0.0925558626404953)
--(axis cs:22,0.0895602419233674)
--(axis cs:23,0.0864687545722167)
--(axis cs:24,0.0839501740497758)
--(axis cs:25,0.0826186015344981)
--(axis cs:26,0.0800063118916811)
--(axis cs:27,0.0772556620426966)
--(axis cs:28,0.0739272513289841)
--(axis cs:29,0.0721861046692106)
--(axis cs:30,0.0692368992028675)
--(axis cs:31,0.0672105644055353)
--(axis cs:32,0.0667106882623138)
--(axis cs:33,0.0650552349660552)
--(axis cs:34,0.0639376700311193)
--(axis cs:35,0.0620206451184199)
--(axis cs:36,0.0607765368097992)
--(axis cs:37,0.0579070180027115)
--(axis cs:38,0.0545703664207188)
--(axis cs:39,0.0538433535900222)
--(axis cs:40,0.0523530619110359)
--(axis cs:41,0.051768342083232)
--(axis cs:42,0.0504241849325001)
--(axis cs:43,0.0481185642402145)
--(axis cs:44,0.045710545161043)
--(axis cs:45,0.045371756538264)
--(axis cs:46,0.044990546049508)
--(axis cs:47,0.0443458527626598)
--(axis cs:48,0.0432809513646924)
--(axis cs:49,0.0420952128386974)
--(axis cs:50,0.0420371336262483)
--(axis cs:50,0.050144703620858)
--(axis cs:50,0.050144703620858)
--(axis cs:49,0.0502754549585824)
--(axis cs:48,0.0516773381990533)
--(axis cs:47,0.05261583875049)
--(axis cs:46,0.0529577185734438)
--(axis cs:45,0.0533003337975661)
--(axis cs:44,0.0538916017878894)
--(axis cs:43,0.0549629757386189)
--(axis cs:42,0.0572175292578913)
--(axis cs:41,0.0587593832348806)
--(axis cs:40,0.0598289930559242)
--(axis cs:39,0.061470196569236)
--(axis cs:38,0.0627638807968933)
--(axis cs:37,0.0655265346730609)
--(axis cs:36,0.0709248101383535)
--(axis cs:35,0.0725343093816172)
--(axis cs:34,0.0748170565986037)
--(axis cs:33,0.075778588619183)
--(axis cs:32,0.0772730696005149)
--(axis cs:31,0.0780187420913705)
--(axis cs:30,0.0805091728505783)
--(axis cs:29,0.0835925110049163)
--(axis cs:28,0.0849067370168797)
--(axis cs:27,0.0917012483844158)
--(axis cs:26,0.0950052049739158)
--(axis cs:25,0.100553781840811)
--(axis cs:24,0.10259593263793)
--(axis cs:23,0.104137924325796)
--(axis cs:22,0.110895913901581)
--(axis cs:21,0.113180826147051)
--(axis cs:20,0.113907891159325)
--(axis cs:19,0.117819473684569)
--(axis cs:18,0.124441617960716)
--(axis cs:17,0.135601617471741)
--(axis cs:16,0.139087329729285)
--(axis cs:15,0.145798767397216)
--(axis cs:14,0.154752245347264)
--(axis cs:13,0.162972124801381)
--(axis cs:12,0.170023564017918)
--(axis cs:11,0.181964722670488)
--(axis cs:10,0.187432317083724)
--(axis cs:9,0.192981497779519)
--(axis cs:8,0.197601683877981)
--(axis cs:7,0.206081075231915)
--(axis cs:6,0.215534939623126)
--(axis cs:5,0.229972229042153)
--cycle;

\path [fill=color6, fill opacity=0.2]
(axis cs:5,0.229972229042153)
--(axis cs:5,0.21710238749041)
--(axis cs:6,0.197933110645406)
--(axis cs:7,0.189291937100327)
--(axis cs:8,0.179402159520036)
--(axis cs:9,0.16949163591136)
--(axis cs:10,0.162308528425315)
--(axis cs:11,0.155903478177593)
--(axis cs:12,0.140229760040631)
--(axis cs:13,0.13121659038493)
--(axis cs:14,0.122081705418005)
--(axis cs:15,0.114970472853538)
--(axis cs:16,0.106128033516187)
--(axis cs:17,0.0959204172252745)
--(axis cs:18,0.0922960317345227)
--(axis cs:19,0.0885679667177011)
--(axis cs:20,0.0869018201470146)
--(axis cs:21,0.0842242409087035)
--(axis cs:22,0.0834570131434878)
--(axis cs:23,0.0828637268076684)
--(axis cs:24,0.0795216346638942)
--(axis cs:25,0.0753099096475641)
--(axis cs:26,0.0690031888826644)
--(axis cs:27,0.065928604785734)
--(axis cs:28,0.0644523410436714)
--(axis cs:29,0.0622955602347303)
--(axis cs:30,0.0600154259655864)
--(axis cs:31,0.0595311898781384)
--(axis cs:32,0.0576769530465846)
--(axis cs:33,0.0558441025514125)
--(axis cs:34,0.0543687464163973)
--(axis cs:35,0.054005315045863)
--(axis cs:36,0.053381735891654)
--(axis cs:37,0.0501053975268719)
--(axis cs:38,0.0490443463140775)
--(axis cs:39,0.0457773223026672)
--(axis cs:40,0.0444437579166671)
--(axis cs:41,0.0412147680572589)
--(axis cs:42,0.0376776940273331)
--(axis cs:43,0.0369846061586345)
--(axis cs:44,0.0359469866775834)
--(axis cs:45,0.0342898724882952)
--(axis cs:46,0.0311633551220264)
--(axis cs:47,0.0305317723342372)
--(axis cs:48,0.027803093403971)
--(axis cs:49,0.0252394262684894)
--(axis cs:50,0.0244029007314614)
--(axis cs:50,0.030590320425863)
--(axis cs:50,0.030590320425863)
--(axis cs:49,0.03169164229898)
--(axis cs:48,0.0346733888087102)
--(axis cs:47,0.0365396748056159)
--(axis cs:46,0.0376011245864357)
--(axis cs:45,0.0388624150932411)
--(axis cs:44,0.0402383989459452)
--(axis cs:43,0.0428975867730728)
--(axis cs:42,0.0436458117869013)
--(axis cs:41,0.0460128109187828)
--(axis cs:40,0.0475262105392332)
--(axis cs:39,0.0490655714237981)
--(axis cs:38,0.051000002213082)
--(axis cs:37,0.0520884369124425)
--(axis cs:36,0.0570678195724984)
--(axis cs:35,0.057803135034045)
--(axis cs:34,0.0581792178423703)
--(axis cs:33,0.0608259873931178)
--(axis cs:32,0.062821133196607)
--(axis cs:31,0.0694077436831555)
--(axis cs:30,0.069973290066966)
--(axis cs:29,0.0717499376075112)
--(axis cs:28,0.0736457107432388)
--(axis cs:27,0.0797681098286559)
--(axis cs:26,0.0824327319376451)
--(axis cs:25,0.0863781892233382)
--(axis cs:24,0.0896348667202973)
--(axis cs:23,0.0918361482954621)
--(axis cs:22,0.0925200246156602)
--(axis cs:21,0.093868879504642)
--(axis cs:20,0.0958379342359319)
--(axis cs:19,0.0979778727332602)
--(axis cs:18,0.10067498144566)
--(axis cs:17,0.103585547326984)
--(axis cs:16,0.116905828737245)
--(axis cs:15,0.129694403672778)
--(axis cs:14,0.138460634149919)
--(axis cs:13,0.150662158055955)
--(axis cs:12,0.156465677433662)
--(axis cs:11,0.168564955791855)
--(axis cs:10,0.179547999326619)
--(axis cs:9,0.186881923897937)
--(axis cs:8,0.19496912778911)
--(axis cs:7,0.202599435171523)
--(axis cs:6,0.212877051397521)
--(axis cs:5,0.229972229042153)
--cycle;

\path [fill=white!49.8039215686275!black, fill opacity=0.2]
(axis cs:5,0.229972229042153)
--(axis cs:5,0.21710238749041)
--(axis cs:6,0.198422311876394)
--(axis cs:7,0.182404420608203)
--(axis cs:8,0.171175190695805)
--(axis cs:9,0.155966761982057)
--(axis cs:10,0.150429333813633)
--(axis cs:11,0.145971168784998)
--(axis cs:12,0.138162514719813)
--(axis cs:13,0.126429625429804)
--(axis cs:14,0.121102311008568)
--(axis cs:15,0.109890728023218)
--(axis cs:16,0.103872156456708)
--(axis cs:17,0.099731937194674)
--(axis cs:18,0.0962836362133676)
--(axis cs:19,0.0911700609661487)
--(axis cs:20,0.0876427896604865)
--(axis cs:21,0.0817234951178875)
--(axis cs:22,0.0775183888517809)
--(axis cs:23,0.0748030917663793)
--(axis cs:24,0.0726557092715857)
--(axis cs:25,0.0710983769869838)
--(axis cs:26,0.069340045669544)
--(axis cs:27,0.0681427310025885)
--(axis cs:28,0.0640272728676596)
--(axis cs:29,0.0618530703904007)
--(axis cs:30,0.0597991388011802)
--(axis cs:31,0.059070002958155)
--(axis cs:32,0.0583991324624368)
--(axis cs:33,0.0543462694123637)
--(axis cs:34,0.053526038177651)
--(axis cs:35,0.0513044444757551)
--(axis cs:36,0.0479850585691134)
--(axis cs:37,0.0474366498732038)
--(axis cs:38,0.0460482969526124)
--(axis cs:39,0.041948819093641)
--(axis cs:40,0.0397182471316713)
--(axis cs:41,0.0384043640993883)
--(axis cs:42,0.0369503543182206)
--(axis cs:43,0.0367939288914322)
--(axis cs:44,0.0367683515397858)
--(axis cs:45,0.034750875866599)
--(axis cs:46,0.0342180466977526)
--(axis cs:47,0.0322672598470712)
--(axis cs:48,0.0321380534715246)
--(axis cs:49,0.0312323600575134)
--(axis cs:50,0.0297799856822928)
--(axis cs:50,0.0365805205556687)
--(axis cs:50,0.0365805205556687)
--(axis cs:49,0.0377249247325394)
--(axis cs:48,0.0391077278220999)
--(axis cs:47,0.0392546498405026)
--(axis cs:46,0.0441005063603495)
--(axis cs:45,0.0443786507624269)
--(axis cs:44,0.0460453520389622)
--(axis cs:43,0.0461781506097178)
--(axis cs:42,0.0464953074729254)
--(axis cs:41,0.048341703973526)
--(axis cs:40,0.0502385898197045)
--(axis cs:39,0.051890735527929)
--(axis cs:38,0.0532969547861117)
--(axis cs:37,0.0550886019396297)
--(axis cs:36,0.056415620456756)
--(axis cs:35,0.0630808748126765)
--(axis cs:34,0.0674416885872101)
--(axis cs:33,0.0682677578094553)
--(axis cs:32,0.0739376173061603)
--(axis cs:31,0.0744455406322768)
--(axis cs:30,0.0755485732779095)
--(axis cs:29,0.0775043148338443)
--(axis cs:28,0.0799948037179423)
--(axis cs:27,0.0828536278059547)
--(axis cs:26,0.0848507684997448)
--(axis cs:25,0.0867318405050581)
--(axis cs:24,0.0897456012503485)
--(axis cs:23,0.092617183601052)
--(axis cs:22,0.0959257978943379)
--(axis cs:21,0.0977928692945899)
--(axis cs:20,0.10253250717398)
--(axis cs:19,0.107866617761485)
--(axis cs:18,0.111453642949157)
--(axis cs:17,0.114607615271283)
--(axis cs:16,0.118723185990698)
--(axis cs:15,0.123076849492475)
--(axis cs:14,0.13176902251874)
--(axis cs:13,0.137649867800307)
--(axis cs:12,0.151170991389077)
--(axis cs:11,0.159852023880272)
--(axis cs:10,0.163912119686737)
--(axis cs:9,0.169854881150956)
--(axis cs:8,0.182363952621547)
--(axis cs:7,0.19698159661006)
--(axis cs:6,0.208091101994526)
--(axis cs:5,0.229972229042153)
--cycle;

\path [fill=color7, fill opacity=0.2]
(axis cs:5,0.229972229042153)
--(axis cs:5,0.21710238749041)
--(axis cs:6,0.196264908284915)
--(axis cs:7,0.179159444411165)
--(axis cs:8,0.158215247791367)
--(axis cs:9,0.138773477847566)
--(axis cs:10,0.133101638740165)
--(axis cs:11,0.12187163882154)
--(axis cs:12,0.114764635492206)
--(axis cs:13,0.109548486635605)
--(axis cs:14,0.107503100045698)
--(axis cs:15,0.100538509716229)
--(axis cs:16,0.0977620952859619)
--(axis cs:17,0.0947922137167023)
--(axis cs:18,0.0897518140452216)
--(axis cs:19,0.0863283187394856)
--(axis cs:20,0.079812692478759)
--(axis cs:21,0.0777490502462248)
--(axis cs:22,0.0747788563970806)
--(axis cs:23,0.0725325371841378)
--(axis cs:24,0.0706083658276488)
--(axis cs:25,0.0679474513671912)
--(axis cs:26,0.0652977637999008)
--(axis cs:27,0.0626060644009678)
--(axis cs:28,0.0606840120136322)
--(axis cs:29,0.0570339558949691)
--(axis cs:30,0.0552337449347227)
--(axis cs:31,0.0537179225225795)
--(axis cs:32,0.0482840168619269)
--(axis cs:33,0.0468740638704665)
--(axis cs:34,0.0442798625780387)
--(axis cs:35,0.0426067460029992)
--(axis cs:36,0.0381247917682028)
--(axis cs:37,0.0379898089428784)
--(axis cs:38,0.036976407597643)
--(axis cs:39,0.0365054357314769)
--(axis cs:40,0.0340722735997495)
--(axis cs:41,0.0338937617391526)
--(axis cs:42,0.032108486074996)
--(axis cs:43,0.0313796559275976)
--(axis cs:44,0.0312384958586265)
--(axis cs:45,0.0300152361162018)
--(axis cs:46,0.0298313915267915)
--(axis cs:47,0.0291882812464132)
--(axis cs:48,0.0290694324867079)
--(axis cs:49,0.0276396947388601)
--(axis cs:50,0.027011674001716)
--(axis cs:50,0.0320049466399175)
--(axis cs:50,0.0320049466399175)
--(axis cs:49,0.0328232538734355)
--(axis cs:48,0.0340191880195649)
--(axis cs:47,0.0342816820394295)
--(axis cs:46,0.0352812375467188)
--(axis cs:45,0.0353684026861457)
--(axis cs:44,0.0361569355520013)
--(axis cs:43,0.0363428779314576)
--(axis cs:42,0.036677656823414)
--(axis cs:41,0.0392586126277007)
--(axis cs:40,0.0394825689569534)
--(axis cs:39,0.0417497984889792)
--(axis cs:38,0.042521608196938)
--(axis cs:37,0.0442050366601581)
--(axis cs:36,0.044418112666756)
--(axis cs:35,0.0503178565669617)
--(axis cs:34,0.0517548356140035)
--(axis cs:33,0.053460862761669)
--(axis cs:32,0.0549110397278945)
--(axis cs:31,0.0581507612486334)
--(axis cs:30,0.0591445864524385)
--(axis cs:29,0.0611755324725529)
--(axis cs:28,0.0657406792942428)
--(axis cs:27,0.0696057127614336)
--(axis cs:26,0.0734465755862433)
--(axis cs:25,0.0768309149349558)
--(axis cs:24,0.0790663630143294)
--(axis cs:23,0.0820636944795517)
--(axis cs:22,0.0854657065863658)
--(axis cs:21,0.087357996154879)
--(axis cs:20,0.0892104800076663)
--(axis cs:19,0.0969583970762643)
--(axis cs:18,0.100046808080343)
--(axis cs:17,0.104130386758315)
--(axis cs:16,0.107580816321894)
--(axis cs:15,0.113314198016437)
--(axis cs:14,0.123076711006694)
--(axis cs:13,0.126379417335512)
--(axis cs:12,0.133721353643156)
--(axis cs:11,0.139586690940326)
--(axis cs:10,0.148689324622315)
--(axis cs:9,0.156830192359553)
--(axis cs:8,0.171631252455669)
--(axis cs:7,0.193927703225881)
--(axis cs:6,0.211627373438375)
--(axis cs:5,0.229972229042153)
--cycle;

\path [fill=color8, fill opacity=0.2]
(axis cs:5,0.229972229042153)
--(axis cs:5,0.21710238749041)
--(axis cs:6,0.160278831014592)
--(axis cs:7,0.130782163847823)
--(axis cs:8,0.116476611076302)
--(axis cs:9,0.0964332893365131)
--(axis cs:10,0.0909970940271918)
--(axis cs:11,0.0848937200910018)
--(axis cs:12,0.0799838492545874)
--(axis cs:13,0.0753719536792764)
--(axis cs:14,0.0725192375026625)
--(axis cs:15,0.0662305828369612)
--(axis cs:16,0.0631898957810735)
--(axis cs:17,0.0597904014311965)
--(axis cs:18,0.0566694875673723)
--(axis cs:19,0.0549615438692448)
--(axis cs:20,0.0524587204684884)
--(axis cs:21,0.0502854282954943)
--(axis cs:22,0.0485177375759072)
--(axis cs:23,0.0454456648919411)
--(axis cs:24,0.0444372309565432)
--(axis cs:25,0.0407273515835956)
--(axis cs:26,0.0384756222783664)
--(axis cs:27,0.0367077372183817)
--(axis cs:28,0.0357680803128359)
--(axis cs:29,0.0338661626496755)
--(axis cs:30,0.0328506051745369)
--(axis cs:31,0.0317215049957481)
--(axis cs:32,0.0288762770262921)
--(axis cs:33,0.0282302502599518)
--(axis cs:34,0.0276126265046207)
--(axis cs:35,0.027171861652372)
--(axis cs:36,0.0259207942838157)
--(axis cs:37,0.0244240034290502)
--(axis cs:38,0.0226805049232498)
--(axis cs:39,0.0225551593726975)
--(axis cs:40,0.0216690437706978)
--(axis cs:41,0.0211328043231009)
--(axis cs:42,0.020691153889118)
--(axis cs:43,0.0195500586840891)
--(axis cs:44,0.019347499671286)
--(axis cs:45,0.0182153763867652)
--(axis cs:46,0.0182153763867652)
--(axis cs:47,0.0181595108243169)
--(axis cs:48,0.0180503633167558)
--(axis cs:49,0.0176131802485801)
--(axis cs:50,0.0174261833006113)
--(axis cs:50,0.0190491751478515)
--(axis cs:50,0.0190491751478515)
--(axis cs:49,0.0192911309590044)
--(axis cs:48,0.0197975457074311)
--(axis cs:47,0.019864637828914)
--(axis cs:46,0.020095003759592)
--(axis cs:45,0.020095003759592)
--(axis cs:44,0.0211877494332459)
--(axis cs:43,0.0214391028494531)
--(axis cs:42,0.0255267584376508)
--(axis cs:41,0.0268282171956197)
--(axis cs:40,0.0272343979720654)
--(axis cs:39,0.0277372755738238)
--(axis cs:38,0.027855769914311)
--(axis cs:37,0.0294444760602151)
--(axis cs:36,0.0305886549822034)
--(axis cs:35,0.0320135843531685)
--(axis cs:34,0.0322725951091343)
--(axis cs:33,0.0326056884009755)
--(axis cs:32,0.0335400814715294)
--(axis cs:31,0.0358461008734134)
--(axis cs:30,0.0368527717228782)
--(axis cs:29,0.0374051929806618)
--(axis cs:28,0.0393245500594882)
--(axis cs:27,0.0404841037423493)
--(axis cs:26,0.0439548761617103)
--(axis cs:25,0.0456213709484763)
--(axis cs:24,0.0480627903424171)
--(axis cs:23,0.0492098431367699)
--(axis cs:22,0.054460064143617)
--(axis cs:21,0.0567772872821245)
--(axis cs:20,0.0588415103598554)
--(axis cs:19,0.0613160108730878)
--(axis cs:18,0.0625221929845047)
--(axis cs:17,0.0652598812884552)
--(axis cs:16,0.0694508371730795)
--(axis cs:15,0.0740021478936112)
--(axis cs:14,0.0787037147583354)
--(axis cs:13,0.0841041113492363)
--(axis cs:12,0.0885640030903338)
--(axis cs:11,0.0949998034039603)
--(axis cs:10,0.100816834998825)
--(axis cs:9,0.108115914008586)
--(axis cs:8,0.133084280281661)
--(axis cs:7,0.145019877547662)
--(axis cs:6,0.175050770556326)
--(axis cs:5,0.229972229042153)
--cycle;

\path [fill=color0, fill opacity=0.2]
(axis cs:5,0.229972229042153)
--(axis cs:5,0.21710238749041)
--(axis cs:6,0.155798026816273)
--(axis cs:7,0.123102902135651)
--(axis cs:8,0.113917307877551)
--(axis cs:9,0.107687405618692)
--(axis cs:10,0.0980923026779431)
--(axis cs:11,0.0893363431119514)
--(axis cs:12,0.0813209067261911)
--(axis cs:13,0.076837672779768)
--(axis cs:14,0.0735844030529145)
--(axis cs:15,0.0655796375150117)
--(axis cs:16,0.0605115438192216)
--(axis cs:17,0.0579995233601734)
--(axis cs:18,0.0558423214004408)
--(axis cs:19,0.0524743811641374)
--(axis cs:20,0.0508619363471092)
--(axis cs:21,0.0491492636232133)
--(axis cs:22,0.0450983039860922)
--(axis cs:23,0.0429462090826068)
--(axis cs:24,0.0410030393429595)
--(axis cs:25,0.0396521537429576)
--(axis cs:26,0.0374339302250458)
--(axis cs:27,0.0359321951970915)
--(axis cs:28,0.0349704747668969)
--(axis cs:29,0.0337971220890256)
--(axis cs:30,0.031514895436772)
--(axis cs:31,0.0309022872935002)
--(axis cs:32,0.0297064539647968)
--(axis cs:33,0.0278288966380216)
--(axis cs:34,0.02680409807476)
--(axis cs:35,0.026083345675027)
--(axis cs:36,0.025212100831817)
--(axis cs:37,0.0245184602938162)
--(axis cs:38,0.024294414603868)
--(axis cs:39,0.0231635550334681)
--(axis cs:40,0.0226553917064914)
--(axis cs:41,0.0225593094010372)
--(axis cs:42,0.0223011822378984)
--(axis cs:43,0.0217759309976266)
--(axis cs:44,0.0216878825172237)
--(axis cs:45,0.0216603454702873)
--(axis cs:46,0.0215851637796201)
--(axis cs:47,0.0214151340243749)
--(axis cs:48,0.0213724569983574)
--(axis cs:49,0.021171693058866)
--(axis cs:50,0.0210557032722952)
--(axis cs:50,0.0261395371043397)
--(axis cs:50,0.0261395371043397)
--(axis cs:49,0.0262133817456179)
--(axis cs:48,0.026606491832663)
--(axis cs:47,0.0266320451665692)
--(axis cs:46,0.0271504652085832)
--(axis cs:45,0.0275640242030838)
--(axis cs:44,0.0275854347726645)
--(axis cs:43,0.0276557622146636)
--(axis cs:42,0.0279664825045331)
--(axis cs:41,0.0281666500760666)
--(axis cs:40,0.028228229677212)
--(axis cs:39,0.0287755320743056)
--(axis cs:38,0.0298173158853254)
--(axis cs:37,0.0299140888730845)
--(axis cs:36,0.0304290392131983)
--(axis cs:35,0.0310671839014509)
--(axis cs:34,0.0315544314795978)
--(axis cs:33,0.0325526578394854)
--(axis cs:32,0.0344877376009545)
--(axis cs:31,0.0367769959348246)
--(axis cs:30,0.0373050200619239)
--(axis cs:29,0.0396968565309007)
--(axis cs:28,0.0403922414727686)
--(axis cs:27,0.0415693519660093)
--(axis cs:26,0.0433126198109958)
--(axis cs:25,0.0447469460910148)
--(axis cs:24,0.0458182445302726)
--(axis cs:23,0.0478206434109306)
--(axis cs:22,0.0511111098022589)
--(axis cs:21,0.0540650647848261)
--(axis cs:20,0.0549005522104057)
--(axis cs:19,0.0569799960883419)
--(axis cs:18,0.06034972303679)
--(axis cs:17,0.0624337419590016)
--(axis cs:16,0.0646839184383747)
--(axis cs:15,0.0692013019408059)
--(axis cs:14,0.0774670783882541)
--(axis cs:13,0.0819812010197471)
--(axis cs:12,0.0866371536323084)
--(axis cs:11,0.0942332586793168)
--(axis cs:10,0.103847416899398)
--(axis cs:9,0.114566745100525)
--(axis cs:8,0.121191619726332)
--(axis cs:7,0.132794493749726)
--(axis cs:6,0.17772078157687)
--(axis cs:5,0.229972229042153)
--cycle;

\addplot [line width=2pt, color0]
table {%
5 0.223537308266282
6 0.215106257648508
7 0.19410934409419
8 0.185400094303649
9 0.17659597044002
10 0.165819737853608
11 0.158818427476327
12 0.151037390517623
13 0.148911009186116
14 0.139632809438272
15 0.13647216363216
16 0.135468225746457
17 0.134956091447036
18 0.134087225477909
19 0.127195817252017
20 0.126622394114084
21 0.126392184436021
22 0.122415381712342
23 0.118961806854965
24 0.117020650667617
25 0.11676936104974
26 0.116287274259848
27 0.115439911462744
28 0.113455031751373
29 0.111093313802576
30 0.110036572043108
31 0.109144733925647
32 0.107942014762115
33 0.107106190899556
34 0.106940490816706
35 0.106805107568487
36 0.106190986857964
37 0.104742698926353
38 0.104580139107721
39 0.104449993254643
40 0.104261804804929
41 0.103601681402623
42 0.103416996057268
43 0.10224322441642
44 0.101811004517742
45 0.101719776673223
46 0.10119023118016
47 0.0999610015056053
48 0.0975283017637233
49 0.0974327740741479
50 0.0968397847220513
};
\addplot [line width=2pt, color1]
table {%
5 0.223537308266282
6 0.209163182901579
7 0.202960832368035
8 0.195023479540642
9 0.180708307084355
10 0.174034748627196
11 0.163272016633155
12 0.150619591582234
13 0.139658551974039
14 0.133834912410882
15 0.126698002199548
16 0.124370514030187
17 0.119191325183156
18 0.113462947752973
19 0.111827787075258
20 0.105799022092879
21 0.0952066661386408
22 0.0903592788267765
23 0.0875272216348163
24 0.0855454436184619
25 0.0838350707011541
26 0.0801145686405212
27 0.0778933034425604
28 0.0752682780078968
29 0.0737828936900674
30 0.0699663733732616
31 0.0677421775928891
32 0.0649705440371463
33 0.0636943507799752
34 0.0612674405862599
35 0.0610975842461438
36 0.0594653881728482
37 0.0580510192259609
38 0.0524431511252887
39 0.049389501124473
40 0.0460436226222198
41 0.0453901071083291
42 0.0421626860856872
43 0.0413614885216454
44 0.0391803722357391
45 0.0376378492191086
46 0.0347659542972145
47 0.0343912444030556
48 0.0341545977924367
49 0.0336225011496011
50 0.0328180733727241
};
\addplot [line width=2pt, color2]
table {%
5 0.223537308266282
6 0.208412633177894
7 0.18407762564871
8 0.151307416592698
9 0.135232983098698
10 0.118824342457717
11 0.112035852340694
12 0.103764537864815
13 0.0973154506301903
14 0.0934387772916905
15 0.0904390028458012
16 0.0851873824368218
17 0.0831466544261707
18 0.0793541504439931
19 0.0760506509671868
20 0.0731110962740774
21 0.069785475149304
22 0.0686972804205654
23 0.0663138915754369
24 0.0643124408966017
25 0.0631674027750305
26 0.0616736530515517
27 0.0607389908850305
28 0.0601437691048003
29 0.0589082145015179
30 0.0580995571505138
31 0.055837205820801
32 0.0549885751686283
33 0.0538994594388228
34 0.0505699187111042
35 0.0488051596764892
36 0.0473089395911094
37 0.0469145117850727
38 0.0460874913717929
39 0.0459883882057849
40 0.045694746496286
41 0.0443873523650657
42 0.0440431372126291
43 0.0437644768640987
44 0.0412395561858622
45 0.0411879368566913
46 0.0406308241126755
47 0.0402065606044973
48 0.0389657795098213
49 0.0386614770649833
50 0.0384893038918701
};
\addplot [line width=2pt, color3]
table {%
5 0.223537308266282
6 0.168551170551198
7 0.143479111088093
8 0.12401794613378
9 0.10361760091146
10 0.0940949335139278
11 0.084031097880096
12 0.0781221412132905
13 0.0724138629506043
14 0.0701114674541989
15 0.0672540779171246
16 0.0649067283962217
17 0.0628358001803174
18 0.0605983444259524
19 0.058005162894315
20 0.0556148169061995
21 0.0539045825184188
22 0.0513868710202452
23 0.0493815505400493
24 0.0478336125517936
25 0.0467754229687245
26 0.0452366786427762
27 0.0417864817662968
28 0.0412797508101065
29 0.0409827985173103
30 0.0405471242495987
31 0.0400666720699541
32 0.0398518735625655
33 0.0389568662610598
34 0.0367571071651087
35 0.0362198532666936
36 0.0350975268125793
37 0.0335737331827646
38 0.0318107587318433
39 0.0313312037672327
40 0.0308083074795569
41 0.0304229849905554
42 0.0299284731943038
43 0.0296050484748145
44 0.0292942330162028
45 0.027920984612312
46 0.0277716317244492
47 0.0274335153698018
48 0.0272909128742581
49 0.0267433662217015
50 0.0265617883383734
};
\addplot [line width=2pt, color4]
table {%
5 0.223537308266282
6 0.199148168083908
7 0.173414372289838
8 0.127437641751415
9 0.115704887100837
10 0.0972214111032524
11 0.0910253564311435
12 0.0863572700446205
13 0.0827128464319992
14 0.0754630656124461
15 0.069895452896262
16 0.0661831626266876
17 0.0633421633532473
18 0.0609017337055315
19 0.0581858423396455
20 0.0569941024361265
21 0.0550779900824178
22 0.0537211884881498
23 0.0536679661857417
24 0.053092060548854
25 0.0501254289713313
26 0.0477615307031864
27 0.0462901678754109
28 0.0447157369877129
29 0.0417676557586737
30 0.0403270455128693
31 0.0382402998359182
32 0.0367546310038317
33 0.0350628332504122
34 0.0348258561151988
35 0.0340134078814417
36 0.0314637784862704
37 0.0302077504417178
38 0.0288694151173123
39 0.0281859847658855
40 0.0276944327332021
41 0.027402390111306
42 0.026796298864668
43 0.0265718680919146
44 0.026397175315935
45 0.0260121399445774
46 0.0256558444622012
47 0.0255008431860473
48 0.0247235276463344
49 0.0242205839715802
50 0.0241868875426742
};
\addplot [line width=2pt, color5]
table {%
5 0.223537308266282
6 0.20845102594898
7 0.199750125898398
8 0.19049184584372
9 0.185106534971639
10 0.179934540163779
11 0.172697536684451
12 0.159262305019254
13 0.151709005221741
14 0.141925098407246
15 0.132405350065851
16 0.125047287155497
17 0.12140568346027
18 0.112579673137999
19 0.106750495898569
20 0.103479961010942
21 0.102868344393773
22 0.100228077912474
23 0.0953033394490063
24 0.0932730533438527
25 0.0915861916876547
26 0.0875057584327985
27 0.0844784552135562
28 0.0794169941729319
29 0.0778893078370635
30 0.0748730360267229
31 0.0726146532484529
32 0.0719918789314144
33 0.0704169117926191
34 0.0693773633148615
35 0.0672774772500185
36 0.0658506734740764
37 0.0617167763378862
38 0.058667123608806
39 0.0576567750796291
40 0.05609102748348
41 0.0552638626590563
42 0.0538208570951957
43 0.0515407699894167
44 0.0498010734744662
45 0.0493360451679151
46 0.0489741323114759
47 0.0484808457565749
48 0.0474791447818729
49 0.0461853338986399
50 0.0460909186235532
};
\addplot [line width=2pt, color6]
table {%
5 0.223537308266282
6 0.205405081021464
7 0.195945686135925
8 0.187185643654573
9 0.178186779904648
10 0.170928263875967
11 0.162234216984724
12 0.148347718737146
13 0.140939374220442
14 0.130271169783962
15 0.122332438263158
16 0.111516931126716
17 0.0997529822761293
18 0.0964855065900913
19 0.0932729197254807
20 0.0913698771914732
21 0.0890465602066727
22 0.087988518879574
23 0.0873499375515653
24 0.0845782506920957
25 0.0808440494354512
26 0.0757179604101547
27 0.072848357307195
28 0.0690490258934551
29 0.0670227489211207
30 0.0649943580162762
31 0.064469466780647
32 0.0602490431215958
33 0.0583350449722651
34 0.0562739821293838
35 0.055904225039954
36 0.0552247777320762
37 0.0510969172196572
38 0.0500221742635797
39 0.0474214468632327
40 0.0459849842279502
41 0.0436137894880209
42 0.0406617529071172
43 0.0399410964658537
44 0.0380926928117643
45 0.0365761437907682
46 0.0343822398542311
47 0.0335357235699265
48 0.0312382411063406
49 0.0284655342837347
50 0.0274966105786622
};
\addplot [line width=2pt, white!49.8039215686275!black]
table {%
5 0.223537308266282
6 0.20325670693546
7 0.189693008609131
8 0.176769571658676
9 0.162910821566507
10 0.157170726750185
11 0.152911596332635
12 0.144666753054445
13 0.132039746615055
14 0.126435666763654
15 0.116483788757847
16 0.111297671223703
17 0.107169776232978
18 0.103868639581262
19 0.099518339363817
20 0.0950876484172333
21 0.0897581822062387
22 0.0867220933730594
23 0.0837101376837157
24 0.0812006552609671
25 0.0789151087460209
26 0.0770954070846444
27 0.0754981794042716
28 0.0720110382928009
29 0.0696786926121225
30 0.0676738560395448
31 0.0667577717952159
32 0.0661683748842985
33 0.0613070136109095
34 0.0604838633824305
35 0.0571926596442158
36 0.0522003395129347
37 0.0512626259064168
38 0.0496726258693621
39 0.046919777310785
40 0.0449784184756879
41 0.0433730340364571
42 0.041722830895573
43 0.041486039750575
44 0.041406851789374
45 0.039564763314513
46 0.039159276529051
47 0.0357609548437869
48 0.0356228906468123
49 0.0344786423950264
50 0.0331802531189807
};
\addplot [line width=2pt, color7]
table {%
5 0.223537308266282
6 0.203946140861645
7 0.186543573818523
8 0.164923250123518
9 0.14780183510356
10 0.14089548168124
11 0.130729164880933
12 0.124242994567681
13 0.117963951985559
14 0.115289905526196
15 0.106926353866333
16 0.102671455803928
17 0.0994613002375085
18 0.0948993110627822
19 0.091643357907875
20 0.0845115862432126
21 0.0825535232005519
22 0.0801222814917232
23 0.0772981158318447
24 0.0748373644209891
25 0.0723891831510735
26 0.069372169693072
27 0.0661058885812007
28 0.0632123456539375
29 0.059104744183761
30 0.0571891656935806
31 0.0559343418856065
32 0.0515975282949107
33 0.0501674633160677
34 0.0480173490960211
35 0.0464623012849805
36 0.0412714522174794
37 0.0410974228015182
38 0.0397490078972905
39 0.0391276171102281
40 0.0367774212783514
41 0.0365761871834266
42 0.034393071449205
43 0.0338612669295276
44 0.0336977157053139
45 0.0326918194011738
46 0.0325563145367552
47 0.0317349816429214
48 0.0315443102531364
49 0.0302314743061478
50 0.0295083103208168
};
\addplot [line width=2pt, color8]
table {%
5 0.223537308266282
6 0.167664800785459
7 0.137901020697743
8 0.124780445678982
9 0.10227460167255
10 0.0959069645130084
11 0.089946761747481
12 0.0842739261724606
13 0.0797380325142564
14 0.0756114761304989
15 0.0701163653652862
16 0.0663203664770765
17 0.0625251413598259
18 0.0595958402759385
19 0.0581387773711663
20 0.0556501154141719
21 0.0535313577888094
22 0.0514889008597621
23 0.0473277540143555
24 0.0462500106494802
25 0.0431743612660359
26 0.0412152492200384
27 0.0385959204803655
28 0.0375463151861621
29 0.0356356778151686
30 0.0348516884487076
31 0.0337838029345808
32 0.0312081792489108
33 0.0304179693304636
34 0.0299426108068775
35 0.0295927230027702
36 0.0282547246330095
37 0.0269342397446327
38 0.0252681374187804
39 0.0251462174732606
40 0.0244517208713816
41 0.0239805107593603
42 0.0231089561633844
43 0.0204945807667711
44 0.020267624552266
45 0.0191551900731786
46 0.0191551900731786
47 0.0190120743266155
48 0.0189239545120935
49 0.0184521556037923
50 0.0182376792242314
};
\addplot [line width=2pt, colorHEBO]
table {%
5 0.223537308266282
6 0.166759404196571
7 0.127948697942688
8 0.117554463801941
9 0.111127075359608
10 0.10096985978867
11 0.0917848008956341
12 0.0839790301792497
13 0.0794094368997576
14 0.0755257407205843
15 0.0673904697279088
16 0.0625977311287982
17 0.0602166326595875
18 0.0580960222186154
19 0.0547271886262396
20 0.0528812442787575
21 0.0516071642040197
22 0.0481047068941756
23 0.0453834262467687
24 0.0434106419366161
25 0.0421995499169862
26 0.0403732750180208
27 0.0387507735815504
28 0.0376813581198327
29 0.0367469893099632
30 0.034409957749348
31 0.0338396416141624
32 0.0320970957828756
33 0.0301907772387535
34 0.0291792647771789
35 0.0285752647882389
36 0.0278205700225076
37 0.0272162745834503
38 0.0270558652445967
39 0.0259695435538868
40 0.0254418106918517
41 0.0253629797385519
42 0.0251338323712158
43 0.0247158466061451
44 0.0246366586449441
45 0.0246121848366856
46 0.0243678144941016
47 0.024023589595472
48 0.0239894744155102
49 0.023692537402242
50 0.0235976201883174
};
\end{axis}

\end{tikzpicture}